\documentclass[a4paper,10pt,twocolumn]{article}

\usepackage{times}
\usepackage[T1]{fontenc} %
\usepackage{array}
\usepackage[numbers]{natbib}
\usepackage{url}
\usepackage{microtype}
\usepackage[]{graphicx}
\usepackage{tikz}
\usetikzlibrary{positioning, shadows.blur, arrows.meta}
\usepackage{wrapfig}
\usepackage[skip=1pt]{caption}
\usepackage{subcaption}
\usepackage{adjustbox}
\usepackage{booktabs}
\usepackage{siunitx}
\usepackage{placeins}
\usepackage{etoc} %
\usepackage{multicol} %
\usepackage{xcolor}
\usepackage{fancyhdr}
\usepackage{hyperref}
\usepackage{amsmath}
\usepackage{amssymb}
\usepackage{mathtools}
\usepackage{pifont}
\usepackage{amsthm}
\usepackage{authblk}
\usepackage{multirow}
\usepackage{balance}

\usepackage[left=1.15in, right=1.15in, top=1.2in, bottom=1in, headsep=.25in]{geometry}

\usepackage[bottom]{footmisc}

\hypersetup{colorlinks=true, linkcolor=blue, citecolor=blue}

\renewenvironment{abstract}
{
    \small
    \begin{quote}
    \centerline{\normalsize \sc \abstractname}
    \vskip -0.2in
    \noindent\par
}
{
    \noindent
    \end{quote}
}


\begin{document}

\newcommand{\SM}{\textbf{SM} }

\newcommand{\eqtext}[1]{\ensuremath{\stackrel{#1}{=}}}
\newcommand{\leqtext}[1]{\ensuremath{\stackrel{#1}{\leq}}}
\newcommand{\R}{\mathbb{R}}
\DeclareRobustCommand{\C}{\mathcal{C}} %
\newcommand{\I}{\pmb{I}}

\newcommand{\funcMain}{u}
\newcommand{\spaceMain}{\mathcal{X}}
\newcommand{\distMain}{\mathfrak{X}}
\newcommand{\funcCoeffsDomain}{a}
\newcommand{\spaceCoeffsDomain}{\mathcal{A}}
\newcommand{\funcCoeffsBoundary}{q}
\newcommand{\funcExtension}{\psi}
\newcommand{\bcDirichlet}{\alpha}
\newcommand{\bcNeumann}{\beta}
\newcommand{\bcSource}{\gamma}
\newcommand{\spaceCoeffsBoundary}{\mathcal{Q}}
\newcommand{\operator}{\mathcal{G}}
\newcommand{\operatorCore}{\Phi}
\newcommand{\operatorExtender}{\Psi}
\newcommand{\params}{\theta}
\newcommand{\paramSpace}{\Theta}
\newcommand{\loss}{\mathcal{L}}
\newcommand{\dataset}{\mathcal{D}}
\newcommand{\DistUniform}{\mathcal{U}}
\newcommand{\error}{e}

\newcommand{\featCoords}{\pmb{\rm x}}
\newcommand{\featCoeffsDomain}{\pmb{\rm \funcCoeffsDomain}}
\newcommand{\featCoeffsBoundary}{\pmb{\rm \funcCoeffsBoundary}}
\newcommand{\featExtension}{\pmb{\rm \funcExtension}}

\newcommand{\funcMainVec}{\pmb{\funcMain}}
\newcommand{\funcSource}{\pmb{S}}

\newcommand{\FF}{\textsl{FF}}
\newcommand{\CA}{\textsl{CA}}

\newcommand{\daedalus}{Daedalus}
\newcommand{\feff}{f_{\mathrm{eff}}} %
\newcommand{\feffp}[1]{f_{\mathrm{eff},#1}} %
\newcommand{\Htrain}{H_{\mathrm{train}}} %
\newcommand{\Hinf}{H_{\mathrm{inf}}} %
\newcommand{\smin}{\sigma_{\min}} %
\newcommand{\indicator}{\text{\usefont{U}{dsrom}{m}{n}1}} %
\newcommand{\relLtwo}{\text{relative-}L^2}
\newcommand{\upmark}{{\color{red!75!black}$\scriptstyle\blacktriangle$}}
\newcommand{\dnmark}{{\color{green!50!black}$\scriptstyle\blacktriangledown$}}
\newcommand{\upval}[1]{{\color{red!75!black}\tiny #1}}
\newcommand{\dnval}[1]{{\color{green!50!black}\tiny #1}}

\newcommand{\encoder}{\mathcal{E}}
\newcommand{\decoder}{\mathcal{D}}
\newcommand{\processor}{\mathcal{P}}

\newcommand{\emb}{\zeta}
\newcommand{\embstat}{\emb_{\mathrm{s}}}
\newcommand{\embmorph}{\emb_{\mathrm{m}}}
\newcommand{\hatembmorph}{\hat{\emb}_{\mathrm{m}}}
\newcommand{\hatembstat}{\hat{\emb}_{\mathrm{s}}}
\newcommand{\embdim}{d_{\emb}}
\newcommand{\latdim}{\embdim} %

\newenvironment{visualizations}
{\clearpage \newgeometry{left=20mm,right=20mm,top=30mm,headheight=10mm,bottom=30mm,footskip=15mm}}
{\restoregeometry}

\graphicspath{{img/}}
\newtheorem{proposition}{Proposition}
\lhead{}
\rhead{}

\title{
    \vskip -0.5in
    \par\noindent\rule{\textwidth}{2pt}
    \bfseries
    Self-Supervised Learning of
    \\ Universal Latent Spaces for Operator Learning
    \vskip -0.2in
    \par\noindent\rule{\textwidth}{2pt}
}
\date{}
\author[1,3]{\bf {Sepehr Mousavi}}
\author[2,3]{\bf Levi Lingsch}
\author[2,3]{\bf Siddhartha Mishra}
\author[1,3]{\bf Laura De Lorenzis}
\affil[1]{Department of Mechanical and Process Engineering, ETH Zurich, Switzerland}
\affil[2]{Seminar for Applied Mathematics, ETH Zurich, Switzerland}
\affil[3]{ETH AI Center, Zurich, Switzerland}

\makeatletter
\twocolumn[%
  {\renewenvironment{tabular}[2][]{\begin{center}}{\end{center}}\@maketitle}%
  \vspace{0.2cm}
]
\makeatother

\begin{abstract}
Neural operators fit their representation of functions to the task they are trained on, so every new task rebuilds a reduced space from labeled data. Following the self-supervised pre-training that made representations shareable in natural language processing and computer vision, we decouple representation learning from operator learning. A functional autoencoder is pre-trained on analytic functions synthesized on the fly, with no solver and no labels, to represent functions on arbitrary point clouds, and the latent space it induces is frozen and reused across tasks. The representation is exactly equivariant to translation, dilation, and amplitude, and reconstructs solutions of partial differential equations (PDEs) and natural images on geometries it never saw.
Being task-agnostic, the latent space can be interpreted in its own right. The whole space is used and decodes to valid fields, it preserves the ordering of distances between physical fields, its geodesics interpolate physical evolutions, and edit directions transfer across fields.
With frozen representation maps, a dataset is encoded once and a neural operator is trained entirely on cached embeddings, which saves memory and training time. On 28 tasks in fluid dynamics, solid mechanics, acoustics, and the Helmholtz and Poisson equations, this paradigm can exceed the accuracy of state-of-the-art end-to-end architectures, with up to an order of magnitude higher throughput and memory savings that fit models nearly two orders of magnitude larger on consumer-grade GPUs. At inference, latent rollouts run up to an order of magnitude faster than regular rollouts, and chunked inference processes functions of over half a billion points on a single consumer-grade GPU.
\end{abstract}

\etocdepthtag.toc{mtbody}
\section{Introduction}
\label{sec:intro}

\begin{figure}[t]
  \centering
  \includegraphics[width=\linewidth]{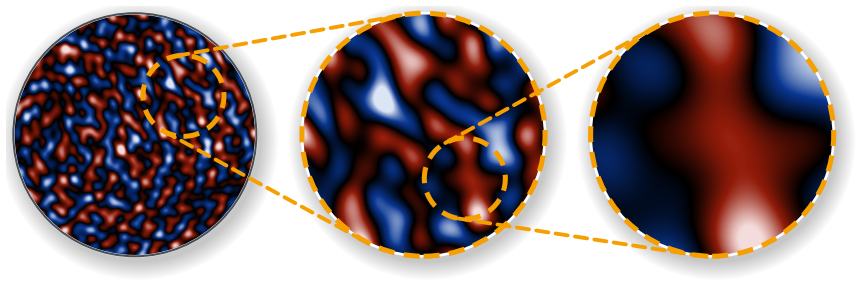}
  \caption{\textbf{Decomposition and rescaling carry a field into a familiar frequency band.} A high-frequency field and partitions of it magnified three and nine times. Their effective spatial frequencies (Appendix~\ref{app:freq}) are $7.83$, $2.53$, and $0.85$ from left to right.}
  \label{fig:hero}
\end{figure}

A wide range of phenomena in science and engineering is governed by partial differential equations (PDEs). Analytical solutions are not available in most practical settings, so such problems are solved numerically. In many applications, such as design optimization, uncertainty quantification, control, and inverse problems, the solver must be run repeatedly under varying inputs. The resulting computational cost is frequently prohibitive. Machine-learned surrogates have therefore become an established alternative, and the dominant paradigm is to learn the solution operator of a PDE-governed system directly from data, using neural operators \citep{lu2021deeponet,li2021fourier,kovachki2023neural}.

The accuracy and efficiency of such a surrogate depend heavily on how a function is represented, and how the neural operator is parameterized. A function is naturally represented in the \emph{physical space} in a discretized format by its values at coordinates on a point cloud, a mesh, or a grid. Very often, proposed parameterizations of neural operators \emph{reduce} this representation to a finite-dimensional \emph{reduced space} via an \emph{encoder} module, approximate the target functions fully or partially in this space via a \emph{processor} module, and finally \emph{reconstruct} the approximated target function back to the physical space via a \emph{decoder} module. An element of the reduced space is an \emph{embedding}, and a processor that maps embeddings to embeddings realizes a \emph{reduced-space operator}. The methods reviewed below differ chiefly in whether the reduced space is fixed in advance or learned, and in whether the reduction or reconstruction maps (jointly the \emph{representation maps}) are linear.

The earliest frameworks inherited the machinery of classical model reduction, in which a basis is extracted from a collection of precomputed solution snapshots, and the coefficients of a new solution are then obtained by solving the PDE restricted to the span of that basis rather than by learning \citep{quarteroni2015reduced}. PCA-Net \citep{bhattacharya2021model} keeps the snapshot basis, taking the leading eigenfunctions of the empirical covariance operator, but replaces the restricted solve with a regression between input and output coefficients. DeepONet \citep{chen1995universal,lu2021deeponet} instead learns the basis itself through a trunk network shared across all inputs, so that every input-dependent quantity resides in the branch coefficients. A parallel line fixed the basis analytically, expanding functions on truncated Fourier modes \citep{li2021fourier}, wavelets localized in space and frequency \citep{tripura2023wavelet}, multiwavelets built from orthogonal polynomials \citep{gupta2021multiwavelet}, pole and residue pairs in the Laplace domain \citep{cao2024laplace}, or Chebyshev and Fourier series \citep{fanaskov2023spectral}. In all of these the representation maps are linear, an explicit basis exists, and the reduced space is a \emph{coefficient space} whose embeddings are interpretable as coefficients in that basis.

Linearity, however, imposes a hard limit on the learning capacity, and several results in the literature delineate it. \citet{peherstorfer2022breaking} reviews how a linear reconstruction on $n$ modes is bounded by the Kolmogorov $n$-width of the solution manifold, and how that width decays slowly for transport-dominated problems with discontinuous data. \citet{lanthaler2023nonlinear} establish a corresponding lower bound for any architecture whose output is confined to a fixed linear span, and \citet{seidman2023variational} observe the same effect empirically on families of translating densities. These and other works \citep{bhattacharya2021model,bunker2025autoencoders} point toward nonlinear representation maps on function spaces, such as deep autoencoders.
The field subsequently moved in that direction. When the representation maps are nonlinear no explicit basis exists, functions are represented as learned latent embeddings, and the reduced space becomes a \emph{latent space}. Certain architectures reduce a field onto a fixed number of learned latent tokens \citep{wu2023solving,wang2024latent,alkin2024universal}, others onto a coarser set of nodes \citep{mousavi2025rigno,wen2025geometry,li2023geometryinformed}, and others reconstruct through neural fields that can be queried at arbitrary coordinates \citep{yin2023continuous,serrano2023operator,serrano2024aroma}. A further group is nonlinear in the same sense but never gives up the indexing of the physical domain, acting either across a hierarchy of resolutions \citep{raonic2024convolutional} or projecting back onto it across layers \citep{wu2024transolver}. In these the reduced representation is transient, rebuilt and discarded layer by layer rather than persisting across the network. These architectures can escape the $n$-width bound, at the cost of the interpretability that a fixed basis provided. In each case, however, the representation maps are fitted to the data of a single task, either jointly with the reduced-space operator against labeled input-output pairs, or by reconstruction on that same dataset beforehand \citep{serrano2023operator,serrano2024aroma}. The reduced space is therefore a by-product of one supervised problem, and the labels pay both for the physics and for learning what functions look like. Little of the representation survives a change of equation, geometry, discretization, or modality, so every new task rebuilds it.

Until a few years ago, representations in natural language processing and in computer vision were coupled to their supervised tasks in much the same way, each model learning its own features for its own objective. Recently, self-supervised pre-training broke that coupling \citep{devlin2019bert,chen2020simple,caron2021emerging,he2022masked}, and enabled representations to be trained at scale. A single model is pre-trained on a pretext objective that requires no labels and is then adapted to each task through a lightweight head, while the backbone remains frozen. What transfers is not the pretext task but the representation the model must construct in order to solve it, and such representations outperform supervised pre-training \citep{he2022masked} and improve with scale \citep{kaplan2020scaling}.

Two analogies carry this recipe over to operator learning. The first concerns the tasks. Semantic segmentation, monocular depth estimation, and image-to-image translation all map an image (a field defined on a pixel grid) to another one, and a PDE-governed problem is of the same type, mapping a given state and conditions to a resulting state. An operator learning task may thus be read as image-to-image translation whose label, the target function, is determined by physics rather than by human perception. The second concerns the data. Self-supervision pays off in vision because labels are expensive while unlabeled inputs are cheap, so a model can learn its representation from far more data than could ever be labeled. In operator learning the labels are expensive as well, since every target requires a numerical simulation, but the unlabeled inputs are not cheap either, because physically measured or simulated fields are themselves scarce. A pretext objective defined on such fields would therefore be limited by the same scarcity it is meant to avoid. Admitting \emph{synthetic} functions as pretext inputs removes this limit. An engineered distribution of analytic functions can be sampled without bound, so the supply of unlabeled inputs becomes unlimited, a stronger position than in vision, where it is bounded by the images that exist.

In this work, we propose a paradigm shift of the same kind for operator learning, by decoupling representation learning from operator learning. We pre-train a task-agnostic functional autoencoder (FAE) on unlabeled synthetic functions that belong to no equation, and the reduced space it induces is then frozen and reused. Because the pretext distribution is engineered rather than sampled from one physics, a single such space serves functions of different origin, on different geometries, at different discretizations, and across modalities, which is what we mean by calling it a \emph{universal latent space}, and we call the framework built on it \daedalus{}\footnote{Project website: \url{https://daedalus.ethz.ch}.}. This places the construction between the two lines above. Its representation maps are nonlinear, yet unlike the latent spaces of existing neural operators, it is the product of no particular task, and the labels of a downstream problem pay only for the physics. An operator is then learned entirely inside the frozen space, on embeddings that can be computed once and cached.
Two ingredients make a representation of this kind transferable. The first is a decomposition of the domain, so that an unfamiliar global field is represented as a collection of local ones (Figure~\ref{fig:hero}). The second is a normalization of each partition into a common frame, which removes its position, size, orientation, and amplitude from the learned embedding. Trained on analytic functions drawn on the fly, with no labels and no solver, a single backbone then reconstructs PDE fields of several kinds as well as natural photographs, on geometries and distributions never seen during training (Appendix~\ref{app:recon}).

\paragraph{Related work.}
Label-free pre-training for PDE surrogates has been proposed and explored by several works, with consistent reported benefits. Masked or corrupted reconstruction improves accuracy and sample efficiency over training from scratch \citep{xu2024self,chen2024data,rahman2024pretraining}, joint embeddings built from Lie point symmetries yield a frozen encoder serving several tasks for one equation \citep{mialon2023self}, and a recent strand drops the solver from the pretext objective altogether \citep{zhang2026cheap,wu2026geopt}. Functional autoencoders formalize the construction \citep{seidman2023variational,bunker2025autoencoders}, frozen neural-field autoencoders carry a processor on their embeddings and train faster and more stably than end-to-end \citep{serrano2023operator,serrano2024aroma}, and the most recent work casts a pre-trained autoencoder as a foundation model, adapted to downstream tasks by parameter-efficient fine-tuning \citep{liu2026tadpole}. In all of them, however, the pre-training remains tied to the task, being fitted on the fields of one equation or dataset and run as a stage of that task's own training pipeline, so every new problem pays for a representation of its own. \daedalus{} instead pre-trains a single representation once, on synthetic functions that belong to no task, and reuses it unchanged across equations, geometries, discretizations, and modalities. The work most closely related to ours is Phaedra \citep{lingsch2026phaedra}. In this work, the authors also separate the pre-training and downstream objectives; however, the focus was to learn discrete representations for cross-entropy based models and lacks a deep investigation into the relationships between latent representations and their continuous physical counterparts.

\begin{figure*}[t]
  \centering
  \definecolor{journey}{HTML}{6741D9}%
  \resizebox{\linewidth}{!}{%
  \begin{tikzpicture}[
      panel/.style={draw=black!10, fill=white, line width=0.4pt, rounded corners=4pt,
                    blur shadow={shadow blur steps=10, shadow xshift=0.6pt, shadow yshift=-0.9pt, shadow opacity=32}},
      badge/.style={anchor=north west, circle, fill=journey, text=white, font=\scriptsize\bfseries, inner sep=0pt, minimum size=11pt},
      lab/.style={anchor=west, font=\footnotesize, text=black!85, inner sep=2.5pt},
      journeystep/.style={draw=journey, line width=1.1pt, -{Stealth[length=3.2pt, width=4pt]}},
    ]
    \def\wA{3.9}\def\wC{5.1}\def\hTop{2.9}\def\hBot{3.55}\def\wD{6.0}\def\gap{0.42}
    \pgfmathsetmacro{\wB}{\wA+\gap+\wC} %
    \pgfmathsetmacro{\hL}{\hTop+\gap+\hBot} %
    \pgfmathsetmacro{\xC}{\wA+\gap} %
    \pgfmathsetmacro{\xD}{\wB+\gap} %
    \pgfmathsetmacro{\yB}{-\hTop-\gap} %
    \draw[panel] (0,0) rectangle (\wA,-\hTop);
    \draw[panel] (\xC,0) rectangle (\xC+\wC,-\hTop);
    \draw[panel] (0,\yB) rectangle (\wB,\yB-\hBot);
    \draw[panel] (\xD,0) rectangle (\xD+\wD,-\hL);
    \node at ({\wA/2},{-\hTop/2}) {\includegraphics[width=2.8cm]{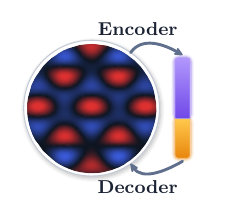}};
    \node at ({\xC+\wC/2},{-\hTop/2}) {\includegraphics[width=4.5cm]{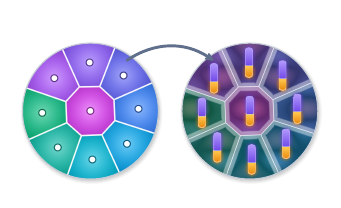}};
    \node at ({\wB/2},{\yB-\hBot/2}) {\includegraphics[width=8.4cm]{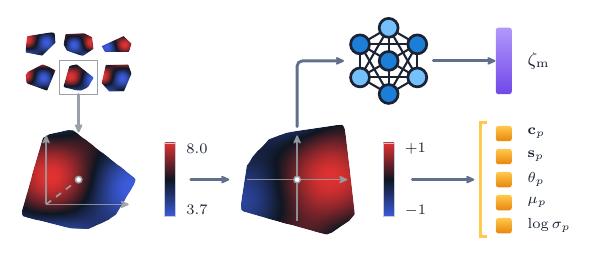}};
    \node at ({\xD+\wD/2},{-\hL/2}) {\includegraphics[width=5.7cm]{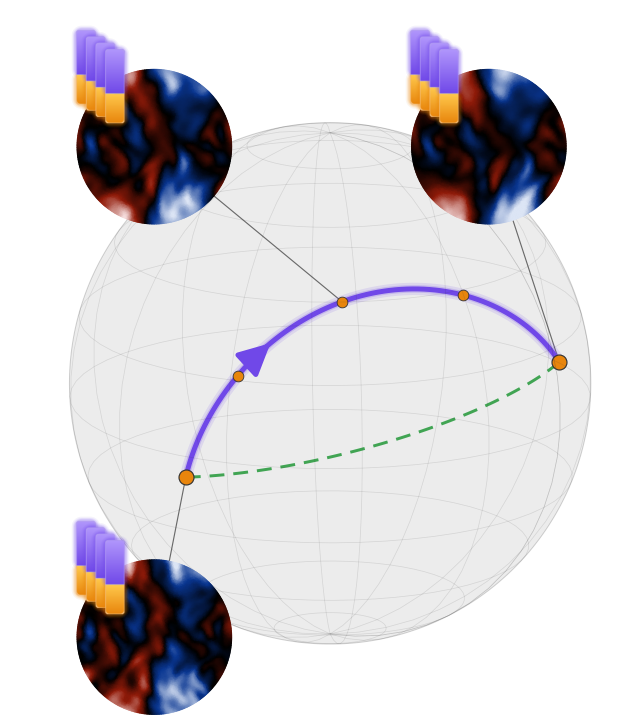}};
    \node[badge] (ba) at (0.12,-0.12) {a}; \node[lab] at (ba.east) {Single-partition FAE};
    \node[badge] (bb) at (\xC+0.12,-0.12) {b}; \node[lab] at (bb.east) {Multi-partition encoding};
    \node[badge] (bc) at (0.12,\yB-0.12) {c}; \node[lab] at (bc.east) {PartNorm};
    \node[badge] (bd) at (\xD+0.12,-0.12) {d}; \node[lab] at (bd.east) {Latent neural operator};
    \draw[journeystep] ({\wA+0.04},{-\hTop/2}) -- ({\xC-0.04},{-\hTop/2});
    \draw[journeystep] ({\xC+\wC/2},{-\hTop-0.04}) -- ({\xC+\wC/2},{\yB+0.04});
    \draw[journeystep] ({\wB+0.04},{\yB-\hBot/2}) -- ({\xD-0.04},{\yB-\hBot/2});
    \definecolor{evolution}{HTML}{7048E8}\definecolor{slerp}{HTML}{2F9E44}%
    \pgfmathsetmacro{\lx}{\xD+\wD-3.15}\pgfmathsetmacro{\ly}{-\hL+0.52}
    \draw[evolution, line width=1.3pt, line cap=round] (\lx,\ly) -- ++(0.5,0);
    \node[anchor=west, font=\scriptsize, text=black!80, inner sep=2pt] at ({\lx+0.55},\ly) {Physical evolution};
    \draw[slerp, line width=1.3pt, dash pattern=on 3pt off 2pt] (\lx,{\ly-0.3}) -- ++(0.5,0);
    \node[anchor=west, font=\scriptsize, text=black!80, inner sep=2pt] at ({\lx+0.55},{\ly-0.3}) {Great circle (SLERP)};
  \end{tikzpicture}}
  \caption{\textbf{Overview of \daedalus{}.} \textbf{(a)} A single-partition FAE maps a function to one embedding and back. \textbf{(b)} A function is decomposed into partitions and represented by one embedding per partition, and these embeddings together form the latent representation on which the processor operates. \textbf{(c)} Before encoding, PartNorm maps each partition into a common local frame. Its center $\mathbf{c}_p$, extent $\mathbf{s}_p$, orientation $\theta_p$, value shift $\mu_p$, and value scale $\sigma_p$ are stored as static features, so the learned embedding carries only the morphology (Appendix~\ref{app:partnorm-frame}). \textbf{(d)} A neural operator acts as motion on the latent hypersphere. Decoding applies these steps in reverse, replaying the stored frame of each partition (Appendix~\ref{app:partnorm-frame}). Panels are illustrative, and two features of \textbf{(d)} are drawing conventions rather than properties of the space. The sphere is drawn as an ordinary two-sphere, whereas the morphologic embeddings live on a hypersphere of far higher dimension, and each marked point stands for the whole stack of $P$ partition embeddings of one snapshot, so an evolution traces $P$ coupled curves rather than the single one shown.}
  \label{fig:teaser}
\end{figure*}

\paragraph{A universal latent space.}
We pre-train a task-agnostic reduced space for functions on arbitrary point clouds, and use a single frozen backbone across several physics, geometries, discretizations, and modalities. The encoder and the decoder that induce it are new architectures, built for this purpose rather than adapted from an operator-learning backbone, accepting clouds of any size and any number of input and output channels, and holding every learned embedding on a hypersphere, where an orthogonalized on-sphere update refines it.

\paragraph{Pre-training without solver-generated data.}
The pretext functions are analytic, synthesized at every step and never reused, so that pre-training is never bounded by data and the observed generalization is free of memorization effects.

\paragraph{Partition normalization (PartNorm).}
We introduce a normalization for functional data that acts on top of the standard input and output normalization. Every partition is carried into a common frame before it is encoded, which imposes equivariance to translation, dilation, amplitude, and optionally rotation exactly and by construction on an unstructured cloud, rather than leaving it to be learned and approximated. It also removes those four degrees of freedom from the latent space, so that none of its capacity is spent re-encoding where a feature sits, how large it is, how it is oriented, or what its magnitude is, and the embedding is left to describe only the morphology of the partition.

\paragraph{Domain decomposition and effective frequency.}
We represent a field as a collection of partitions, and identify a single scalar of a normalized partition, its effective spatial frequency (Appendix~\ref{app:freq}), that governs how accurately the partition can be represented. We show that the accuracy attainable on functional data is dictated by this quantity rather than by the particular form of the functions the model was trained on, which recasts out-of-distribution transfer as decomposition into familiar local problems rather than extrapolation, and makes the number of partitions an inference-time accuracy control.

\paragraph{An interpretable latent space.}
Because the space is fixed and independent of any operator learning task, its structure can be studied in its own right. We propose several ways of interpreting it, namely its atoms and great circles, the correspondence between latent and physical distances, transferable directions, geodesic morphing, and family embeddings (Appendix~\ref{app:latent}), and show through them that it is a suitable space for operator learning. Its distances preserve the order of distances between the decoded fields, it composes locally, edit directions transfer between unrelated fields, and geodesics track local physical evolution.

\paragraph{Latent operator learning with input and output caching.}
Since the representation maps are frozen, both sides of an operator dataset can be encoded once and cached, so that the operator of each task is trained wholly inside the reduced space, at a cost that no longer depends on the resolution of the data. The bulk of this training runs on cached embeddings with a \emph{latent loss}, evaluated between embeddings. The training ends with a brief \emph{cooldown} phase on a \emph{physical loss}, evaluated on the decoded fields, which closes the gap left by the latent proxy at close to the cost of latent training. A single backbone, pre-trained once on synthetic functions, serves every downstream task of this work as it is, with no further training and therefore at no extra cost per task. On a benchmark of 28 problems, the resulting operators are more accurate than those of the state-of-the-art end-to-end framework \citep{wen2025geometry} on 23 of them, and train on average 2.1 times faster per problem. On functions of about a million points, latent training runs up to 14 times faster than the training of the end-to-end baseline. The memory it saves goes further. On a single 24\,GB consumer-grade GPU, latent training fits a processor of 101 million parameters on functions of 262 thousand points at a batch of 32, where the end-to-end baseline fits a model of at most about one million parameters. Latent training also runs at a batch of 32 on functions of 4.2 million points, 64 times as large as the end-to-end baseline reaches on variable point clouds.

\paragraph{Fast inference on very large problems.}
The frozen space also pays off at inference. On the same GPU, a 1.6B processor model maps functions of 262 thousand points in about the time the end-to-end baseline needs with five times fewer parameters. A \emph{latent rollout} encodes the initial state once, advances it in the reduced space, and decodes only the final state, which makes a long rollout up to 10 times faster than one that decodes and re-encodes at every step. Since the encoder and the decoder act on every partition independently of the others, \emph{chunked inference} encodes and decodes the points of a function one chunk at a time while the processor still acts on all partitions at once, and maps a function of 537 million points in a single inference, 16 times as large as the GPU holds without chunks. All of these measurements are on variable point clouds and include the cost of partitioning.

\section{Construction}
\label{sec:construction}

Our task-agnostic latent space rests on three pieces. An FAE reduces and reconstructs any discretized scalar function on a single partition, and is trained on nothing but synthetic functions drawn from predefined analytic distributions. Domain decomposition reuses the FAE on a field of arbitrary size, reducing every scalar channel on every one of its partitions. A processor then learns the task, on an operator dataset and entirely inside the frozen space. The rest of this section explains each step in more detail.

\paragraph{The reconstruction task.}
The object the FAE is trained on is one discretized scalar function, a pair $(\mathbf{x},\funcMain)$ of coordinates $\mathbf{x}\in\R^{N\times D}$ in $D$ spatial dimensions and values $\funcMain\in\R^{N}$, one scalar per point, in which the number of points and their locations are arbitrary and may differ from one function to the next. An encoder $\encoder$ performs the reduction, carrying such a pair into a single embedding $\emb$, and a decoder $\decoder$ performs the reconstruction, returning the value $\decoder(\emb,\mathbf{y})$ of the represented function at any query coordinate $\mathbf{y}$ of the domain. The training objective is the mismatch between the reconstructed and the observed values, so the task boils down to a near-lossless compression of any discretized scalar function into a single embedding vector (Figure~\ref{fig:teaser}a). By targeting scalar-valued functions we avoid tying the FAE to a fixed number of channels, since a vector-valued field is reduced one component at a time. The representation is therefore blind to the physical meaning of what it encodes, a property the rest of this paper exploits.
With this formulation, the input and the output coordinates are fully detached. What the decoder produces is a pure neural field attached to an embedding rather than a value list tied to a discretization, so it can be queried anywhere. The decoded field therefore has the advantages of an implicit neural representation \citep{essakine2025where}, namely evaluation at any coordinate and resolution, a size set by its embeddings rather than by its discretization, and exact spatial derivatives (Appendix~\ref{app:derivatives}). Implicit neural representations are commonly fitted to each signal anew \citep{mildenhall2020nerf}, or obtain the code of a new signal by optimizing it against a shared decoder \citep{park2019deepsdf,yin2023continuous}, whereas here a new field is represented by one pass of the encoder and the decoder is never retrained. The coordinates at which a field is observed therefore need not coincide with those at which it is reconstructed, and the two clouds share neither their size nor their placement.
The pretext objective is the reconstruction of the uncorrupted function, which departs from the masked and corrupted objectives that dominate self-supervised pre-training in vision and in the PDE surrogates that have adopted it \citep{xu2024self,chen2024data,rahman2024pretraining}. Corruption is effective on natural images because what has been removed can only be filled in by learning the statistics of natural images, so the objective deliberately installs a bias toward a data distribution, and that bias is what transfers. Such a bias is unfavorable for a task-agnostic functional autoencoder. An arbitrary function carries no semantics to be inferred, what the downstream operator needs is a faithful and invertible compression, and a preference for any particular family of fields is therefore a liability rather than an asset. Joint-embedding objectives are unsuitable for a second reason as well \citep{mialon2023self}. They are built to discard nuisance factors between two views of one object, and for a generic function, with no governing equation to supply further symmetries, the available views amount to shifts, dilations, and rotations, which the PartNorm presented below removes exactly, so such an objective would be asked to learn by approximation an invariance the representation already holds by construction. It would also make no promise that an embedding can be turned back into a field, which is precisely what operator learning requires.

\paragraph{From one embedding to many.}
A single embedding per function is a bottleneck. Its size is fixed at $\embdim$ numbers, the same for a smooth field as for one with fine structure everywhere, so beyond some level of detail a whole field no longer fits into it. We therefore decompose the domain (Figure~\ref{fig:teaser}b). A \emph{partitioner} splits the $N$ points into $P$ \emph{partitions}. The same encoder is applied to each partition independently and returns one embedding per partition, and a query coordinate is answered from the embeddings of the partitions that cover it. The reduced representation of a field is accordingly a collection of $P$ embeddings, one for each partition.
The construction mirrors domain decomposition methods for PDEs \citep{quarteroni1999domain,dolean2015introduction}, which split a large problem on a complex domain into subproblems on simpler subdomains, each small enough to be handled by one local solver, and recover the global solution by coupling the subdomains along their interfaces. Here the local solver is the encoder and the decoder, applied to every partition alike, and overlapping partitions play the role of the overlap between subdomains in Schwarz-type methods, through which neighboring pieces agree. The coupling between partitions that a domain decomposition solver iterates over is left to the processor.
The move from one global embedding to many spatially anchored local ones is likewise a transition computer vision has already made, and for the reason we also observe, namely that a single global embedding oversmooths and does not scale to complex signals \citep{peng2020convolutional,chen2021learning,chabra2020deep,bauer2023spatial}. Local representations also make neural fields far cheaper to train and to evaluate \citep{reiser2021kilonerf,mueller2022instant}.
More partitions mean smaller neighborhoods and a finer latent resolution, so $P$ governs the fidelity of the representation and is a task-specific choice (Appendix~\ref{app:sweep}). Fewer partitions do not degrade a field at random but keep its low-frequency content and drop the finer detail. Our design allows a fast post-hoc tuning of $P$ that requires no training and can be done very fast by attending to the reconstruction accuracy at several candidate values.

\begin{figure}[t]\centering
  \includegraphics[width=\linewidth]{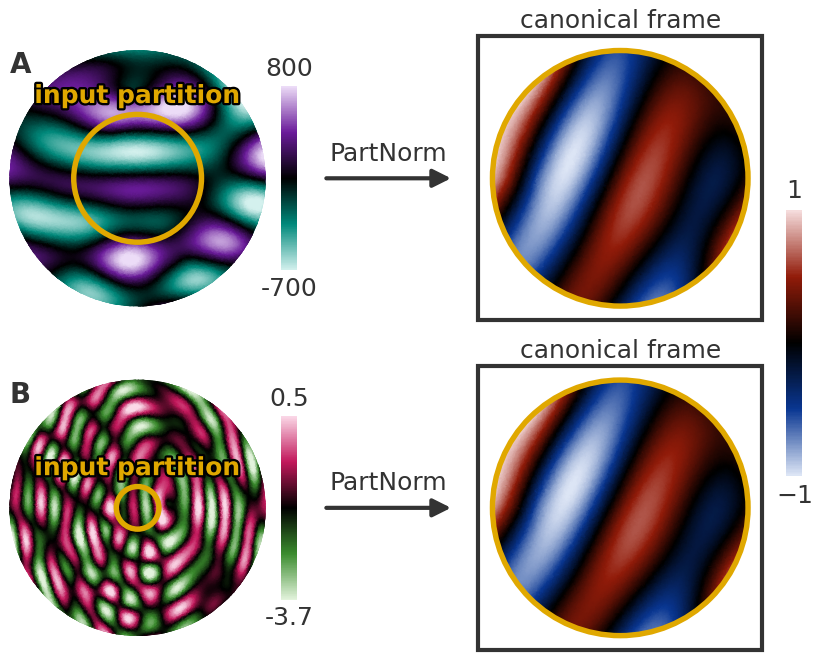}
  \caption{\textbf{PartNorm carries different inputs into one frame.} Input A takes values in the hundreds, while input B oscillates three times faster, is rotated, and takes values of order one, as their own color ranges show, and its partition is three times smaller. Each input has a colormap of its own, since the two can be entirely different functions. PartNorm carries both partitions into the canonical frame, with the optional rotation, where their points fill the same disk and their values the range $[-1,1]$, and the two become the same normalized morphology. The removed position, size, orientation, and range are kept as static features. Appendix~\ref{app:partnorm-cases} shows further cases together with the embeddings.}
  \label{fig:partnorm-main}
\end{figure}

\paragraph{PartNorm.}
What makes one encoder usable on partitions of every position, size, and amplitude is a normalization applied before the encoding (Figure~\ref{fig:teaser}c). Each partition is carried into a canonical local frame. The coordinates of its points are shifted so that the partition center lies at the origin and scaled per axis by the extent of the partition so that the points fill a symmetric box, while the values are min-max normalized within the partition. Optionally, each partition is also rotated into a canonical orientation read from the moments of its values, which makes the representation exactly equivariant to rotation, a strategy successfully applied to gridded data \citep{wei2026revit}. The quantities defining this frame are not discarded but retained as the \emph{static} part $\embstat$ of the embedding, which the decoder replays to invert the frame exactly. The frame therefore travels losslessly alongside the learned part, and the \emph{morphologic} part $\embmorph$ is left to describe only the normalized shape of the partition, so that an embedding is the pair $\emb=(\embstat,\embmorph)$.
Two properties follow, and together they are the reason a model trained on one narrow synthetic distribution reconstructs fields it was never shown. First, the representation is exactly equivariant to translation, dilation, amplitude, and optionally rotation, by construction rather than by approximation, on an unstructured cloud where no architectural device offers the same guarantee. Second, since these quantities are kept in the static part, the morphologic embedding spends none of its capacity on where a feature sits, how large it is, how it is oriented, or what its magnitude is.
Normalization carries structurally similar partitions to nearly identical inputs whatever their global context (Figure~\ref{fig:partnorm-main}), so their embeddings coincide (Appendix~\ref{app:partnorm-cases}).
A high-frequency field observed in a small partition normalizes to almost the same input as a low-frequency field in a large one and receives almost the same embeddings, since within a partition absolute scale is immaterial and only the number of oscillations across the partition, its \emph{effective spatial frequency} (ESF), survives (Appendix~\ref{app:freq}). The model therefore need not have seen a field of a given physical type, only partitions of comparable normalized morphology, which the synthetic distribution supplies in abundance. Difficulty becomes a property of normalized morphology rather than of function class, which is what the ESF measures, and an unfamiliar global field is decomposed into familiar local ones rather than extrapolated to.
The factorization of a partition into a normalized morphology and a set of frame parameters has a direct predecessor in the shape-gain decomposition of Phaedra \citep{lingsch2026phaedra}, which separates the shape of a feature from its magnitude on a regular grid. PartNorm carries that idea from a scalar gain on a grid to a full local frame, comprising position, per-axis scale, orientation, and value range, attached to each partition of an arbitrary cloud. Appendix~\ref{app:partnorm-frame} gives the construction, and Figure~\ref{fig:teaser}c an illustration.

\paragraph{Synthetic functions.}
The FAE is pre-trained without access to any PDE solutions. Every batch is synthesized on the fly on the GPU from predefined analytic distributions, multi-mode anisotropic sinusoidal functions with random frequencies, with no dataset on disk and no solver in the loop. Three arguments recommend this over a fixed corpus. First, generation costs no storage and no input pipeline, so that pre-training is bounded by compute rather than by data and may be run for as long as it keeps improving. Second, no function is ever presented twice, so the model fits a distribution rather than a finite sample, and memorization of solver output cannot explain the transfer we report, since no solver output is ever seen. Third, the distribution is entirely under our control, which turns in-distribution and out-of-distribution status into a statement about effective spatial frequency rather than about physical type.

\paragraph{Architecture.}
The encoder and the decoder are new architectures, designed around the requirements a universal latent space imposes rather than adapted from an existing operator-learning backbone. Four features carry that design. Both maps are built around a single partition, the encoder taking many points to one embedding and the decoder one embedding to many points, and the encoder aggregates over the points of a partition by a mean, which leaves it indifferent to their number and to their order and is what lets one trained model accept clouds of any size, while the decoder answers every query coordinate independently of every other. A root-mean-square normalization (RMSNorm) \citep{zhang2019root} applied after every layer returns the morphologic embedding to a fixed magnitude, which is the mechanism that confines it to the hypersphere and makes it a direction vector rather than an arbitrary point. The embedding is then refined rather than replaced, the update being orthogonalized against the current embedding and applied under a learned scale, which discards the component that would merely rescale the embedding and be undone by the normalization, removing at once a redundancy and a source of training instability. The construction is close to the normalized transformer of \citet{loshchilov2025ngpt}, which likewise re-normalizes the hidden state onto the unit hypersphere after every layer, there with a learned per-dimension step size on the update, but does not orthogonalize it.
The decoder, finally, is much shallower than the encoder, since answering one query coordinate from an embedding is a lighter problem than compressing a partition into one (Appendix~\ref{app:architecture}).

\paragraph{Latent operator learning.}
Once pre-trained, the backbone is frozen, and every downstream task reuses it as it is. An operator is learned on top of the collection of embeddings, entirely inside the reduced space (Figure~\ref{fig:teaser}d). A processor $\processor$ is placed between $\encoder$ and $\decoder$ and maps the embeddings of the input field to the embeddings of the output field. What reaches it is a small collection of embeddings rather than a point cloud, so the discretization of the field, its size and its sampling alike, is left entirely outside the operator.
Learning a map between reduced representations, rather than between fields, is how several of the earliest data-driven surrogates were built. When the representation maps are linear the reduced space is a coefficient space that exists before any training, so the map to be learned is a map between coefficient vectors. This is what the projected solve of a reduced-basis method computes \citep{quarteroni2015reduced}, what non-intrusive reduced-order models regress with a neural network in place of that solve \citep{hesthaven2018non}, what PCA-Net learns between input and output coefficients \citep{bhattacharya2021model}, and what the proper orthogonal decomposition (POD) variant of DeepONet learns from the input samples to the output coefficients \citep{lu2022comprehensive}. The practice was lost in the move to nonlinear representation maps, not because it ceased to be desirable but because those maps are fitted jointly with the operator against labels in the physical space, which puts the reduction and the reconstruction back inside every training step. \citet{alkin2024universal} recover part of it at inference, advancing an evolution problem by repeated application of the processor without returning to the physical space between steps, a \emph{latent rollout} in their terminology, which buys a large speed-up at inference. The loss, however, is still evaluated in the physical space, so the reduction and the reconstruction remain inside training.

\paragraph{Input and output caching.}
Freezing the representation maps removes the reduction and the reconstruction from training as well, and what it buys is caching. Because the encoder never changes, the embeddings of the training inputs can be computed once and stored, and because the decoder never changes either, a loss evaluated in the reduced space allows the targets to be cached in the same way. Training then reads embeddings from a table and touches neither the physical space nor the two dense maps that reach it (Appendix~\ref{app:caching}). Three gains follow. The reduction and the reconstruction, which dominate both the memory and the time of an end-to-end step, are removed from every step rather than merely amortized. The dataset is read and encoded once, so the input pipeline leaves the training loop altogether. Finally, the whole optimization budget is spent on the operator rather than partly on relearning what functions look like.
Two conditions make this possible. The first is architectural, namely that the physical space be crossed exactly twice. Architectures that never leave the physical space, or that return to it at every layer, admit no such treatment \citep{pfaff2021learning,raonic2024convolutional,wu2024transolver}.
The second condition is on the latent space itself, and is easy to overlook. A loss evaluated on embeddings is a proxy for the error in the physical space, and it is a faithful proxy only if the two agree on what is close, that is, only if embeddings near one another in the reduced space decode to functions near one another in the physical space. A functional autoencoder may compress and reconstruct with arbitrarily high accuracy and still fail this requirement, in which case a processor trained to match embeddings optimizes something only loosely related to the field it is meant to predict. The property is therefore measured directly for the space constructed here (Appendix~\ref{app:latent-metric}).

\paragraph{Processor architecture.}
The architecture of the processor is a free choice. The embeddings are anchored at the partition centers and therefore form a point cloud of their own, so any architecture that maps such a set to a set of the same kind is admissible. Because the backbone is frozen and the embeddings are cached, the processor can be replaced without retraining the FAE. We use a transformer over the partition embeddings, with the same on-sphere orthogonalized update as the encoder and rotary position embeddings \citep{su2024roformer} extended to two dimensions (Appendix~\ref{app:architecture}).

\paragraph{Training with a cooldown.}
Caching makes the latent loss a cheap objective for training the processor, but the latent loss only remains a proxy for the error of the decoded field. We therefore train on it for most of the epochs, on the frozen and cached backbone, and switch to the physical loss for a short final \emph{cooldown} phase, with the backbone still frozen. The cooldown closes the gap that the proxy leaves, since it weighs the static and the morphologic part of a prediction by their actual effect on the decoded field, while the total cost stays close to that of latent training (Section~\ref{sec:operator}). As an option, the cooldown can also unfreeze the encoder and the decoder and train them with the processor, which gains accuracy on one task at the cost of the universality of the backbone and of the invertibility of its two maps. Appendix~\ref{app:operator-cooldown} analyzes why the cooldown works, and Appendix~\ref{app:benchmark-regimes} compares it with training on the physical loss from the start.

\section{Analysis and Interpretability}
\label{sec:analysis}

This section studies the trained FAE from two complementary angles. The first concerns its reach, namely how far the representation maps carry beyond the narrow distribution they were trained on and which property of the input determines their accuracy. The question bears directly on operator learning, since a processor placed on top of a representation cannot recover what the representation fails to capture. The second concerns the latent space itself. Because this space is fixed and belongs to no particular task, it can be examined in its own right, which is not possible for the latent spaces of existing neural operators, whose geometry is shaped by a single operator and entangled with it. Throughout, we evaluate a single trained backbone ($\embdim{=}128$, see \hyperref[tab:backbones]{appendix prelude}) with the inference partitioner of Appendix~\ref{app:partitioner} and report median relative $L^2$ errors. The operators of Section~\ref{sec:operator} run on a smaller backbone ($\embdim{=}8$) which has a lower capacity and needs more partitions for the same reconstruction accuracy.

\begin{figure*}[t]\centering
  \includegraphics[width=\textwidth]{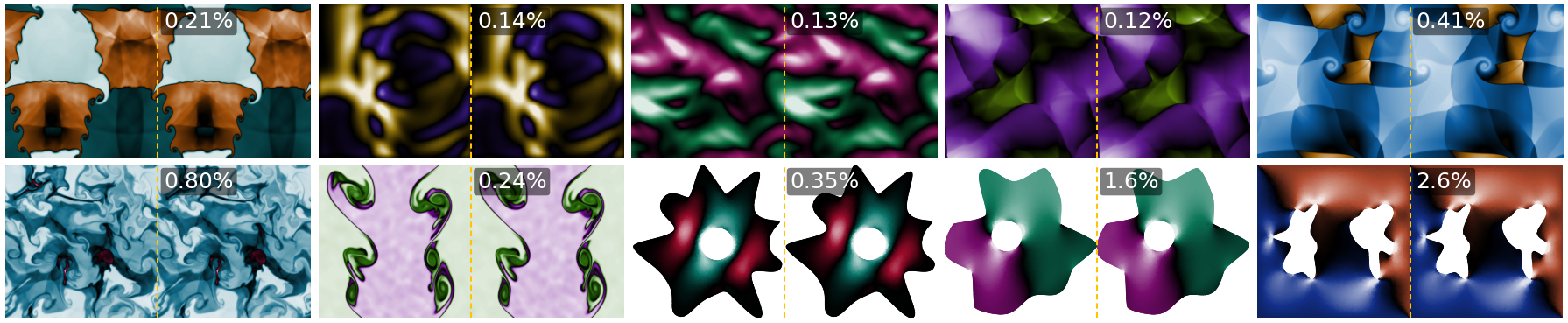}
  \caption{\textbf{Reconstruction across domains, from a synthetic-only model.} Every field is followed directly to its right by its \daedalus{} reconstruction, on the same color scale, with the relative $L^2$ error of the reconstruction in its top-left corner. First row, left to right: the compressible-Euler density $\rho$ of CEU-RiemannKH; the acoustic wave of AWA-Layer; the incompressible-flow velocity $u$ of IEU-PwC; and the density $\rho$ of CEU-Gauss and of CEU-Riemann. Second row: the density $\rho$ of CEU-RiemannCurved and of CEU-KH; the displacement $u_y$ of one sample and the Cauchy stress $\sigma_{yy}$ of another of SEQ-CloudCircle-Hyp, on a domain with a hole; and the shear stress $\sigma_{xy}$ of SEQ-SquareCloud-Lin, a linear-elastic body on a square domain with two holes. The SEQ fields are encoded with $128$ partitions and all fields with $1024$ partitions. Dataset names are those of Table~\ref{tab:datasets}. None of these inputs resembles the analytic training functions of Figure~\ref{fig:train-dist}.}
  \label{fig:gallery}
\end{figure*}

\paragraph{Generalization to unseen functions and geometries.}
\label{sec:recon}
The FAE is trained exclusively on analytic functions of a narrow frequency band, drawn on a fixed unit disk (Figure~\ref{fig:train-dist}). One would therefore expect it to reconstruct faithfully only functions that resemble this distribution. This is not the case. The same frozen model reconstructs fields of much higher frequency, fields with localized spikes, hyperelastic displacement and Cauchy-stress fields on a domain with a hole, incompressible flow, compressible flow with shocks, acoustic waves (Figure~\ref{fig:gallery}), and natural photographs (Appendix~\ref{app:recon-images}), and in every case the reconstruction recovers fine texture and not merely the coarse layout. None of these inputs resembles the training functions in its physics or its geometry, and not a single PDE solution, image, or other natural field is seen during training. The transfer therefore cannot be explained by memorization, and it is exactly this property that allows one backbone to serve the heterogeneous inputs of many operator-learning tasks.

The explanation lies in the structure imposed on the representation rather than in the training data. When a field oscillates faster than the training functions, the domain is split into more partitions, and PartNorm stretches the content of each, now smaller, partition over the canonical frame. The number of oscillations within a partition is thereby pulled back into the range the FAE has seen (Figure~\ref{fig:hero}). An arbitrary geometry is handled by the same mechanism, since the decomposition cuts a complex domain into small pieces, each of which is geometrically simple once expressed in its own frame. The partition sweeps of Appendix~\ref{app:sweep-count} confirm this empirically. On synthetic fields as well as on incompressible flow and on the stress field of a hyperelastic body, the error decreases steadily with the partition count, from several percent or more with a single partition to a fraction of a percent with 256. The partition count therefore acts as a resolution control that is set at inference, per input and per task, without any retraining. Moreover, scored at points the encoder did not see, the error keeps falling as the number of partitions grows (Appendix~\ref{app:sweep-sparsity}), which confirms that an embedding describes a field defined continuously in space rather than a particular set of samples. Such a field can also be differentiated. Since the decoder reads every query point from its own coordinate, the exact spatial gradient of any decoded field follows from one forward-mode pass of automatic differentiation per axis, and on the smooth fields we examined it is accurate to within $2\%$ from a single partition, although no derivative enters the training (Appendix~\ref{app:derivatives}). Derived fields such as strains and stresses can thus be read off a reconstruction, and physics-informed losses can be imposed on the decoded output of a latent operator.

\paragraph{Effective spatial frequency.}
\label{sec:freq}
The partition sweeps reveal a regularity. Whatever the origin of a field, additional partitions improve its reconstruction, and a field that is hard to reconstruct at a given partition count becomes easy once its partitions are small enough. This observation led us to ask whether the difficulty of a field could be measured directly, and whether such a measure would distinguish between function families at all. The global frequency content of a field is the obvious candidate, but it ignores that the encoder never sees the global field. Each partition is normalized and encoded independently, and what matters is therefore how much variation remains inside a partition once it has been normalized. We accordingly propose a metric for the effective spatial frequency (ESF) of a partition, which compares the second derivatives of the normalized function over that partition with its first derivatives and is calibrated so that it recovers the frequency of a single sine. The ESF of a field cut into partitions averages those of its partitions by their share of the variation of the field (Appendix~\ref{app:freq}). One field thus has a lower ESF when it is cut into more partitions.

\begin{figure}[t]\centering
  \includegraphics[width=\linewidth]{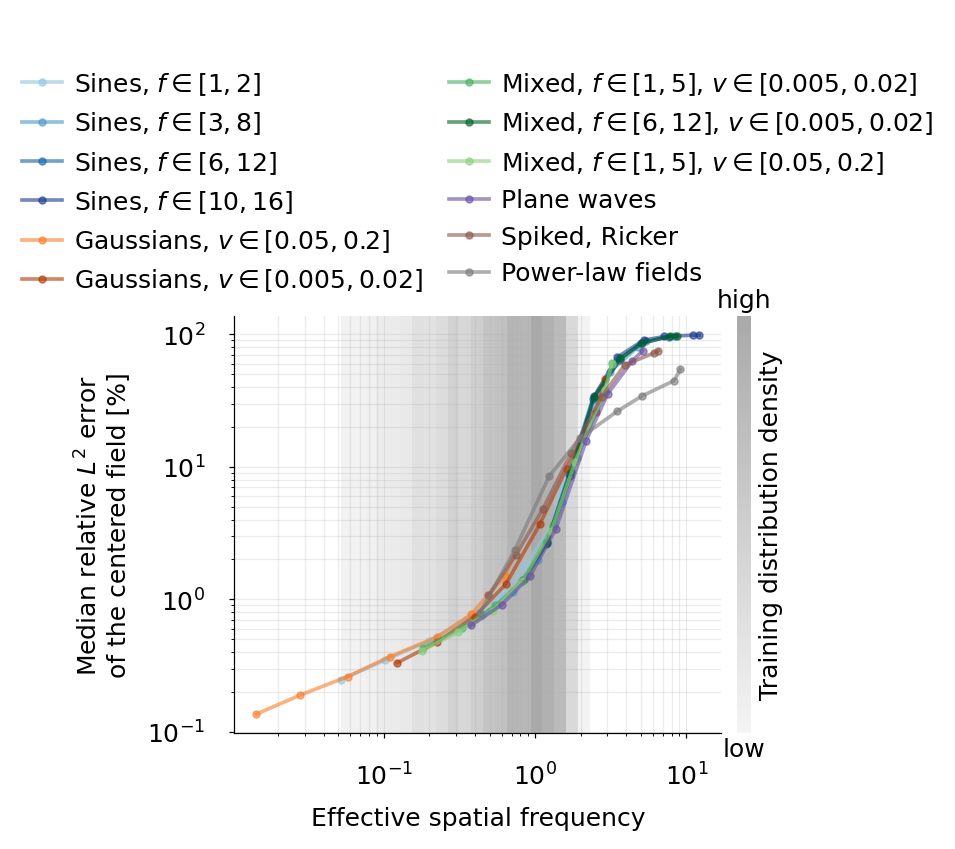}
  \caption{\textbf{Error tracks effective frequency, not function class.} Median relative $L^2$ error of the centered field, $\lVert\hat\funcMain-\funcMain\rVert_{L^2(\Omega)}/\lVert\funcMain-\bar\funcMain\rVert_{L^2(\Omega)}$ with $\bar\funcMain$ the mean of the field, against the effective spatial frequency of the partitioned field (Appendix~\ref{app:freq}). Each curve follows one fixed set of $64$ functions of a synthetic family (Figure~\ref{fig:freq-families}), decomposed into $1$ to $128$ partitions at a fixed number of $256$ points per partition on average, so that frequency is not confounded with undersampling and moving along a curve changes only the partition count. More partitions lower the effective frequency (Figure~\ref{fig:feff}). The curves of different synthetic families collapse onto one another. The gray bands give the density of the effective frequency over the training functions.}
  \label{fig:freq}
\end{figure}

Figure~\ref{fig:freq} plots the reconstruction error against the ESF for several unrelated synthetic families. The curves collapse onto a single trend. Once difficulty is measured in the frame in which the encoder operates, the family a function belongs to has little bearing on how hard it is to represent, and the ESF alone largely determines the error. The result puts the transfer described above on a quantitative footing, since an unfamiliar global field is not extrapolated to but decomposed into partitions whose ESF lies within the band seen during training. It also identifies the limitation of the approach. What bounds the representation is a frequency band fixed by the training distribution, and a partition whose ESF lies outside this band cannot be represented accurately, however the domain is partitioned. This is in line with \citet{fanaskov2023spectral}, who find that neural operators learn only the frequencies present in their training data, and our measurements locate the same effect one level lower, in the representation of the input itself.

\paragraph{Partial observations and lower-dimensional manifolds.}
A related question is what the model does with a field that is observed on only part of its domain. The same model, trained only on two-dimensional functions that fill a disk, generalizes zero-shot to functions observed on one-dimensional curves embedded in the plane (Appendix~\ref{app:recon-curves}). Along the curve, the reconstruction tracks the observed values closely. Away from it, the embedding still describes a two-dimensional function, one that matches the observation on the curve and continues it smoothly into its surroundings. Different curves through one field accordingly yield different but equally plausible continuations, and these approach the true field as the curves cover more of the domain. Two experiments on a single partition probe the same tolerance from opposite directions (Appendix~\ref{app:recon-masked}). When only the inner half of a disk is observed and the embedding is decoded on the whole disk, the field is extrapolated accurately for a short distance beyond the observed region, with an error of about $5\%$ in the first tenth of the radius, before the continuation becomes unconstrained. Conversely, when only an outer annulus is observed, the masked interior is filled in, an inpainting task, with an error of about $2\%$ for one or two radial modes that grows to about $20\%$ for five and ten, as the unobserved region becomes increasingly under-determined by the observed one. Where the observation leaves the field unconstrained, the completion resembles the training functions, a bias toward the training distribution that another training family would shift toward its own functions (Appendix~\ref{app:recon-bias}). This tolerance to partial observations follows from the design, since the encoder aggregates whatever points a partition contains and the decoder can be queried at any coordinate. It is of practical relevance for operator learning, where inputs and outputs are frequently defined on sets of different dimension, such as boundary conditions prescribed on the boundary of a domain or quantities of interest evaluated on a surface. \citet{mousavi2026imposing} show that extending boundary data into the interior of the domain with a dedicated learned module is an effective way of conditioning a neural operator on such data. A single latent space shared by functions on a domain and on its boundary provides this extension without any additional module.

\begin{figure}[t]\centering
  \includegraphics[width=\linewidth]{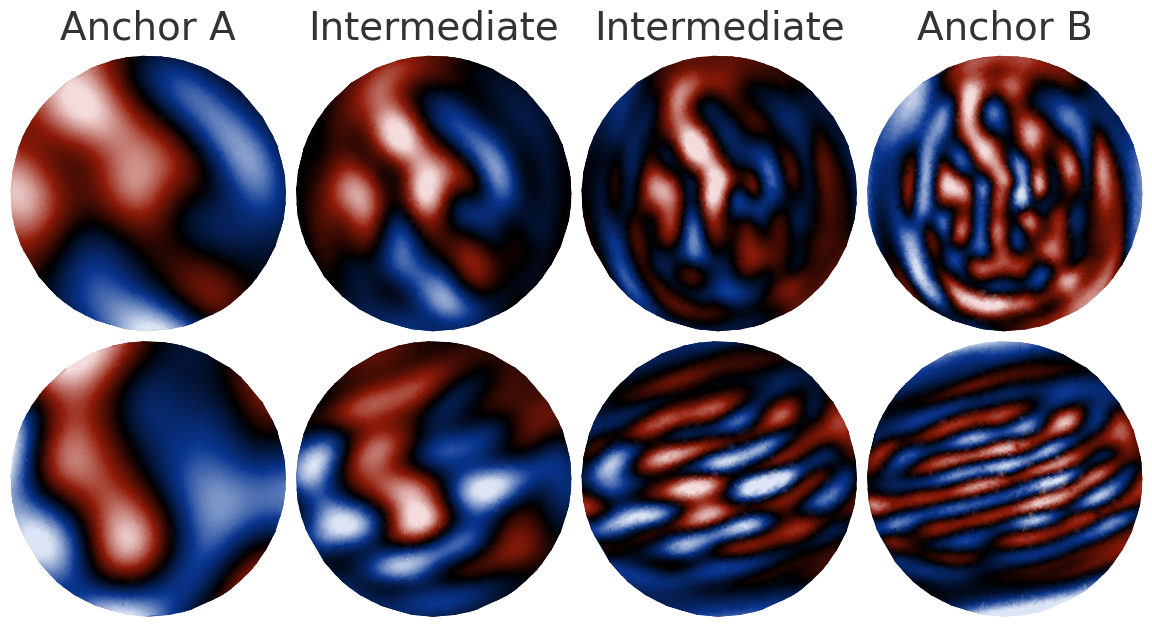}
  \caption{\textbf{Geodesic morphing.} Each row decodes points on the great circle between the embeddings of a low-frequency radial-sine field (anchor A) and a high-frequency one (anchor B). The intermediates are coherent fields rather than noise. Single-partition embeddings, with further examples in Appendix~\ref{app:latent-morph}.}
  \label{fig:morph-main}
\end{figure}

\paragraph{Exploring the latent space.}
\label{sec:latent}
We now turn to the latent space. Every morphologic embedding is a point on a fixed hypersphere that decodes to a field, and both the encoder and the decoder are continuous and differentiable maps. The space can therefore be explored directly, by choosing points and paths on the hypersphere and decoding them (Appendix~\ref{app:latent}). The first finding is that the entire hypersphere decodes to valid fields, rather than only a thin region around the embeddings of training functions. The components of the latent space themselves decode to elementary morphologies (Appendix~\ref{app:latent-atoms}), and walking on the sphere, whether along one angle of its spherical parametrization or along the great circle between two embeddings, changes the decoded field gradually and through coherent intermediates (Figure~\ref{fig:morph-main}). The transition from one function to another can thus be followed step by step, which indicates that the space is organized around the structure of functions rather than around the identity of individual samples. The space is, moreover, coherent in a stronger sense, since flipping the sign of a component changes the field while preserving the integrity of its morphology (Appendix~\ref{app:latent-walk}). No small set of components is therefore responsible for holding a field together.

The geometry of the latent space furthermore agrees with that of the physical space, in that the relative distance between two decoded fields nearly equals the chord distance between the unit directions of their embeddings (Appendix~\ref{app:latent-metric}). This is the property on which a loss evaluated in the reduced space relies (Section~\ref{sec:construction}). Without it, a processor trained to match embeddings would optimize a quantity only loosely related to the error of the fields it predicts. Embeddings also compose, but only locally. The on-sphere sum of two nearby embeddings decodes to nearly the normalized mean of their two fields, and the agreement degrades smoothly with their distance as the nonlinearity of the decoder takes over, so the space is locally linear and globally curved (Appendix~\ref{app:latent-superpos}). Finally, the capacity of the space is used in full. The spectrum of the linearized decode-encode map, which decodes an embedding and encodes the result again, shows that each individual embedding occupies about half of the $128$ dimensions, but the discarded directions differ from one embedding to the next, so that the population as a whole occupies all of them (Appendices~\ref{app:latent-spectrum} and~\ref{app:latent-usage}). Among the directions it occupies, spatial frequency is the dominant axis along which families of functions are organized (Appendix~\ref{app:latent-families}), consistent with the role the ESF plays in the difficulty of reconstruction.

\paragraph{Transferable directions and physical dynamics.}
The observations so far establish that the latent geometry is smooth and faithful to distances. Two further experiments indicate that it is also meaningful, in the sense that simple geometric operations on embeddings correspond to interpretable operations on fields.

The first experiment concerns time. We encode two snapshots of a physical evolution, move halfway along the geodesic that connects their embeddings, and decode the result. The geodesic has no knowledge of the governing equations, and nothing is fitted, time-stepped, or supervised in the process. Whether the decoded midpoint resembles the true intermediate state is therefore not obvious a priori. It is plausible, however, because an evolution over a short interval changes a field gradually, and a geometry that preserves distances may map such a gradual change to a short and nearly straight path. We compare the interpolant with the true physical state at the intermediate time and with the baseline of simply keeping one of the two snapshots (Appendix~\ref{app:timeinterp}). Two time indices apart on incompressible flow from a Brownian-bridge initial condition, the interpolant has an error of $1.4\%$, against $7.9\%$ for keeping a snapshot and a reconstruction floor of $0.32\%$, so that the geodesic recovers most of the change between the two states. The advantage narrows as the interval grows. It persists over every interval we examined on all eleven evolution datasets we measured, of incompressible, compressible, and acoustic flow, shocked compressible flow included, although the margin is narrowest on compressible flow. The latent space thus admits a zero-shot interpolation of physical dynamics in time, and this suggests that the path a trained processor has to follow between two states is short and well behaved in this geometry.

The second experiment concerns edits. We take a single pair of examples that differ by one operation, such as a blur, an anisotropic stretch, or the split of one feature into two, and read off the difference between their embeddings as a direction. Stepping along this direction beyond either example continues the edit, and stepping in the opposite direction reverses it. Adding the same direction to the embedding of an unrelated field, moreover, applies the same edit to that field (Appendix~\ref{app:latent-dirs}). An operation therefore occupies a consistent direction across the hypersphere, rather than being tied to the pair it was read from. This is the behavior one would hope for in a representation that is meant to host operators, since it shows that at least some meaningful transformations act on embeddings in a simple and transferable way, independently of the particular field to which they are applied.

\section{Operator Learning}
\label{sec:operator}

With a frozen latent space, we can take advantage of input and output caching and train a processor module that operates entirely in the latent space. All operators of this section are learned on one and the same low-dimensional backbone ($\embdim{=}8$, \hyperref[tab:backbones]{appendix prelude}), pre-trained once on synthetic functions and used for every problem as it is. No problem re-trains the backbone, so a new task adds no cost to it, and the cost of a task is that of its processor alone. This section presents the most important findings. Details of the experiments and detailed benchmarks against baseline architectures are in Appendix~\ref{app:benchmark}. We compare against two end-to-end frameworks that set a high bar. On 28 benchmark problems across diverse PDEs, GAOT \citep{wen2025geometry} reaches the highest accuracy and robustness among the neural operators it is compared with while being the most efficient of them, and it attains state-of-the-art results on large-scale three-dimensional industrial flow datasets. RIGNO \citep{mousavi2025rigno} is reported to be markedly more accurate than earlier neural operators and efficient on problems of moderate size, and it generalizes robustly to unseen resolutions in space and time.

\begin{figure*}[t]\centering
  \begin{subfigure}[b]{0.37\textwidth}\centering
    \includegraphics[width=\textwidth]{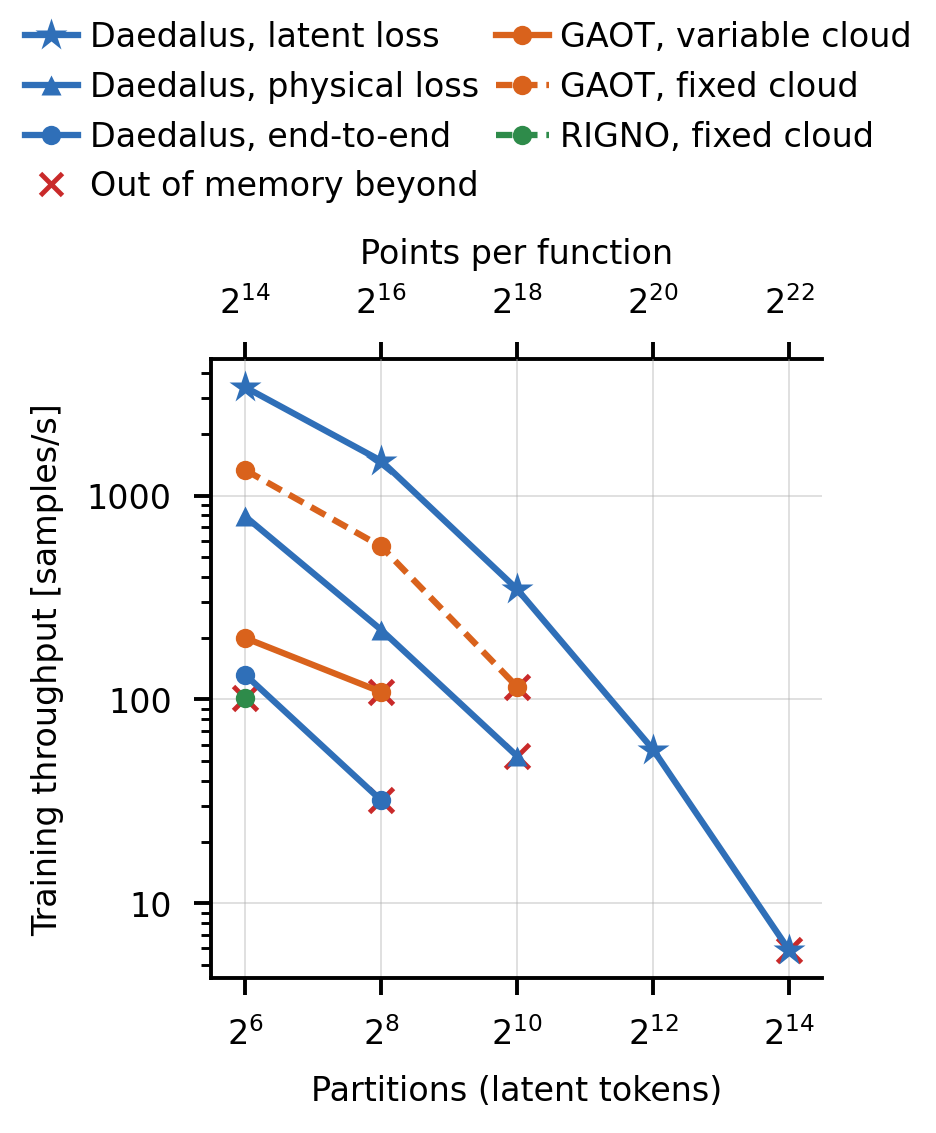}
    \caption{Performance.}
  \end{subfigure}\hfill
  \begin{subfigure}[b]{0.6\textwidth}\centering
    \includegraphics[width=\textwidth]{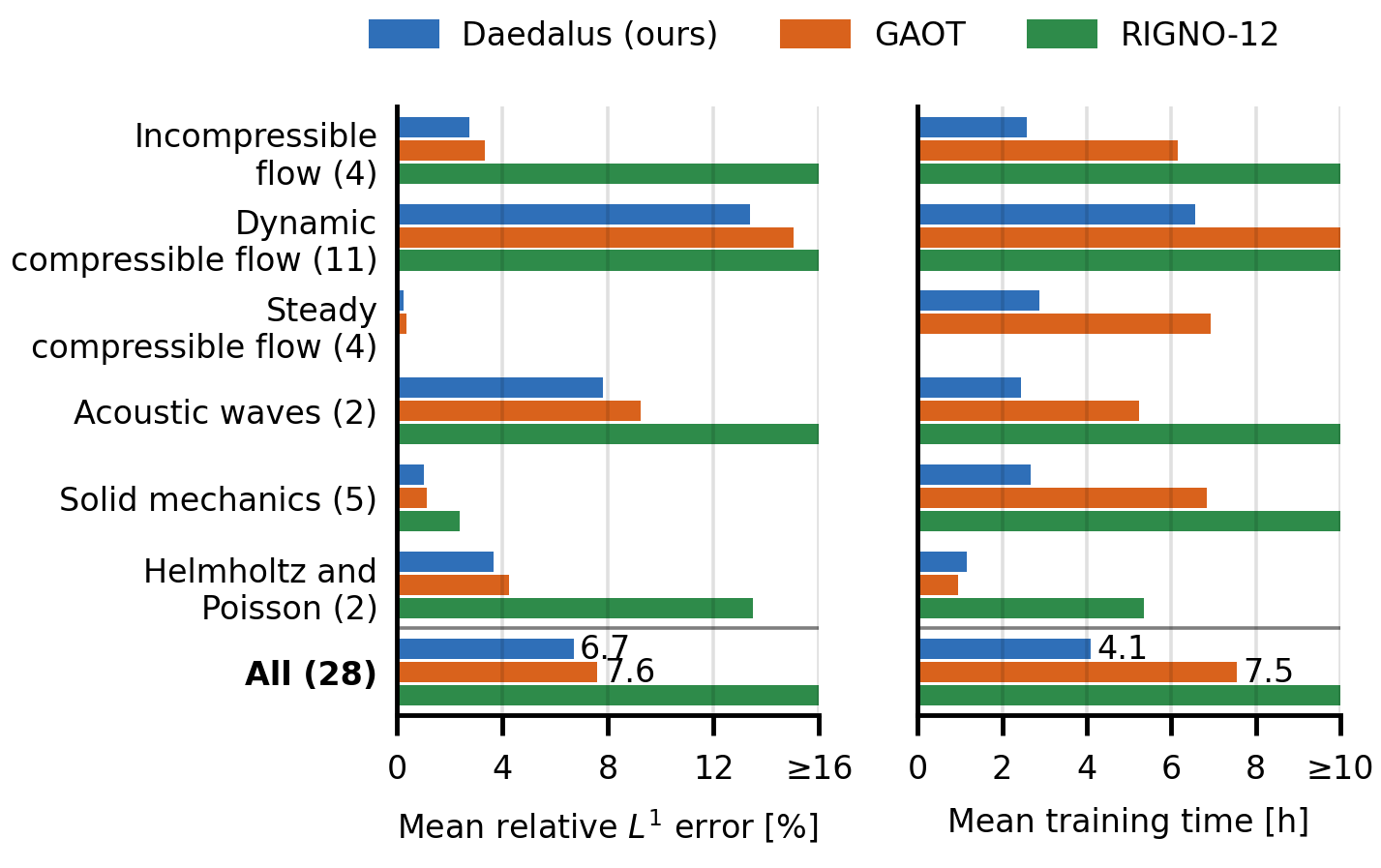}
    \caption{Accuracy and training time.}
  \end{subfigure}
  \caption{\textbf{Cost and accuracy of operators on the frozen backbone.} (a) Training throughput on one 24\,GB RTX~4090 against the number of partitions, at 256 points per partition, so that the points per function (top axis) grow with the number of partitions, and at a batch of 32. The blue lines are \daedalus{} trained on the latent loss with input and output caches (stars), on the physical loss with an input cache (triangles), and end to end, partitioning and encoding the points at every step (circles). The orange lines are GAOT trained end to end on a point cloud of its own for every sample (solid) and on one point cloud shared by the batch (dashed), and the green point is RIGNO trained end to end on a shared point cloud. Every line ends at a red cross, beyond which its model runs out of memory, so that RIGNO fits only the smallest problem and the latent loss reaches problems 16 times larger than any other setting. (b) The median relative $L^1$ test error and the total training time of \daedalus{}, GAOT, and RIGNO-12 (the smaller RIGNO variant), averaged over the problems of every family of the accuracy benchmark, with the number of problems in parentheses, and over all 28 problems. A bar that reaches the end of an axis is cut there. RIGNO-12 is not trained on the steady compressible flows, so its average over all problems is over the other 24. Appendices~\ref{app:perf} and~\ref{app:benchmark} give the complete performance and accuracy benchmarks, and Table~\ref{tab:benchmark} the result on every problem.}
  \label{fig:benchmark-summary}
\end{figure*}

\paragraph{\daedalus{} is substantially more efficient to train and infer.} A processor trained in the latent space reads cached embeddings and never touches the points. The cost of a training step is therefore independent of the resolution of the data. From 4 thousand to 67 million points per function, latent training runs at the same rate and within the same memory (Figure~\ref{fig:perf-space-size}). No existing end-to-end framework can do this, since each encodes and decodes the full point cloud at every step. We compare training and inference with two state-of-the-art end-to-end frameworks, GAOT \citep{wen2025geometry} and RIGNO-12 \citep{mousavi2025rigno} (Appendix~\ref{app:perf}). For a fair comparison, the number of partitions, latent tokens, or regional nodes grows with the problem at a fixed number of points per partition. The processor is also deepened until it no longer fits, which locates the limits of every architecture. \daedalus{} is more efficient with fixed and with variable geometries. Its advantage is largest with variable geometries, where GAOT cannot reuse the geometric preprocessing that it otherwise shares across a batch.
All of the following holds on a single consumer GPU with 24\,GB of memory and includes the cost of partitioning. \daedalus{} trains on problems of up to 4.2 million points in 16384 partitions at a batch of 32 (Figure~\ref{fig:benchmark-summary}a). This is 64 times larger than the largest problem on which GAOT trains with variable geometries, and 16 times larger with a fixed geometry. On that largest problem, \daedalus{} trains 13.6 and 3.0 times faster, respectively (Figure~\ref{fig:perf-problem-size-training}). On problems of 262 thousand points in 4096 partitions, it trains a processor of 101 million parameters, 92 times larger than the largest GAOT that fits. At the size of that GAOT, \daedalus{}  trains 4.9 times faster (Figure~\ref{fig:perf-model-size-training}). At inference it runs a processor of 1.6 billion parameters on 32 such problems at once, whereas the largest GAOT that fits has 285 million (Figure~\ref{fig:perf-model-size-latency}). The encoder and the decoder act on every partition independently, so a function can be encoded and decoded in chunks while the processor still sees the whole domain. In this way a single GPU maps a function of 537 million points, 16 times more than fits in one piece (Figure~\ref{fig:perf-chunks}). Finally, a \emph{latent rollout} encodes the initial state once, advances it in the latent space, and decodes only the states that are needed. Long autoregressive rollouts thus run up to 10 times faster than when every step is decoded and re-encoded, and up to 9.6 times faster than GAOT, which steps in the physical space (Figure~\ref{fig:perf-rollout}).

\paragraph{These savings significantly shift the scale of what is feasible in operator learning.} At the same compute budget, these numbers change the scale at which foundation models for physics can be pre-trained and fine-tuned, and at which task-specific models can be trained from scratch. End-to-end pre-training passes every step through the encoder and the decoder of the full point cloud, so the cost of a step grows with the resolution of the data, and dense corpora have to be downsampled to fit. On cached embeddings a step costs the processor alone at any resolution, so the same budget pre-trains a far larger processor, 92 times larger than GAOT on one GPU, on corpora kept at full resolution. Since the memory of latent training grows linearly with the number of partitions and with the depth of the processor, by about 1.2\,GB per thousand partitions and 0.27\,GB per unit at a batch of 32 (Figures~\ref{fig:perf-problem-size-training} and~\ref{fig:perf-model-size-training}), this carries over to a single node. Eight RTX~Pro~6000 GPUs hold 768\,GB, 32 times the memory of one RTX~4090, and if that memory is pooled by sharding the model and the batch across the GPUs, the measured growth projects pre-training at a batch of 32 on problems of about 160 million points in more than half a million partitions, or a processor of about 4 billion parameters on problems of 262 thousand points. Fine-tuning such a model for a new task reads the same frozen representation, so it runs on cached embeddings of the new data and fits on a single consumer GPU rather than on the hardware that pre-trained it. A task-specific model trained from scratch on one consumer GPU, finally, handles problems of millions of points, 64 times larger than end-to-end training fits, and trains up to 13.6 times faster, so that large models can be trained in a few hours rather than days.

\paragraph{The cooldown closes the proxy gap at little extra cost.} We trained the same processor in four ways on eight of the problems, on the latent loss throughout, on the physical loss throughout, with a cooldown, and with a cooldown that also unfreezes the backbone (Appendix~\ref{app:benchmark-regimes}). On the steady problems the cooldown closes the proxy gap, the excess error that training on the latent proxy leaves, with an error 31\% below that of the latent loss and only 3\% above that of training on the physical loss throughout. On the evolution problems it does no harm, its error being 8\% below that of the latent loss. Because the bulk of the training still runs on cached embeddings, it adds only 16\% to the training time, whereas the physical loss throughout takes $1.7$ times as long as the latent loss, since it forgoes the output cache. Unfreezing the encoder and the decoder during the cooldown lowers the error of the frozen cooldown by a further 8\% on the steady problems and leaves the evolution problems unchanged, at $2.4$ times its training time, since an unfrozen backbone admits no cache. The pre-trained modules still help in this mode, since the bulk of the training runs on cached embeddings and the unfrozen encoder and decoder start from their pre-trained weights, but the gain costs the invertibility of the encoding and decoding and the interpretability of the embeddings, because the FAE then adapts to the task at hand.

\paragraph{\daedalus{} exceeds state-of-the-art accuracy at significantly lower training times.} We benchmark the processor on the frozen backbone against GAOT and RIGNO-12 on 28 problems under one protocol, with the same data, splits, rollout, and metric for every model (Appendix~\ref{app:benchmark}). The problems span compressible and incompressible fluid dynamics, steady compressible flow past airfoils and bluff bodies, solid mechanics, acoustic waves, and the Helmholtz and Poisson equations, seventeen of them time dependent and eleven steady, on fixed grids as well as on point clouds whose geometry varies between samples. Trained mostly in the latent space and cooled down on the physical loss, \daedalus{} matches or exceeds the accuracy of both baselines at a fraction of their training cost (Figure~\ref{fig:benchmark-summary}b and Table~\ref{tab:benchmark}). Its relative $L^1$ error is lower than that of GAOT on 23 of the 28 problems, and the per-problem error of GAOT exceeds ours by 23\% on average. Training is on average $2.1$ times faster per problem than for GAOT. GAOT runs in its fast mode on the fixed-geometry problems, where its geometric preprocessing is computed once and shared across all samples. That saving exists only when every sample shares one geometry, whereas cached embeddings serve variable geometries just as well, and on the nine problems with a varying geometry its training time is $2.7$ times ours. Our training time excludes the pre-training of the backbone, which is paid once for all problems and does not use any dataset.

\section{Conclusion}

We have shown that the representation of functions can be learned once and decoupled from operator learning, without sacrificing accuracy and with large gains in memory and training speed, much as natural language processing and computer vision decoupled their representations from their tasks to train them at scale. A functional autoencoder pre-trained on analytic functions alone, with no solver in the loop, induces a reduced space that is frozen and reused, and the same backbone reconstructs PDE fields of several kinds and natural photographs, on geometries and samplings it never saw. This adds a nonlinear yet task-agnostic representation to the succession of operator-learning frameworks.

The transfer follows from structure rather than from data. Domain decomposition and PartNorm reduce an unfamiliar field to partitions in a common frame, and the effective spatial frequency of a partition, rather than the family of the field, determines how accurately it is represented. The same quantity also sets the limit of the approach, since a partition whose effective frequency lies above the band of the training distribution cannot be represented accurately. The accuracy of the representation thereby becomes an inference-time decision, since the number of partitions is chosen per input and per task without retraining. Because the encoder aggregates whatever points a partition holds and the decoder answers any coordinate, the same space also holds fields observed only on curves or on part of their domain, and so offers a shared representation for data defined on a domain and on its boundary. A latent space that belongs to no specific family of functions can, moreover, be examined in its own right. The whole hypersphere decodes to valid fields, distances on it track distances between the decoded fields, geodesics between nearby states of an evolution approximate the states in between, and edit directions read from one pair of fields transfer to unrelated ones.

Freezing the representation maps makes input and output caching possible. A dataset encoded once is a dataset compressed once, so an operator trains on stored embeddings without touching the physical space, at a cost that no longer depends on the resolution of the data. The savings in memory and time grow with the density of the problem, dense training data need not be downsampled, and a single consumer GPU trains large processors on problems far denser than end-to-end training can hold. These gains can be decisive at every stage of a foundation model for physics. Pre-training a large processor on many datasets becomes far cheaper when every dataset is compressed once and a step runs the processor alone, adapting such a model to a downstream task needs neither the data at full resolution in the loop nor a data-center GPU, and at inference latent rollouts and chunked encoding and decoding let large models advance over long horizons and on problems of hundreds of millions of points on a consumer-grade GPU. Trained mostly in the latent space and finished by a short cooldown on the physical loss, an operator on the frozen backbone matches or exceeds the accuracy of the state-of-the-art end-to-end models across fluid dynamics, solid mechanics, and acoustics at a fraction of their training cost, and the architecture of the processor becomes an independent choice that can be replaced without retraining the backbone. Scaling operator learning asks for the same shift that natural language processing and computer vision have already made, from a model per task to a representation shared across tasks, and a backbone that encodes one scalar channel at a time on any point cloud is indifferent to the number of variables a field carries, to the physics behind them, and to the discretization on which they are observed.

The same structure that makes the representation transferable also sets its limits. A high-frequency field is captured only if it is partitioned finely enough for the effective spatial frequency of every partition to fall into the training band, and its high-frequency features are lost otherwise (Appendix~\ref{app:freq}). How finely a field can be partitioned is in turn bounded by its sampling, since a partition needs enough points to be represented (Appendix~\ref{app:sweep}). Overlapping partitions reuse points and relax this bound, but each partition then covers a larger area and carries a higher effective frequency. Rotation, moreover, is handled only optionally, since full rotation equivariance is available but not enforced in the trained backbones (Appendix~\ref{app:equivariance}). Finally, the design of \daedalus{} targets latent operator learning as a whole, and on specific tasks it can fall behind end-to-end frameworks fitted to that task alone, as on four problems of the accuracy benchmark, where GAOT is more than 5\% more accurate (Table~\ref{tab:benchmark}).

These limitations also point to the next steps. The present backbones are two-dimensional, and three-dimensional backbones are the natural continuation, where the savings of caching grow with the number of points. Such a backbone could in turn serve two- and three-dimensional tasks alike, as well as mixed modalities, and computer vision tasks, which we have not explored, offer a further application of the same paradigm. Beyond the dimension, the distribution of the partitions and the dimension of the embedding are currently fixed for a whole input, and adapting both to the local content of a field would spend the capacity of the representation where a field needs it.

\FloatBarrier

\balance
\bibliographystyle{abbrvnat}
\bibliography{references}

\begin{thebibliography}{65}
\providecommand{\natexlab}[1]{#1}
\providecommand{\url}[1]{\texttt{#1}}
\expandafter\ifx\csname urlstyle\endcsname\relax
  \providecommand{\doi}[1]{doi: #1}\else
  \providecommand{\doi}{doi: \begingroup \urlstyle{rm}\Url}\fi

\bibitem[Alkin et~al.(2024)Alkin, F\"{u}rst, Schmid, Gruber, Holzleitner, and Brandstetter]{alkin2024universal}
B.~Alkin, A.~F\"{u}rst, S.~Schmid, L.~Gruber, M.~Holzleitner, and J.~Brandstetter.
\newblock {Universal Physics Transformers}: A framework for efficiently scaling neural operators.
\newblock In \emph{Advances in Neural Information Processing Systems}, 2024.

\bibitem[Bauer et~al.(2023)Bauer, Dupont, Brock, Rosenbaum, Schwarz, and Kim]{bauer2023spatial}
M.~Bauer, E.~Dupont, A.~Brock, D.~Rosenbaum, J.~R. Schwarz, and H.~Kim.
\newblock Spatial {Functa}: Scaling {Functa} to {ImageNet} classification and generation, 2023.
\newblock URL \url{https://arxiv.org/abs/2302.03130}.

\bibitem[Bhattacharya et~al.(2021)Bhattacharya, Hosseini, Kovachki, and Stuart]{bhattacharya2021model}
K.~Bhattacharya, B.~Hosseini, N.~B. Kovachki, and A.~M. Stuart.
\newblock Model reduction and neural networks for parametric {PDE}s.
\newblock \emph{The SMAI journal of computational mathematics}, 7:\penalty0 121--157, 2021.

\bibitem[Bunker et~al.(2025)Bunker, Girolami, Lambley, Stuart, and Sullivan]{bunker2025autoencoders}
J.~Bunker, M.~Girolami, H.~Lambley, A.~M. Stuart, and T.~J. Sullivan.
\newblock Autoencoders in function space.
\newblock \emph{Journal of Machine Learning Research}, 26\penalty0 (165):\penalty0 1--54, 2025.

\bibitem[Cao et~al.(2024)Cao, Goswami, and Karniadakis]{cao2024laplace}
Q.~Cao, S.~Goswami, and G.~E. Karniadakis.
\newblock Laplace neural operator for solving differential equations.
\newblock \emph{Nature Machine Intelligence}, 6\penalty0 (6):\penalty0 631--640, 2024.

\bibitem[Caron et~al.(2021)Caron, Touvron, Misra, J\'egou, Mairal, Bojanowski, and Joulin]{caron2021emerging}
M.~Caron, H.~Touvron, I.~Misra, H.~J\'egou, J.~Mairal, P.~Bojanowski, and A.~Joulin.
\newblock Emerging properties in self-supervised vision transformers.
\newblock In \emph{IEEE/CVF International Conference on Computer Vision}, 2021.

\bibitem[Chabra et~al.(2020)Chabra, Lenssen, Ilg, Schmidt, Straub, Lovegrove, and Newcombe]{chabra2020deep}
R.~Chabra, J.~E. Lenssen, E.~Ilg, T.~Schmidt, J.~Straub, S.~Lovegrove, and R.~Newcombe.
\newblock Deep local shapes: Learning local {SDF} priors for detailed {3D} reconstruction.
\newblock In \emph{European Conference on Computer Vision}, 2020.

\bibitem[Chen and Chen(1995)]{chen1995universal}
T.~Chen and H.~Chen.
\newblock Universal approximation to nonlinear operators by neural networks with arbitrary activation functions and its application to dynamical systems.
\newblock \emph{IEEE Transactions on Neural Networks}, 6\penalty0 (4):\penalty0 911--917, 1995.

\bibitem[Chen et~al.(2020)Chen, Kornblith, Norouzi, and Hinton]{chen2020simple}
T.~Chen, S.~Kornblith, M.~Norouzi, and G.~Hinton.
\newblock A simple framework for contrastive learning of visual representations.
\newblock In \emph{International Conference on Machine Learning}, 2020.

\bibitem[Chen et~al.(2024)Chen, Song, Ren, Subramanian, Morozov, and Mahoney]{chen2024data}
W.~Chen, J.~Song, P.~Ren, S.~Subramanian, D.~Morozov, and M.~W. Mahoney.
\newblock Data-efficient operator learning via unsupervised pretraining and in-context learning.
\newblock In \emph{Advances in Neural Information Processing Systems}, 2024.

\bibitem[Chen et~al.(2021)Chen, Liu, and Wang]{chen2021learning}
Y.~Chen, S.~Liu, and X.~Wang.
\newblock Learning continuous image representation with local implicit image function.
\newblock In \emph{IEEE/CVF Conference on Computer Vision and Pattern Recognition}, 2021.

\bibitem[{Daedalus}(2026)]{daedaluscredits2026}
{Daedalus}.
\newblock Image credits, 2026.
\newblock URL \url{https://daedalus.ethz.ch/credits.html}.
\newblock [Online; accessed 20-October-2026].

\bibitem[Devlin et~al.(2019)Devlin, Chang, Lee, and Toutanova]{devlin2019bert}
J.~Devlin, M.-W. Chang, K.~Lee, and K.~Toutanova.
\newblock {BERT}: Pre-training of deep bidirectional transformers for language understanding.
\newblock In \emph{Conference of the North American Chapter of the Association for Computational Linguistics}, 2019.

\bibitem[Dolean et~al.(2015)Dolean, Jolivet, and Nataf]{dolean2015introduction}
V.~Dolean, P.~Jolivet, and F.~Nataf.
\newblock \emph{An Introduction to Domain Decomposition Methods}.
\newblock Society for Industrial and Applied Mathematics, 2015.
\newblock ISBN 9781611974065.
\newblock \doi{10.1137/1.9781611974065}.

\bibitem[Essakine et~al.(2025)Essakine, Cheng, Cheng, Zhang, Deng, Zhu, Sch{\"o}nlieb, and Aviles-Rivero]{essakine2025where}
A.~Essakine, Y.~Cheng, C.-W. Cheng, L.~Zhang, Z.~Deng, L.~Zhu, C.-B. Sch{\"o}nlieb, and A.~I. Aviles-Rivero.
\newblock Where do we stand with implicit neural representations? a technical and performance survey.
\newblock \emph{Transactions on Machine Learning Research}, 2025.
\newblock ISSN 2835-8856.
\newblock Survey Certification.

\bibitem[Fanaskov and Oseledets(2023)]{fanaskov2023spectral}
V.~S. Fanaskov and I.~V. Oseledets.
\newblock Spectral neural operators.
\newblock \emph{Doklady Mathematics}, 108\penalty0 (S2):\penalty0 S226--S232, 2023.
\newblock \doi{10.1134/s1064562423701107}.

\bibitem[Gupta et~al.(2021)Gupta, Xiao, and Bogdan]{gupta2021multiwavelet}
G.~Gupta, X.~Xiao, and P.~Bogdan.
\newblock Multiwavelet-based operator learning for differential equations.
\newblock In \emph{Advances in Neural Information Processing Systems}, 2021.

\bibitem[He et~al.(2022)He, Chen, Xie, Li, Doll\'ar, and Girshick]{he2022masked}
K.~He, X.~Chen, S.~Xie, Y.~Li, P.~Doll\'ar, and R.~Girshick.
\newblock Masked autoencoders are scalable vision learners.
\newblock In \emph{IEEE/CVF Conference on Computer Vision and Pattern Recognition}, 2022.

\bibitem[Heo et~al.(2024)Heo, Park, Han, and Yun]{heo2024rotary}
B.~Heo, S.~Park, D.~Han, and S.~Yun.
\newblock Rotary position embedding for vision transformer.
\newblock In \emph{European Conference on Computer Vision}, 2024.

\bibitem[Herde et~al.(2024)Herde, Raoni\'{c}, Rohner, K\"{a}ppeli, Molinaro, de~B\'{e}zenac, and Mishra]{herde2024poseidon}
M.~Herde, B.~Raoni\'{c}, T.~Rohner, R.~K\"{a}ppeli, R.~Molinaro, E.~de~B\'{e}zenac, and S.~Mishra.
\newblock Poseidon: Efficient foundation models for {PDE}s.
\newblock In \emph{Advances in Neural Information Processing Systems}, 2024.

\bibitem[Hesthaven and Ubbiali(2018)]{hesthaven2018non}
J.~Hesthaven and S.~Ubbiali.
\newblock Non-intrusive reduced order modeling of nonlinear problems using neural networks.
\newblock \emph{Journal of Computational Physics}, 363:\penalty0 55--78, 2018.

\bibitem[Kaplan et~al.(2020)Kaplan, McCandlish, Henighan, Brown, Chess, Child, Gray, Radford, Wu, and Amodei]{kaplan2020scaling}
J.~Kaplan, S.~McCandlish, T.~Henighan, T.~B. Brown, B.~Chess, R.~Child, S.~Gray, A.~Radford, J.~Wu, and D.~Amodei.
\newblock Scaling laws for neural language models.
\newblock \emph{CoRR}, abs/2001.08361, 2020.

\bibitem[Kovachki et~al.(2023)Kovachki, Li, Liu, Azizzadenesheli, Bhattacharya, Stuart, and Anandkumar]{kovachki2023neural}
N.~Kovachki, Z.~Li, B.~Liu, K.~Azizzadenesheli, K.~Bhattacharya, A.~Stuart, and A.~Anandkumar.
\newblock Neural operator: Learning maps between function spaces with applications to {PDE}s.
\newblock \emph{Journal of Machine Learning Research}, 24\penalty0 (89):\penalty0 1--97, 2023.

\bibitem[Lanthaler et~al.(2023)Lanthaler, Molinaro, Hadorn, and Mishra]{lanthaler2023nonlinear}
S.~Lanthaler, R.~Molinaro, P.~Hadorn, and S.~Mishra.
\newblock Nonlinear reconstruction for operator learning of {PDE}s with discontinuities.
\newblock In \emph{International Conference on Learning Representations}, 2023.

\bibitem[Li et~al.(2021)Li, Kovachki, Azizzadenesheli, Liu, Bhattacharya, Stuart, and Anandkumar]{li2021fourier}
Z.~Li, N.~B. Kovachki, K.~Azizzadenesheli, B.~Liu, K.~Bhattacharya, A.~Stuart, and A.~Anandkumar.
\newblock {F}ourier neural operator for parametric partial differential equations.
\newblock In \emph{International Conference on Learning Representations}, 2021.

\bibitem[Li et~al.(2023)Li, Kovachki, Choy, Li, Kossaifi, Otta, Nabian, Stadler, Hundt, Azizzadenesheli, and Anandkumar]{li2023geometryinformed}
Z.~Li, N.~B. Kovachki, C.~Choy, B.~Li, J.~Kossaifi, S.~P. Otta, M.~A. Nabian, M.~Stadler, C.~Hundt, K.~Azizzadenesheli, and A.~Anandkumar.
\newblock Geometry-informed neural operator for large-scale {3D} {PDE}s.
\newblock In \emph{Advances in Neural Information Processing Systems}, 2023.

\bibitem[Li et~al.(2024)Li, Zheng, Kovachki, Jin, Chen, Liu, Azizzadenesheli, and Anandkumar]{li2024physics}
Z.~Li, H.~Zheng, N.~Kovachki, D.~Jin, H.~Chen, B.~Liu, K.~Azizzadenesheli, and A.~Anandkumar.
\newblock Physics-informed neural operator for learning partial differential equations.
\newblock \emph{ACM / IMS Journal of Data Science}, 1\penalty0 (3):\penalty0 1–27, May 2024.
\newblock ISSN 2831-3194.
\newblock \doi{10.1145/3648506}.

\bibitem[Lingsch et~al.(2026)Lingsch, Kissas, Jakubik, and Mishra]{lingsch2026phaedra}
L.~Lingsch, G.~Kissas, J.~Jakubik, and S.~Mishra.
\newblock Phaedra: Learning high-fidelity discrete tokenization for the physical science, 2026.
\newblock URL \url{https://arxiv.org/abs/2602.03915}.

\bibitem[Liu et~al.(2026)Liu, Koehler, Holzschuh, and Thuerey]{liu2026tadpole}
Q.~Liu, F.~Koehler, B.~Holzschuh, and N.~Thuerey.
\newblock Tadpole: Autoencoders as foundation models for {3D} {PDE}s with online learning, 2026.
\newblock URL \url{https://arxiv.org/abs/2605.15284}.

\bibitem[Lloyd(1982)]{lloyd1982least}
S.~Lloyd.
\newblock Least squares quantization in {PCM}.
\newblock \emph{IEEE Transactions on Information Theory}, 28\penalty0 (2):\penalty0 129--137, 1982.

\bibitem[Loshchilov et~al.(2025)Loshchilov, Hsieh, Sun, and Ginsburg]{loshchilov2025ngpt}
I.~Loshchilov, C.-P. Hsieh, S.~Sun, and B.~Ginsburg.
\newblock {nGPT}: Normalized transformer with representation learning on the hypersphere.
\newblock In \emph{International Conference on Learning Representations}, 2025.

\bibitem[Lu et~al.(2021)Lu, Jin, Pang, Zhang, and Karniadakis]{lu2021deeponet}
L.~Lu, P.~Jin, G.~Pang, Z.~Zhang, and G.~E. Karniadakis.
\newblock Learning nonlinear operators via {DeepONet} based on the universal approximation theorem of operators.
\newblock \emph{Nature Machine Intelligence}, 2021.

\bibitem[Lu et~al.(2022)Lu, Meng, Cai, Mao, Goswami, Zhang, and Karniadakis]{lu2022comprehensive}
L.~Lu, X.~Meng, S.~Cai, Z.~Mao, S.~Goswami, Z.~Zhang, and G.~E. Karniadakis.
\newblock A comprehensive and fair comparison of two neural operators (with practical extensions) based on {FAIR} data.
\newblock \emph{Computer Methods in Applied Mechanics and Engineering}, 393:\penalty0 114778, 2022.

\bibitem[Mialon et~al.(2023)Mialon, Garrido, Lawrence, Rehman, LeCun, and Kiani]{mialon2023self}
G.~Mialon, Q.~Garrido, H.~Lawrence, D.~Rehman, Y.~LeCun, and B.~Kiani.
\newblock Self-supervised learning with {L}ie symmetries for partial differential equations.
\newblock In \emph{Advances in Neural Information Processing Systems}, 2023.

\bibitem[Mildenhall et~al.(2020)Mildenhall, Srinivasan, Tancik, Barron, Ramamoorthi, and Ng]{mildenhall2020nerf}
B.~Mildenhall, P.~P. Srinivasan, M.~Tancik, J.~T. Barron, R.~Ramamoorthi, and R.~Ng.
\newblock {NeRF}: Representing scenes as neural radiance fields for view synthesis.
\newblock In \emph{European Conference on Computer Vision}, 2020.

\bibitem[Mousavi et~al.(2025)Mousavi, Wen, Lingsch, Herde, Raonić, and Mishra]{mousavi2025rigno}
S.~Mousavi, S.~Wen, L.~Lingsch, M.~Herde, B.~Raonić, and S.~Mishra.
\newblock {RIGNO}: A graph-based framework for robust and accurate operator learning for {PDE}s on arbitrary domains.
\newblock In \emph{Advances in Neural Information Processing Systems}, 2025.

\bibitem[Mousavi et~al.(2026)Mousavi, Mishra, and Lorenzis]{mousavi2026imposing}
S.~Mousavi, S.~Mishra, and L.~D. Lorenzis.
\newblock Imposing boundary conditions on neural operators via learned function extensions.
\newblock In \emph{International Conference on Machine Learning}, volume~43, 2026.

\bibitem[M{\"u}ller et~al.(2022)M{\"u}ller, Evans, Schied, and Keller]{mueller2022instant}
T.~M{\"u}ller, A.~Evans, C.~Schied, and A.~Keller.
\newblock Instant neural graphics primitives with a multiresolution hash encoding.
\newblock \emph{ACM Transactions on Graphics}, 41\penalty0 (4):\penalty0 1--15, 2022.
\newblock \doi{10.1145/3528223.3530127}.

\bibitem[Park et~al.(2019)Park, Florence, Straub, Newcombe, and Lovegrove]{park2019deepsdf}
J.~J. Park, P.~Florence, J.~Straub, R.~Newcombe, and S.~Lovegrove.
\newblock {DeepSDF}: Learning continuous signed distance functions for shape representation.
\newblock In \emph{IEEE/CVF Conference on Computer Vision and Pattern Recognition}, 2019.

\bibitem[Peherstorfer(2022)]{peherstorfer2022breaking}
B.~Peherstorfer.
\newblock Breaking the {K}olmogorov barrier with nonlinear model reduction.
\newblock \emph{Notices of the American Mathematical Society}, 69\penalty0 (5):\penalty0 725--733, 2022.

\bibitem[Peng et~al.(2020)Peng, Niemeyer, Mescheder, Pollefeys, and Geiger]{peng2020convolutional}
S.~Peng, M.~Niemeyer, L.~Mescheder, M.~Pollefeys, and A.~Geiger.
\newblock Convolutional occupancy networks.
\newblock In \emph{European Conference on Computer Vision}, 2020.

\bibitem[Perez et~al.(2018)Perez, Strub, De~Vries, Dumoulin, and Courville]{perez2018film}
E.~Perez, F.~Strub, H.~De~Vries, V.~Dumoulin, and A.~Courville.
\newblock {FiLM}: Visual reasoning with a general conditioning layer.
\newblock In \emph{AAAI Conference on Artificial Intelligence}, 2018.

\bibitem[Pfaff et~al.(2021)Pfaff, Fortunato, Sanchez-Gonzalez, and Battaglia]{pfaff2021learning}
T.~Pfaff, M.~Fortunato, A.~Sanchez-Gonzalez, and P.~Battaglia.
\newblock Learning mesh-based simulation with graph networks.
\newblock In \emph{International Conference on Learning Representations}, 2021.

\bibitem[Quarteroni and Valli(1999)]{quarteroni1999domain}
A.~Quarteroni and A.~Valli.
\newblock \emph{Domain Decomposition Methods for Partial Differential Equations}.
\newblock Oxford University Press, 1999.
\newblock ISBN 9780198501787.
\newblock \doi{10.1093/oso/9780198501787.001.0001}.

\bibitem[Quarteroni et~al.(2015)Quarteroni, Manzoni, and Negri]{quarteroni2015reduced}
A.~Quarteroni, A.~Manzoni, and F.~Negri.
\newblock \emph{Reduced basis methods for partial differential equations: an introduction}, volume~92.
\newblock Springer, 2015.

\bibitem[Rahman et~al.(2024)Rahman, George, Elleithy, Leibovici, Li, Bonev, White, Berner, Yeh, Kossaifi, et~al.]{rahman2024pretraining}
A.~Rahman, R.~J. George, M.~Elleithy, D.~Leibovici, Z.~Li, B.~Bonev, C.~White, J.~Berner, R.~A. Yeh, J.~Kossaifi, et~al.
\newblock Pretraining codomain attention neural operators for solving multiphysics {PDE}s.
\newblock In \emph{Advances in Neural Information Processing Systems}, 2024.

\bibitem[Raonic et~al.(2024)Raonic, Molinaro, De~Ryck, Rohner, Bartolucci, Alaifari, Mishra, and de~B{\'e}zenac]{raonic2024convolutional}
B.~Raonic, R.~Molinaro, T.~De~Ryck, T.~Rohner, F.~Bartolucci, R.~Alaifari, S.~Mishra, and E.~de~B{\'e}zenac.
\newblock Convolutional neural operators for robust and accurate learning of {PDE}s.
\newblock In \emph{Advances in Neural Information Processing Systems}, 2024.

\bibitem[Reiser et~al.(2021)Reiser, Peng, Liao, and Geiger]{reiser2021kilonerf}
C.~Reiser, S.~Peng, Y.~Liao, and A.~Geiger.
\newblock {KiloNeRF}: Speeding up neural radiance fields with thousands of tiny {MLPs}.
\newblock In \emph{IEEE/CVF International Conference on Computer Vision}, 2021.

\bibitem[Seidman et~al.(2023)Seidman, Kissas, Pappas, and Perdikaris]{seidman2023variational}
J.~H. Seidman, G.~Kissas, G.~J. Pappas, and P.~Perdikaris.
\newblock Variational autoencoding neural operators.
\newblock In \emph{International Conference on Machine Learning}, 2023.

\bibitem[Serrano et~al.(2023)Serrano, Le~Boudec, Kassa{\"\i}~Koupa{\"\i}, Wang, Yin, Vittaut, and Gallinari]{serrano2023operator}
L.~Serrano, L.~Le~Boudec, A.~Kassa{\"\i}~Koupa{\"\i}, T.~X. Wang, Y.~Yin, J.-N. Vittaut, and P.~Gallinari.
\newblock Operator learning with neural fields: Tackling {PDE}s on general geometries.
\newblock In \emph{Advances in Neural Information Processing Systems}, 2023.

\bibitem[Serrano et~al.(2024)Serrano, Wang, Le~Naour, Vittaut, and Gallinari]{serrano2024aroma}
L.~Serrano, T.~X. Wang, E.~Le~Naour, J.-N. Vittaut, and P.~Gallinari.
\newblock {AROMA}: Preserving spatial structure for latent {PDE} modeling with local neural fields.
\newblock In \emph{Advances in Neural Information Processing Systems}, 2024.

\bibitem[Su et~al.(2024)Su, Ahmed, Lu, Pan, Bo, and Liu]{su2024roformer}
J.~Su, M.~Ahmed, Y.~Lu, S.~Pan, W.~Bo, and Y.~Liu.
\newblock {RoFormer}: Enhanced transformer with rotary position embedding.
\newblock \emph{Neurocomputing}, 568:\penalty0 127063, 2024.

\bibitem[Tripura and Chakraborty(2023)]{tripura2023wavelet}
T.~Tripura and S.~Chakraborty.
\newblock Wavelet neural operator for solving parametric partial differential equations in computational mechanics problems.
\newblock \emph{Computer Methods in Applied Mechanics and Engineering}, 404:\penalty0 115783, 2023.
\newblock \doi{10.1016/j.cma.2022.115783}.

\bibitem[Wang et~al.(2021)Wang, Wang, and Perdikaris]{wang2021learning}
S.~Wang, H.~Wang, and P.~Perdikaris.
\newblock Learning the solution operator of parametric partial differential equations with physics-informed {DeepONet}s.
\newblock \emph{Science Advances}, 7\penalty0 (40), Oct. 2021.
\newblock ISSN 2375-2548.
\newblock \doi{10.1126/sciadv.abi8605}.

\bibitem[Wang and Wang(2024)]{wang2024latent}
T.~Wang and C.~Wang.
\newblock Latent neural operator for solving forward and inverse {PDE} problems.
\newblock In \emph{Advances in Neural Information Processing Systems}, 2024.

\bibitem[Wei et~al.(2026)Wei, List, and Thuerey]{wei2026revit}
H.~Wei, B.~List, and N.~Thuerey.
\newblock {ReViT}: Rotational-equivariant vision transformers for neural {PDE} solvers.
\newblock In \emph{International Conference on Machine Learning}, 2026.

\bibitem[Wen et~al.(2025)Wen, Kumbhat, Lingsch, Mousavi, Zhao, Chandrashekar, and Mishra]{wen2025geometry}
S.~Wen, A.~Kumbhat, L.~Lingsch, S.~Mousavi, Y.~Zhao, P.~Chandrashekar, and S.~Mishra.
\newblock Geometry aware operator transformer as an efficient and accurate neural surrogate for {PDE}s on arbitrary domains.
\newblock In \emph{Advances in Neural Information Processing Systems}, 2025.

\bibitem[{Wikimedia Commons}(2026)]{wikimediacommons2026}
{Wikimedia Commons}.
\newblock {Wikimedia Commons}{,} the free media repository, 2026.
\newblock URL \url{https://commons.wikimedia.org}.
\newblock [Online; accessed 15-October-2026].

\bibitem[Wu et~al.(2023)Wu, Hu, Luo, Wang, and Long]{wu2023solving}
H.~Wu, T.~Hu, H.~Luo, J.~Wang, and M.~Long.
\newblock Solving high-dimensional {PDE}s with latent spectral models.
\newblock In \emph{International Conference on Machine Learning}, 2023.

\bibitem[Wu et~al.(2024)Wu, Luo, Wang, Wang, and Long]{wu2024transolver}
H.~Wu, H.~Luo, H.~Wang, J.~Wang, and M.~Long.
\newblock Transolver: A fast {T}ransformer solver for {PDE}s on general geometries.
\newblock In \emph{International Conference on Machine Learning}, 2024.

\bibitem[Wu et~al.(2026)Wu, Guo, Li, Dou, Long, He, and Matusik]{wu2026geopt}
H.~Wu, M.~Guo, Z.~Li, Z.~Dou, M.~Long, K.~He, and W.~Matusik.
\newblock {GeoPT}: Scaling physics simulation via lifted geometric pre-training.
\newblock In \emph{International Conference on Machine Learning}, 2026.

\bibitem[Xu et~al.(2024)Xu, Zhou, and Bian]{xu2024self}
B.~Xu, Y.~Zhou, and X.~Bian.
\newblock Self-supervised learning based on {T}ransformer for flow reconstruction and prediction.
\newblock \emph{Physics of Fluids}, 36\penalty0 (2), 2024.
\newblock \doi{10.1063/5.0188998}.

\bibitem[Yin et~al.(2023)Yin, Kirchmeyer, Franceschi, Rakotomamonjy, and Gallinari]{yin2023continuous}
Y.~Yin, M.~Kirchmeyer, J.-Y. Franceschi, A.~Rakotomamonjy, and P.~Gallinari.
\newblock Continuous {PDE} dynamics forecasting with implicit neural representations.
\newblock In \emph{International Conference on Learning Representations}, 2023.

\bibitem[Zhang and Sennrich(2019)]{zhang2019root}
B.~Zhang and R.~Sennrich.
\newblock Root mean square layer normalization.
\newblock In \emph{Advances in Neural Information Processing Systems}, 2019.

\bibitem[Zhang et~al.(2026)Zhang, Wu, Zhang, and Wang]{zhang2026cheap}
Z.~Zhang, Y.~Wu, K.~Zhang, and Y.~Wang.
\newblock From cheap geometry to expensive physics: A physics-agnostic pretraining framework for neural operators.
\newblock In \emph{International Conference on Learning Representations}, 2026.

\end{thebibliography}

\onecolumn
\appendix
\renewcommand\thefigure{\thesection.\arabic{figure}}
\renewcommand\thetable{\thesection.\arabic{table}}
\captionsetup{labelsep=space}
\etocdepthtag.toc{mtappendix}
\begin{center}
  {\Huge {\bf Appendix}}
\end{center}
\phantomsection
\label{app:overview}

The appendix covers the partitioning and the local frame, the reconstructions of the pre-trained FAE, the geometry of its latent space, the operators learned on the frozen backbone with their benchmarks, and the architecture.

\begingroup
\etocsettagdepth{mtbody}{none}
\etocsettagdepth{mtappendix}{subsection}
\etocsetstyle{section}{}{\leavevmode\leftskip 0pt\rightskip 2em plus 1fil}
  {\vspace{0.25em}\textbf{\etocnumber\enspace\etocname}\hfill\textbf{\etocpage}\par}{}
\etocsetstyle{subsection}{}{\leavevmode\leftskip 1.2em}{\etocnumber\enspace\etocname\hfill\etocpage\par}{}
\etocsettocstyle{\paragraph{Contents.}\mbox{}\par\small\begin{multicols}{2}}{\end{multicols}}
\tableofcontents
\endgroup

\paragraph{Backbones.} Two backbones are trained on multi-mode anisotropic radial sines with random frequencies on the unit disk, drawn anew at every step. Depth counts the applied units, weight-shared repetitions included. The high-dimensional backbone serves every analysis (Section~\ref{sec:analysis}, Appendices~\ref{app:accuracy} and~\ref{app:latent}) and the low-dimensional one every operator (Section~\ref{sec:operator}, Appendix~\ref{app:operators}), being cheaper to train on (Appendix~\ref{app:caching-modes}) but needing more partitions for the same accuracy.

\vspace{-0.6em}
\begin{center}\small
  \captionof{table}{\textbf{The two backbones.} Both are trained only on synthetic radial sines, the high-dimensional one for the analysis and the low-dimensional one for the operators.}
  \renewcommand{\arraystretch}{1.05}\setlength{\tabcolsep}{8pt}
  \phantomsection\label{tab:backbones}
  \begin{tabular}{lcc}
    \toprule
     & high-dimensional & low-dimensional \\
    \midrule
    latent dimension & $128$ & $8$ \\
    encoder / decoder depth & $128$ / $16$ & $32$ / $4$ \\
    stored parameters & $4.83$\,M & $0.10$\,M \\
    training frequencies & up to $5$ & up to $1.5$ \\
    used for & analysis & operators \\
    \bottomrule
  \end{tabular}
\end{center}
\vspace{-0.6em}

\paragraph{Losses and metrics.} A point in several partitions takes the mean of their reconstructions. The FAE is trained on the \emph{physical loss}, the relative $L^1$ error of the reconstructed field per function and channel, averaged over a batch. A processor is trained either on the same physical loss, evaluated on its decoded prediction, or on the \emph{latent loss}, which compares its predicted embeddings with the encoded targets without decoding. The latent loss is the weighted mean $(\mathcal{L}_{\mathrm{m}}+w_{\mathrm{s}}\,\mathcal{L}_{\mathrm{s}})/(1+w_{\mathrm{s}})$ of a morphological term $\mathcal{L}_{\mathrm{m}}$, the cosine distance $1-\cos(\hatembmorph,\embmorph)$ between the predicted and the target morphologic embedding of every partition and channel, and a static term $\mathcal{L}_{\mathrm{s}}$, a logarithmic relative error $\log_2(1+\lVert\hatembstat-\embstat\rVert_2/\lVert\embstat\rVert_2)$ of the predicted static features $\embstat$ of a function, of which the processor predicts the value frames, with $w_{\mathrm{s}}=0.3$. Reconstructions are scored by the relative $L^2$ error, and photographs by the peak signal-to-noise ratio. Operators are scored on the raw fields by the relative $L^1$ error of every output channel, averaged over the channels, and a table reports its median over the test samples.

\FloatBarrier
\clearpage

\section{Partitioning}
\label{app:partnorm}
\setcounter{figure}{0}
\setcounter{table}{0}

An input is given as a point cloud that samples a function on its domain, and the encoder receives it as a set of partitions (Figure~\ref{fig:teaser}b,c). A $k$-means partitioner decomposes the domain by clustering the points \citep{lloyd1982least}, and every partition is mapped into a normalized local frame in which it is encoded and from which it is decoded. This appendix describes the partitioner (Appendix~\ref{app:partitioner}), gives the frame explicitly (Appendix~\ref{app:partnorm-frame}), shows the equivariances the frame confers on the encode-decode map (Appendix~\ref{app:equivariance}), and defines the effective spatial frequency, which measures how hard a normalized partition is to reconstruct (Appendix~\ref{app:freq}).

\FloatBarrier
\subsection{The partitioner}
\label{app:partitioner}

\begin{figure}[!htbp]\centering
  \includegraphics[width=\textwidth]{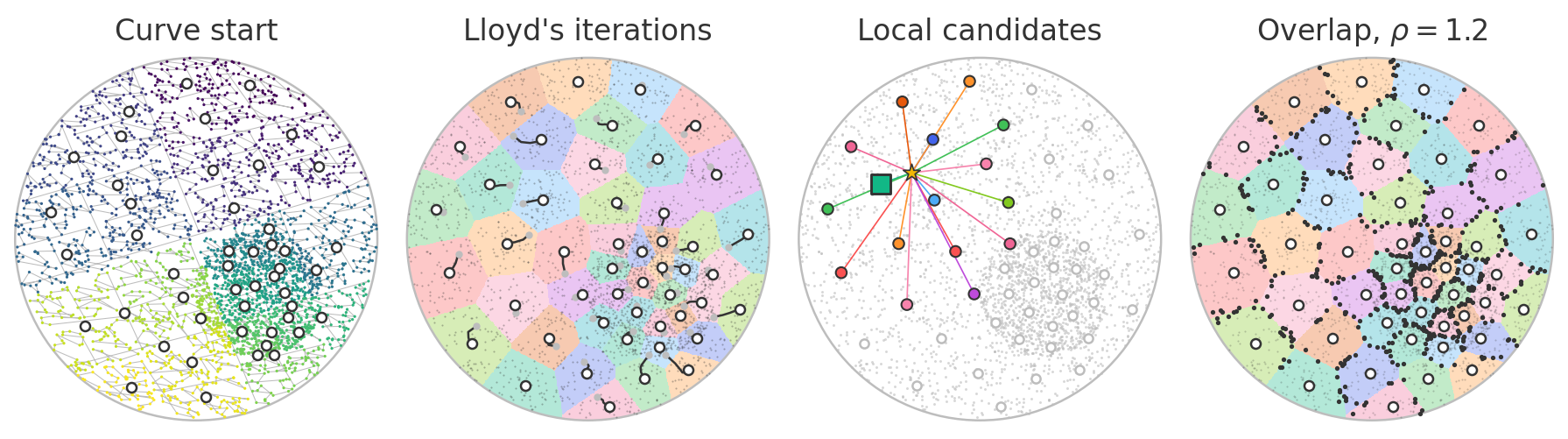}
  \caption{\textbf{The partitioner.} From left to right, the points of a non-uniform cloud ordered along a Z-order curve after a random rotation and colored by their rank on it, with the initial centers, the means of equal runs of consecutive points; the partitions after five of Lloyd's iterations, with the path every center took from its start; the local search of one point (star), which is compared only with the 16 centers nearest to its current center (square); and the partitions at an overlap factor $\rho=1.2$, whose further assignments (dark points), the smallest distance ratios of Equation~\eqref{eq:overlap}, lie along the partition boundaries. An illustration on 3000 points and 48 partitions.}
  \label{fig:partitioning}
\end{figure}

The partitions are the clusters of a $k$-means clustering of the coordinates, computed anew for every input and independently of the values, so the partitioner has to stay cheap on dense clouds and at large partition counts. A balanced start, local iterations, and chunked distances keep it so (Figure~\ref{fig:partitioning}).

Lloyd's algorithm \citep{lloyd1982least} starts from the means of $P$ equal runs of the points along a space-filling curve. The points are rotated by a random angle, quantized on a grid, and sorted by the Morton code that interleaves the bits of their two grid indices, which orders them along a Z-order curve, and each run of $N/P$ consecutive points gives one initial center. These centers already follow the density of the cloud and the clusters they induce are balanced, so a few iterations reach what many do from random points. Within an iteration a point is compared only with the 16 centers nearest to its current one, read from a table of neighboring centers built when the iterations begin, rather than with all $P$ centers, since the centers move little from one iteration to the next, and the cost of an iteration therefore does not grow with the partition count. At inference we run five such iterations.

The final assignment uses a table rebuilt from the final centers and walks every point from its last center to ever nearer neighboring centers until none is nearer, which on every cloud we checked finds the exact nearest center. The center of a partition is then recomputed once as the mean of the points nearest to it. Distances are evaluated over chunks of at most $2^{24}$ point-to-center pairs, so the full $N\times P$ distance matrix is never formed and the memory of the partitioner grows linearly with the number of points. On very large clouds the clustering can also be fitted on a random subsample of the points, after which the remaining points walk to their centers from a first guess along the curve.

By default every point belongs to exactly one partition, the one whose center is nearest, so there are $N$ point-to-partition assignments. An overlap factor $\rho\ge1$ raises their number to $N_{\mathrm{asg}}=\lfloor\rho N\rfloor$. Every point keeps its own partition, and the $N_{\mathrm{asg}}-N$ further assignments go to the pairs of a point $i$ and a partition $p$ with the smallest distance ratio
\begin{equation}
  r_{ip} = \frac{\lVert\mathbf{x}_i-\mathbf{c}_p\rVert^2}{\min_{q}\lVert\mathbf{x}_i-\mathbf{c}_q\rVert^2}\;\ge\;1,
  \label{eq:overlap}
\end{equation}
ranked over the whole cloud. A ratio near one marks a point that lies almost as close to the center of a neighboring partition as to its own, so the surplus extends every partition across its boundary into its neighbors. Because the ranking is global rather than per partition, the partitions grow unevenly, and the further assignments go where partitions are crowded and therefore small, which evens out the partition sizes. The candidates of a point are the 16 partitions whose centers lie nearest to its own center, and only its 8 best-ranked candidates enter the global ranking, so the ranking never scores all $NP$ pairs. A point that lies in several partitions is decoded by each, and their estimates are averaged (Equation~\eqref{eq:partnorm-decode}). Inference uses $\rho=1.2$, which adds a fifth to the number of assignments.

\FloatBarrier
\subsection{PartNorm}
\label{app:partnorm-frame}

Each partition is mapped into a canonical local frame before it is encoded (Figure~\ref{fig:teaser}c). The coordinates are shifted so the partition center is at the origin and scaled by the per-axis extent into a symmetric box, and the function values are min-max normalized within the partition. The five resulting quantities, the center $\mathbf{c}_p$, the axis scales $\mathbf{s}_p$, the orientation angle $\theta_p$, the value shift $\mu_p$, and the value scale $\sigma_p$, are retained as the \emph{static} features $\embstat$ and replayed by the decoder to invert the frame exactly, so the learned embedding $\embmorph$ is left to describe only the normalized \emph{morphology} of the partition.

\begin{figure}[!t]\centering
  \includegraphics[width=\textwidth]{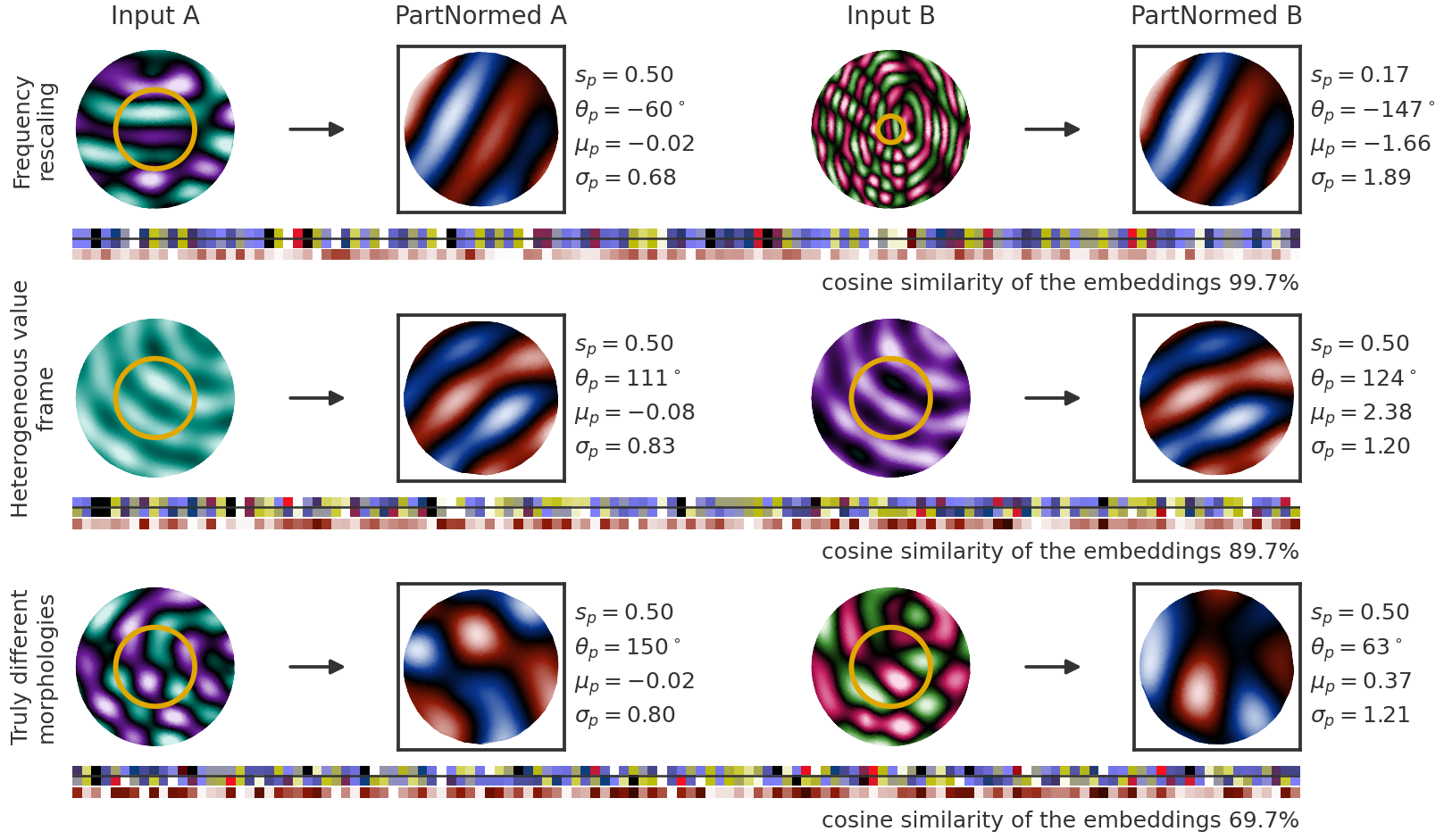}
  \caption{\textbf{PartNorm on three pairs of inputs.} Each row follows two inputs A and B, with their partitions circled, into the canonical frame (boxed), and lists the static features of every normalized partition, namely the largest component $s_p$ of its axis scales, the angle $\theta_p$ of the canonical rotation, the value shift $\mu_p$, and the value scale $\sigma_p$. The strip under each row shows the morphologic embeddings of A (top) and B (middle) and the magnitude of their difference (bottom), with their cosine similarity.}
  \label{fig:partnorm-cases}
\end{figure}

This local frame is the reason a model trained on one narrow synthetic distribution reconstructs fields it was never shown. Normalization removes the location, size, and amplitude of a partition, mapping structurally similar partitions to nearly identical inputs regardless of their global context, so their learned embeddings coincide (Figure~\ref{fig:partnorm-cases}). In particular a high-frequency field observed in a small partition can normalize to nearly the same input as a low-frequency field in a large partition, and the encoder returns nearly the same embedding, since within a partition absolute scale is immaterial and only the number of oscillations across the partition remains. The model therefore does not need to have seen a field of a given physical type, only partitions of comparable normalized morphology, which the synthetic distribution supplies in abundance. This is exactly what the effective spatial frequency of Appendix~\ref{app:freq} measures, and it is why an out-of-distribution input is decomposed into in-distribution partitions rather than extrapolated to.

Let $\mathcal{I}_p$ be the points assigned to partition $p$, let $\mathbf{c}_p$ be its center, the mean of the points nearest to it, and write $\funcMain_i=\funcMain(\mathbf{x}_i)$. The coordinate frame shifts the points to the center and scales every axis by the largest distance from it,
\begin{equation}
  \tilde{\mathbf{x}}_i = (\mathbf{x}_i-\mathbf{c}_p)\oslash\mathbf{s}_p,
  \qquad
  s_{p,d} = \max_{i\in\mathcal{I}_p}\,\lvert x_{i,d}-c_{p,d}\rvert,
  \label{eq:partnorm-x}
\end{equation}
with $\oslash$ the elementwise division, so that every partition fills the box $[-1,1]^2$ whatever its size and shape. The value frame maps the values of the partition onto $[-1,1]$ through their mid-range and half-range,
\begin{equation}
  \tilde{\funcMain}_i = \frac{\funcMain_i-\mu_p}{\sigma_p},
  \qquad
  \mu_p = \tfrac12\Big(\max_{i\in\mathcal{I}_p}\funcMain_i+\min_{i\in\mathcal{I}_p}\funcMain_i\Big),
  \qquad
  \sigma_p = \max\Big\{\tfrac12\Big(\max_{i\in\mathcal{I}_p}\funcMain_i-\min_{i\in\mathcal{I}_p}\funcMain_i\Big),\,\smin\Big\},
  \label{eq:partnorm-u}
\end{equation}
where the floor $\smin=10^{-8}$ keeps the scale positive. The encoder reads the pairs $(\tilde{\mathbf{x}}_i,\tilde{\funcMain}_i)$ of a partition and returns its morphologic embedding $\emb_{\mathrm{m},p}$, and the frame is kept beside it as the static features $\emb_{\mathrm{s},p}=(\mathbf{c}_p,\mathbf{s}_p,\theta_p,\mu_p,\log\sigma_p)$, the scale stored as its logarithm. The decoder $\decoder$ evaluates the normalized field at the frame coordinates of a query point and maps it back through the value frame, and a point that lies in several partitions averages their estimates,
\begin{equation}
  \funcMain(\mathbf{x}) \approx \frac{1}{\lvert\mathcal{A}(\mathbf{x})\rvert}\sum_{p\in\mathcal{A}(\mathbf{x})}\Big(\mu_p+\sigma_p\,\decoder\big(\emb_{\mathrm{m},p},\,(\mathbf{x}-\mathbf{c}_p)\oslash\mathbf{s}_p\big)\Big),
  \label{eq:partnorm-decode}
\end{equation}
with $\mathcal{A}(\mathbf{x})$ the partitions that contain $\mathbf{x}$. Since the frame is replayed exactly, the error of encoding and decoding a partition is the error of the normalized decode multiplied by $\sigma_p$.

Two degenerate partitions are guarded separately. An empty partition has no extrema, and its frame is set to $\mu_p=0$ and $\sigma_p=1$. A constant partition has a well-defined mid-range but no range, and only its scale is raised to the floor, so the decoded value is $\mu_p$ up to a term of order $\smin$ and the constant is reproduced to floating-point precision whatever the network returns.

The encoder can also rotate the coordinates of every partition into a canonical orientation, by an angle $\theta_p$ read from moments of the normalized partition weighted by its values. The dipole $\mathbf{m}_p=\sum_{i\in\mathcal{I}_p}\tilde{\funcMain}_i\tilde{\mathbf{x}}_i$ gives a full direction, $\theta_p=\operatorname{atan}(m_{p,2}/m_{p,1})$ taken in the quadrant of $\mathbf{m}_p$, wherever its length is at least $5\%$ of $\sum_{i}\lvert\tilde{\funcMain}_i\rvert$ times the radius of the partition. Where it is weaker, the quadrupole $\sum_{i}\tilde{\funcMain}_i\,\big(\tilde x_{i,1}^2-\tilde x_{i,2}^2,\;2\tilde x_{i,1}\tilde x_{i,2}\big)$ gives the axis of elongation as half its polar angle, with the sign of the axis set by the dipole, and a partition with neither moment above the threshold is left unrotated. The decoder replays $\theta_p$ like the rest of the frame, so encoding followed by decoding would remain exact. The trained backbones nonetheless do not apply this rotation, and $\theta_p=0$ throughout. We found the encode-decode map approximately equivariant to rotations without it (Appendix~\ref{app:equivariance}), and therefore keep the orientation of a partition in its morphological embedding rather than moving it into the static features.

\paragraph{Different inputs, one normalized partition.}
\label{app:partnorm-cases}
Figure~\ref{fig:partnorm-cases} follows three pairs of inputs through the frame, every partition encoded as a single partition by the analysis backbone with the optional rotation switched on. In the first pair, input B oscillates three times faster than input A and is rotated, rescaled, and shifted in its values, and its partition is three times smaller. The two normalized partitions coincide, and so do their embeddings, with a cosine similarity of $99.7\%$. In the second pair the two inputs are one function on partitions of the same size, without a rotation between them, but input B is input A scaled by $1.5$ and shifted by $2.5$, so that on one color scale the two inputs take its opposite ends, with a value scale and a value shift that also vary mildly across the partition. The frame removes one value shift and one value scale per partition, so the difference of the two frames disappears, while the mild variation across the partition survives, and the normalized partitions agree closely but not exactly, with a cosine similarity of $89.7\%$. In the third pair the inputs are two independent functions of truly different morphologies, and their normalized partitions and embeddings stay different, with a cosine similarity of $69.7\%$. Normalization therefore identifies partitions that differ only by the transformations it removes, and keeps apart partitions that differ in their morphology.

\FloatBarrier
\subsection{Equivariance}
\label{app:equivariance}

The frame makes the encode-decode map, which encodes a field and decodes it again, equivariant to the transformations it normalizes away. A translation or a dilation of the coordinates moves the centers and the extents of the partitions with the points, $\mathbf{c}_p\mapsto s\,\mathbf{c}_p+\mathbf{c}$ and $\mathbf{s}_p\mapsto s\,\mathbf{s}_p$ for a translation by $\mathbf{c}$ and a dilation by $s$, and leaves the normalized coordinates of Equation~\eqref{eq:partnorm-x} unchanged. A shift or a positive scale of the values moves the mid-range and the half-range, $\mu_p\mapsto\sigma\mu_p+\mu$ and $\sigma_p\mapsto\sigma\sigma_p$ for a value scale $\sigma>0$ and a value shift $\mu$, and leaves the normalized values of Equation~\eqref{eq:partnorm-u} unchanged. The encoder reads only the normalized pairs, so the morphological embedding $\emb_{\mathrm{m},p}$ is invariant to all four transformations, while the static features take them up, and the decoder, which replays the static features (Equation~\eqref{eq:partnorm-decode}), returns the transformed reconstruction. The encode-decode map is therefore equivariant to these transformations, since the embedding records where a partition lies, how large it is, and how large its values are in the static features, and what the partition looks like in the morphological embedding.

\begin{figure}[!htbp]\centering
  \includegraphics[width=\textwidth]{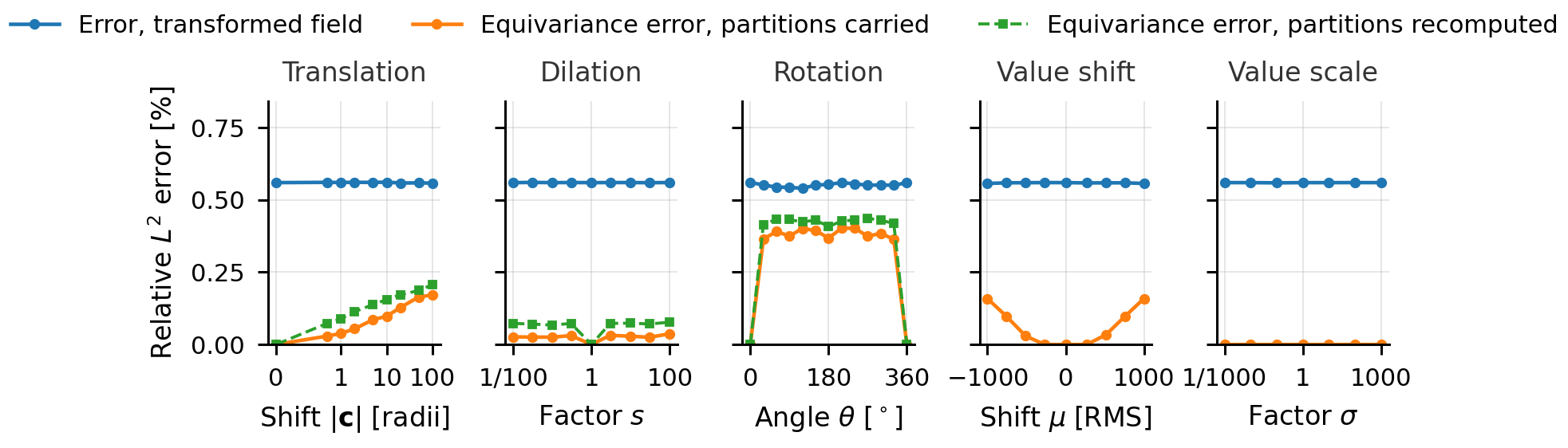}
  \caption{\textbf{Equivariance of the encode-decode map.} For each transformation, a translation by $\mathbf{c}$, a dilation by $s$, a rotation by $\theta$, a value scale $\sigma$, and a value shift $\mu$, the relative $L^2$ error of the reconstructed transformed field, and the equivariance error, the relative $L^2$ distance between the reconstruction of the transformed field and the transformed reconstruction, with the partitions carried along with the transformation and recomputed on the transformed cloud. Analysis backbone, 16 functions of the evaluation preset Sines (eval), three radial-sine modes with frequencies $f\in[1,4]$, at 8192 points and 64 partitions.}
  \label{fig:equivariance}
\end{figure}

Figure~\ref{fig:equivariance} measures this on the analysis backbone, for 16 functions of the evaluation preset Sines (eval), three radial-sine modes with frequencies $f\in[1,4]$, at 8192 points and 64 partitions, over translations up to 100 disk radii, dilations between $1/100$ and $100$, value shifts up to $\pm1000$ times the root mean square of the field, and value scales between $1/1000$ and $1000$. Two quantities are recorded. The first is the error of the transformed field, the relative $L^2$ distance between the reconstruction of a transformed field and that field itself, which stays constant if the transformation does not affect the accuracy. The second is the \emph{equivariance error}, the relative $L^2$ distance between two fields that coincide for an exactly equivariant map, namely the reconstruction of the transformed field and the transformed reconstruction of the original field. Over all four ranges the error of the transformed field does not change, and the equivariance error is negligible, well below the reconstruction error, whether the partitions are carried along with the transformation or recomputed on the transformed cloud.

Rotation is the exception, since the trained backbones do not rotate the partitions into a canonical orientation (Appendix~\ref{app:partnorm-frame}). The morphological embeddings then change with the angle, yet over a full turn the error of the rotated field does not change, and the equivariance error stays below the reconstruction error. The encode-decode map is thus equivariant to rotations up to the reconstruction error, with the orientation of every partition carried in its morphological embedding.

\subsection{Effective spatial frequency}
\label{app:freq}

That fields of unrelated type transfer comparably is explained by measuring difficulty in the normalized frame rather than in the physical one. The coordinate frame of a partition $p$ (Equation~\eqref{eq:partnorm-x}) maps its region $\Omega_p$ onto a region $\tilde\Omega_p\subset[-1,1]^2$, and by the chain rule the gradient and the Hessian of the field with respect to the normalized coordinates are $\nabla_{\tilde{\mathbf{x}}}\funcMain=\mathbf{s}_p\odot\nabla\funcMain$ and $\nabla_{\tilde{\mathbf{x}}}^2\funcMain=(\mathbf{s}_p\mathbf{s}_p^{\!\top})\odot\nabla^2\funcMain$, with $\nabla^2\funcMain$ the Hessian of $\funcMain$ and $\odot$ the elementwise product. We define the effective spatial frequency of the partition as
\begin{equation}
  \feffp{p} \;=\; \frac{1}{2\pi}\left(\frac{\int_{\tilde\Omega_p}\lVert\nabla_{\tilde{\mathbf{x}}}^2\funcMain(\tilde{\mathbf{x}})\rVert_F^2\,d\tilde{\mathbf{x}}}{\int_{\tilde\Omega_p}\lVert\nabla_{\tilde{\mathbf{x}}}\funcMain(\tilde{\mathbf{x}})\rVert_2^2\,d\tilde{\mathbf{x}}}\right)^{1/2},
  \label{eq:feff}
\end{equation}
a Rayleigh quotient of the second derivatives over the first that measures how quickly the slope of the normalized field changes across the partition. The value frame $(\mu_p,\sigma_p)$ of Equation~\eqref{eq:partnorm-u} drops out, since the shift has no derivative and the scale multiplies both integrals by $\sigma_p^2$, so $\feffp{p}$ is a property of the normalized partition the encoder reads, and a partition on which the field is affine, a plain slope, has frequency zero. The normalization by $2\pi$ makes the quotient recover the frequency of a single sine. On a single partition covering the unit disk, which is its own normalized frame ($\mathbf{c}_p=\mathbf{0}$, $\mathbf{s}_p=\mathbf{1}$), the sine $\funcMain(\mathbf{x})=a\sin\phi(\mathbf{x})$ of frequency $f$, with phase $\phi(\mathbf{x})=2\pi f\,\mathbf{n}^{\!\top}\mathbf{x}+\phi_0$ along a unit direction $\mathbf{n}$, has $\nabla\funcMain=2\pi fa\cos\phi\,\mathbf{n}$ and $\nabla^2\funcMain=-(2\pi f)^2a\sin\phi\,\mathbf{n}\mathbf{n}^{\!\top}$, and since $\lVert\mathbf{n}\mathbf{n}^{\!\top}\rVert_F=1$,
\begin{equation}
  \feffp{p} \;=\; f\left(\frac{\int_{\Omega_p}\sin^2\phi\,d\mathbf{x}}{\int_{\Omega_p}\cos^2\phi\,d\mathbf{x}}\right)^{1/2}.
\end{equation}
Over a region that holds whole periods along $\mathbf{n}$ the two integrals are equal and $\feffp{p}=f$ exactly.

A field decomposed into partitions has one frequency per partition, whereas its reconstruction error is measured on the whole field, and relating the two requires a single frequency of the partitioned field. We weight the partitions by their share of the variation of the field. For disjoint partitions $\Omega_1,\dots,\Omega_P$ of the domain $\Omega$, with $\bar\funcMain$ the mean of the field over $\Omega$ and $\bar\funcMain_p$ its mean over $\Omega_p$, the energy of the centered field splits exactly into the variation inside the partitions and the variation of their means,
\begin{equation}
  \lVert\funcMain-\bar\funcMain\rVert_{L^2(\Omega)}^2
  \;=\; \sum_{p=1}^{P}\lVert\funcMain-\bar\funcMain_p\rVert_{L^2(\Omega_p)}^2
  \;+\; \sum_{p=1}^{P}\lvert\Omega_p\rvert\,\big(\bar\funcMain_p-\bar\funcMain\big)^2,
  \label{eq:feff-split}
\end{equation}
since $\funcMain-\bar\funcMain_p$ integrates to zero over $\Omega_p$ and the cross terms vanish. The first sum is the variation the embeddings have to represent, and the second the variation the frames carry through their value shifts. With the weights $\omega_p=\lVert\funcMain-\bar\funcMain_p\rVert_{L^2(\Omega_p)}^2/\lVert\funcMain-\bar\funcMain\rVert_{L^2(\Omega)}^2$, which sum to one less the share $\omega_0$ of the second term, the effective spatial frequency of the partitioned field is the root mean square
\begin{equation}
  \feff \;=\; \Big(\sum_{p=1}^{P}\omega_p\,\feffp{p}^2\;+\;\omega_0\cdot 0\Big)^{1/2},
  \label{eq:feff-field}
\end{equation}
in which the share of the frames counts at frequency zero. On a single partition $\omega_1=1$ and $\feff=\feffp{1}$.

These weights are the ones with which the partitions enter the error of the whole field. Writing $\hat\funcMain$ for the reconstruction and $e_p=\lVert\hat\funcMain-\funcMain\rVert_{L^2(\Omega_p)}/\lVert\funcMain-\bar\funcMain_p\rVert_{L^2(\Omega_p)}$ for the relative error of partition $p$, the relative error of the centered field obeys
\begin{equation}
  \frac{\lVert\hat\funcMain-\funcMain\rVert_{L^2(\Omega)}^2}{\lVert\funcMain-\bar\funcMain\rVert_{L^2(\Omega)}^2}
  \;=\; \sum_{p=1}^{P}\omega_p\,e_p^2
\end{equation}
exactly when the partitions are disjoint. Since the frame is replayed exactly, the reconstruction error of a partition is $\sigma_p$ times the error of its normalized decode (Appendix~\ref{app:partnorm-frame}), so $e_p$ does not depend on the value frame and is, like $\feffp{p}$, a property of the normalized partition, while the variation between the partition means is reproduced by the frames and adds no error. Equation~\eqref{eq:feff-field} averages the frequencies of the partitions with the same weights and in the same squares as the error averages their errors, and counts the share of the frames where it contributes to the error, at zero. A plain mean or median over the partitions would instead weigh a nearly flat partition, which contributes almost nothing to the error, as heavily as one that holds most of the variation of the field, and would ignore that a field whose variation lies mostly between its partitions is reproduced by the frames.

\begin{figure}[!hbtp]\centering
  \begin{subfigure}{0.6\textwidth}\centering
    \includegraphics[width=\textwidth]{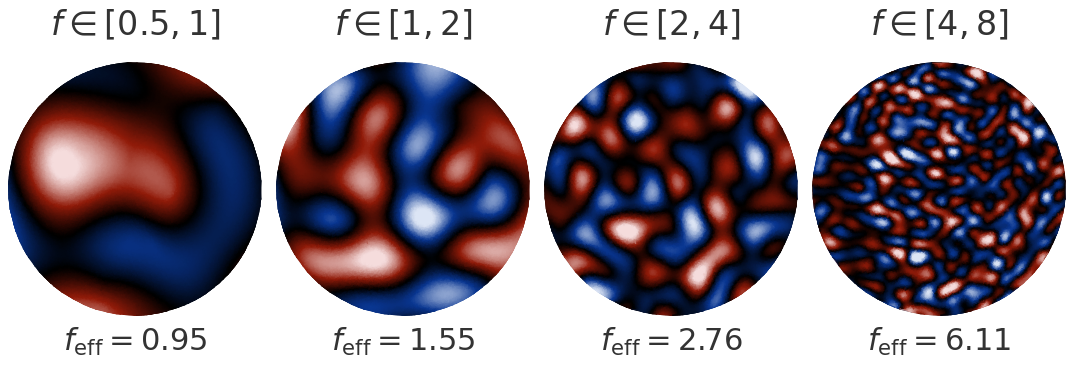}
    \caption{Each field as one partition.}
  \end{subfigure}\\[0.4em]
  \begin{subfigure}{0.6\textwidth}\centering
    \includegraphics[width=\textwidth]{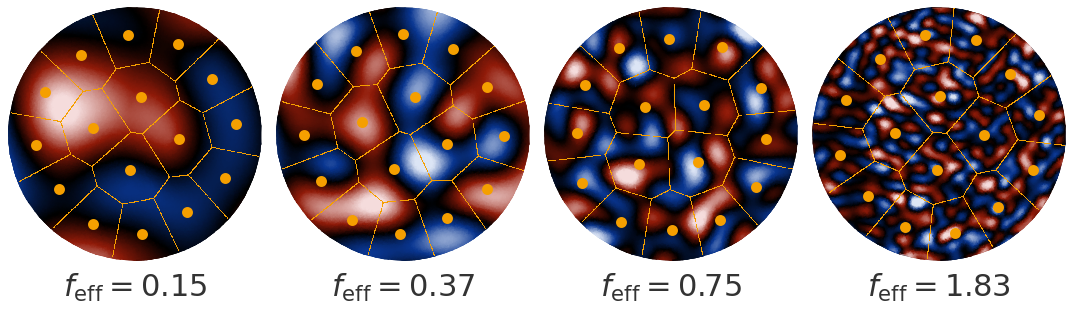}
    \caption{The same fields cut into 16 partitions.}
  \end{subfigure}
  \caption{\textbf{Effective spatial frequency.} Four radial-sine fields of 25 modes on the unit disk, the modes drawn from the frequency range above each column, annotated with their effective spatial frequency $\feff$ (Equation~\eqref{eq:feff-field}). (a) The whole disk as a single partition. (b) The same fields cut into 16 partitions by the inference partitioner without overlap (orange boundaries and centers). Every field has a lower effective frequency in (b), since each partition holds fewer of its oscillations and part of the variation of the field is carried by the frames.}
  \label{fig:feff}
\end{figure}

Figure~\ref{fig:feff} shows the quantity on four radial-sine fields of 25 modes each. On the whole disk it rises with the frequency range of the modes, from $0.95$ to $6.11$, and cut into 16 partitions the same fields fall to between $0.15$ and $1.83$. For one and the same function, more partitions thus shrink every partition and lower its effective frequency, as the nested partitions of Figure~\ref{fig:hero} also show, where a field of effective frequency $7.83$ falls to $2.53$ and then $0.85$ on two successively smaller regions. Plotting reconstruction error against $\feff$ for several synthetic families, each a fixed set of functions decomposed into more and more partitions at a fixed number of points per partition so that frequency is isolated from undersampling, collapses the families onto one trend (Figure~\ref{fig:freq}). Difficulty is therefore a property of normalized morphology and not of function class, and the partitioning acts on it directly, since more partitions lower $\feff$ and improve accuracy (Appendix~\ref{app:sweep}). Out-of-distribution transfer is, in these terms, not generalization but the reduction of an unfamiliar field to partitions whose effective frequency the model has seen.

\begin{figure}[!hbtp]\centering
  \includegraphics[width=\textwidth]{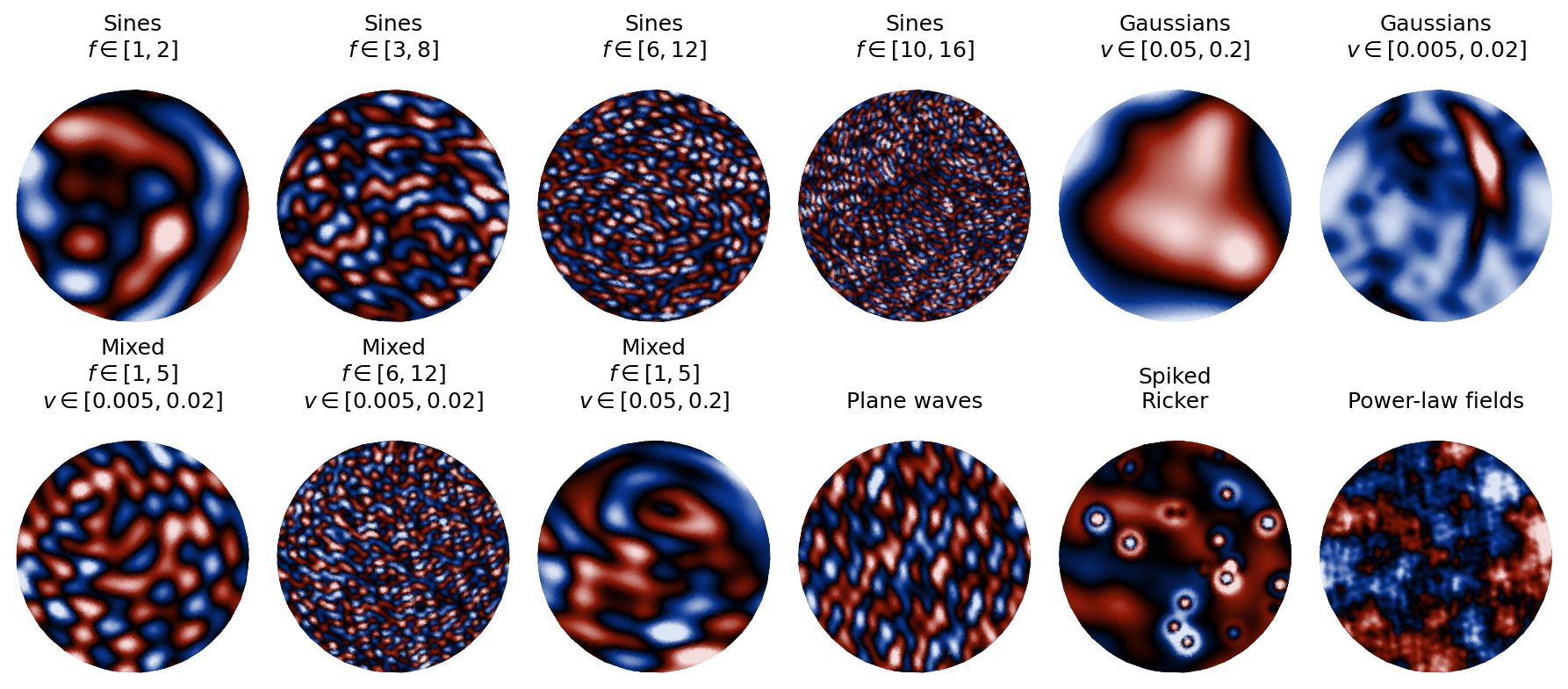}
  \caption{\textbf{The synthetic families of Figure~\ref{fig:freq}.} One function of every family, the first of the 64 that the figure uses, on the unit disk, each on its own color scale. The labels give the frequency range $f$ of the modes or the variance range $v$ of the Gaussian bumps.}
  \label{fig:freq-families}
\end{figure}

Figure~\ref{fig:freq} is obtained with the analysis backbone as follows. It covers twelve synthetic families, shown by one sample each in Figure~\ref{fig:freq-families}, namely radial sines of 25 modes with frequencies in $[1,2]$, $[3,8]$, $[6,12]$, and $[10,16]$, Gaussian bumps of 100 modes with variances in $[0.05,0.2]$ and $[0.005,0.02]$, three equal mixtures of Gaussian bumps with radial sines of 25 modes, the narrower bumps with sines of frequencies in $[1,5]$ or $[6,12]$ and the wider bumps with sines of frequencies in $[1,5]$, plane waves of 10 modes with frequencies in $[1,8]$, spiked fields of 20 Ricker spikes of width $0.05$ on radial sines with frequencies in $[1,3]$, and power-law fields of 200 modes with spectral exponent $3$ between the wavenumbers $0.5$ and $16$. Every family contributes a fixed set of 64 functions, and every function is sampled once, at $32{,}768$ points drawn uniformly on the unit disk. For $P\in\{1,2,4,\dots,128\}$ partitions the function is reconstructed from the first $256P$ of its points, which are again a uniform sample of the disk, so that a partition holds $256$ points on average at every partition count. The inference partitioner runs without overlap, which makes the partitions disjoint, as Equation~\eqref{eq:feff-split} requires. All families are given in closed form, and the gradients and Hessians entering Equation~\eqref{eq:feff} are those of the closed form, evaluated exactly at the points, with the integrals of Equations~\eqref{eq:feff} to \eqref{eq:feff-field} evaluated as sums over the points of each partition. The radial sines have a cusp at the center of every mode, where the second derivatives grow as the inverse distance and the integral of their square diverges logarithmically. The sums over a finite sample stay finite, and removing the points within $0.02$ of every mode center lowers the median effective frequency of the radial sines with frequencies in $[1,2]$ by $3\%$, and that of the higher ranges by less than $1\%$. A point of a curve is the median over the 64 functions of the relative $L^2$ error of the centered field, $\lVert\hat\funcMain-\funcMain\rVert_{L^2(\Omega)}/\lVert\funcMain-\bar\funcMain\rVert_{L^2(\Omega)}$, against the median of their effective frequency $\feff$, both taken on the same points and partitions. This figure alone measures the error relative to the centered field, since the frames reproduce a constant offset of a field exactly, and a field with a large offset would otherwise read as accurate whatever its shape. The gray bands along the frequency axis give the density of the effective frequency of $1024$ functions of the training distribution of the analysis backbone as seen by it during training.

The same statement read in the other direction gives the limitation. What bounds the representation is a frequency \emph{band}, fixed by the training distribution. More partitions pull the effective frequency of every partition down into that band, and this is what recovers a high-frequency field, but only as long as each partition keeps enough points to be represented (Appendix~\ref{app:sweep-sparsity}). A field that cannot be partitioned finely enough for its partitions to reach the band, because its sampling is too sparse, is not recovered. \citet{fanaskov2023spectral} report the corresponding effect for operators, finding that a neural operator learns only the frequencies present in its training data. The collapse in Figure~\ref{fig:freq} locates that effect one level down, in the representation itself, which suggests that part of what is read as an operator failure is in fact a failure to represent the input.

\FloatBarrier
\clearpage

\section{Generalization}
\label{app:accuracy}
\setcounter{figure}{0}
\setcounter{table}{0}

This appendix collects the reconstruction results of the analysis backbone. The first part reconstructs synthetic, physical, and natural fields that the model never saw in training, including functions observed on curves and on parts of their domain, the second measures how the error depends on the number of partitions and on the resolution of the input, and the third examines the spatial derivatives of decoded fields.

\FloatBarrier
\subsection{Reconstruction and out-of-distribution transfer}
\label{app:recon}

\begin{figure}[!htbp]\centering
  \includegraphics[width=\textwidth]{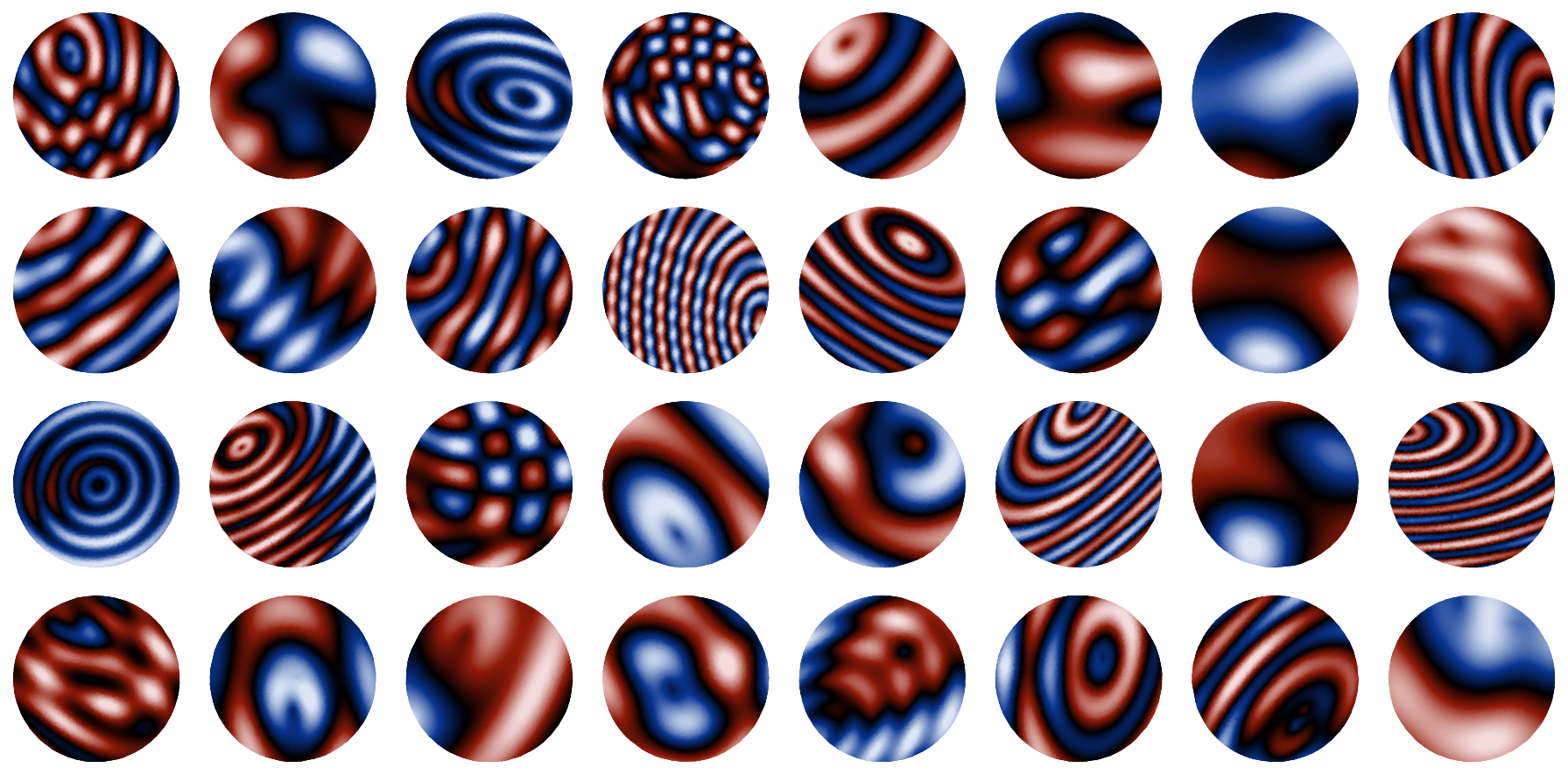}
  \caption{\textbf{Samples of the training distribution.} Independent draws of the training preset of the analysis backbone, radial sines of three modes on the unit disk, the only functions and the only domain the backbone is trained on. The frequencies are comparatively low and each function carries only a few modes.}
  \label{fig:train-dist}
\end{figure}

The analysis backbone is trained only on radial sines on the unit disk, sums of three anisotropic elliptical wavefronts with random frequencies between $0.2$ and $5$, drawn anew at every step (Figure~\ref{fig:train-dist}). These are the only functions and the only domain the model has seen. Every other field in this appendix, from the higher-frequency and spiked synthetic functions to the PDE solutions on other domains and the natural images, lies outside this distribution, and the samples of Figure~\ref{fig:train-dist} set it apart from the fields the model transfers to.

\subsubsection{Unseen distributions and geometries}
\label{app:recon-gallery}
\label{app:recon-fields}

The informative test of the representation is not accuracy on the distribution it was fitted to but transfer to fields of a different origin. On held-out synthetic functions the reconstruction is near-exact for low-frequency sines and Gaussian bumps and degrades gracefully for high-frequency and spiked inputs. The same frozen model reconstructs hyperelastic displacement and Cauchy-stress fields, turbulent and shear-layer flow, compressible-flow density with shocks, and acoustic wave fields, each on geometries and samplings absent from training. Figure~\ref{fig:gallery} places ten fields from nine physics datasets next to their reconstructions. Seven are fields on regular grids, flow and wave fields sampled on a uniform grid of pixels, each taken at the middle snapshot of a trajectory and encoded with $1024$ partitions, namely compressible-Euler density fields with shocks and shear layers from five initial conditions, an acoustic wave in a layered medium, and an incompressible-flow velocity. The other three are fields on point clouds, fields of elastic bodies sampled at the irregularly placed nodes of domains with holes, the displacement of a hyperelastic body on a disk-like domain and a Cauchy stress of another, and the shear stress of a linear-elastic body on a square domain with two holes, encoded with $128$ partitions. Trained only on analytic functions on a disk, the model recovers fine texture and not merely the coarse layout, on domains, samplings, and content with no counterpart in training.

\subsubsection{Natural images}
\label{app:recon-images}

\begin{figure}[!htbp]\centering
  \includegraphics[width=\textwidth]{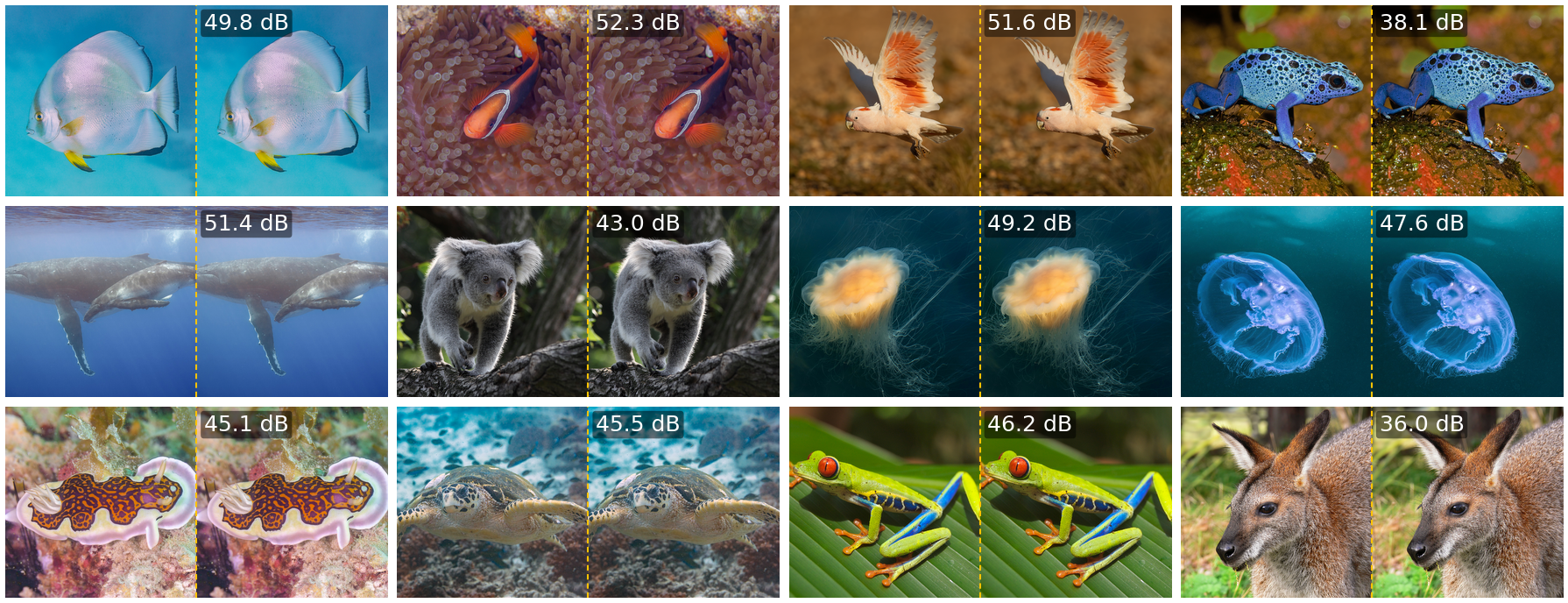}
  \caption{\textbf{Natural images, animals.} Every photograph, cut to its central square, is followed directly to its right by its \daedalus{} reconstruction with $4096$ partitions per color channel, whose PSNR is printed in its top-left corner. Photographs from Wikimedia Commons~\citep{wikimediacommons2026}, credited individually on the project website~\citep{daedaluscredits2026}.}
  \label{fig:img-animals}
\end{figure}

\begin{figure}[!htbp]\centering
  \includegraphics[width=\textwidth]{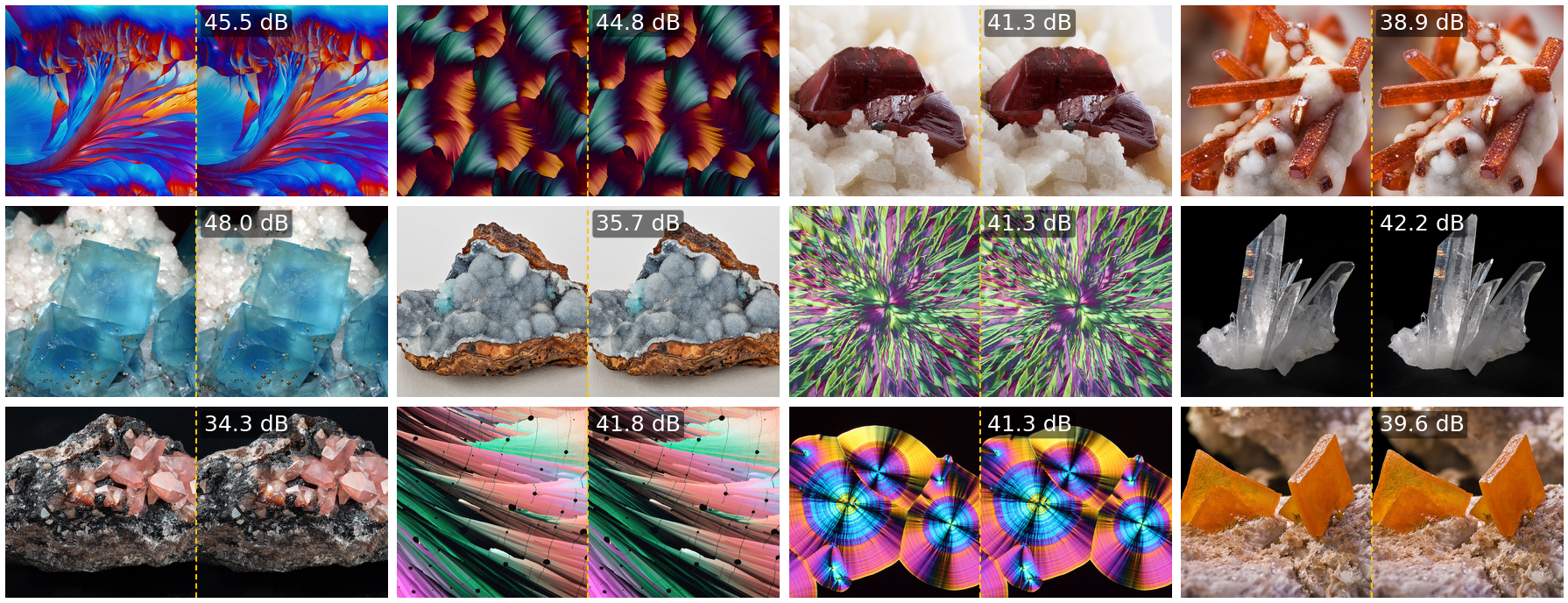}
  \caption{\textbf{Natural images, minerals and crystals.} As in Figure~\ref{fig:img-animals}, every photograph is followed directly to its right by its reconstruction with $4096$ partitions per color channel, with the PSNR in its top-left corner. Photographs from Wikimedia Commons~\citep{wikimediacommons2026}, credited individually on the project website~\citep{daedaluscredits2026}.}
  \label{fig:img-minerals}
\end{figure}

\begin{figure}[!htbp]\centering
  \includegraphics[width=\textwidth]{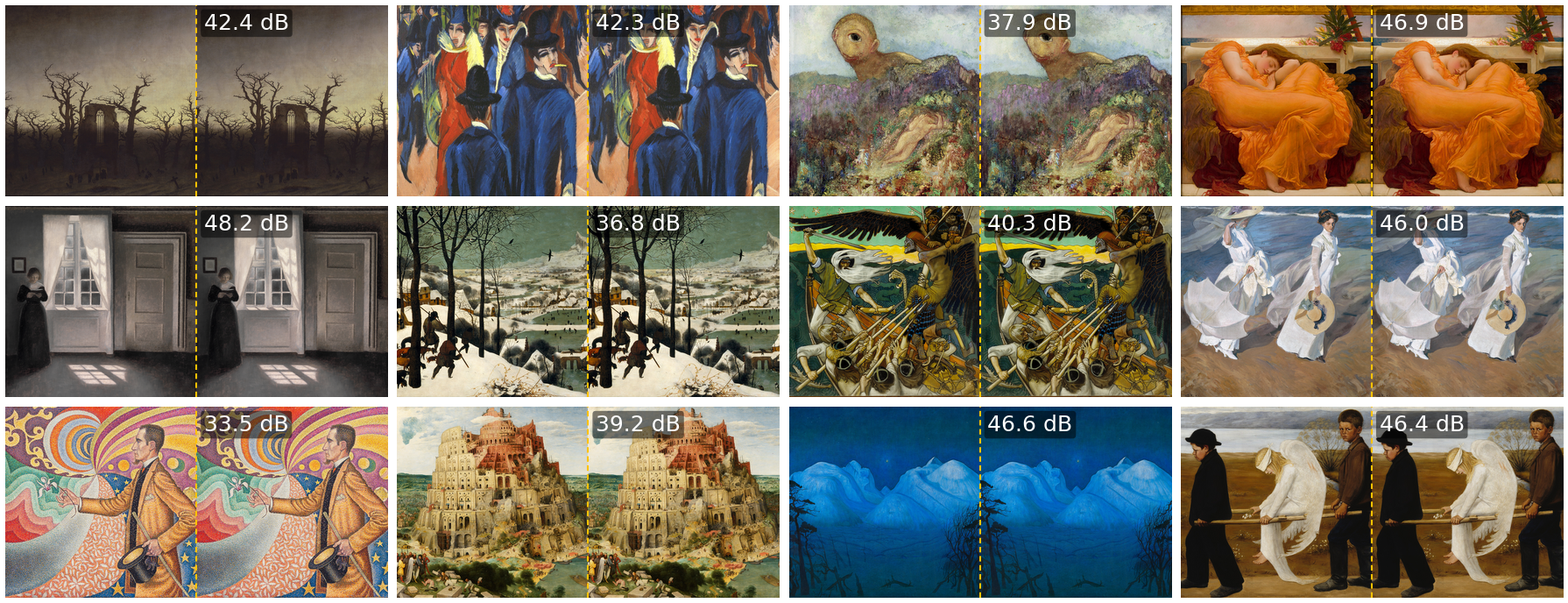}
  \caption{\textbf{Natural images, paintings.} As in Figure~\ref{fig:img-animals}, every photograph is followed directly to its right by its reconstruction with $4096$ partitions per color channel, with the PSNR in its top-left corner. Photographs from Wikimedia Commons~\citep{wikimediacommons2026}, credited individually on the project website~\citep{daedaluscredits2026}.}
  \label{fig:img-paintings}
\end{figure}

\begin{figure}[!htbp]\centering
  \includegraphics[width=\textwidth]{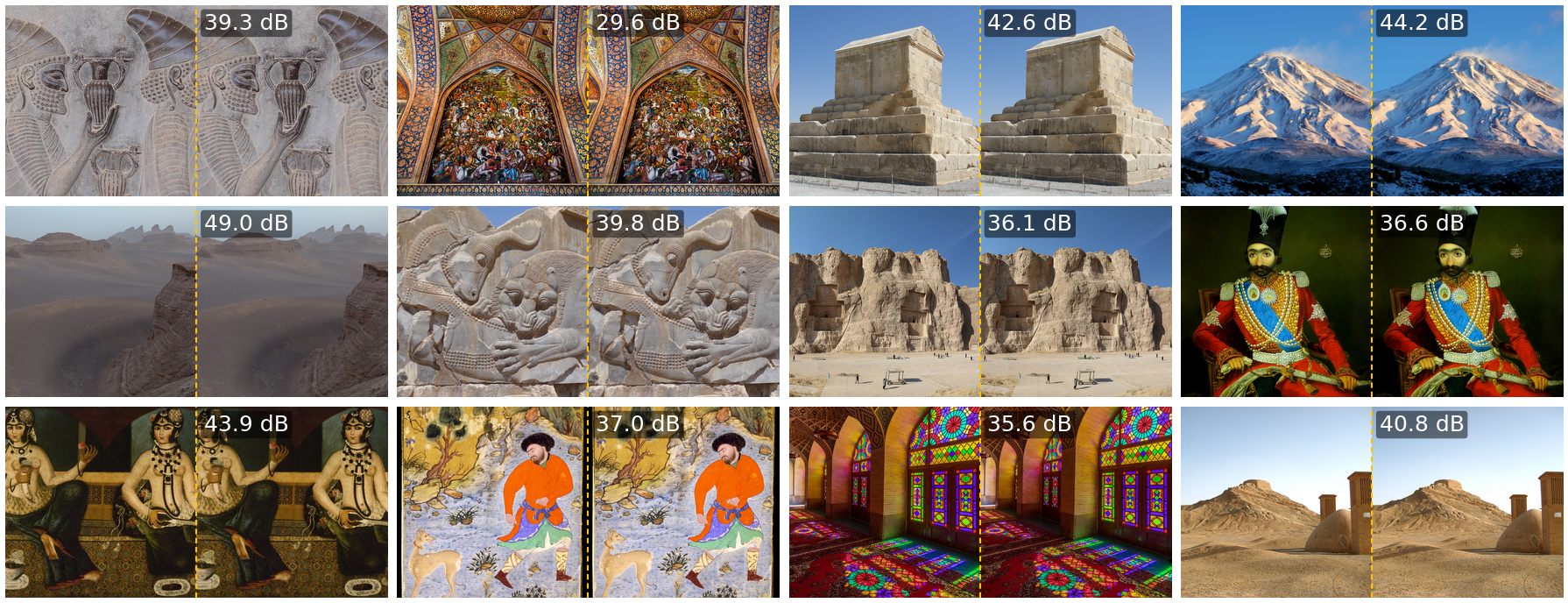}
  \caption{\textbf{Natural images, Persian art and places.} As in Figure~\ref{fig:img-animals}, every photograph is followed directly to its right by its reconstruction with $4096$ partitions per color channel, with the PSNR in its top-left corner. Photographs from Wikimedia Commons~\citep{wikimediacommons2026}, credited individually on the project website~\citep{daedaluscredits2026}.}
  \label{fig:img-persian}
\end{figure}

A photograph is three scalar functions on a pixel grid, one per color channel, and the frozen model encodes and decodes each channel like any other input. Figures~\ref{fig:img-animals}--\ref{fig:img-persian} reconstruct $48$ photographs in four groups, animals, minerals and crystals, paintings, and Persian art and places, most of them Featured Pictures of Wikimedia Commons~\citep{wikimediacommons2026}, with the credit of every photograph on the project website~\citep{daedaluscredits2026}. Every photograph is cut to its central square, resampled to $512\times512$ pixels, and reconstructed with $4096$ partitions per channel, about $64$ pixels per partition, and the peak signal-to-noise ratio (PSNR) of each reconstruction is printed on it. The median PSNR over the $48$ photographs is $42.3$~dB, with a range from $29.6$ to $52.3$~dB. The lowest values belong to texture finer than a partition, such as a densely painted fresco, a pointillist painting, or rhodochrosite crystals on a granular rock, and the highest to content dominated by smooth regions, such as whales in open water or a clownfish before a blurred anemone.

\FloatBarrier
\subsubsection{Functions on one-dimensional curves}
\label{app:recon-curves}

\begin{figure}[!htbp]\centering
  \includegraphics[width=.8\textwidth]{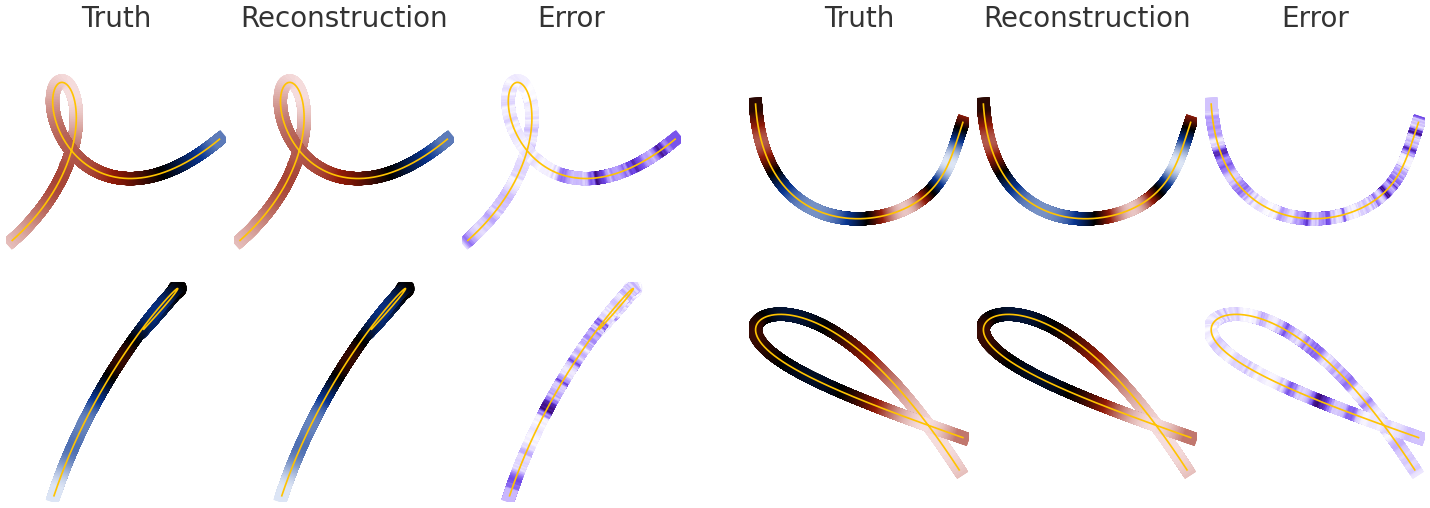}
  \caption{\textbf{Functions on curves.} Four smooth random curves in the plane (yellow), each carrying a scalar function drawn as a band colored by its value (left of each triple), reconstructed with $16$ partitions (middle), and the absolute error, averaged over $17$ neighboring points along the curve (right).}
  \label{fig:curves-recon}
\end{figure}

The FAE is trained only on functions that fill a two-dimensional disk, yet the same frozen model accepts a function sampled on a one-dimensional curve embedded in the plane, such as a boundary trace. Figure~\ref{fig:curves-recon} encodes smooth random curves with $16$ partitions and decodes on the same points, and the reconstruction follows the observed values along each curve with a small, unstructured error. The encoder aggregates whatever points a partition holds, so a partition that contains only a piece of a curve is as valid an input as one that fills an area. The same backbone, with the same weights and without retraining, therefore encodes functions on one-dimensional curves and on two-dimensional domains into one latent space. This is of use for multi-modal tasks whose inputs or outputs live on manifolds of different dimension, such as an operator that maps a boundary or surface function to the field inside the domain, or the field to its trace on the boundary, since one encoder and one decoder then represent both sides.

\begin{figure}[!htbp]\centering
  \includegraphics[width=0.46\textwidth]{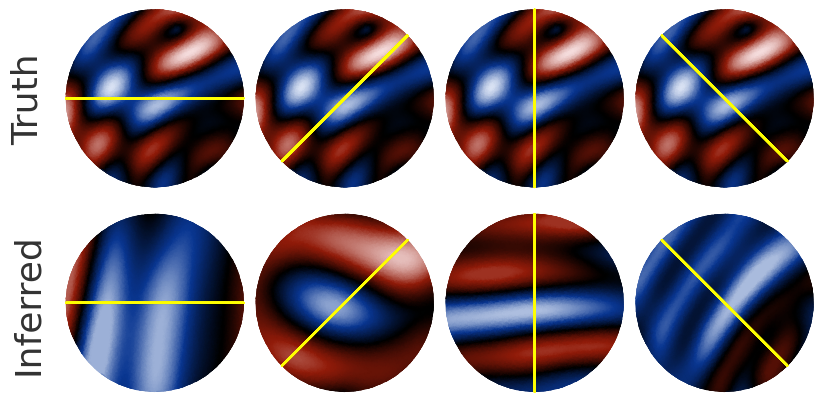}\\[0.2em]
  \includegraphics[width=0.66\textwidth]{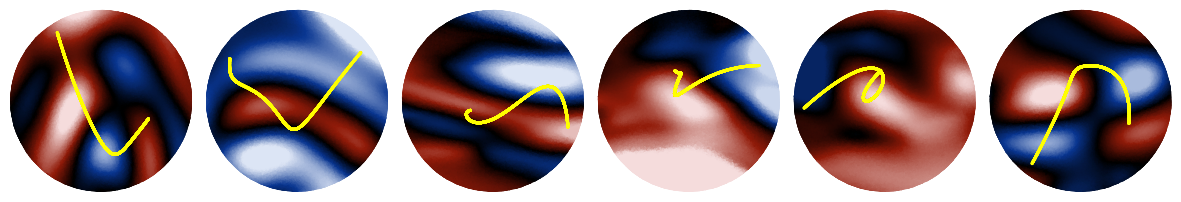}
  \caption{\textbf{One-dimensional inputs.} Top, one field (Truth) observed on straight lines through the disk at four angles (yellow) and the field decoded from each line on a single partition (Inferred). Along the line the decoded field tracks the truth, and off the line the decoder fills in a plausible field. Bottom, six curves through one underlying field give six different plausible completions, so the encoder does not return a memorized training function.}
  \label{fig:curves}
\end{figure}

\begin{figure}[!htbp]\centering
  \includegraphics[width=0.8\textwidth]{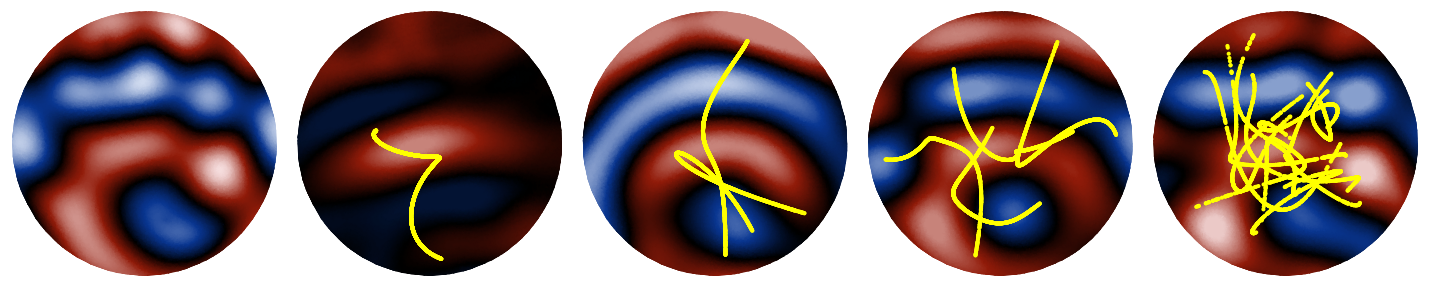}
  \caption{\textbf{Coverage by curves.} One underlying field (left) observed on an increasing number of curves (yellow). The more of the domain the curves cover, the more of the field the observations constrain.}
  \label{fig:curves-coverage}
\end{figure}

A curve under-determines the field around it, and the decoder, which can be queried anywhere, answers with a two-dimensional function nonetheless. Observed on a straight line through the disk, a field is tracked along the line, while off the line the decoder returns a plausible smooth continuation that the observation does not constrain, a different one for every direction of the line (Figure~\ref{fig:curves}, top). Different curves through the same field likewise give different plausible completions (Figure~\ref{fig:curves}, bottom), which is direct evidence that the encoder does not return a memorized training function, and as more curves cover the domain the observations constrain more of it and the true field re-emerges (Figure~\ref{fig:curves-coverage}). This is what a shared latent space for functions on a domain and on its boundary requires, since an operator conditioned on boundary data then receives it already extended into the interior.

\FloatBarrier
\subsubsection{Extrapolation and inpainting}
\label{app:recon-masked}

\begin{figure}[!htbp]\centering
  \includegraphics[width=0.6\textwidth]{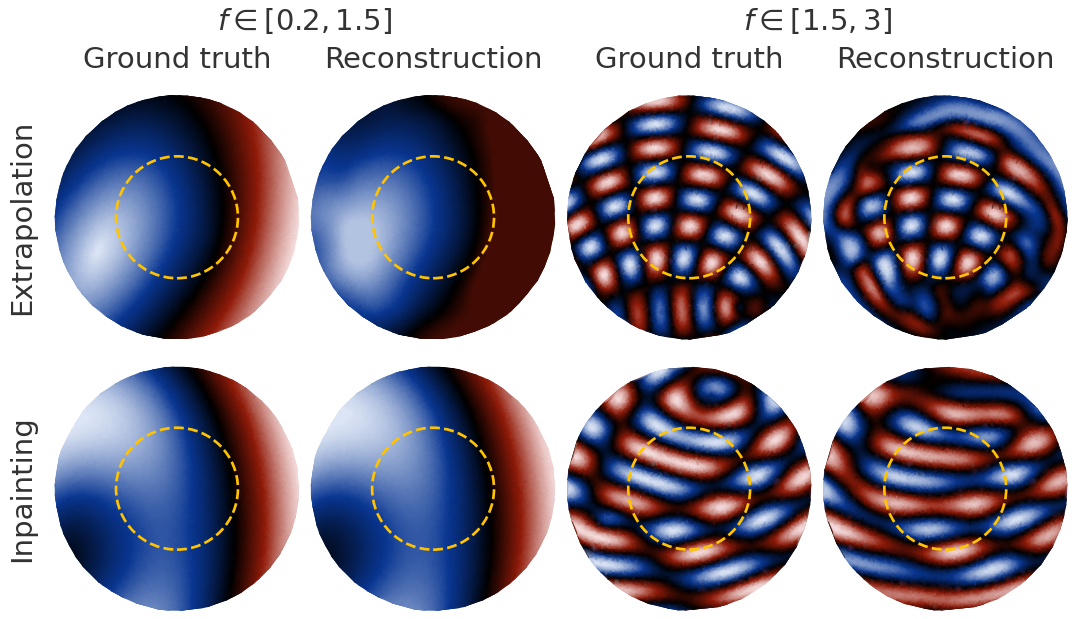}
  \caption{\textbf{Extrapolation and inpainting.} A two-mode radial-sine field encoded with one partition from the inner disk only (top) or the outer annulus only (bottom) and decoded on the whole disk, with frequencies in the range of the error sweep of Figure~\ref{fig:masked-errors} (left pair of columns) and in a higher range (right pair). The dashed circle marks the mask at $r=0.5$.}
  \label{fig:masked-examples}
\end{figure}

\begin{figure}[!htbp]\centering
  \includegraphics[width=0.9\textwidth]{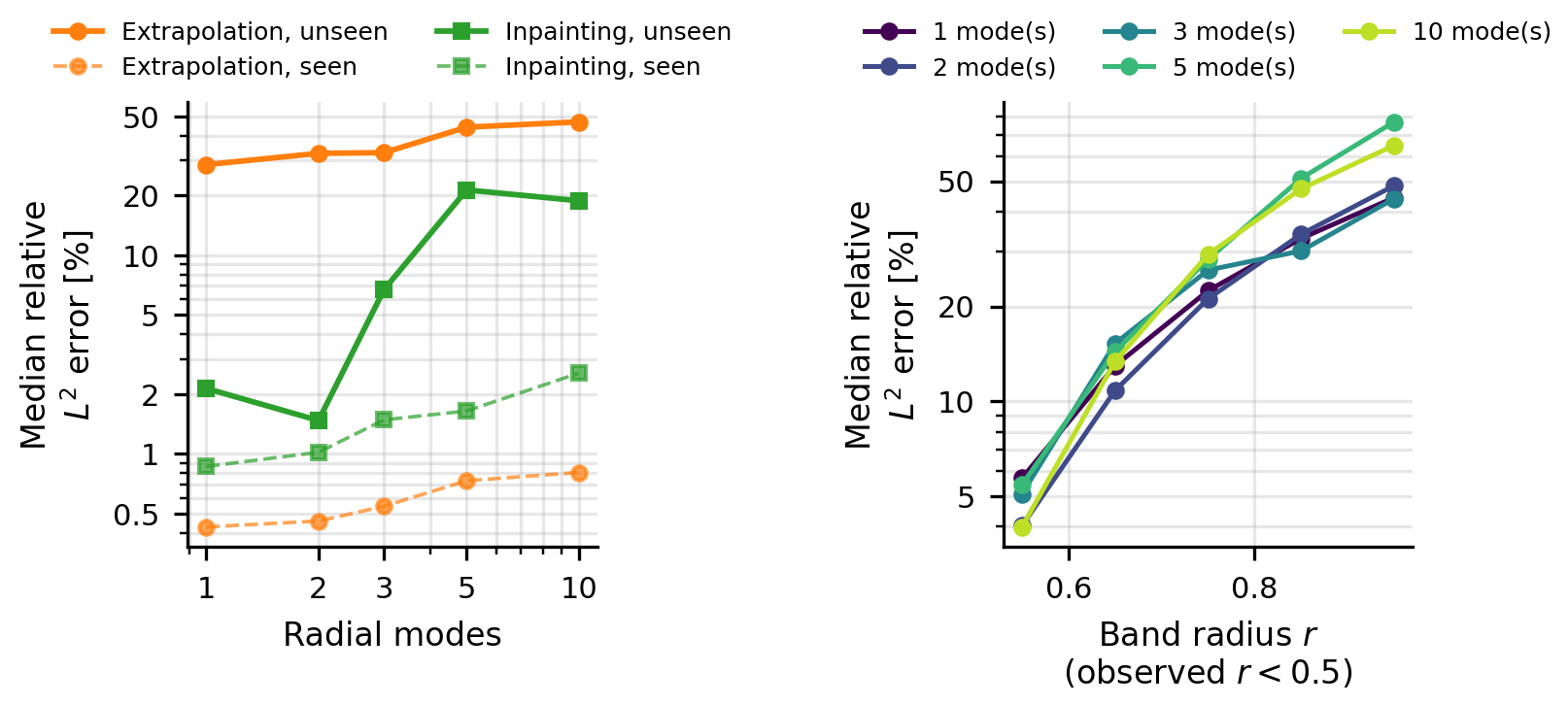}
  \caption{\textbf{Errors under partial observation.} Left, median relative $L^2$ error on the unseen (solid) and the observed (dashed) region against the number of radial modes, for extrapolation and inpainting. Right, the extrapolation error in radial bands beyond the observed disk. Eight fields per setting.}
  \label{fig:masked-errors}
\end{figure}

Two experiments probe partial observations from opposite directions, both with a single partition on the unit disk and a mask at radius $0.5$, for radial-sine fields with $1$ to $10$ modes whose frequencies lie in the training band (Figure~\ref{fig:masked-examples}). In the first, the encoder observes only the inner disk $r<0.5$ and the embedding is decoded on the whole disk. The observed region is reconstructed to below $1\%$, and the continuation remains accurate close to the mask, with a median error of about $5\%$ in the band $0.5<r<0.6$, before it becomes unconstrained further out and reaches $44$ to $77\%$ in the outermost band (Figure~\ref{fig:masked-errors}, right). This is expected, since the local frame is built from the observed extent, and the rim of the disk therefore lies at twice the normalized radius the encoder has ever been asked to describe. In the second, the encoder observes only the annulus $0.5<r<1$ and the masked inner disk is filled in, an inpainting task. The frame now spans the whole domain, so no coordinate is out of range, and the question is only whether the embedding determines a region for which it holds no samples. For one or two radial modes the inner disk is recovered to about $2\%$, and the error grows with the complexity of the field to about $20\%$ for five and ten modes (Figure~\ref{fig:masked-errors}, left), as an unobserved region is increasingly under-determined by the observed one.


\subsubsection{Bias toward the training distribution}
\label{app:recon-bias}

Where an observation leaves part of a field unconstrained, the decoder still returns a value there, and the completions of the two previous experiments show what shapes it. Off a curve the decoder fills in a smooth field of the kind it was trained on (Figure~\ref{fig:curves}). Beyond an observed disk the extrapolation continues the observed oscillations as radial wavefronts, and an inpainted disk is filled with oscillations of the same kind, which match the truth near the observed annulus and depart from it toward the center, at low and at higher frequencies alike (Figure~\ref{fig:masked-examples}). Under partial observation the model thus fills the missing information with functions that resemble its training functions, the sums of radial sines of Figure~\ref{fig:train-dist}. This bias is a property of the training distribution rather than of the architecture. A backbone trained on another family of functions would complete an unobserved region with functions of that family, so the training distribution acts as the prior of every partially observed reconstruction and should be chosen to match the structure expected where an application observes nothing.

\FloatBarrier
\subsection{Partition count and input resolution}
\label{app:sweep}

\subsubsection{Partitioning sets the accuracy}
\label{app:sweep-count}

\begin{figure}[!htbp]\centering
  \begin{subfigure}{\textwidth}\centering
    \includegraphics[width=0.85\textwidth]{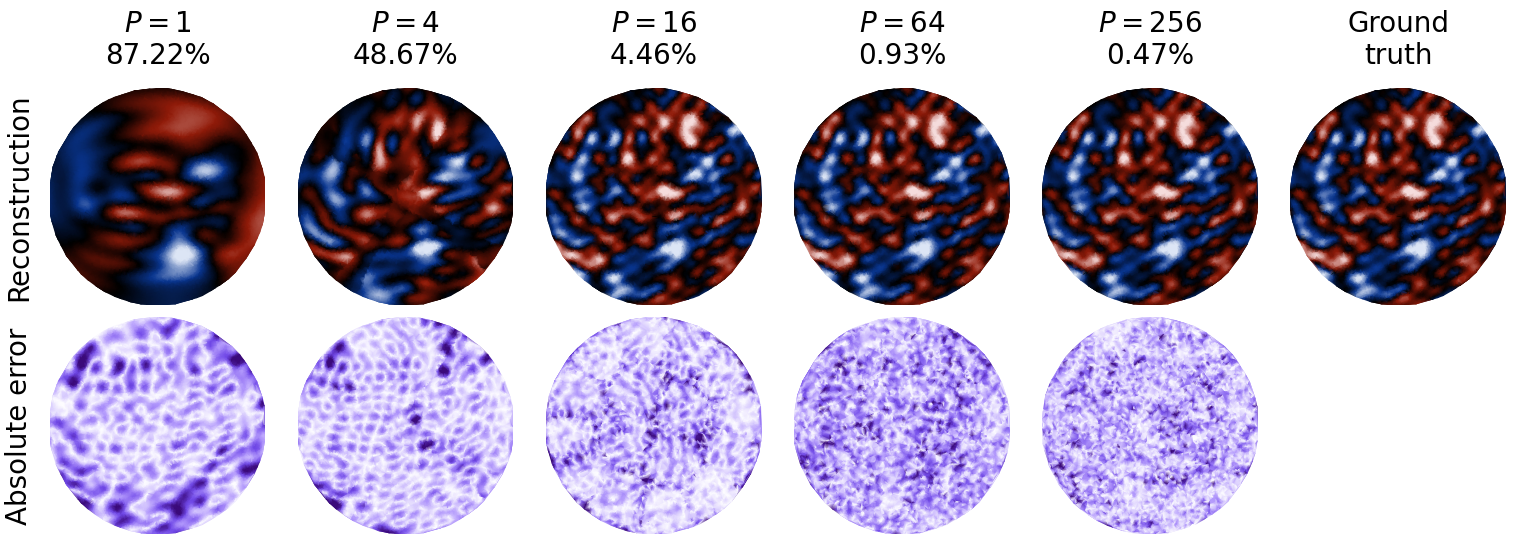}
    \caption{A high-frequency radial-sine field.}
    \label{fig:sweep-synth}
  \end{subfigure}\\[0.6em]
  \begin{subfigure}{\textwidth}\centering
    \includegraphics[width=0.85\textwidth]{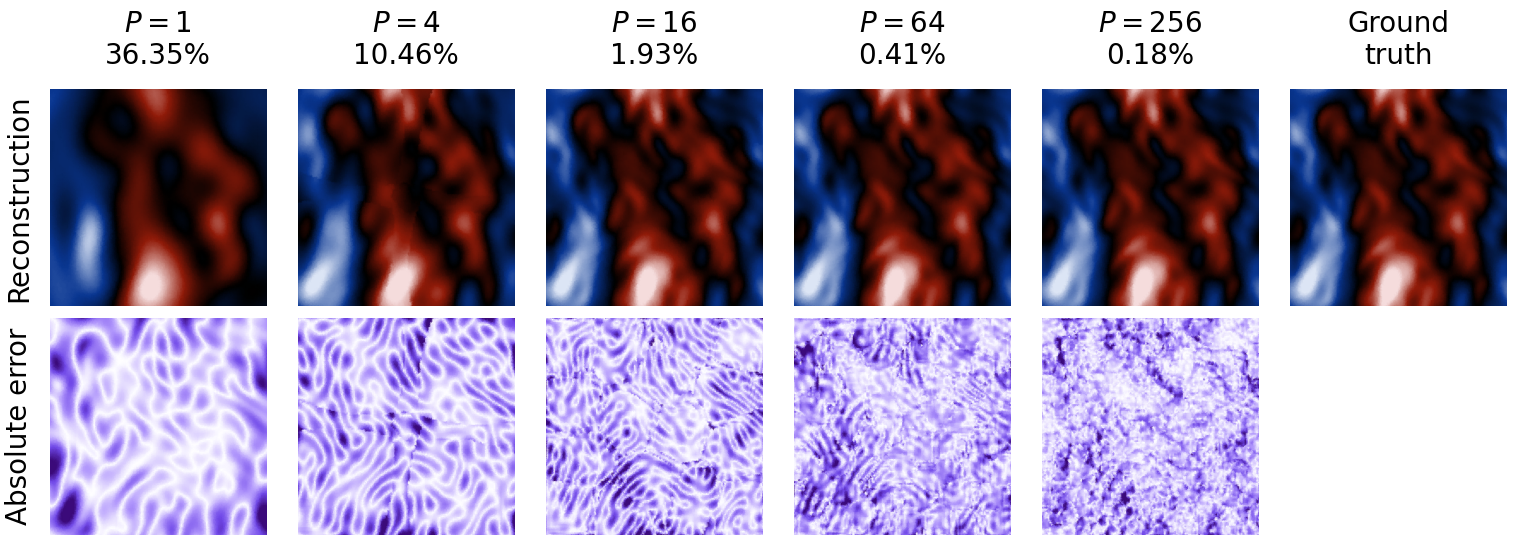}
    \caption{One velocity component of an incompressible-flow snapshot from a Brownian-bridge initial condition ($128{\times}128$ grid).}
    \label{fig:sweep-ns}
  \end{subfigure}\\[0.6em]
  \begin{subfigure}{\textwidth}\centering
    \includegraphics[width=0.85\textwidth]{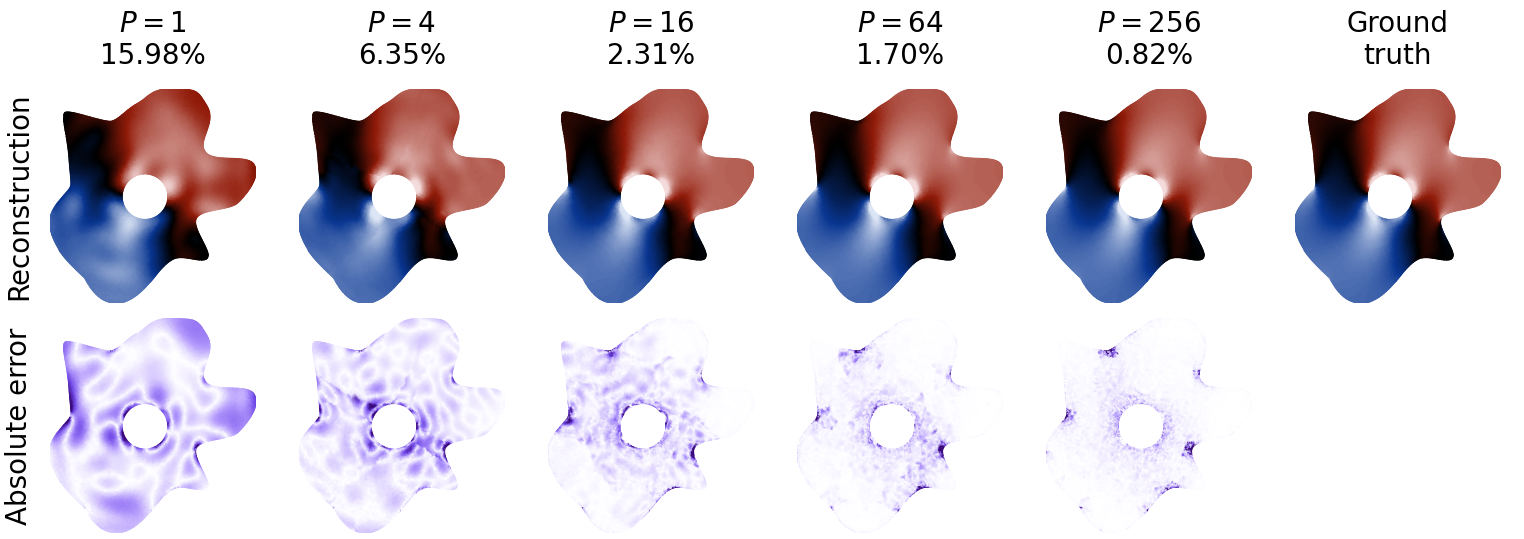}
    \caption{The Cauchy normal stress $\sigma_{yy}$ of a hyperelastic body on a domain with a hole.}
    \label{fig:sweep-stress}
  \end{subfigure}
  \caption{\textbf{Partition sweeps.} Three fields reconstructed with $1$ to $256$ partitions (top row of each panel, with the relative $L^2$ error above every reconstruction and the ground truth on the right) and the absolute error (bottom row). Larger partitions capture only the low-frequency content, and more partitions recover the detail. In (c) the error concentrates at the hole and at the corners of the boundary, where the stress varies fastest, and recedes as the partitions there shrink.}
  \label{fig:sweep}
\end{figure}

More partitions shrink the size of each individual one, lower its effective frequency, and raise accuracy. A simple field needs only a few partitions and a complex one needs many, but in every case we examined the error decreases steadily as the count grows. Figure~\ref{fig:sweep-synth} shows this for a high-frequency synthetic field, where large partitions capture only its low-frequency content and more partitions recover the detail. The coarse reconstructions are not noise. With few partitions, and therefore few latent dimensions for the whole field, the representation keeps the low-frequency features of the field and drops the finer ones, as a truncated spectral expansion would, and every further partition adds detail on top of a faithful coarse version. The same behavior holds for real fields the model never saw in training. For an incompressible-flow velocity field the error falls from $36.3\%$ with a single partition to $0.18\%$ with $256$ (Figure~\ref{fig:sweep-ns}), and for the Cauchy stress $\sigma_{yy}$ of a hyperelastic body on a domain with a hole it falls from $16.0\%$ to $0.82\%$ over the same range (Figure~\ref{fig:sweep-stress}). For the stress field the error concentrates around the hole and at the corners of the boundary, where the stress varies fastest, and shrinks there as the partitions get smaller. At few partitions the partition boundaries are also visible in the error maps of the flow and the stress field, and they fade as the count grows. Reconstructing each partition independently and stitching the results, in the manner of a divide-and-conquer scheme, is what turns an unfamiliar global field into a set of familiar local problems.

\FloatBarrier
\subsubsection{Robustness to resolution and sparsity}
\label{app:sweep-sparsity}

\begin{figure}[!htbp]\centering
  \begin{subfigure}{0.34\textwidth}\centering
    \includegraphics[width=\linewidth]{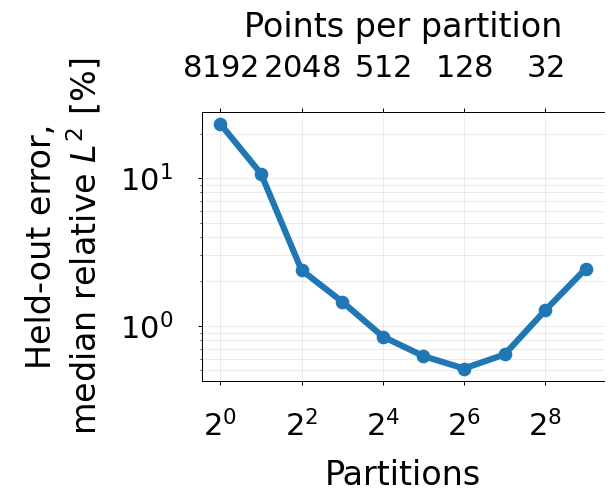}
    \caption{A fixed cloud of $8192$ points.}
  \end{subfigure}\hspace{0.1\textwidth}
  \begin{subfigure}{0.34\textwidth}\centering
    \includegraphics[width=\linewidth]{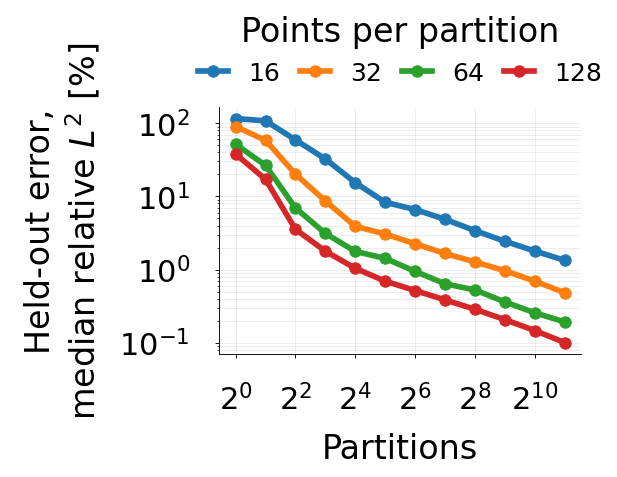}
    \caption{A cloud that grows with the number of partitions.}
  \end{subfigure}
  \caption{\textbf{Points per partition.} The median relative $L^2$ error at points the encoder did not see, on the evaluation preset Sines (eval), three radial-sine modes with frequencies $f\in[1,4]$, against the partition count. The top axis gives the points per partition. (a) With the cloud fixed, more partitions leave fewer points in each, and past $64$ partitions the error grows. (b) With the cloud grown in proportion to the partitions, at $16$ to $128$ points per partition, the error keeps falling.}
  \label{fig:sparsity}
\end{figure}

The partition count fixes how many embeddings describe a field, and the input resolution how many points each embedding is computed from.
The experiment of Figure~\ref{fig:sparsity} separates the two, by letting the points per partition shrink while the number of partitions grows. With the cloud held at $8192$ points of the evaluation distribution, more partitions eventually starve each partition of points (Figure~\ref{fig:sparsity}a). The error is scored at points the encoder did not see, and it is lowest at $64$ partitions of $128$ points and grows almost fivefold by $512$ partitions of $16$ points. Scored at the encoded points themselves it would keep falling instead, since an embedding can store the few values it receives, so only the held-out score exposes the loss. Growing the cloud with the partition count keeps the points per partition constant, and the held-out error then keeps falling at every density we tried, from $16$ to $128$ points per partition, to between $1.34\%$ and $0.10\%$ at $2048$ partitions, the denser partitions lower throughout (Figure~\ref{fig:sparsity}b). More partitions therefore cost accuracy only when they leave too few points in each, and not by themselves. Too few points per partition also make the frame itself unreliable. The center, the extent, and the value range of a partition are statistics of its own points, and with a handful of points they describe the sampled points rather than the field around them, so the normalized frame, and with it the decoded field between the points, carries large errors that no embedding can correct.

\FloatBarrier
\subsection{Derivatives}
\label{app:derivatives}

The decoder evaluates a query point from its own coordinate, expressed in the frame of each partition that contains it, and from the embeddings of those partitions (Equation~\eqref{eq:partnorm-decode}). No other query point enters, so the decoded field $\hat\funcMain$ is a neural field on the domain, and for every coordinate axis $i$ the Jacobian of the decoded values with respect to the $i$-th coordinates of the query points is diagonal. A single forward-mode pass of automatic differentiation that seeds the same unit tangent $\mathbf{e}_i$ at every query point therefore returns $\partial\hat\funcMain/\partial x_i$ at all points at once, and one such pass per axis, two in the plane, gives the exact gradient $\nabla\hat\funcMain$ of the decoded field. The Jacobian itself is never formed. A forward-mode pass returns the product of the Jacobian with the seeded tangent, which by this structure is its diagonal, so the memory and the computation are those of a decode that carries one tangent per axis, both growing linearly with the number of query points. The full Jacobian of $N$ decoded values with respect to their $2N$ coordinates instead holds $2N^2$ entries, about $34$\,GB in single precision for $N=65536$. A decoder whose output at one query point depends on the others would have a dense Jacobian, and a single forward-mode pass would then return a mixture of the derivatives at different points, so that the derivative at every point would require a pass per point or the full Jacobian. The value frame of every partition lies inside the differentiated map, so the gradient is in physical units. No grid, no difference stencil, and no per-point evaluation are involved, and the cost is that of one decode per axis that carries a tangent.

Physics-informed operator learning has so far obtained derivatives in other ways. PINO \citep{li2024physics} considers numerical differentiation of the output on its grid by finite differences or in Fourier space, which is fast but requires a uniform grid and, for the spectral route, a smooth or periodic field, pointwise automatic differentiation, which the authors describe as exact but slower and more memory-consuming, and an explicit differentiation of the Fourier layers. Its released training code computes the residuals of its Darcy and Navier--Stokes problems with finite differences and Fourier differentiation on the grid. Physics-informed DeepONets \citep{wang2021learning} differentiate the network with respect to the input of its trunk by automatic differentiation, which is exact because a DeepONet also evaluates every query coordinate on its own. In both cases the derivatives enter a physics-informed training loss of a model built for one family of PDEs. Nothing in the training of the backbone involves derivatives. It is trained only to reconstruct the values of analytic functions and is never asked for a smooth field or an accurate gradient.

\begin{figure}[!htbp]\centering
  \includegraphics[width=\textwidth]{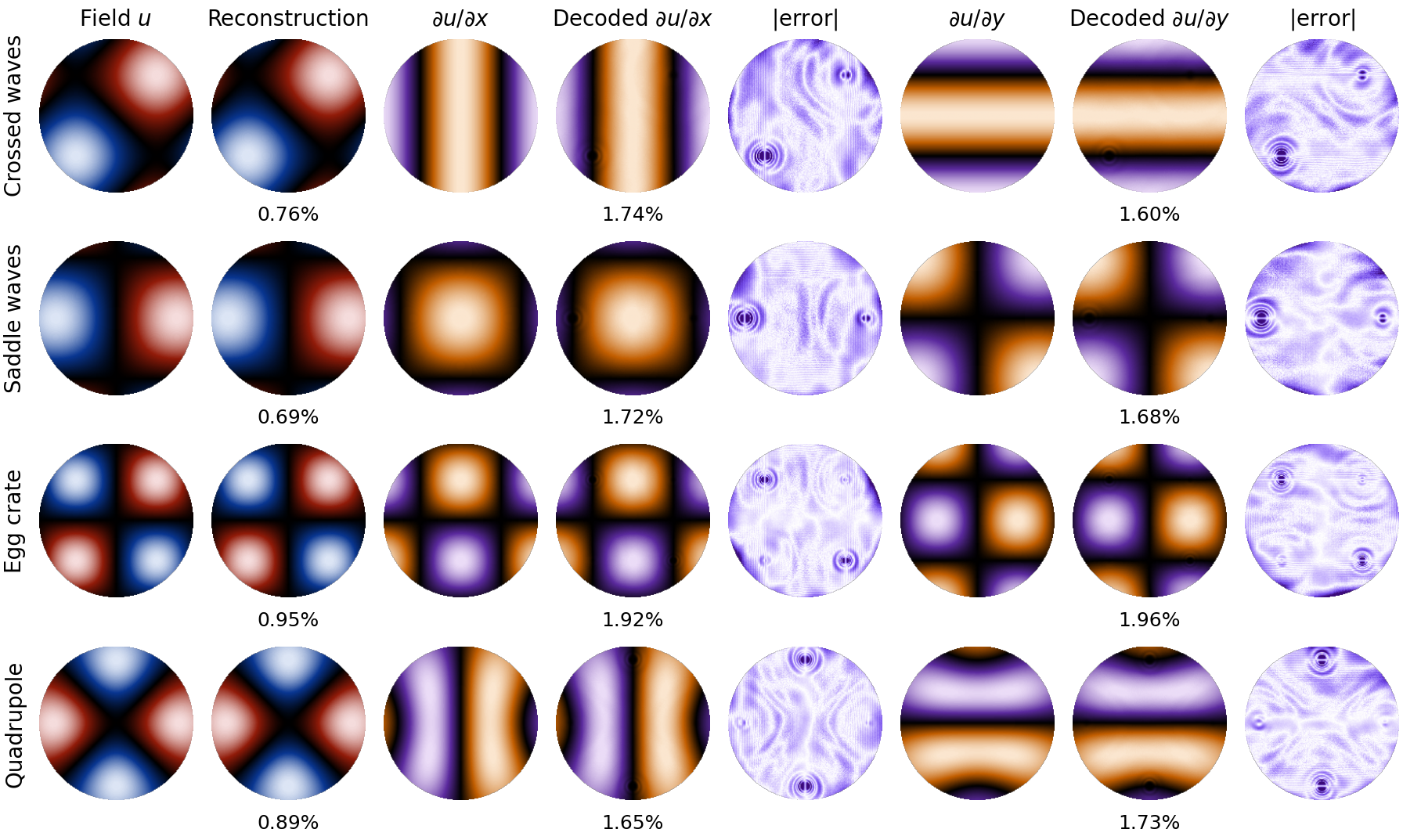}
  \caption{\textbf{Derivatives of decoded fields.} Four analytic fields on the unit disk, each encoded as a single partition from $16384$ random points by the analysis backbone and decoded on a pixel grid. From left to right, the field $\funcMain$, its reconstruction with the relative $L^2$ error, and for each axis the analytic derivative, the derivative of the decoded field computed by forward-mode automatic differentiation of the decoder with its relative $L^2$ error, and their absolute difference. The derivatives are drawn in their own colormap. The backbone is trained to reconstruct values only.}
  \label{fig:derivatives}
\end{figure}

Figure~\ref{fig:derivatives} compares the decoded gradient with the analytic one for four smooth fields on the unit disk, each encoded as a single partition from $16384$ random points and decoded on a pixel grid of the disk. The values are reconstructed to within $1\%$ and every derivative to within $2\%$, with the error spread thinly over the disk apart from a few small spots. Two consequences follow. The accuracy of the derivatives indicates that a decoded field is a smooth function of position rather than an interpolant of the encoded samples, and hence that the embeddings it comes from are meaningful points of a space of smooth functions. Derivative fields are moreover available from any reconstruction or prediction at little cost. Strains and stresses follow from a decoded displacement field and fluxes from a decoded potential, and a physics-informed residual can be imposed on the decoded output of a latent operator without a grid or a stencil. With several partitions the gradient remains exact for the decoded field, which is then assembled from the pieces of the partitions that contain each point, so its accuracy near partition boundaries depends on the overlap that averages those pieces (Appendix~\ref{app:partitioner}).

\FloatBarrier
\clearpage

\section{Interpretability}
\label{app:latent}
\setcounter{figure}{0}
\setcounter{table}{0}

The encoder and decoder are deterministic, continuous, and differentiable maps, and every morphologic embedding is a direction on a hypersphere that decodes to a field. The latent space is therefore an object that can be explored in its own right. This appendix collects that exploration. It starts from the elementary morphologies the basis directions of the latent space decode to and the great circles between them (Appendix~\ref{app:latent-atoms}). It then establishes the correspondence between the latent and the physical space, in that distances on the sphere track distances between the decoded fields, embeddings compose linearly only locally, small edits of an embedding change its field coherently, and the linearized decode-encode map and the spread of whole families show how much of the sphere the embeddings use (Appendix~\ref{app:latent-corr}). It reads edit directions from single pairs of examples and transfers them to unrelated fields (Appendix~\ref{app:latent-dirs}), follows geodesics between embeddings, measures how far a geodesic carries a real physical evolution, and continues a geodesic past its endpoint (Appendix~\ref{app:latent-morph}), and closes with how families of functions are arranged on the sphere (Appendix~\ref{app:latent-families}).

\FloatBarrier
\subsection{Atoms and great circles}
\label{app:latent-atoms}

\begin{figure}[!h]\centering
  \includegraphics[width=\textwidth]{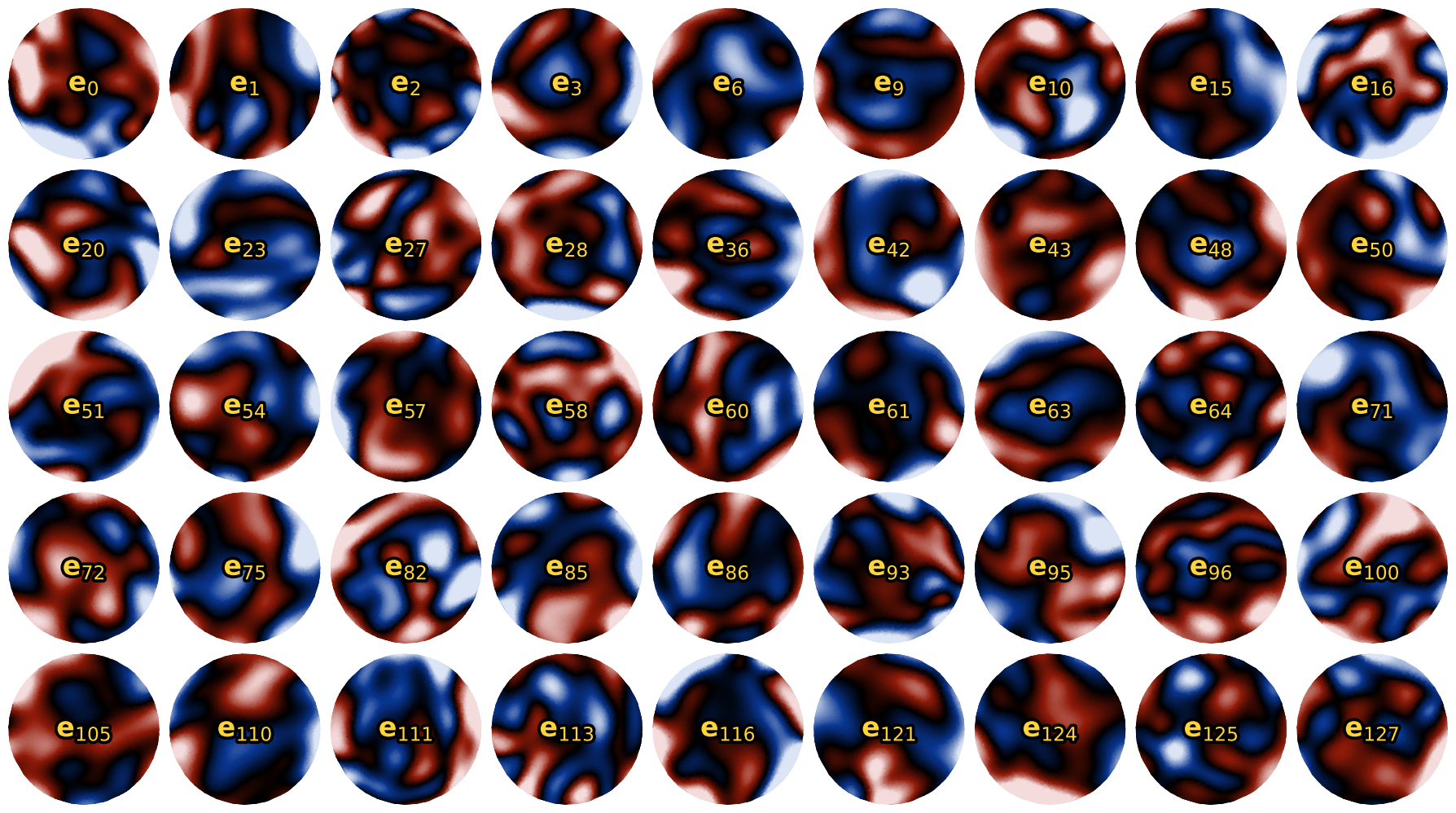}
  \caption{\textbf{Latent atoms.} The fields decoded from the basis directions $\mathbf{e}_i$ of $45$ randomly chosen dimensions, each on one partition of the unit disk with the identity frame, with the direction $\mathbf{e}_i$ at the center of its field.}
  \label{fig:atoms}
\end{figure}

\begin{figure}[!htbp]\centering
  \includegraphics[width=0.8\textwidth]{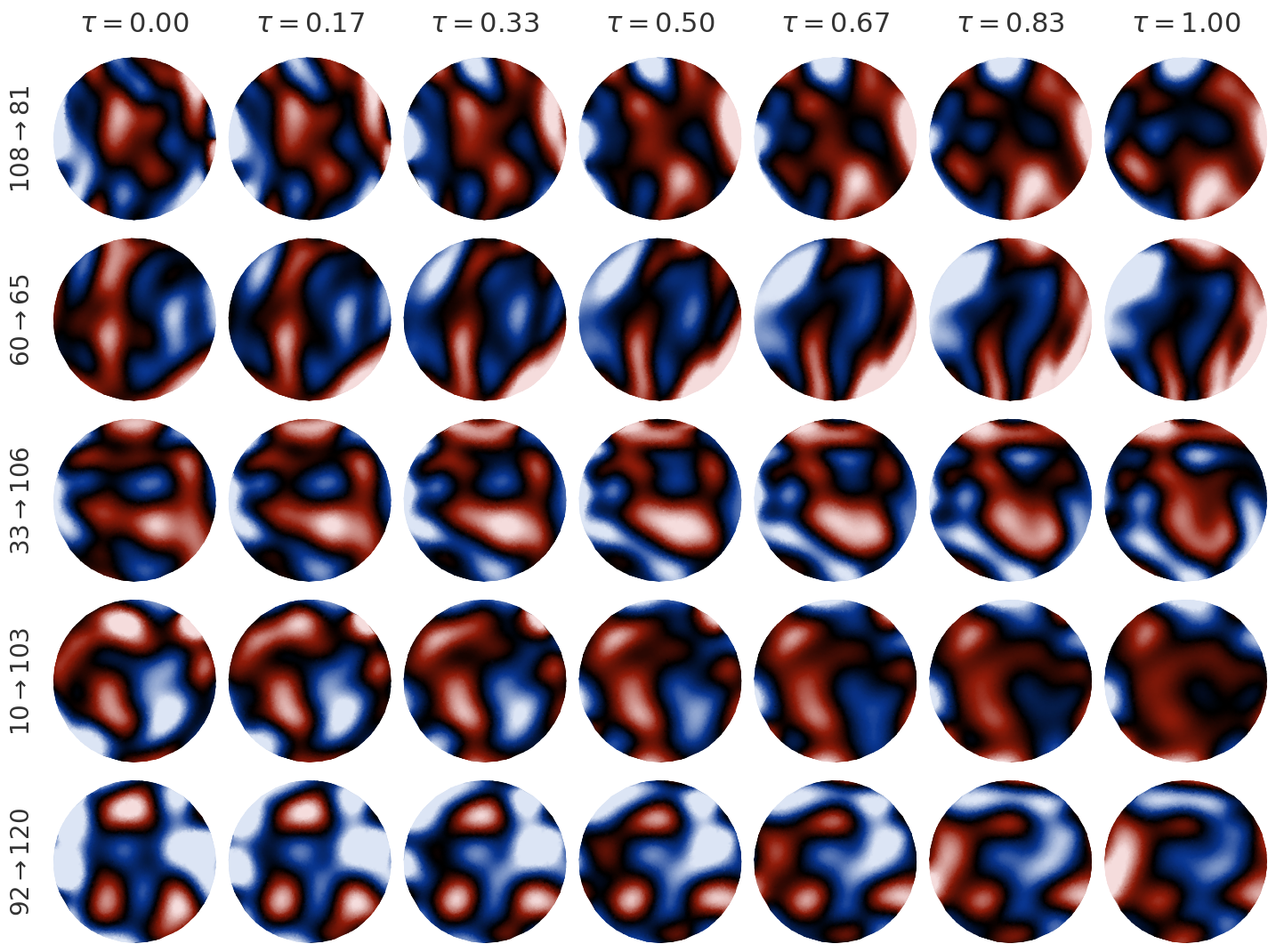}
  \caption{\textbf{Great circles between atoms.} Decoding along the great circle from $\mathbf{e}_i$ to $\mathbf{e}_j$ for five pairs of axes, at the fraction $\tau$ of the arc above each column.}
  \label{fig:greatcircle}
\end{figure}

The first question is whether the basis directions of the latent space carry meaning by themselves. We place a single partition on the unit disk with the identity frame, set its morphological embedding to a signed basis direction $\pm\mathbf{e}_i$ scaled to the radius $\sqrt{\latdim}$ of the sphere, and decode it on the disk. Every axis decodes to an elementary morphology, a smooth field of a few lobes and oscillations (Figure~\ref{fig:atoms}), and the opposite sign of the same axis decodes to another field of the same kind. The second question is whether the regions between the axes are populated as well. Spherical interpolation from $\mathbf{e}_i$ to $\mathbf{e}_j$ walks the quarter of a great circle that joins them, and decoding along it morphs one atom into the other through plausible fields at every step, never through noise or a blank field (Figure~\ref{fig:greatcircle}). The sphere is thus covered by valid fields rather than by isolated points that happen to decode well.

\FloatBarrier
\subsection{Latent--physical correspondence}
\label{app:latent-corr}

This part asks how faithfully the geometry of the sphere reflects the geometry of the fields it decodes to. It compares distances and sums of embeddings with those of their fields, edits single embeddings and follows how their fields change, and then reads from the linearized decode-encode map and from whole families of embeddings how much of the sphere the model uses.

\subsubsection{Correlated distances}
\label{app:latent-metric}

We first ask whether distances on the sphere mean the same as distances between the decoded fields, which decides whether the latent space can stand in for the physical space in a loss or an optimization. Pairs of embeddings are built at controlled cosine similarities, both are decoded, and the relative $L^2$ distance between the two fields is compared with the chord distance $d_{\mathrm{chord}}=2\sin(\theta/2)=\sqrt{2(1-\cos\theta)}$ between the unit directions of the embeddings, with $\theta$ the angle between them. The two distances agree nearly one to one with a narrow spread, the median staying within $20\%$ of the line of slope one over the whole range (Figure~\ref{fig:metric}), so a distance measured on the sphere predicts the distance between the fields, and an objective defined on the embeddings measures what the corresponding objective on the fields would.

\subsubsection{Local superposition}
\label{app:latent-superpos}

\begin{figure}[!htbp]\centering
  \begin{subfigure}{0.39\textwidth}\centering
    \includegraphics[width=\linewidth]{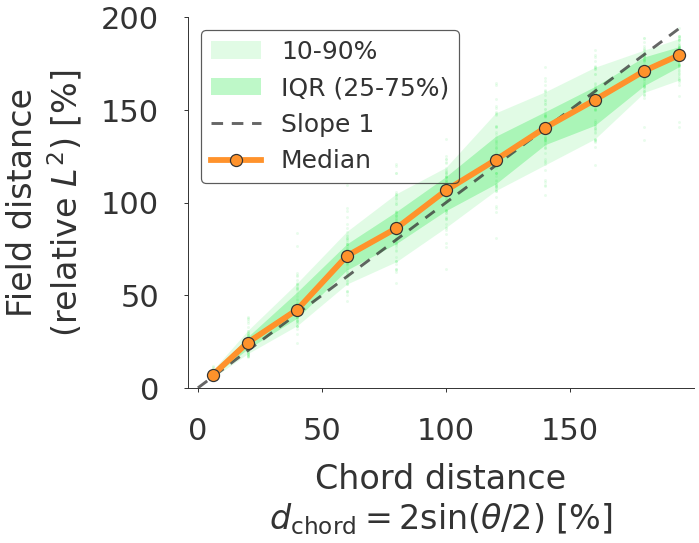}
    \caption{Function-space distance against chord distance.}
    \label{fig:metric}
  \end{subfigure}\hspace{0.02\textwidth}
  \begin{subfigure}{0.39\textwidth}\centering
    \includegraphics[width=\linewidth]{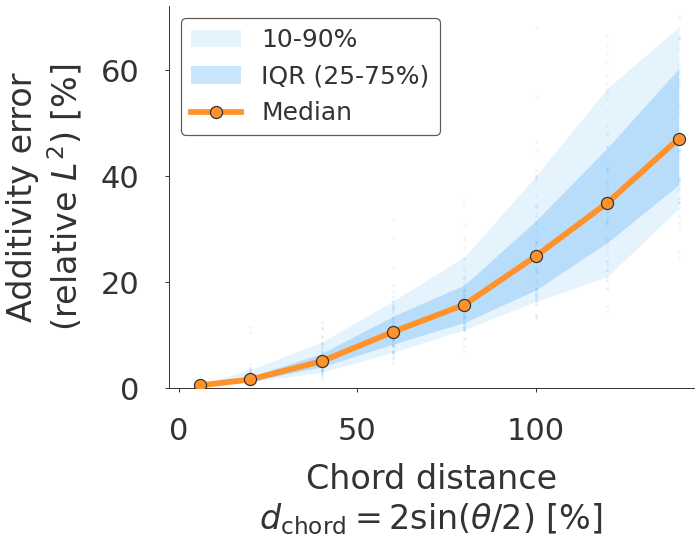}
    \caption{Additivity error against chord distance.}
    \label{fig:superpos}
  \end{subfigure}
  \caption{\textbf{A shared metric and local superposition.} (a) The relative $L^2$ distance between the decoded fields of embedding pairs at controlled similarity against their chord distance on the sphere. (b) The additivity error between the latent and the physical composition of two embeddings against their chord distance. Both panels show the median over $60$ pairs at every chord distance (line), their interquartile range (IQR, the dark band between the $25$th and the $75$th percentile), the band between the $10$th and the $90$th percentile (light), and the single pairs (dots). The dashed line in (a) has slope one.}
  \label{fig:metric-superpos}
\end{figure}

Superposition asks whether embeddings add the way fields do. Two embeddings are composed once in the latent space, by decoding their sum projected back onto the sphere, and once in the physical space, as the min--max normalized mean of the two decoded fields. The two compositions agree when the embeddings are close and diverge as they move apart, and their relative $L^2$ distance, the additivity error, starts near zero for aligned embeddings and grows smoothly with the chord distance between them (Figure~\ref{fig:superpos}). The gap is the nonlinearity of the decoder. Superposition is therefore local, and the latent space is locally linear and globally curved.

\subsubsection{Local edits}
\label{app:latent-walk}

\begin{figure}[!htbp]\centering
  \begin{subfigure}{0.49\textwidth}\centering
    \includegraphics[width=\linewidth]{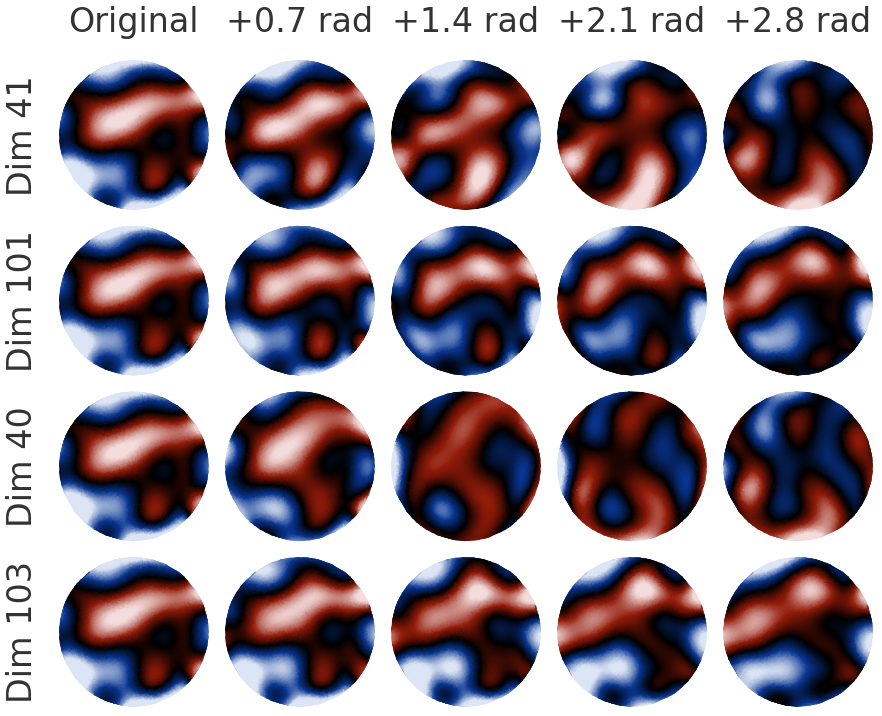}
    \caption{Walking one angle.}
  \end{subfigure}\hfill
  \begin{subfigure}{0.49\textwidth}\centering
    \includegraphics[width=\linewidth]{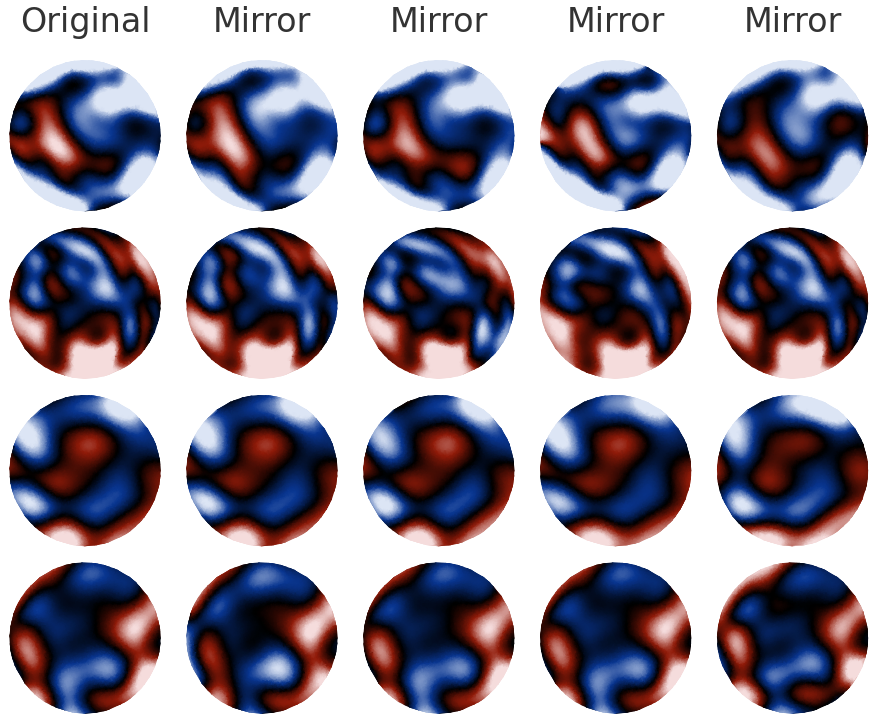}
    \caption{Negating one component.}
  \end{subfigure}
  \caption{\textbf{Local edits.} (a) One embedding walked along one angle in steps of $0.7$ radians, a different angle in every row. (b) Four random embeddings, one per row, with one component negated, a different one in every column. Every edit keeps the morphology coherent.}
  \label{fig:walk}
\end{figure}

This part edits single embeddings in place and asks whether the edits stay coherent. Walking moves an embedding along one angle of its spherical parametrization, in steps of $0.7$ radians, and mirroring negates one of its components (Figure~\ref{fig:walk}). A mirrored embedding decodes to a field whose morphology has changed in a definite way, with lobes reshaped, merged, or displaced, while its overall structure stays coherent and the field remains recognizably related to the original. A perturbation of the embedding, even a large one along a few components, therefore changes the field rather than destroying it, and the latent space is robust to noise and perturbations of the embeddings, with no small set of components responsible for holding a field together. The walk changes the morphology gradually, and every step can be associated with the one before it, so the edits accumulate rather than jump. This is the correspondence of the shared metric once more, seen on single steps rather than on pairs, since embeddings that lie close on the sphere decode to fields that lie close in the physical space.

\subsubsection{The spectrum of the decode-encode map}
\label{app:latent-spectrum}

The edits above probe a few directions around an embedding. The spectrum of the decode-encode map, which decodes an embedding and encodes the decoded field again, probes all of them at once, and asks which directions around an embedding the model represents faithfully. Write $\encoder$ for the encoder, which maps a field on a single partition to its embedding, and $\decoder$ for the decoder, which maps an embedding back to a field. Decoding an embedding $\embmorph$ and encoding the result again returns an embedding $\encoder(\decoder(\embmorph))$, which for a well-reconstructed field is close to $\embmorph$ itself. A small perturbation $\delta\embmorph$ of the embedding is carried through this map by its Jacobian,
\begin{equation}
  \encoder\big(\decoder(\embmorph+\delta\embmorph)\big) \approx \encoder\big(\decoder(\embmorph)\big) + J(\embmorph)\,\delta\embmorph,
  \qquad
  J(\embmorph) = J_{\encoder}\big(\decoder(\embmorph)\big)\,J_{\decoder}(\embmorph),
\end{equation}
the product of the Jacobian $J_{\decoder}$ of the decoder at $\embmorph$ and the Jacobian $J_{\encoder}$ of the encoder at the decoded field. An eigenvector of $J(\embmorph)$ is a direction around $\embmorph$ that the decode-encode map maps onto itself, and its eigenvalue $\lambda$ tells what becomes of a perturbation along it. An eigenvalue near one marks a direction that survives decoding and re-encoding unchanged, which the model therefore represents faithfully. An eigenvalue near zero marks a direction that leaves no trace in the decoded field, so that the encoder cannot recover it, a direction the bottleneck discards. Eigenvalues between zero and one mark contracted directions, whose perturbation the decode-encode map partly undoes, and eigenvalues above one amplified ones. The number of eigenvalues near one thus counts the directions around an embedding that carry information about its field, an effective rank of the embedding.

\begin{figure}[!htbp]\centering
  \includegraphics[width=\textwidth]{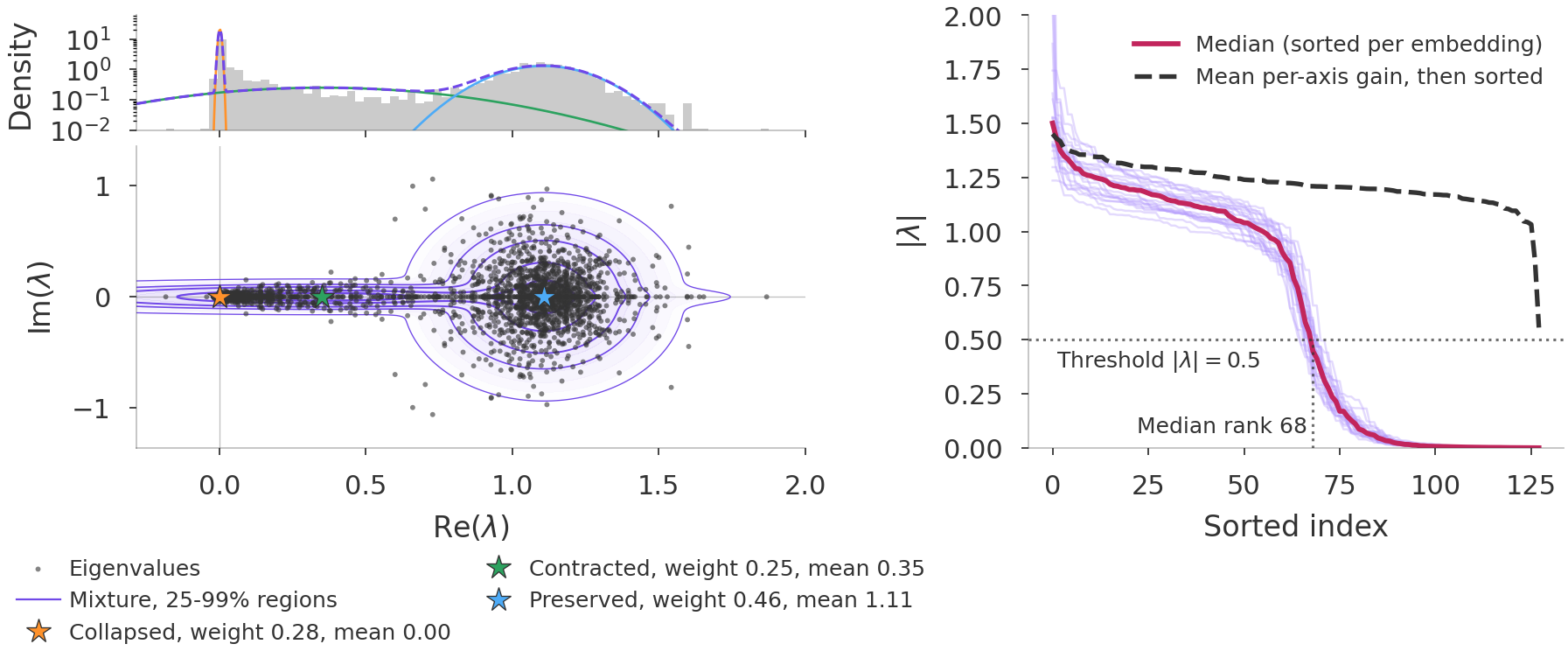}
  \caption{\textbf{Spectrum of the decode-encode map.} Left, the eigenvalues of its Jacobian at many real embeddings in the complex plane, with the density of a three-component Gaussian mixture fitted to all of them, shaded and with the contours that enclose $25$, $50$, $75$, $90$, $95$, and $99\%$ of the eigenvalues, and the means of its three components (stars), each with its weight and the real part of its mean. Above it, the marginal along the real axis, the histogram of the eigenvalues with the three weighted components of the mixture and their sum (dashed), on a logarithmic scale. Right, the eigenvalue magnitudes sorted per embedding, their median, and the mean gain of every fixed axis, sorted afterward. The effective rank of an embedding counts its magnitudes above $0.5$ (dotted), and its median over the embeddings is $68$.}
  \label{fig:spectrum}
\end{figure}

Over many real low-frequency embeddings the eigenvalues fall into two groups, with a spread of contracted values between them (Figure~\ref{fig:spectrum}, left). A three-component Gaussian mixture fitted to them in the complex plane places a tight collapsed component at $0$, which holds $28\%$ of the eigenvalues, a broad preserved component centered at $1.11$, which holds $46\%$, and between the two a wide contracted component centered at $0.35$, which holds $25\%$. The spread of the preserved component along the imaginary axis shows that the Jacobian of the decode-encode map is not symmetric. Sorted per embedding, the magnitudes drop from about one to about zero at a knee, which gives a per-embedding effective rank near $68$ of the $128$ dimensions (Figure~\ref{fig:spectrum}, right). The collapsed directions differ from one embedding to the next, so no axis is globally redundant, and the population of embeddings uses all $128$ dimensions.

\subsubsection{Latent usage}
\label{app:latent-usage}

\begin{figure}[!htbp]\centering
  \begin{subfigure}{0.49\textwidth}\centering
    \includegraphics[width=\linewidth]{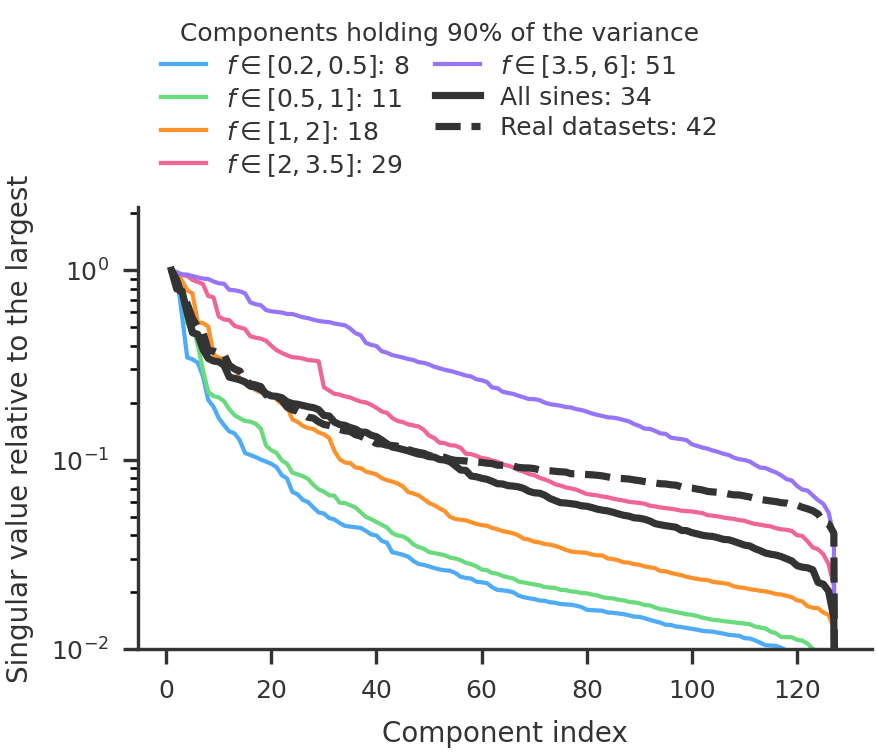}
    \caption{Effective dimension.}
  \end{subfigure}\hfill
  \begin{subfigure}{0.49\textwidth}\centering
    \includegraphics[width=\linewidth]{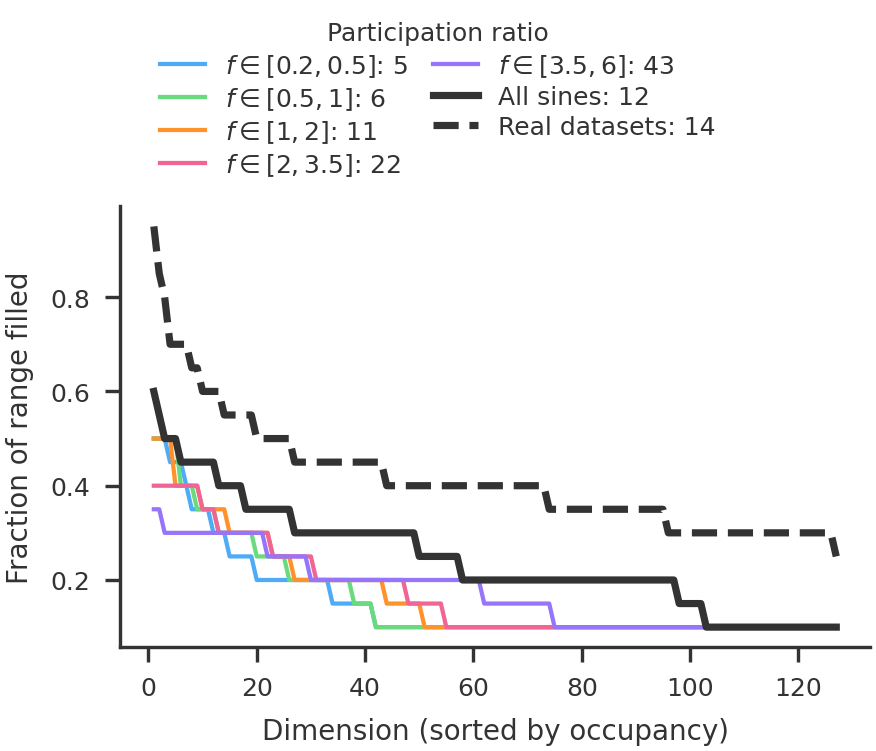}
    \caption{Per-dimension occupancy.}
  \end{subfigure}
  \caption{\textbf{Latent usage} of radial sines in five frequency bands, of their union, and of real fields of $73$ dataset channels (dashed), with $16$ partitions per function and only channels reconstructed to below $15\%$. (a) The singular values of the embeddings of every family in the tangent space at their mean direction, relative to the largest and sorted, with the number of components that hold $90\%$ of the variance. (b) The fraction of the angular range filled in every dimension, sorted, with the participation ratio of every family.}
  \label{fig:usage}
\end{figure}

The spectrum describes one embedding at a time. How much of the sphere a whole family of functions occupies is measured on the single-partition morphologic embeddings of the family, two ways (Figure~\ref{fig:usage}). Since the embeddings lie on a sphere, a principal component analysis of the raw vectors would mix the curvature of the sphere into the variance. They are therefore first mapped to the tangent plane of the sphere at their normalized mean direction $\bar{\emb}_{\mathrm{m}}$ by the logarithmic map, which sends an embedding $\embmorph$ at angle $\theta$ from $\bar{\emb}_{\mathrm{m}}$ to its component orthogonal to $\bar{\emb}_{\mathrm{m}}$ rescaled to length $\theta$. The map preserves the distances to $\bar{\emb}_{\mathrm{m}}$ exactly and the others approximately, as long as the family does not spread over too large a part of the sphere, and the tangent plane is a linear space, where a singular value decomposition of the centered tangent vectors is justified. The number of its components that capture $90\%$ of the variance is the effective dimension of the family. The second measure, the occupancy, expresses every embedding, rotated into the basis of those components, in hyperspherical coordinates, $127$ angles, the first $126$ ranging over $[0,\pi]$ and the last over $[0,2\pi)$, divides the range of every angle into $20$ bins, and counts a bin as filled when an embedding lands in it. The occupancy of a dimension is its fraction of filled bins. For radial sines the effective dimension grows monotonically with the frequency, from $8$ components for the lowest of five bands to $51$ for the highest, while every band fills $15$ to $17\%$ of the angular range of a dimension on average, and the union of the bands needs $34$ components and fills $24\%$. Real fields, $73$ channels of $24$ datasets of incompressible and compressible flow, acoustic waves, the Helmholtz and the Poisson equation, and elasticity, taken together as one family of the same size as the union of the sines, need $42$ components and fill $42\%$ of the angular range. They spread over more of the sphere than the sines do, so the embeddings of fields of other origin are not confined to the region the synthetic functions occupy, and no family fills the sphere alone.

\FloatBarrier
\subsection{Transferable directions}
\label{app:latent-dirs}

A latent space that hosts operators should make at least some transformations of fields simple to express. This part asks whether a transformation of a field, such as a blur or a stretch, corresponds to a fixed direction on the sphere, one that applies the same transformation to any embedding it is added to, rather than to an operation tied to the field it was observed on. If it does, a single example of a transformation defines it everywhere, and the latent space carries operations on fields as well as the fields themselves. The test reads a direction from one example and applies it elsewhere. Two crops of a field that differ by exactly one operation, a blur, an anisotropic stretch, or a split of one feature into two, are each encoded to a single-partition embedding, and the difference $\eta=\embmorph^{B}-\embmorph^{A}$ of their embeddings is read off. The direction is then applied to an unrelated embedding $\embmorph$ by a step along it, scaled by a real coefficient $c$ whose sign sets the sense of the edit, and a projection back onto the sphere,
\begin{equation}
  \embmorph(c) \;=\; \sqrt{\latdim}\,\mathrm{normalize}(\embmorph + c\,\eta),
  \label{eq:latent-dir}
\end{equation}
and the result is decoded. If the decoded field undergoes the same edit as the pair, the operation occupies a consistent direction across the sphere.

\begin{figure}[!htbp]\centering
  \begin{subfigure}{0.49\textwidth}\centering
    \includegraphics[width=\linewidth]{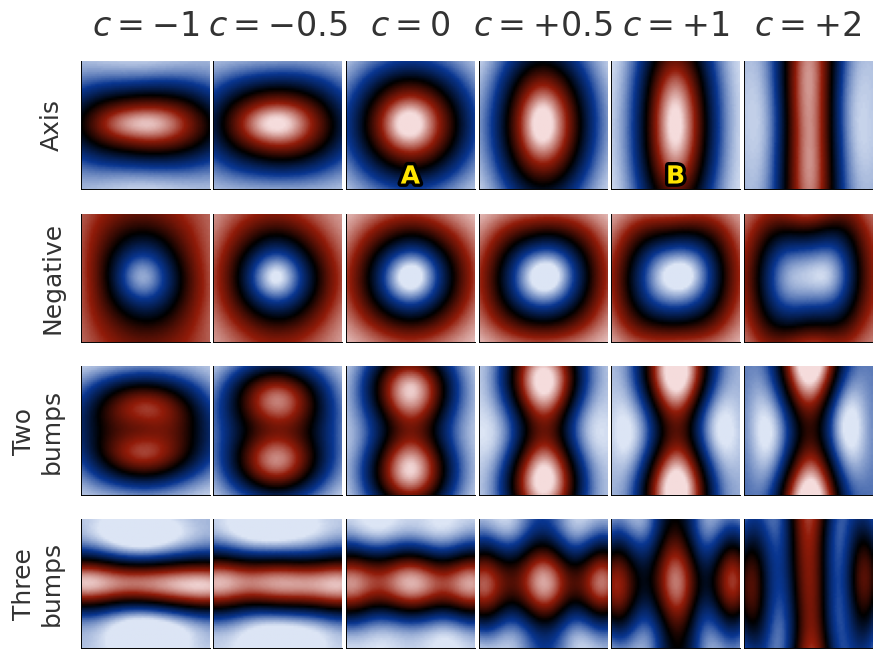}
    \caption{Anisotropic stretch, vertical.}
    \label{fig:dir-anisotropy-vertical}
  \end{subfigure}\hfill
  \begin{subfigure}{0.49\textwidth}\centering
    \includegraphics[width=\linewidth]{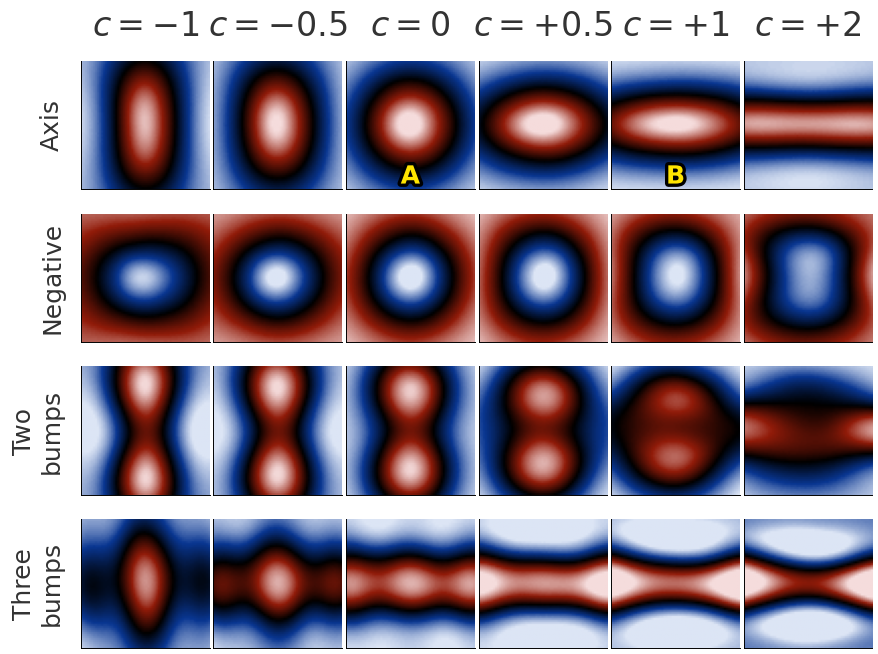}
    \caption{Anisotropic stretch, horizontal.}
    \label{fig:dir-anisotropy-horizontal}
  \end{subfigure}
  \caption{\textbf{Transferred directions, anisotropic stretch.} Each vector, read from a single pair of anchors (top row), turns a round bump into (a) a tall or (b) a wide ellipse. Applied to unrelated embeddings (rows below), it performs the corresponding edit for positive steps $c$, and the opposite edit, a stretch along the other axis, for negative ones.}
  \label{fig:dir-stretch}
\end{figure}

\begin{figure}[!htbp]\centering
  \begin{subfigure}{0.49\textwidth}\centering
    \includegraphics[width=\linewidth]{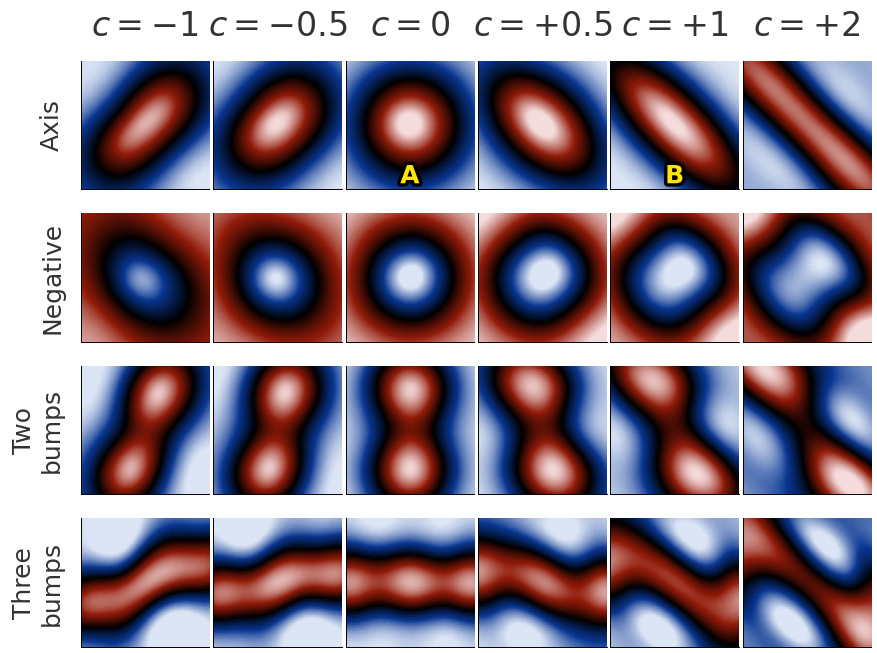}
    \caption{Anisotropic stretch, diagonal.}
    \label{fig:dir-anisotropy-diagonal}
  \end{subfigure}\hfill
  \begin{subfigure}{0.49\textwidth}\centering
    \includegraphics[width=\linewidth]{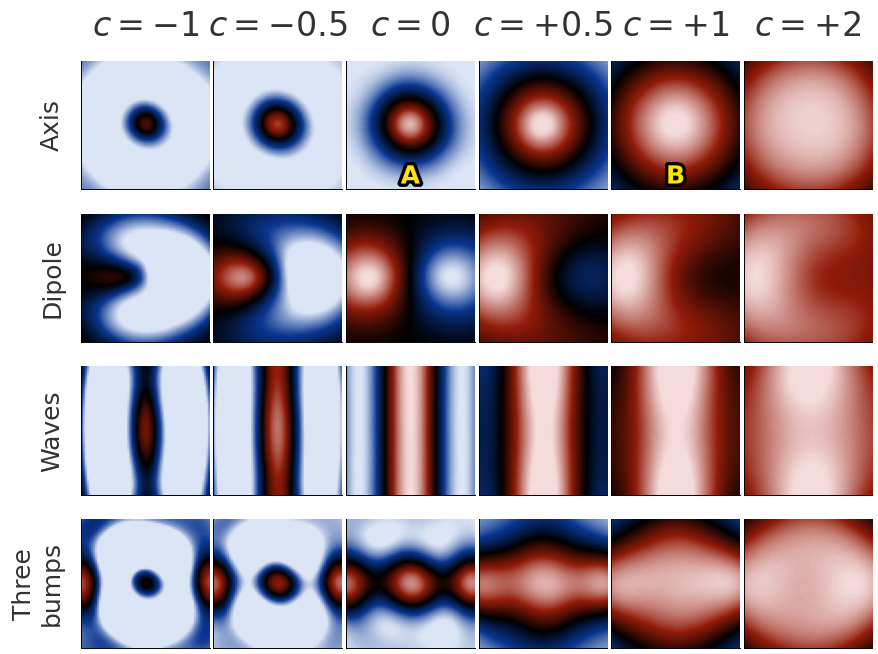}
    \caption{Diffusion.}
    \label{fig:dir-diffusion}
  \end{subfigure}
  \caption{\textbf{Transferred directions, diagonal stretch and diffusion.} (a) A vector read from a single pair of anchors (top row) turns a round bump into an ellipse along the diagonal, and its negative stretches the bump along the other diagonal. (b) A vector that turns a sharp bump into a broad one, and whose negative sharpens it. Applied to unrelated embeddings (rows below), each performs the corresponding edit.}
  \label{fig:dir-diag-diffusion}
\end{figure}

\begin{figure}[!htbp]\centering
  \begin{subfigure}{0.49\textwidth}\centering
    \includegraphics[width=\linewidth]{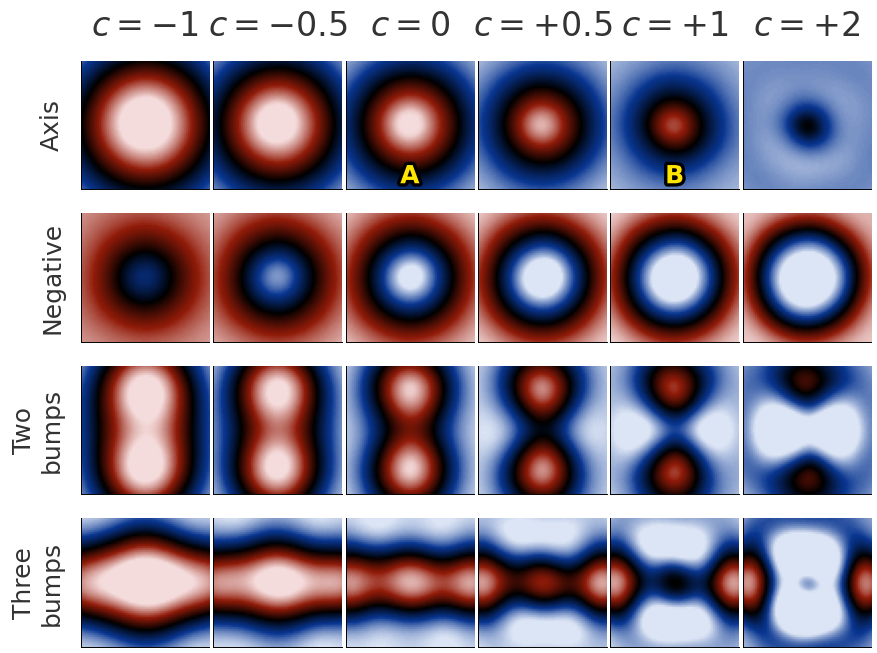}
    \caption{Peakedness.}
    \label{fig:dir-peakedness}
  \end{subfigure}\hfill
  \begin{subfigure}{0.49\textwidth}\centering
    \includegraphics[width=\linewidth]{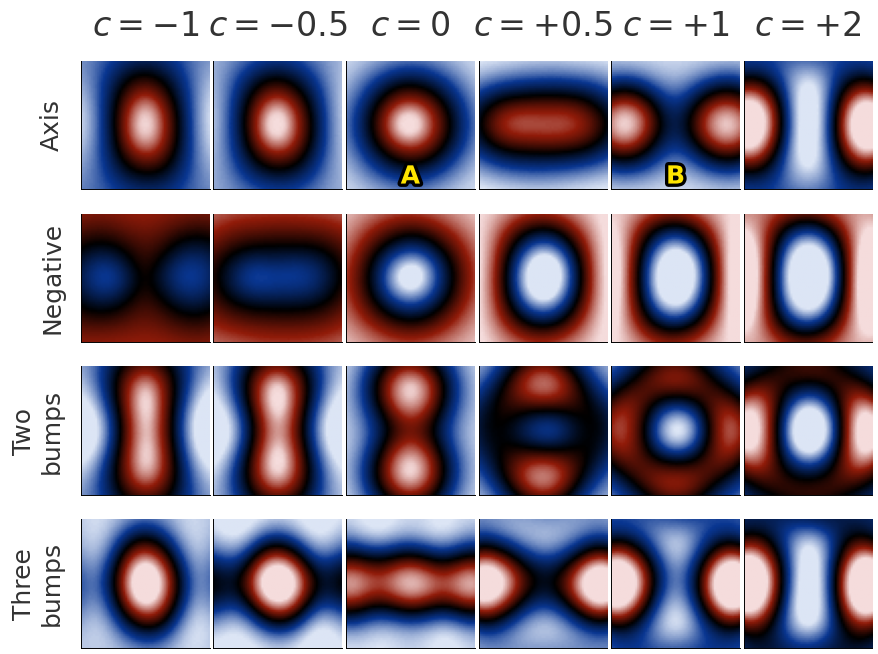}
    \caption{Splitting.}
    \label{fig:dir-splitting}
  \end{subfigure}
  \caption{\textbf{Transferred directions, peakedness and splitting.} (a) A vector read from a single pair of anchors (top row) turns a smooth bump into a sharp peak, and its negative flattens a bump into a plateau. (b) A vector that turns one bump into two, and whose negative merges neighboring bumps. Applied to unrelated embeddings (rows below), each performs the corresponding edit.}
  \label{fig:dir-peak-split}
\end{figure}

Figures~\ref{fig:dir-stretch}--\ref{fig:dir-peak-split} show six such directions. Each was read from one pair of anchors, drawn in the top row of its figure as the base at $c=0$ and the edited field at $c=1$, and is applied with the steps $c$ given above the columns to the unrelated embeddings in the rows below, such as a sign-flipped version of the anchor and fields of two or three bumps. The three anisotropy directions stretch a round bump into a tall, a wide, or a diagonal ellipse, diffusion broadens a sharp bump, peakedness turns a smooth bump into a sharp peak, and splitting divides one bump into two. In every case the vector read from a single pair performs the same edit on the unrelated fields. Extrapolating to $c=2$ continues each edit beyond the second anchor. Stepping to negative $c$, which neither anchor shows, performs the opposite edit, and does so just as coherently. The negated vertical stretch widens a bump, negated diffusion sharpens it, negated peakedness flattens it into a plateau, and the negated split merges two or three bumps into one. A single pair of examples thus defines an operation together with its inverse. The transfer holds for operations that act in place, while an edit that would move structure across the domain does not transfer, for the same reason the geodesic between the states of an evolution fails on transport (Appendix~\ref{app:latent-morph-evo}).

These six directions are ones a human recognizes and can name, and each was found by constructing the corresponding pair of examples. The difference between any two embeddings defines a direction in the same way, so the sphere presumably holds many more transferable directions that correspond to more abstract operations, which no simple description captures and which are not examined here.

\FloatBarrier
\subsection{Geodesic morphing and what it integrates}
\label{app:latent-morph}

\begin{figure}[!htbp]\centering
  \includegraphics[width=\textwidth]{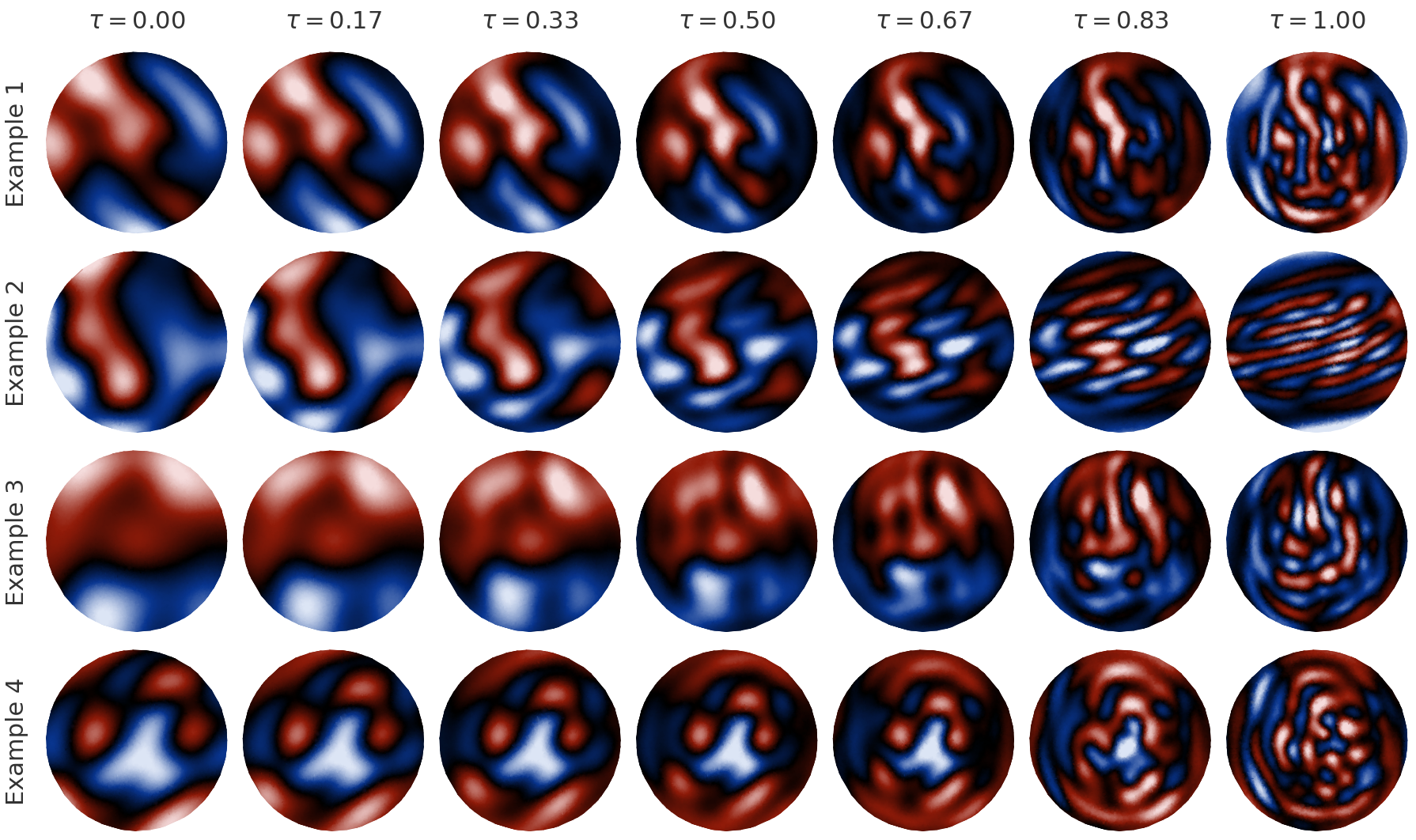}
  \caption{\textbf{Geodesic morphing between two sines.} Decoding the great-circle path between two embeddings, at the fractions $\tau$ above the columns, morphs one field into the other through coherent intermediates, never through noise, with the frequency increasing gradually along the path. Every row is a different pair.}
  \label{fig:latent-morph}
\end{figure}

\begin{figure}[!htbp]\centering
  \includegraphics[width=\textwidth]{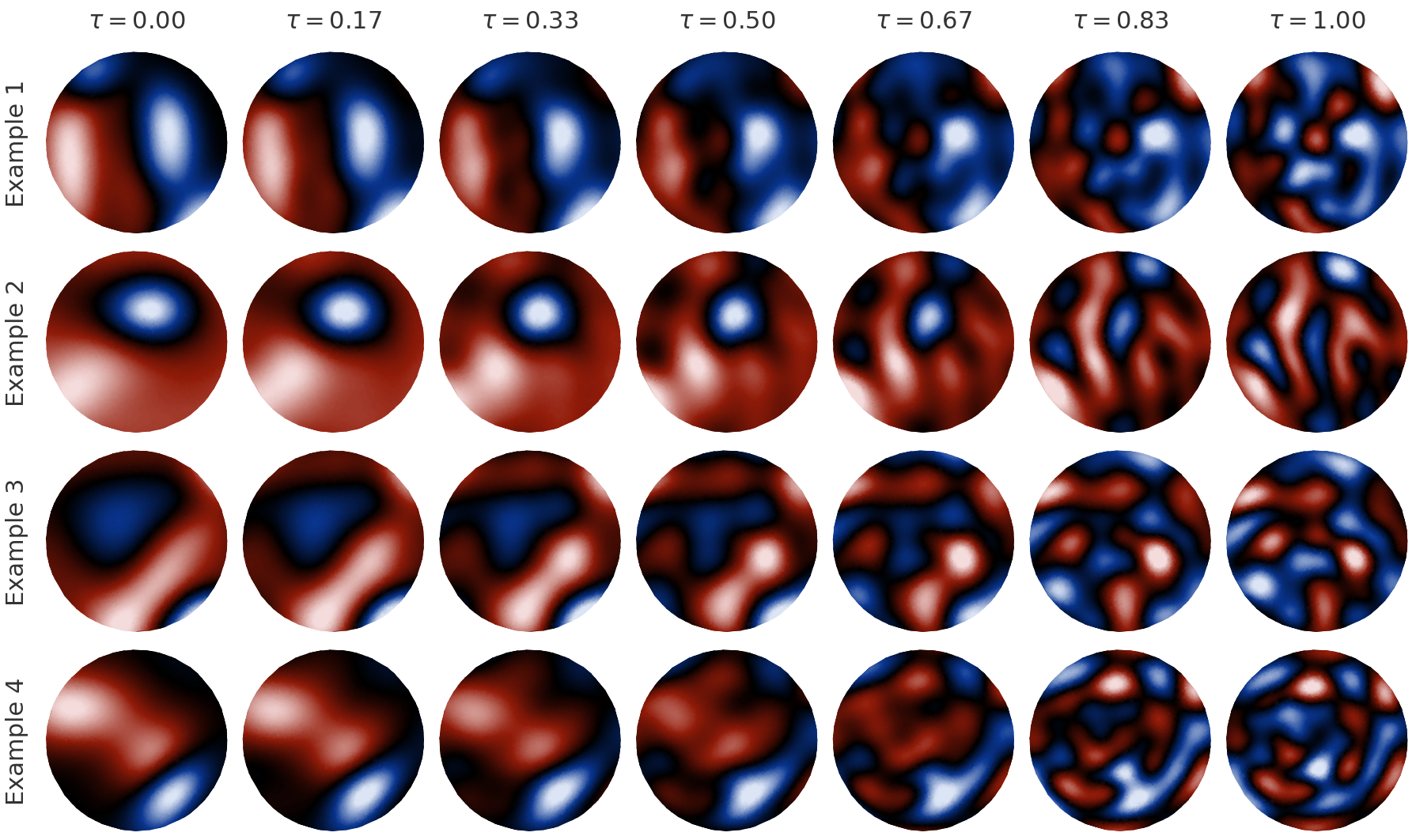}
  \caption{\textbf{Geodesic morphing from a bump to a sine.} A Gaussian bump develops the oscillations of a sine field gradually, never passing through noise or a blank field.}
  \label{fig:morph}
\end{figure}

Embeddings lie on a sphere of radius $\sqrt{\latdim}$, so the great circle, spherical linear interpolation, is the natural path between two of them. For two embeddings $\embmorph^{0}$ and $\embmorph^{1}$ at angle $\theta$ it is $\embmorph(\tau)=\big(\sin((1-\tau)\theta)\,\embmorph^{0}+\sin(\tau\theta)\,\embmorph^{1}\big)/\sin\theta$, which stays on the sphere for every $\tau$, where the straight chord would cut through interior points that the decoder never meets in training. Decoding along it morphs one function into another through coherent intermediates (Figures~\ref{fig:latent-morph} and~\ref{fig:morph}). The transition is gradual, every intermediate lying close to its neighbors. Between a low- and a high-frequency radial sine the frequency increases step by step along the path, ring by ring, rather than switching from one field to the other at some point of the path, and a Gaussian bump develops the oscillations of a sine field in the same gradual manner.

\subsubsection{The geodesic against a physical evolution}
\label{app:latent-morph-evo}

\begin{figure}[!htbp]\centering
  \begin{subfigure}{0.49\textwidth}\centering
    \includegraphics[width=\linewidth]{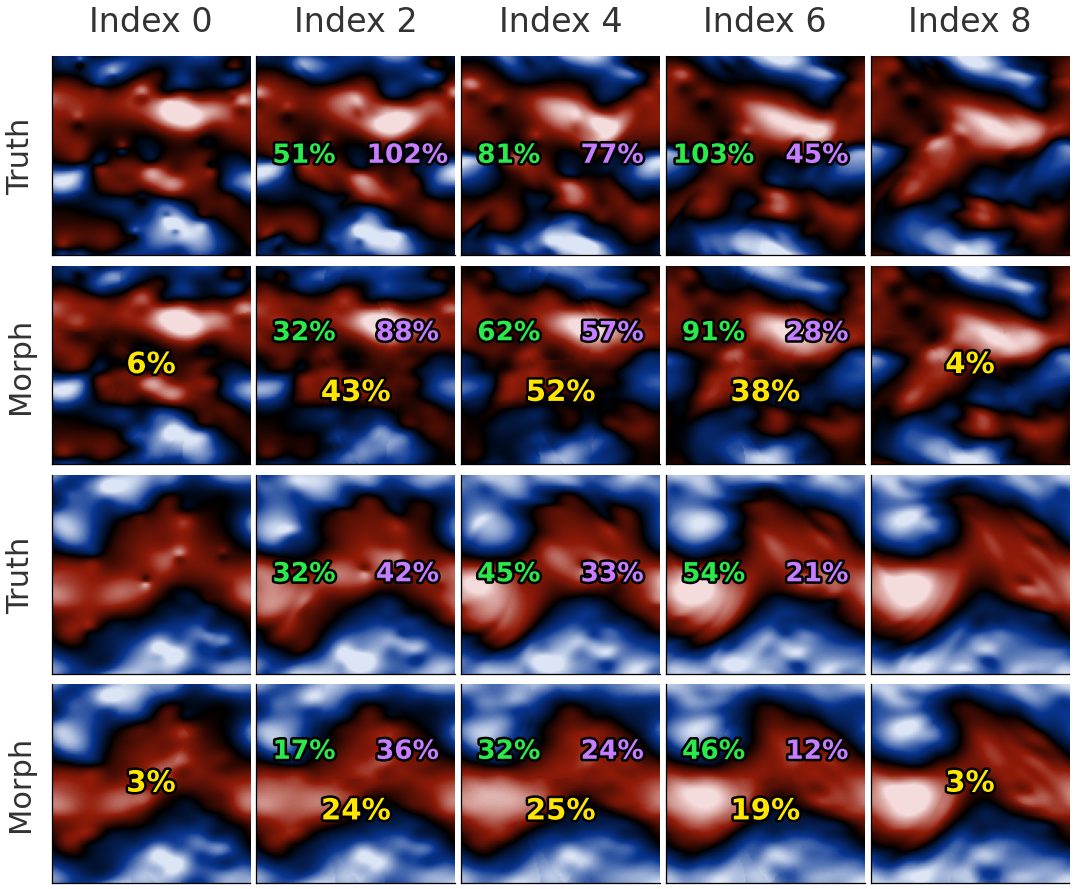}
    \caption{First trajectories.}
  \end{subfigure}\hfill
  \begin{subfigure}{0.49\textwidth}\centering
    \includegraphics[width=\linewidth]{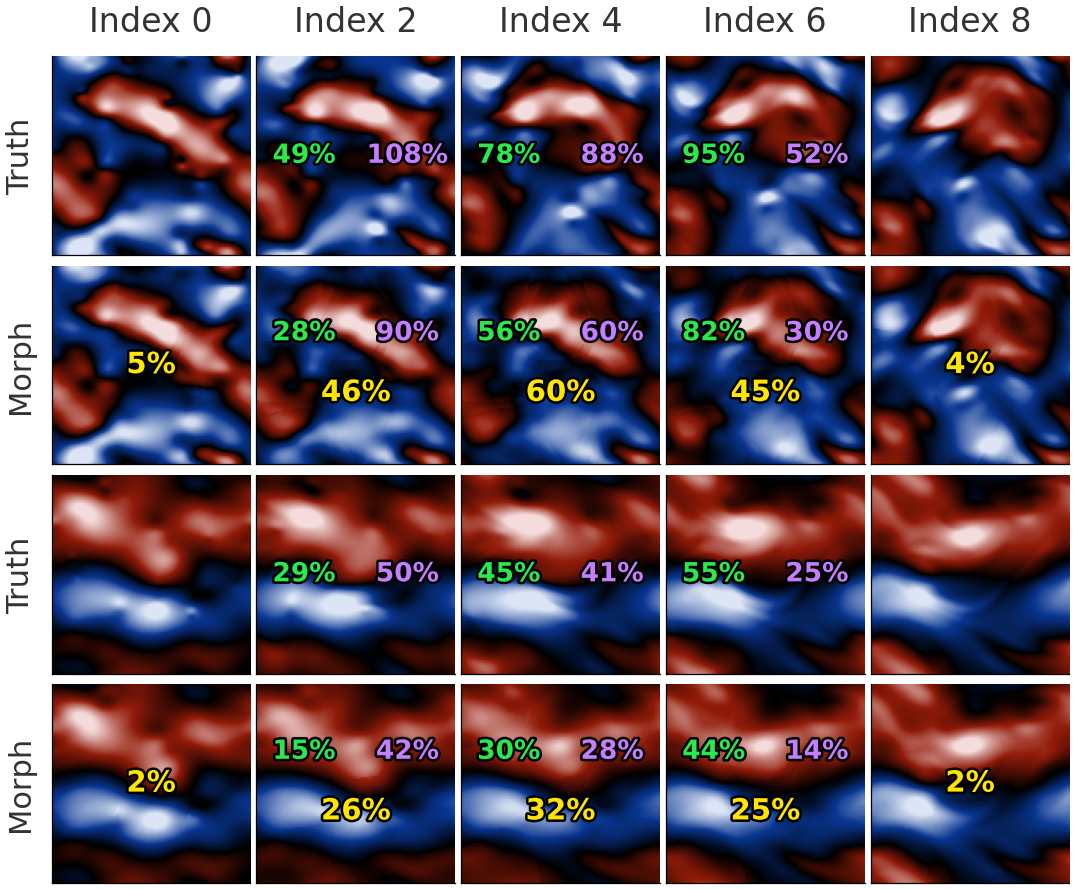}
    \caption{Further trajectories.}
  \end{subfigure}
  \caption{\textbf{Geodesic against true evolution, incompressible flow from a Gaussian initial condition, velocity $v$.} For every trajectory, the true states (Truth) and the decoded geodesic between the embeddings of the first and the last state (Morph) at the time indices above the columns. Every intermediate panel gives the relative $L^2$ distance of its field to the first (green) and to the last (violet) true state, and every Morph panel its distance to the true state at the same time (yellow), the reconstruction error at the endpoints. Only the two endpoints are fixed.}
  \label{fig:morph-evo-gauss-v}
\end{figure}

\begin{figure}[!htbp]\centering
  \begin{subfigure}{0.49\textwidth}\centering
    \includegraphics[width=\linewidth]{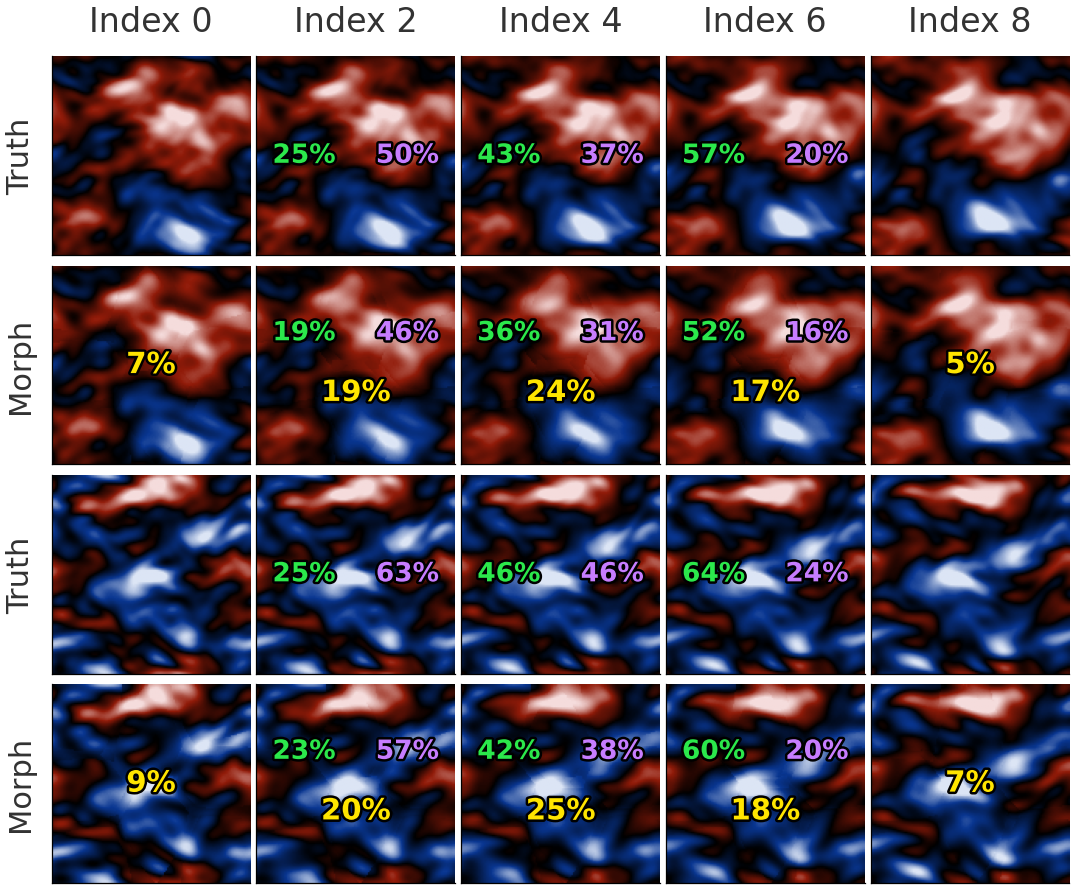}
    \caption{First trajectories.}
  \end{subfigure}\hfill
  \begin{subfigure}{0.49\textwidth}\centering
    \includegraphics[width=\linewidth]{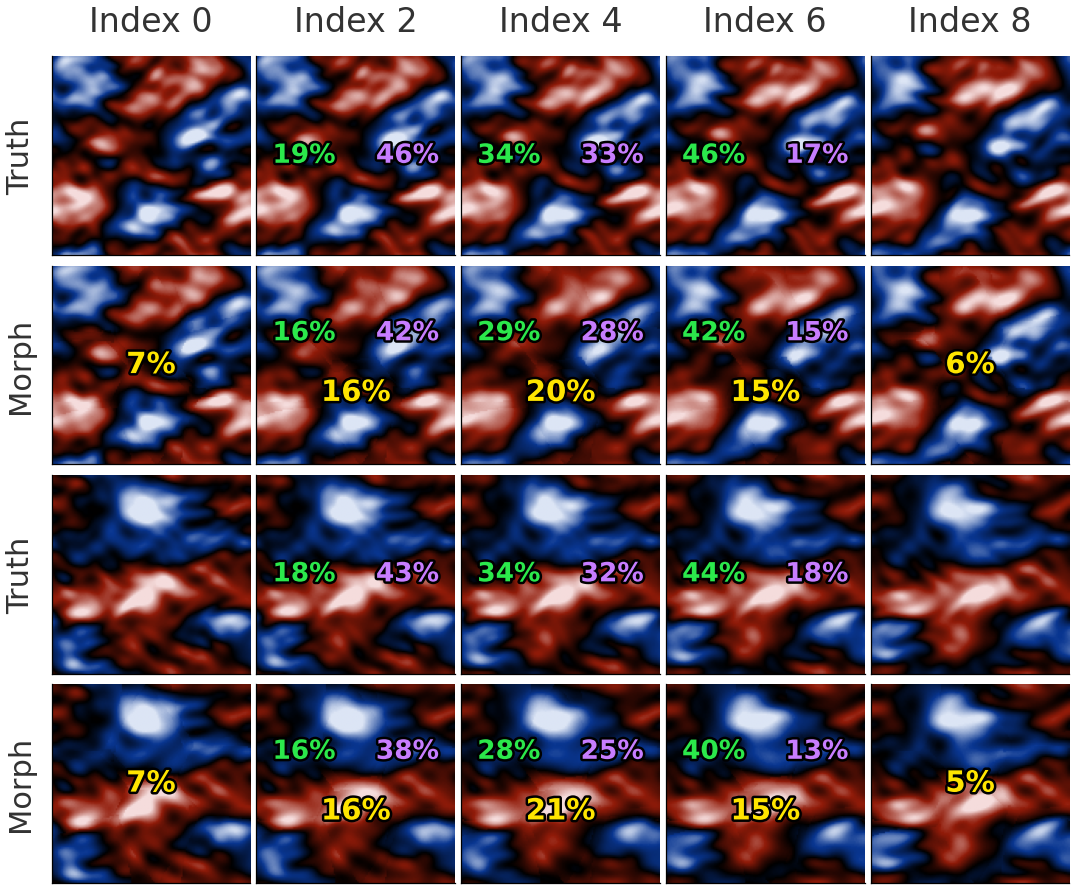}
    \caption{Further trajectories.}
  \end{subfigure}
  \caption{\textbf{Geodesic against true evolution, incompressible flow from a Brownian-bridge initial condition, velocity $v$,} annotated as in Figure~\ref{fig:morph-evo-gauss-v}.}
  \label{fig:morph-evo-bb-v}
\end{figure}

\begin{figure}[!htbp]\centering
  \begin{subfigure}{0.49\textwidth}\centering
    \includegraphics[width=\linewidth]{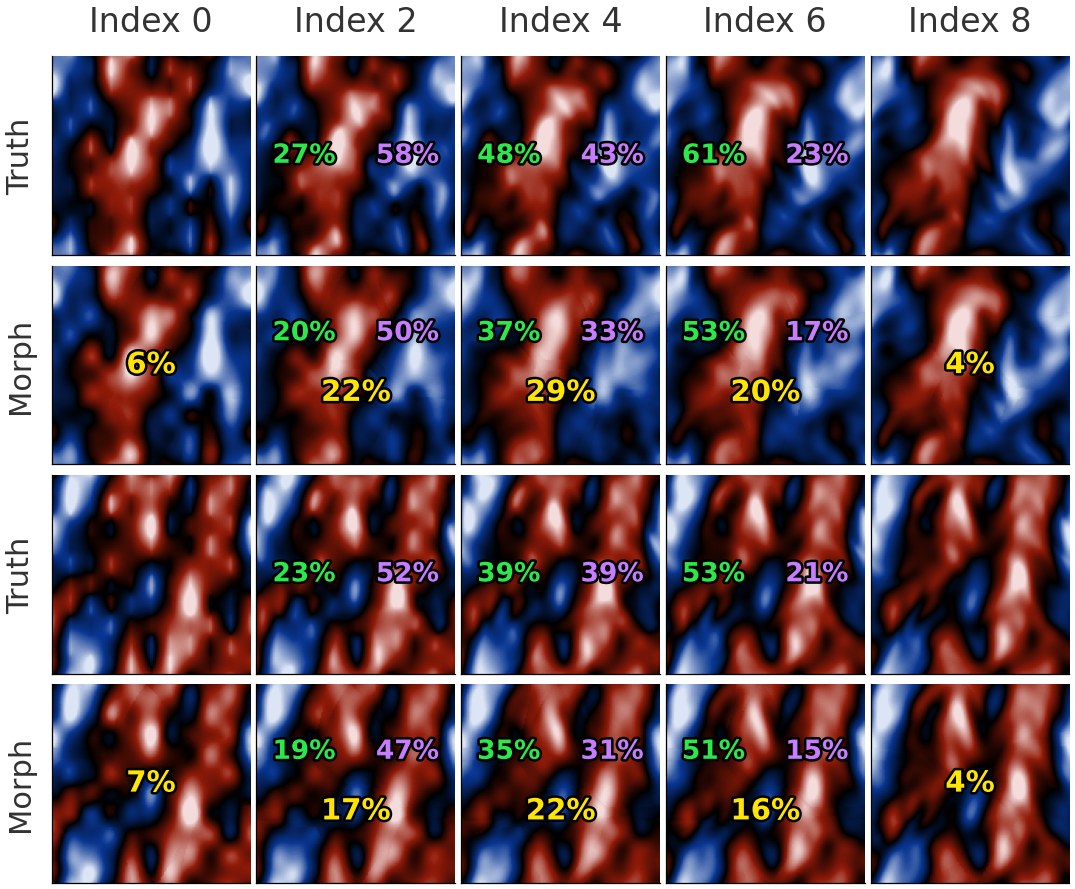}
    \caption{First trajectories.}
  \end{subfigure}\hfill
  \begin{subfigure}{0.49\textwidth}\centering
    \includegraphics[width=\linewidth]{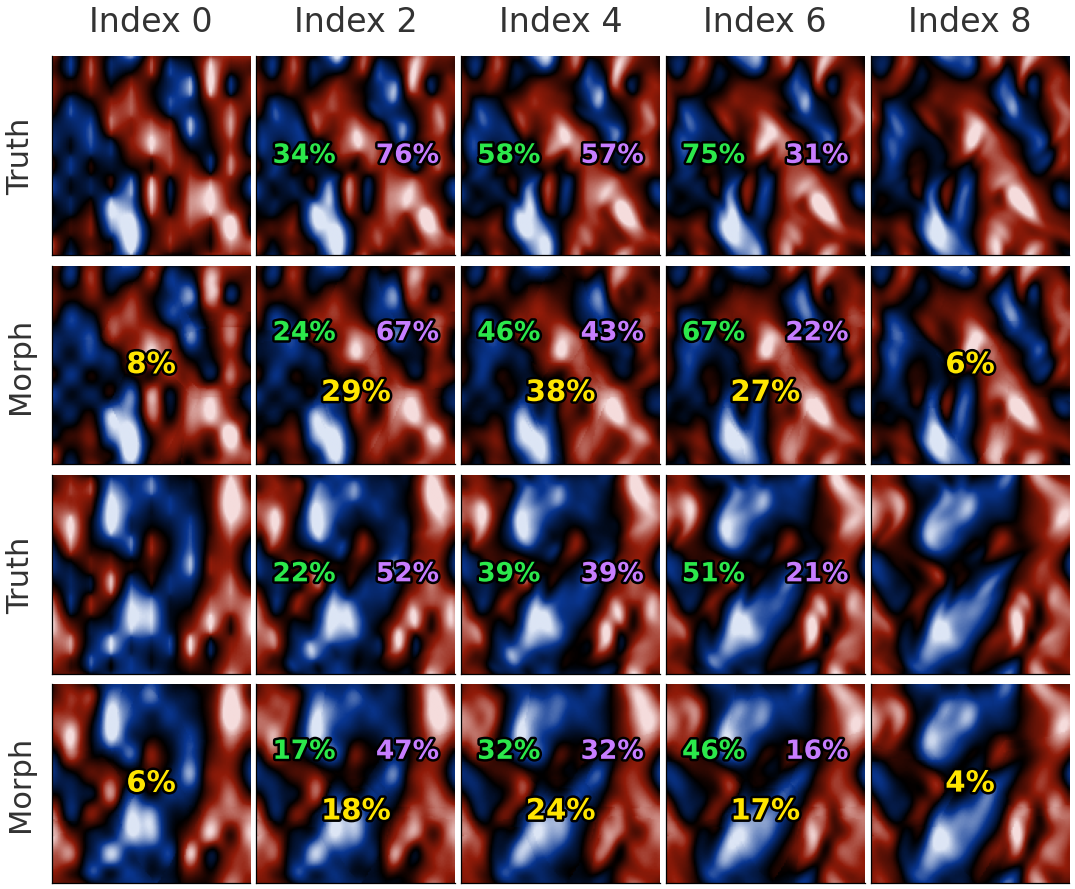}
    \caption{Further trajectories.}
  \end{subfigure}
  \caption{\textbf{Geodesic against true evolution, incompressible flow from a piecewise-constant initial condition, velocity $u$,} annotated as in Figure~\ref{fig:morph-evo-gauss-v}.}
  \label{fig:morph-evo-pwc-u}
\end{figure}

\begin{figure}[!htbp]\centering
  \begin{subfigure}{0.49\textwidth}\centering
    \includegraphics[width=\linewidth]{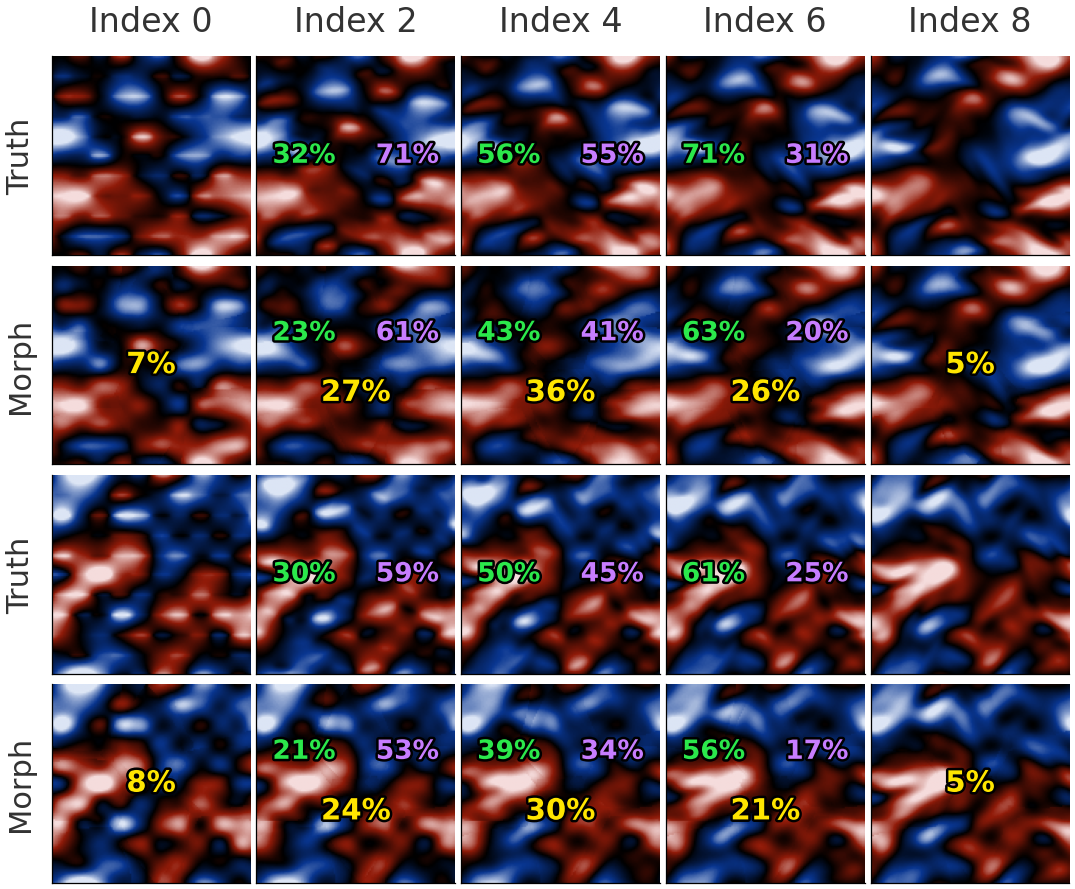}
    \caption{First trajectories.}
  \end{subfigure}\hfill
  \begin{subfigure}{0.49\textwidth}\centering
    \includegraphics[width=\linewidth]{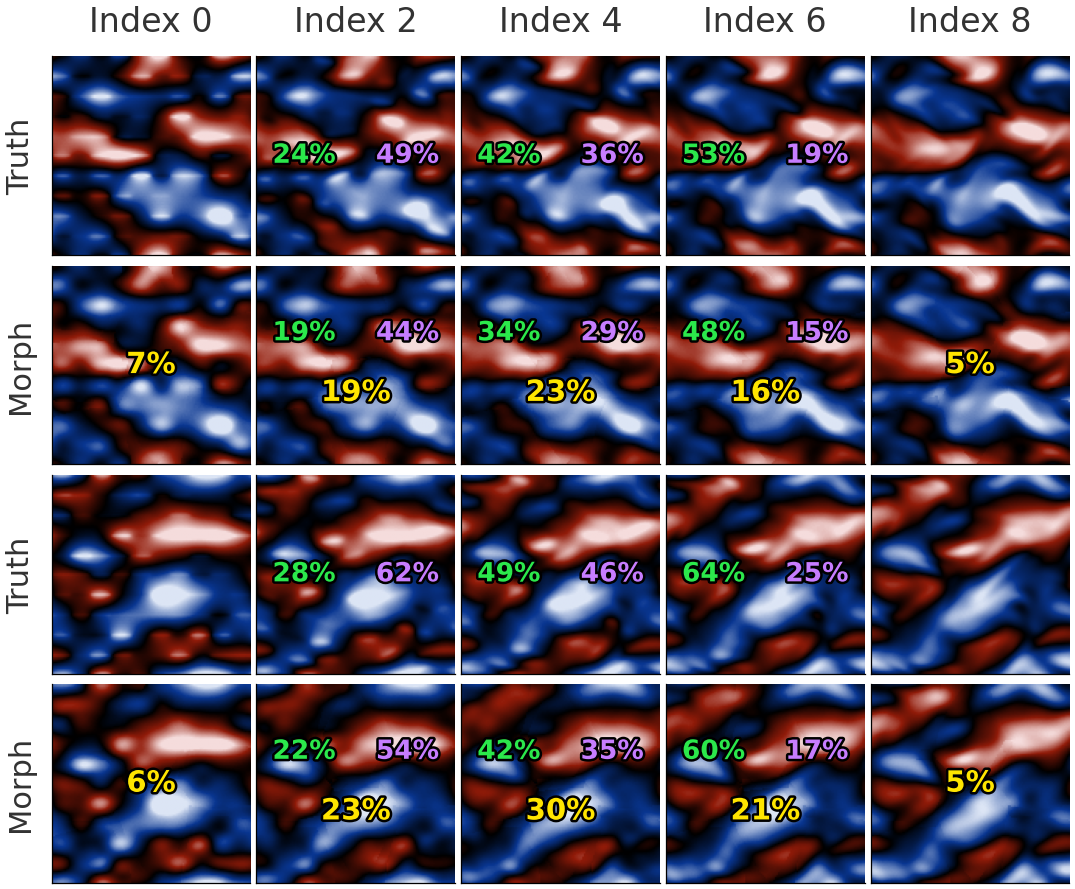}
    \caption{Further trajectories.}
  \end{subfigure}
  \caption{\textbf{Geodesic against true evolution, incompressible flow from a piecewise-constant initial condition, velocity $v$,} annotated as in Figure~\ref{fig:morph-evo-gauss-v}.}
  \label{fig:morph-evo-pwc-v}
\end{figure}

When the endpoints are two states of a physical evolution, the geodesic is unsupervised, since only the two endpoints are fixed, with no time stepping and no labels, and it can be compared with the true intermediate states. Figures~\ref{fig:morph-evo-gauss-v}--\ref{fig:morph-evo-pwc-v} do so for incompressible flows from three families of initial conditions. Every panel carries relative $L^2$ distances, green to the first endpoint, violet to the last one, and, on a decoded state of the geodesic, yellow to the true state at the same time, which at the two endpoints is the reconstruction error. The green and the violet number of a true state measure how far the evolution has moved from either endpoint. They are large in every figure, so the two endpoints differ markedly, the evolution between them is substantial, and neither endpoint is a good stand-in for the states between them, which makes the interpolation a hard test. A yellow number below or comparable to those two numbers therefore means that the decoded geodesic lies at least as close to the true state as either endpoint does. It lies below them at every intermediate time, with a single tie, and most clearly at the middle of the interval, where the true state is about as far from one endpoint as from the other. The agreement requires the change between the endpoints to be local. We observed that it does not hold for transport equations, such as the acoustic wave equation, whose solutions carry structure across the domain.

\subsubsection{How far a latent geodesic carries a physical evolution}
\label{app:timeinterp}

The comparison above can be turned into a measurement over many trajectories. For a pair of snapshots $\Delta t$ time indices apart we encode both, move halfway, decode, and compare against the true state at the midpoint index. Only the two anchors are ever seen, and nothing is fitted or stepped.
Halfway is taken component-wise on the two parts of the embedding. The morphologic part follows the great circle with its norm blended linearly. The static features are blended linearly in the units the encoder stores them in. The two anchors share a point cloud and a partitioning, so the coordinate frame $(\mathbf{c}_p,\mathbf{s}_p,\theta_p)$ is identical at both ends and its interpolation is an identity, which leaves the value frame as the part that moves. There, $\mu_p$, a partition's value midpoint, interpolates arithmetically, while $\sigma_p$ is stored as a logarithm, so a linear blend interpolates the partition's half-width \emph{geometrically} and the amplitude sweeps multiplicatively rather than through zero.

Every admissible pair at a given $\Delta t$ contributes, and we report the per-trajectory mean reduced by a median over $32$ trajectories per dataset, each read at every time index, with $P=128$ partitions, $N=16384$ points, and an overlap that adds half to the number of point-to-partition assignments. Two references bound the curve. The first is the \emph{anchor similarity}, the mean relative $L^2$ distance of the true midpoint to the two \emph{true} anchors, which is what keeping an anchor instead of interpolating would cost, and which is plotted as a second curve per dataset. The second is the \emph{reconstruction floor}, the error of encoding and decoding the true states, averaged over all time indices, which no interpolation can go below, and which is written as a corner note rather than a line.

\begin{figure}[!htbp] \centering
  \includegraphics[width=0.6\textwidth]{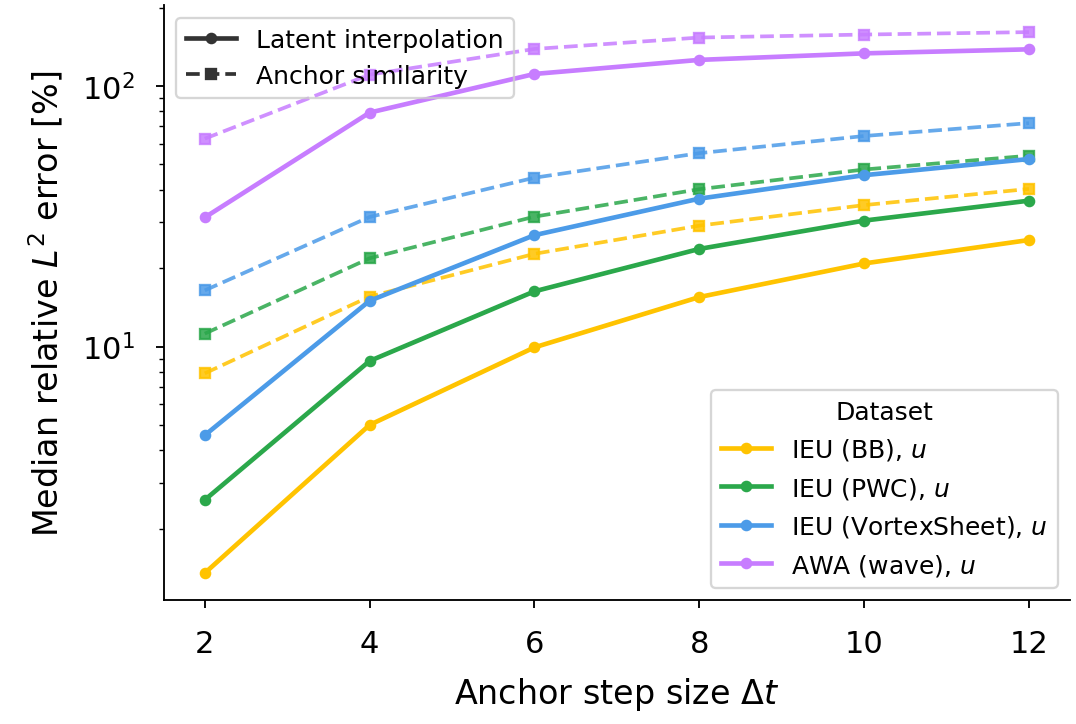}
  \caption{\textbf{The latent midpoint across datasets.} Color is the dataset, solid the latent interpolation, and dashed its anchor-similarity baseline. Every dataset stays below its own baseline, and the gap narrows as $\Delta t$ grows. Three incompressible-flow datasets and the acoustic wave equation, 32 trajectories each, $P=128$, $N=16384$, all 21 time indices (16 for AWA), up to $\Delta t=12$.}
  \label{fig:time-interp-all}
\end{figure}

The measurement covers eleven evolution datasets, five of incompressible flow, five of compressible flow and one of acoustic waves, and Figure~\ref{fig:time-interp-all} collects four of them on one set of axes. At $\Delta t=2$ the interpolant sits far below its baseline, at $1.4\%$ against $7.9\%$ on the incompressible Brownian-bridge data, only a few times its floor of $0.32\%$, and at $4.6\%$ against $16.4\%$ on the vortex sheet. The margin closes as the interval lengthens, and at $\Delta t=20$, beyond the range of the figure, it reaches $39.8\%$ against $55.9\%$ and $68.4\%$ against $87.5\%$. No dataset crosses its baseline at any interval. The compressible Euler datasets keep the narrowest margins without losing them, and even compressible Riemann density, with its traveling shocks, starts at $9.7\%$ against $15.7\%$ and still sits below its baseline at $\Delta t=20$, at $41.0\%$ against $45.2\%$.

\subsubsection{Extrapolation past an anchor}
\label{app:latent-extrap}

\begin{figure}[!htbp]\centering
  \begin{subfigure}{0.49\textwidth}\centering
    \includegraphics[width=\linewidth]{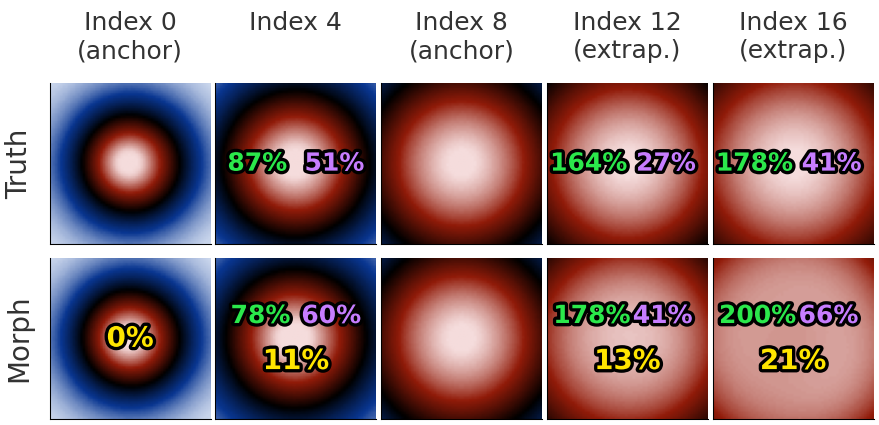}
    \caption{Isotropic diffusion.}
  \end{subfigure}\hfill
  \begin{subfigure}{0.49\textwidth}\centering
    \includegraphics[width=\linewidth]{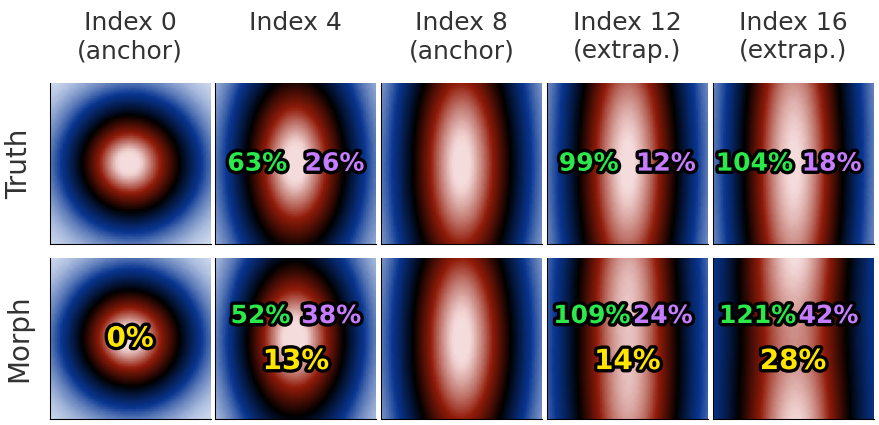}
    \caption{Anisotropic diffusion.}
  \end{subfigure}
  \caption{\textbf{Extrapolating past an anchor, isotropic and anisotropic diffusion.} For every evolution, the true states (top) and the geodesic between the embeddings of the indices $0$ and $8$, continued past the second anchor to the indices $12$ and $16$ (Morph, bottom). The numbers are relative $L^2$ distances, to the first (green) and the last (violet) anchor, and of the decoded field to the true state (yellow). Continuing the geodesic keeps following the true evolution beyond the second anchor.}
  \label{fig:extrap}
\end{figure}

\begin{figure}[!htbp]\centering
  \begin{subfigure}{0.49\textwidth}\centering
    \includegraphics[width=\linewidth]{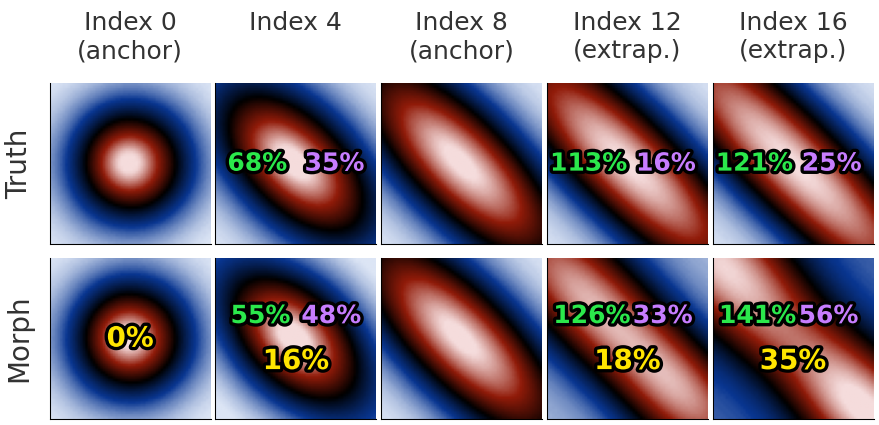}
    \caption{Diagonal diffusion.}
  \end{subfigure}\hfill
  \begin{subfigure}{0.49\textwidth}\centering
    \includegraphics[width=\linewidth]{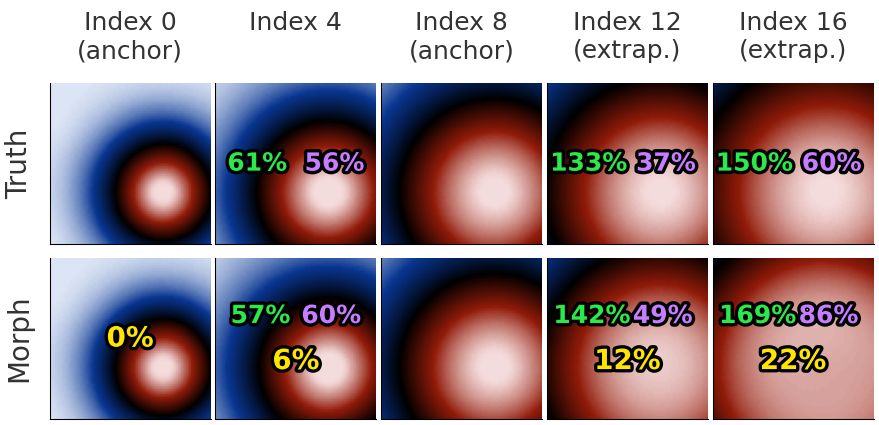}
    \caption{Off-center diffusion.}
  \end{subfigure}
  \caption{\textbf{Extrapolating past an anchor, diagonal and off-center diffusion,} annotated as in Figure~\ref{fig:extrap}.}
  \label{fig:extrap-b}
\end{figure}

\begin{figure}[!htbp]\centering
  \begin{subfigure}{0.49\textwidth}\centering
    \includegraphics[width=\linewidth]{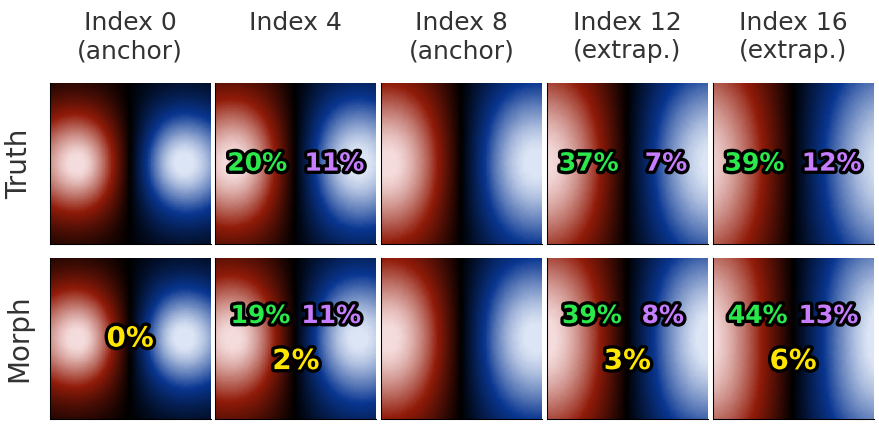}
    \caption{A spreading dipole.}
  \end{subfigure}\hfill
  \begin{subfigure}{0.49\textwidth}\centering
    \includegraphics[width=\linewidth]{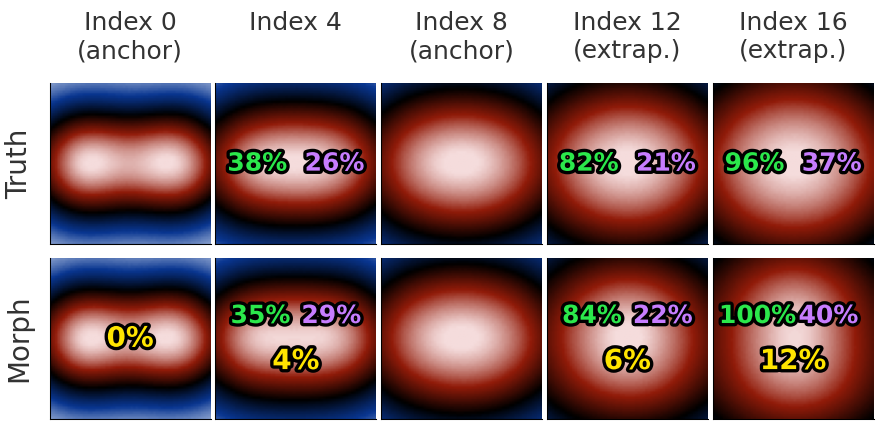}
    \caption{Merging bumps.}
  \end{subfigure}
  \caption{\textbf{Extrapolating past an anchor, a spreading dipole and merging bumps,} annotated as in Figure~\ref{fig:extrap}.}
  \label{fig:extrap-c}
\end{figure}

A geodesic need not stop at its second endpoint. Continuing it beyond the second anchor, at the constant angular speed set by the two anchors, is the simplest extrapolation there is, linear in the angle along the great circle, and it uses no information beyond the two anchors. On manufactured evolutions with exact solutions, taken on a single partition and compared in the frame-free morphology, this linear continuation carries the evolution well past the second anchor (Figures~\ref{fig:extrap}--\ref{fig:extrap-c}). Isotropic, anisotropic, diagonal, and off-center diffusion, a spreading dipole, and two merging bumps are anchored at the time indices $0$ and $8$ and extrapolated to the indices $12$ and $16$, half the spacing of the anchors and the full spacing beyond the second one. At the index $12$ the extrapolated field errs by $3$ to $18\%$, and it is closer to the true state than the last anchor on four of the six evolutions, by a factor of two to three and a half for isotropic and off-center diffusion, the dipole, and the merging bumps, and about as close on the other two. At the index $16$, as far beyond the second anchor as that anchor is beyond the first, the error is $6$ to $34\%$, still below that of keeping the last anchor on the same four evolutions. On these four the extrapolated field is closer to a state that neither anchor shows than the nearer anchor is, using nothing but the two anchors. Their true path in the latent space is therefore close to the continued geodesic, though not equal to it. Nonlinear and more complex evolutions are not expected to follow a geodesic so closely, and capturing them is what the processor is trained for. That a straight continuation on the sphere predicts states that neither anchor shows indicates that a local evolution traces a path of nearly constant direction in the latent space, which the latent space thus allows to be extrapolated, and not only interpolated. The extrapolation is limited to evolutions that change structure in place. The time interpolation above degrades on shocked compressible flow, where a discontinuity travels across the domain, and a transferred direction applies an edit that acts in place but not one that moves structure (Appendix~\ref{app:latent-dirs}), and a continued geodesic likewise cannot transport structure.

\FloatBarrier
\subsection{Family embeddings}
\label{app:latent-families}

\begin{figure}[!htbp]\centering
  \begin{subfigure}{0.325\textwidth}\centering
    \includegraphics[width=\linewidth]{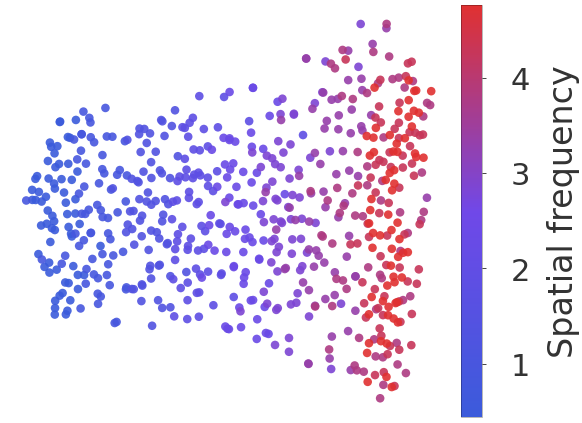}
    \caption{t-SNE over ten frequency bands.}
    \label{fig:tsne-freq}
  \end{subfigure}\hfill
  \begin{subfigure}{0.325\textwidth}\centering
    \includegraphics[width=\linewidth]{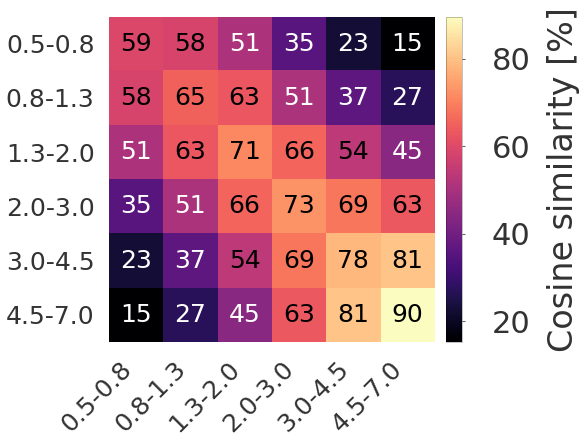}
    \caption{Cosine similarity between the bands.}
    \label{fig:crossfam}
  \end{subfigure}\hfill
  \begin{subfigure}{0.325\textwidth}\centering
    \includegraphics[width=\linewidth]{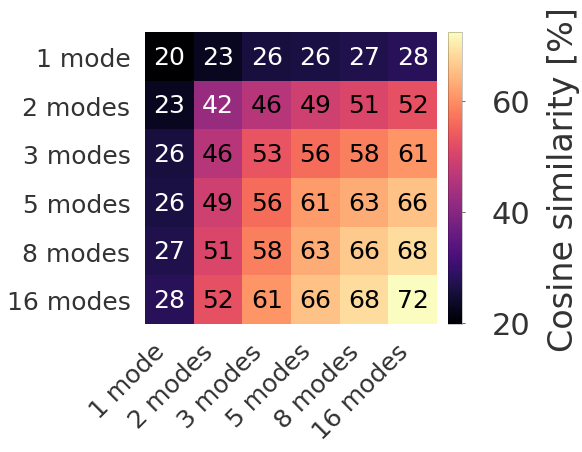}
    \caption{Cosine similarity across mode counts.}
    \label{fig:cf-modes}
  \end{subfigure}
  \caption{\textbf{Family embedding of radial sines.} (a) t-SNE under the chord distance of one embedding per function, colored by the spatial frequency of its band, ordered into a smooth frequency gradient, so the embeddings are arranged by the frequency of their fields. (b) The cosine similarity between the mean embeddings of six frequency bands, falling off smoothly with the distance between the bands, the same arrangement read from the means. (c) The cosine similarity between the mean embeddings of radial sines with different numbers of modes, at frequencies from $1$ to $3$, a weaker axis than the frequency, along which sines with more modes are more similar to one another.}
  \label{fig:tsne}
\end{figure}

How the latent space is organized globally is read from many embeddings at once, one per function on a single partition. Placed under t-distributed stochastic neighbor embedding (t-SNE) with the chord distance, mixed radial sines from ten frequency bands order continuously by spatial scale into a smooth frequency gradient, with the highest band splitting off (Figure~\ref{fig:tsne-freq}). The embeddings are thus arranged on the sphere by the frequency content of their fields, although nothing in training labels the frequency. The cosine similarity between the mean embeddings of six frequency bands shows the same organization in a different form (Figure~\ref{fig:crossfam}). It falls off smoothly with the distance between two bands, so that neighboring bands share most of their mean direction and distant bands little of it, and spatial frequency is the dominant axis of the arrangement.

\begin{figure}[!htbp]\centering
  \includegraphics[width=\textwidth]{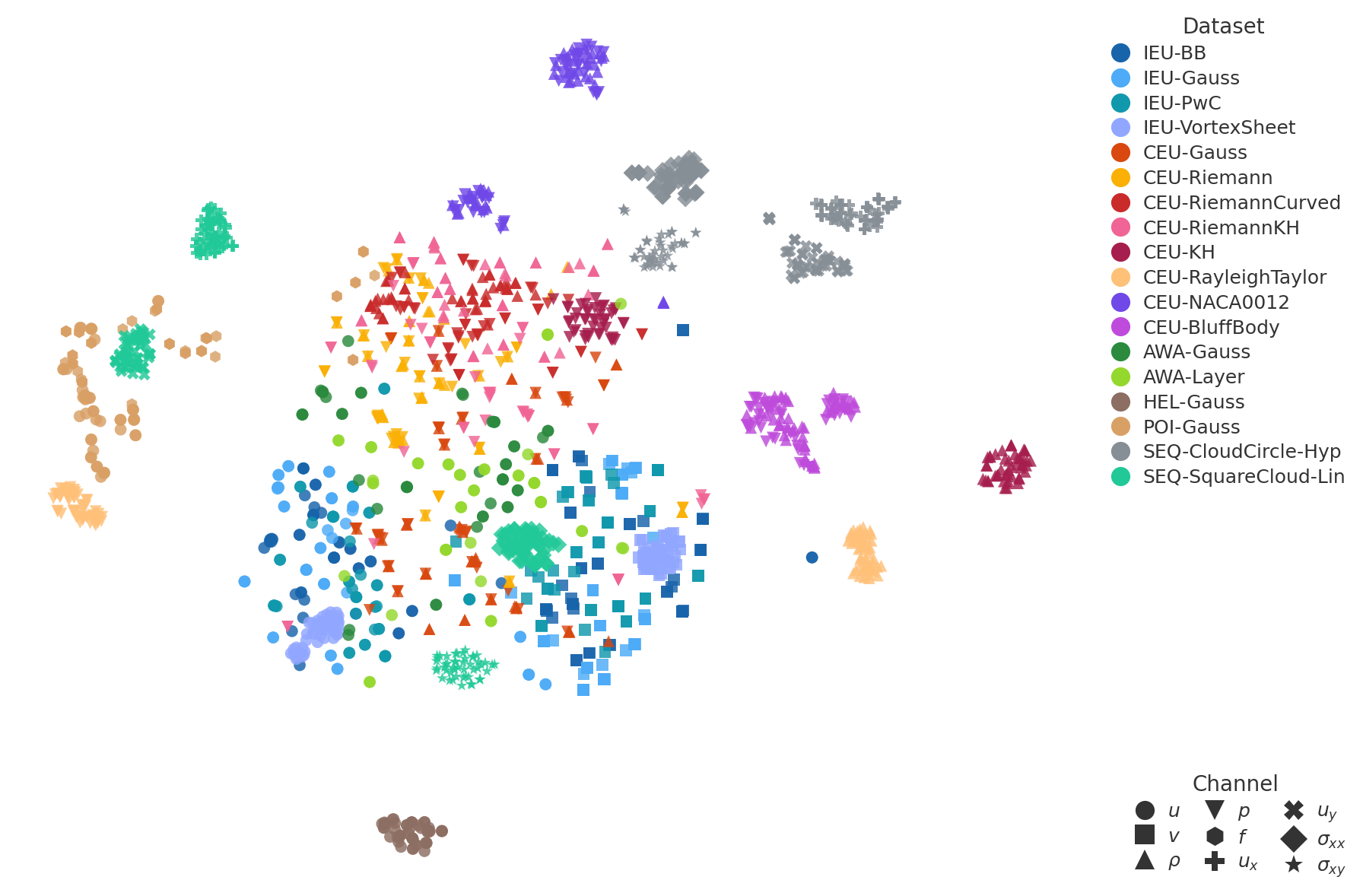}
  \caption{\textbf{Family embedding of real fields.} t-SNE under the chord distance of one embedding per function on a single partition, for $48$ samples of each of $37$ channels of $18$ datasets of incompressible and compressible flow, acoustic waves, the Helmholtz and the Poisson equation, and elasticity, colored by dataset and marked by channel.}
  \label{fig:tsne-variants}
\end{figure}

The number of modes is a weaker axis (Figure~\ref{fig:cf-modes}). Its similarities lie closer together than those across frequency bands, and they grow with the number of modes, from $20\%$ between functions of one mode to $72\%$ between functions of sixteen. Sines with more modes are thus more similar to one another in the latent space, as they are in the physical space, where a sum of many modes at similar frequencies approaches a statistically homogeneous texture and different draws look alike, whereas a single mode is set apart by its own orientation and phase. Real physical fields organize the same way, into coherent families by dataset and channel (Figure~\ref{fig:tsne-variants}).

\FloatBarrier
\clearpage

\section{Operator learning}
\label{app:operators}
\setcounter{figure}{0}
\setcounter{table}{0}

This appendix describes how operators are learned on a frozen backbone, what freezing the backbone makes possible, and how the resulting operators compare in accuracy and in efficiency with baselines trained end to end. It first presents the training strategies, namely the two training objectives, the frozen and the unfrozen backbone, and the caching that a frozen backbone admits (Appendix~\ref{app:operator-loss}), then the phases of a training run and the cooldown that closes it (Appendix~\ref{app:operator-phases}), and finally the accuracy benchmark and the performance benchmark (Appendices~\ref{app:benchmark} and~\ref{app:perf}). Every operator in this appendix is learned on the low-dimensional backbone, whose morphological embeddings have eight dimensions (\hyperref[app:overview]{appendix prelude}). It is cheaper to train on in the physical and the end-to-end modes than the analysis backbone, at the price of more partitions for the same accuracy.

\FloatBarrier
\subsection{Training strategies}
\label{app:operator-loss}
\label{app:operator-terms}

\begin{figure}[!htbp]\centering
  \begin{subfigure}{0.44\textwidth}\centering
    \includegraphics[width=\linewidth]{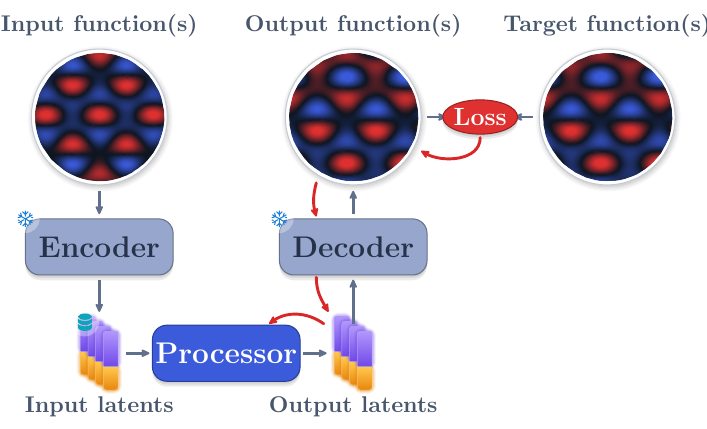}
    \caption{Physical loss.}
  \end{subfigure}
  \hspace*{0.06\textwidth}
  \begin{subfigure}{0.44\textwidth}\centering
    \includegraphics[width=\linewidth]{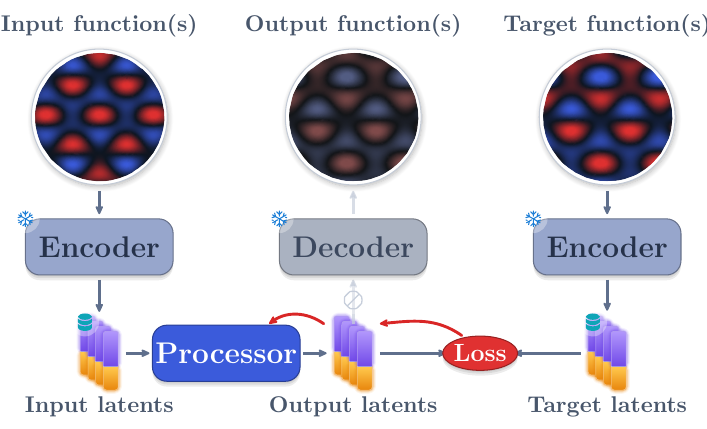}
    \caption{Latent loss.}
  \end{subfigure}
  \caption{\textbf{Training on a frozen backbone.} Only the processor trains, the snowflakes mark the frozen encoder and decoder, and the red arrows mark the backward pass. (a) Under the physical loss the decoder maps the predicted embeddings back to a field, which is compared with the target field. (b) Under the latent loss the prediction is compared with the encoded target, so the decoder is not used in training. The cylinders mark the embeddings that can be computed once and cached (Appendix~\ref{app:caching}).}
  \label{fig:downstream}
\end{figure}

An operator maps input functions to output functions, and operator learning approximates such a map from pairs of input and output functions \citep{kovachki2023neural}. With a pre-trained backbone the operator is learned in the latent space of the backbone rather than on the functions themselves (Figure~\ref{fig:downstream}). For a training pair, the encoder maps the input fields, on their partitions, to the input embeddings, and the target fields, on the same partitions, to the target embeddings. A trainable network, the processor, maps the input embeddings to output embeddings, one per partition, which the decoder turns back into fields. In this work, the processor is a windowed transformer, and other architectures can take its place (Appendix~\ref{app:architecture}).

\paragraph{Training objectives.} The processor can be trained on two kinds of objectives. A \emph{physical loss} is evaluated on fields. The output embeddings are decoded at the points of the target field, and the decoded field is compared with the target field by an error of fields. A \emph{latent loss} is evaluated on embeddings and decodes nothing. It compares the output embeddings with the target embeddings directly, by a distance in the latent space. The physical loss measures the error that matters, while the latent loss is a proxy for it, and a faithful one only to the extent that distances between embeddings track distances between the decoded fields (Appendix~\ref{app:latent-metric}). We call the excess physical error of a processor trained on the latent loss over that of a processor trained on the physical loss the \emph{proxy gap}. It depends on the task and on how closely the latent distance follows the error of the fields, and Appendix~\ref{app:operator-cooldown} describes how a final phase on the physical loss closes it. The two losses used in this work are defined in the \hyperref[app:overview]{appendix prelude}.

\paragraph{Frozen and unfrozen backbone.} The backbone can be kept \emph{frozen} or be \emph{unfrozen}. A frozen backbone keeps the pre-trained weights of the encoder and the decoder, and only the processor is trained. An unfrozen backbone trains the encoder and the decoder together with the processor, so the backbone is fine-tuned on the task at hand. An unfrozen encoder moves the embeddings it produces, so an encoded target does not stay fixed during training, and the unfrozen backbone is therefore combined with the physical loss only, whereas a frozen backbone admits both objectives. This gives three training regimes that share the processor, a frozen backbone under the latent loss, a frozen backbone under the physical loss, and an unfrozen backbone under the physical loss. They differ in what a step computes and in what can be cached, and therefore widely in their cost.

\subsubsection{Input and output caching}
\label{app:caching}
\label{app:caching-modes}

With the backbone frozen, the partitioner, the encoder, and the decoder are fixed maps, so whatever they compute from the training data alone can be computed once, before training, and stored. Two conditions make this possible. The first is architectural. The physical space must be crossed exactly twice, once by the encoder at the input and once by the decoder at the output, so that everything in between acts on embeddings that the training data determine once and for all. An architecture that never leaves the physical space, or that returns to it at every layer, has no such embeddings to store. The second condition concerns the output cache only. A cached target is an embedding, and training on it means training on the latent loss, which stands in for the physical error faithfully only if nearby embeddings decode to nearby fields (Appendix~\ref{app:latent-metric}).

What can be cached depends on the training regime, and Figure~\ref{fig:downstream} marks it with cylinders. Under the physical loss with a frozen backbone (Figure~\ref{fig:downstream}a), the partitions and the input embeddings of the whole training set can be stored in an \emph{input cache}, together with the coordinates, the partition assignments, and the target values that the decoder needs to evaluate the loss. A step then reads the input embeddings, applies the processor, and decodes its output, and the encoder never runs. Under the latent loss (Figure~\ref{fig:downstream}b), the target embeddings can be stored as well, in an \emph{output cache}, and a step reads both embeddings and runs the processor alone, neither encoding nor decoding. An unfrozen backbone changes its embeddings at every step, so it admits no cache, and every step partitions, encodes, processes, and decodes. The three regimes are thus trained in three \emph{training modes}, the latent mode with both caches, the physical mode with an input cache, and the uncached end-to-end mode.

The caches are written once into memory-mapped arrays on a local disk and read back by row, so that the operating system keeps the frequently read rows in memory and the training set never has to fit into the host memory as a whole. For a task without time pairing a row holds one training function. Every function is partitioned once, when the cache is built, so a cached phase sees a single partitioning of each training function, whereas an uncached run draws a new partitioning at every visit. For a time-continuous operator (Appendix~\ref{app:operator-time}) the pairs drawn from a trajectory change with every epoch, but the snapshots behind them do not, so the cache holds one row per snapshot, indexed by the trajectory and the time index, and a pair is two lookups into that table. All snapshots of one trajectory are encoded on the same partitions, so the input and the target embeddings of a pair share their partitions, which is also what allows a rollout to feed a prediction back as an input.

The caches make training faster in two ways. The input cache saves the partitioning and the forward pass of the encoder in every step. The backward pass skips the frozen encoder in any case, since no trainable parameter precedes it, so the input cache saves time but little memory. The output cache also removes the decoder from the step, together with its backward pass, and leaves the processor alone. The decoder dominates the memory of a step under the physical loss. It lies between the processor and the loss, so the backward pass runs through it and its activations at every point of every partition, which scale with the number of points times its depth, have to be stored. Removing it is therefore primarily a memory gain, which grows with the number of points and matters most for large point clouds. A step of the latent mode never touches the points, so its cost does not depend on the resolution of the data at all, whereas every mode that encodes or decodes the points slows down and eventually runs out of memory as they grow. The memory saved can be spent on larger batches at a fixed problem size, or on larger processors and larger problems at a fixed batch. Outside the step, the training set is read and encoded once, before training, so neither data loading nor preprocessing remains in the training loop. Appendix~\ref{app:perf} measures these gains.

\subsubsection{Time-continuous operators}
\label{app:operator-time}

A time-dependent dataset would otherwise reduce to one static map between two fixed time indices. As in recent neural operators for time-dependent problems \citep{herde2024poseidon,mousavi2025rigno,wen2025geometry}, the processor is instead trained as a time-continuous operator that maps the state at a time $t$ to the state at $t+\tau$ for any admissible lead time $\tau$. Pairs are formed on the time grid, with lead times up to a training window of $\Htrain$ time steps. An epoch draws one input-output pair per trajectory whatever the window, so that widening the window enlarges the set of admissible pairs without lengthening an epoch, and covering that set takes many epochs. The input time and the lead time of a trajectory in a given epoch are drawn uniformly, so every input time is trained equally often and a batch mixes input times and lead times. The lead time enters the processor normalized by the span of the time window.

\paragraph{Marching strategies.}
\label{app:operator-marching}
Following \citet{mousavi2025rigno}, the output of a time-continuous processor can stand for three quantities, namely the state $\funcMain(t+\tau)$ itself, the residual $\funcMain(t+\tau)-\funcMain(t)$, or the difference quotient $(\funcMain(t+\tau)-\funcMain(t))/\tau$, which we refer to as the derivative. The state is recovered from the latter two by adding the residual, or the derivative times the lead time, to the input state, and the error is always computed on the recovered state, so it measures the same quantity under all three strategies. The residual and the derivative give a better-posed regression target. They vary far less across lead times than the state does, so one set of output statistics fits the whole training window rather than being stretched across it, and the identity at $\tau\to0$ is built into the update rather than learned, which \citet{mousavi2025rigno} report to improve the accuracy with variable lead times. Their cost is that their target cache grows with the training window, and that they cannot be rolled out in the latent space. \daedalus{} predicts the state itself. The advantages reported by \citet{mousavi2025rigno} for the residual and the derivative were measured under a loss in the physical space and with end-to-end training, where the regression target is a field whose scale and structure change strongly with the lead time. Under the latent loss the target is an embedding, normalized by construction, and the embeddings of the states along a trajectory vary smoothly, so the regression is already well posed without a difference target. A state target also keeps the input and the output of the processor in the same distribution, the embeddings of states, whereas a residual or a derivative has a distribution of its own. Together with the latent rollout and the small output cache, this decides the choice for \daedalus{}. In the benchmarks of Appendix~\ref{app:benchmark}, RIGNO and GAOT are trained with the derivative, as in their original works.

\label{app:caching-marching}
Under the latent loss the marching strategy also changes what the output cache (Appendix~\ref{app:caching}) holds, and therefore its size. A state target does not depend on the input, so a trajectory of $S$ snapshots admits at most $S-1$ distinct targets whatever the training window, all of them rows of the snapshot table. A residual or derivative target depends on both times, and since the encoder is nonlinear its embedding cannot be assembled from the embeddings of the two states it is built from. Every admissible pair then needs its own entry, and a training window of $\Htrain$ steps admits $(S-1)\Htrain-\Htrain(\Htrain-1)/2$ of them per trajectory, far more than the $S$ snapshots. The training step is unaffected and still runs entirely in the latent space, matching the processor output against a pre-encoded residual or derivative. A physical loss avoids the growth altogether, since the target field is then plain arithmetic on two cached snapshots.

\paragraph{Latent rollouts.} An autoregressive rollout reaches a target time by repeated application of the processor. A state prediction is itself a state embedding, so it feeds straight back in as the next input, and the whole rollout runs in the latent space, encoding once at the start and decoding once at the end, as in \citet{alkin2024universal}. Its cost is measured in Appendix~\ref{app:perf-results} (Figure~\ref{fig:perf-rollout}). A residual or derivative prediction cannot be unrolled in this way, because its update adds fields, and the sum of two embeddings does not decode to the sum of the two fields, the decoder being nonlinear, so every update decodes, adds, and re-encodes, which reinstates the cost that caching removes from training. Within an inference window of $\Hinf$ steps each step is predicted directly from the anchor state at the corresponding lead time, and the anchor is reset to the latest prediction only every $\Hinf$ steps, so that a single application per window accumulates error. Setting $\Hinf$ to one step recovers plain autoregression, and setting it to the whole span gives a single direct prediction. An evaluation that compares the state at the target time only is affected by $\Hinf$ through the path along which that state is reached, not through the quantity that is measured.

\FloatBarrier
\subsection{Training phases}
\label{app:operator-phases}

A training run is divided into up to three phases. A warmup opens the run on small batches, which take many steps per function while the learning rate rises linearly from zero. The main phase follows on the full batch and carries the bulk of the training, during which the learning rate decays to a small value. A cooldown closes the run. It spans the final epochs at a constant small learning rate, and it may use another training regime than the phases before it, with its own objective and its own choice of a frozen or an unfrozen backbone.

\subsubsection{The cooldown}
\label{app:operator-cooldown}

The training regimes trade cost against fidelity. The latent loss with a frozen and cached backbone is the cheapest regime there is, since a step involves the processor alone, but it optimizes a proxy. The physical loss optimizes the error itself, but every step decodes, and an unfrozen backbone also encodes. A cooldown combines the two. The bulk of the epochs trains on the latent loss with both caches, and the final epochs switch to the physical loss. Since the expensive steps are confined to the cooldown, the total cost of training stays close to that of latent training. A frozen cooldown trains on an input cache built when it starts, whereas an unfrozen cooldown trains without a cache.

The switch to the physical loss serves two purposes. The first is to close the proxy gap. A processor trained on the latent loss approaches the target embeddings in the latent distance, which weighs every deviation by its size in the latent space. The decoded field, however, responds to deviations along different directions of the embedding differently, so the embedding that is optimal under the latent distance is not necessarily the one with the smallest error of the decoded field. This difference is the proxy gap of Appendix~\ref{app:operator-loss}. The physical loss evaluates every deviation by its actual effect on the decoded field, so a cooldown that starts from the latent solution moves the predictions toward the embeddings that decode best, and the gap closes. The second purpose is to balance the static and the morphological parts of a prediction against each other. The latent loss weighs the two by a fixed weight and asks for exactly the encoded target, whereas the effect of each part on the decoded field differs from partition to partition. Under the physical loss the processor may deviate slightly from the encoded embeddings wherever such a deviation decodes to a field that suits the task better. An error in the predicted value frame of a partition, for instance, shifts or rescales the whole partition, and with independent errors from one partition to the next the decoded field may show a honeycomb pattern. The physical loss penalizes the discontinuities at the cell boundaries, so the processor trades small deviations of the frames and the morphologies for a field whose partitions join, and the pattern disappears.

The cooldown can also unfreeze the backbone. This lifts the restriction to a fixed representation, at the cost of the two properties that make the backbone worth sharing. An unfrozen backbone is specialized to one task, so it is no longer universal, and since only the decoded output of the processor is supervised, nothing keeps the encoder and the decoder inverse to each other, so a field encoded and decoded by them is no longer reproduced faithfully. An unfrozen backbone therefore belongs to its task and is stored with the processor.

Two details make the switch well behaved. If the backbone is unfrozen in the cooldown only, it needs an optimizer state of its own, started when the cooldown starts. A single optimizer over all parameters would reach the cooldown with zero moment estimates and a large step count for the backbone, which amounts to Adam without bias correction and takes steps several times larger than the learning rate at first. In addition, the running average of the loss that guards against loss spikes is restarted when the objective changes, since it would otherwise compare the new loss with the scale of the old one.

\FloatBarrier
\subsection{Accuracy benchmark}
\label{app:benchmark}

We compare the operator learned on the frozen backbone with two baselines trained end to end, RIGNO-12 \citep{mousavi2025rigno} and GAOT \citep{wen2025geometry}, on 28 problems (Table~\ref{tab:benchmark}). The three models are trained and evaluated through one data pipeline, with the same inputs, targets, splits, rollout, and metric, and each model is otherwise free in its architecture, normalization, loss, time marching strategy, optimizer, and training length.

\paragraph{Problems.} The benchmark covers incompressible Euler flow (IEU, four families of initial conditions), dynamic compressible Euler flow (CEU, five problems on $128^2$ grids in primitive variables, each also on its native $512^2$ grid in conservative variables, and a Rayleigh--Taylor instability under gravity), steady compressible flow past airfoils and bluff bodies, acoustic waves (AWA), the Helmholtz equation (HEL), the Poisson equation (POI), and the static equilibrium of linear and hyperelastic bodies (SEQ). Seventeen problems are time dependent and eleven are steady. The problems on regular grids share one geometry across their samples, whereas the five elasticity problems and the four steady compressible-flow problems are point clouds on a geometry that changes from sample to sample. Static inputs, such as a wave speed, a signed distance function (SDF), or a source, are given to every model together with the state. Table~\ref{tab:datasets} describes every dataset briefly.

\paragraph{Data.} The first 128 samples of a dataset form its test split, the next 128, or 96 for the Rayleigh--Taylor problem, its validation split, and the training split follows with between 1024 and 8192 samples. Grids of $128^2$ points are used whole, a $512^2$ grid through a fixed subset of 131072 points per sample, and a point cloud is subsampled without replacement to 8192 or 16384 points, except the clouds of the steady compressible flows, which are used whole, with 16384 points around an airfoil and 32768 around a bluff body. Coordinates are mapped to $[-1,1]^2$, and no model uses periodic boundaries. On the $512^2$ grids RIGNO-12 and GAOT train on a fixed random subset of 16384 of these points per sample, for cost, and are validated and tested on all of them, whereas \daedalus{} trains on all of them, and is nonetheless faster to train there (Table~\ref{tab:benchmark}).

\begin{table}[!htbp]\centering\scriptsize
  \renewcommand{\arraystretch}{1.1}\setlength{\tabcolsep}{4pt}
  \caption{\textbf{Datasets.} The equation, discretization, and a short description of every dataset of the benchmark, with the inputs and outputs of every family in parentheses. The five dynamic compressible-flow problems appear on $128^2$ grids in the primitive variables, the density $\rho$, the velocity $(u,v)$ and the pressure $p$, and on their native $512^2$ grids in the conservative variables, the density, the momentum $(m_x,m_y)$ and the total energy $E$. The steady compressible flows take the Mach number $M$ and the angle of attack $\alpha$ of the free stream, the elasticity problems output the displacement $u$ and the stress $\sigma$, the acoustic waves have the wave speed $c$, the Helmholtz equation the coefficient $a$, and the Poisson equation the source $f$, with the solution $u$ in all three. A dagger marks a geometry that varies between samples. The steady compressible flows are resampled from datasets of GAOT \citep{wen2025geometry}, so their errors do not compare with those reported there. The incompressible, the dynamic compressible, and the acoustic problems are time dependent, and the others steady.}
  \label{tab:datasets}
  \begin{tabular}{@{}l l l p{6.4cm}@{}}
    \toprule
    Dataset & Equation & Points & Description \\
    \midrule
    \multicolumn{4}{l}{\emph{Incompressible flow} ($u, v \to u, v$)} \\
    IEU-BB & incompressible Euler & $128^2$ grid & Flow from a Brownian-bridge random initial condition. \\
    IEU-Gauss & incompressible Euler & $128^2$ grid & Flow from Gaussian vortices. \\
    IEU-PwC & incompressible Euler & $128^2$ grid & Flow from a piecewise-constant initial condition. \\
    IEU-VortexSheet & incompressible Euler & $128^2$ grid & Flow from a perturbed vortex sheet. \\
    \midrule
    \multicolumn{4}{l}{\emph{Dynamic compressible flow} ($\rho, u, v, p \to \rho, u, v, p$ on $128^2$ grids, $\rho, m_x, m_y, E \to \rho, m_x, m_y, E$ on $512^2$ grids)} \\
    CEU-Gauss & compressible Euler & $128^2$, $512^2$ grid & Smooth flow from Gaussian initial data. \\
    CEU-Riemann & compressible Euler & $128^2$, $512^2$ grid & Riemann problems, with shocks and contact discontinuities. \\
    CEU-RiemannCurved & compressible Euler & $128^2$, $512^2$ grid & Riemann problems with curved initial interfaces. \\
    CEU-RiemannKH & compressible Euler & $128^2$, $512^2$ grid & Riemann problems that seed Kelvin--Helmholtz instabilities. \\
    CEU-KH & compressible Euler & $128^2$, $512^2$ grid & Shear layers rolling up in a Kelvin--Helmholtz instability. \\
    CEU-RayleighTaylor & compressible Euler & $128^2$ grid & A Rayleigh--Taylor instability under gravity. \\
    \midrule
    \multicolumn{4}{l}{\emph{Steady compressible flow} ($M\cos\alpha, M\sin\alpha$, SDF $\to \rho$)} \\
    CEU-NACA0012$^\dagger$ & compressible Euler & cloud, $16384$ & Flow past a NACA0012 airfoil of perturbed shape, at a varying Mach number and angle of attack. \\
    CEU-NACA2412$^\dagger$ & compressible Euler & cloud, $16384$ & As above, past a NACA2412 airfoil. \\
    CEU-RAE2822$^\dagger$ & compressible Euler & cloud, $16384$ & As above, past an RAE2822 airfoil. \\
    CEU-BluffBody$^\dagger$ & compressible Euler & cloud, $32768$ & Flow past one of ten bluff bodies, circles, ellipses, squares, rectangles, a cone, and a semicircle. \\
    \midrule
    \multicolumn{4}{l}{\emph{Acoustic waves} ($u, c \to u$)} \\
    AWA-Gauss & acoustic wave & $128^2$ grid & Waves in a medium of smoothly varying wave speed $c$. \\
    AWA-Layer & acoustic wave & $128^2$ grid & Waves in a layered medium of piecewise-constant wave speed $c$. \\
    \midrule
    \multicolumn{4}{l}{\emph{Solid mechanics} (SDF $\to u, \sigma$)} \\
    SEQ-CloudCircle-Hyp$^\dagger$ & quasi-static elasticity & cloud & A hyperelastic body with a perturbed circular outline and a hole. \\
    SEQ-CloudCircle-Lin0$^\dagger$ & quasi-static elasticity & cloud & The same geometries, linear elastic. \\
    SEQ-CloudCircle-Lin1$^\dagger$ & quasi-static elasticity & cloud & As Lin0, under other boundary conditions. \\
    SEQ-SquareCloud-Hyp$^\dagger$ & quasi-static elasticity & cloud & A hyperelastic square with two holes of random shape. \\
    SEQ-SquareCloud-Lin$^\dagger$ & quasi-static elasticity & cloud & The same geometries, linear elastic. \\
    \midrule
    \multicolumn{4}{l}{\emph{Helmholtz and Poisson equations} ($a$ and the boundary value $\to u$ for HEL, $f \to u$ for POI)} \\
    HEL-Gauss & Helmholtz & $128^2$ grid & Standing waves in a medium with Gaussian inclusions in the coefficient $a$, under a constant boundary value. \\
    POI-Gauss & Poisson & $128^2$ grid & A source $f$ that is a sum of Gaussians. \\
    \bottomrule
  \end{tabular}
\end{table}

\paragraph{Time.} A time-dependent problem keeps every second snapshot up to the time index 14, or 10 for the Rayleigh--Taylor problem. Training may pair any two retained snapshots, with the lead time given as a fraction of the span. Evaluation starts from the initial snapshot, rolls out in blocks of eight time indices, and scores the final snapshot only.

\begin{table}[!htbp]\centering\scriptsize
  \renewcommand{\arraystretch}{1.08}\setlength{\tabcolsep}{2pt}
  \caption{\textbf{Accuracy and training cost on 28 problems.} Median relative $L^1$ test error in percent and total training time (TTT) in hours on four RTX~4090 GPUs with \daedalus{} as the anchor. A red \upmark{} marks a baseline above \daedalus{} by more than 5\% and a green \dnmark{} one below it, followed by the relative difference of the errors, $(e-e_{\textrm{Daedalus}})/e_{\textrm{Daedalus}}$, or by the factor between the training times, and a value within 5\% carries no mark. \daedalus{} is trained with a cooldown, the last fifth of its epochs on the physical loss with the backbone frozen (Appendix~\ref{app:operator-cooldown}), and its TTT excludes the pre-training of the backbone, which is shared by all problems. A dagger marks a geometry that varies between samples, and a dash a problem on which RIGNO-12 is not trained. The mean row averages the per-problem relative differences of the errors and the per-problem ratios of the training times, and the last row counts the problems on which a baseline's error lies above, within 5\% of, and below that of \daedalus{}, for RIGNO-12 over the 24 problems on which it is trained.}
  \label{tab:benchmark}
  \resizebox{0.98\textwidth}{!}{%
  \begin{tabular}{l r r @{\hspace{14pt}}
                  r @{\;}c@{\,}r r @{\;}c@{\,}r @{\hspace{14pt}}
                  r @{\;}c@{\,}r r @{\;}c@{\,}r}
    \toprule
     & \multicolumn{2}{c}{\daedalus{} (ours)} & \multicolumn{6}{c}{GAOT} & \multicolumn{6}{c}{RIGNO-12} \\
    \cmidrule(lr){2-3}\cmidrule(lr){4-9}\cmidrule(lr){10-15}
    Problem & {Error [\%]} & {TTT [h]} & \multicolumn{3}{c}{Error [\%]} & \multicolumn{3}{c}{TTT [h]} & \multicolumn{3}{c}{Error [\%]} & \multicolumn{3}{c}{TTT [h]} \\
    \midrule
    \multicolumn{15}{l}{\emph{Incompressible flow}} \\
    IEU-BB & 2.00 & 2.5 & 2.83 & \upmark & \upval{42\%} & 8.7 & \upmark & \upval{$\times$3.5} & 12.8 & \upmark & \upval{540\%} & 26 & \upmark & \upval{$\times$10.6} \\
    IEU-Gauss & 4.63 & 2.7 & 6.97 & \upmark & \upval{51\%} & 5.1 & \upmark & \upval{$\times$1.9} & 37.6 & \upmark & \upval{713\%} & 26 & \upmark & \upval{$\times$9.7} \\
    IEU-PwC & 2.72 & 2.5 & 3.01 & \upmark & \upval{11\%} & 5.1 & \upmark & \upval{$\times$2.0} & 27.4 & \upmark & \upval{907\%} & 26 & \upmark & \upval{$\times$10.5} \\
    IEU-VortexSheet & 1.66 & 2.6 & 0.578 & \dnmark & \dnval{65\%} & 5.7 & \upmark & \upval{$\times$2.2} & 5.78 & \upmark & \upval{248\%} & 31 & \upmark & \upval{$\times$12.0} \\
    \midrule
    \multicolumn{15}{l}{\emph{Dynamic compressible flow}} \\
    CEU-Gauss ($128^2$) & 4.95 & 2.8 & 6.77 & \upmark & \upval{37\%} & 5.1 & \upmark & \upval{$\times$1.8} & 23.7 & \upmark & \upval{379\%} & 28 & \upmark & \upval{$\times$9.8} \\
    CEU-Gauss ($512^2$) & 3.45 & 11 & 5.94 & \upmark & \upval{72\%} & 14 & \upmark & \upval{$\times$1.3} & 27.1 & \upmark & \upval{687\%} & 28 & \upmark & \upval{$\times$2.6} \\
    CEU-Riemann ($128^2$) & 4.06 & 3.0 & 3.94 &  &  & 5.0 & \upmark & \upval{$\times$1.7} & 11.3 & \upmark & \upval{179\%} & 28 & \upmark & \upval{$\times$9.4} \\
    CEU-Riemann ($512^2$) & 3.68 & 11 & 5.72 & \upmark & \upval{55\%} & 17 & \upmark & \upval{$\times$1.6} & 17.5 & \upmark & \upval{376\%} & 28 & \upmark & \upval{$\times$2.6} \\
    CEU-RiemannCurved ($128^2$) & 29.2 & 2.8 & 34.1 & \upmark & \upval{17\%} & 5.1 & \upmark & \upval{$\times$1.8} & 46.2 & \upmark & \upval{58\%} & 32 & \upmark & \upval{$\times$11.4} \\
    CEU-RiemannCurved ($512^2$) & 28.6 & 11 & 40.0 & \upmark & \upval{40\%} & 16 & \upmark & \upval{$\times$1.5} & 51.1 & \upmark & \upval{79\%} & 31 & \upmark & \upval{$\times$2.8} \\
    CEU-RiemannKH ($128^2$) & 8.86 & 2.8 & 11.2 & \upmark & \upval{26\%} & 5.4 & \upmark & \upval{$\times$1.9} & 26.0 & \upmark & \upval{194\%} & 30 & \upmark & \upval{$\times$10.7} \\
    CEU-RiemannKH ($512^2$) & 8.38 & 11 & 13.6 & \upmark & \upval{62\%} & 24 & \upmark & \upval{$\times$2.2} & 31.5 & \upmark & \upval{276\%} & 29 & \upmark & \upval{$\times$2.6} \\
    CEU-KH ($128^2$) & 13.6 & 2.9 & 12.3 & \dnmark & \dnval{9\%} & 5.0 & \upmark & \upval{$\times$1.7} & 14.9 & \upmark & \upval{10\%} & 29 & \upmark & \upval{$\times$10.1} \\
    CEU-KH ($512^2$) & 11.8 & 11 & 16.0 & \upmark & \upval{35\%} & 13 & \upmark & \upval{$\times$1.2} & 25.2 & \upmark & \upval{113\%} & 29 & \upmark & \upval{$\times$2.7} \\
    CEU-RayleighTaylor ($128^2$) & 30.7 & 3.0 & 15.7 & \dnmark & \dnval{49\%} & 2.8 & \dnmark & \dnval{$\times$1.1} & 32.3 & \upmark & \upval{5\%} & 16 & \upmark & \upval{$\times$5.1} \\
    \midrule
    \multicolumn{15}{l}{\emph{Steady compressible flow}} \\
    CEU-NACA0012$^\dagger$ & 0.246 & 2.2 & 0.389 & \upmark & \upval{58\%} & 8.4 & \upmark & \upval{$\times$3.8} & {\textendash} &  &  & {\textendash} &  &  \\
    CEU-NACA2412$^\dagger$ & 0.302 & 2.2 & 0.417 & \upmark & \upval{38\%} & 7.6 & \upmark & \upval{$\times$3.4} & {\textendash} &  &  & {\textendash} &  &  \\
    CEU-RAE2822$^\dagger$ & 0.264 & 2.2 & 0.393 & \upmark & \upval{49\%} & 7.6 & \upmark & \upval{$\times$3.4} & {\textendash} &  &  & {\textendash} &  &  \\
    CEU-BluffBody$^\dagger$ & 0.180 & 4.9 & 0.241 & \upmark & \upval{34\%} & 4.1 & \dnmark & \dnval{$\times$1.2} & {\textendash} &  &  & {\textendash} &  &  \\
    \midrule
    \multicolumn{15}{l}{\emph{Acoustic waves}} \\
    AWA-Gauss & 8.54 & 2.5 & 9.80 & \upmark & \upval{15\%} & 5.2 & \upmark & \upval{$\times$2.1} & 24.3 & \upmark & \upval{184\%} & 26 & \upmark & \upval{$\times$10.4} \\
    AWA-Layer & 7.06 & 2.4 & 8.67 & \upmark & \upval{23\%} & 5.3 & \upmark & \upval{$\times$2.2} & 19.3 & \upmark & \upval{173\%} & 26 & \upmark & \upval{$\times$11.0} \\
    \midrule
    \multicolumn{15}{l}{\emph{Solid mechanics}} \\
    SEQ-CloudCircle-Hyp$^\dagger$ & 0.768 & 2.4 & 0.857 & \upmark & \upval{12\%} & 6.3 & \upmark & \upval{$\times$2.6} & 2.23 & \upmark & \upval{190\%} & 25 & \upmark & \upval{$\times$10.3} \\
    SEQ-CloudCircle-Lin0$^\dagger$ & 0.589 & 2.5 & 1.04 & \upmark & \upval{76\%} & 5.8 & \upmark & \upval{$\times$2.3} & 1.65 & \upmark & \upval{180\%} & 26 & \upmark & \upval{$\times$10.3} \\
    SEQ-CloudCircle-Lin1$^\dagger$ & 1.09 & 2.6 & 1.11 &  &  & 7.5 & \upmark & \upval{$\times$2.9} & 3.80 & \upmark & \upval{248\%} & 25 & \upmark & \upval{$\times$9.8} \\
    SEQ-SquareCloud-Hyp$^\dagger$ & 1.74 & 2.9 & 1.75 &  &  & 7.6 & \upmark & \upval{$\times$2.6} & 2.39 & \upmark & \upval{38\%} & 40 & \upmark & \upval{$\times$13.7} \\
    SEQ-SquareCloud-Lin$^\dagger$ & 0.979 & 3.0 & 1.01 &  &  & 7.0 & \upmark & \upval{$\times$2.3} & 1.78 & \upmark & \upval{82\%} & 34 & \upmark & \upval{$\times$11.2} \\
    \midrule
    \multicolumn{15}{l}{\emph{Helmholtz and Poisson equations}} \\
    HEL-Gauss & 5.97 & 1.2 & 7.24 & \upmark & \upval{21\%} & 0.99 & \dnmark & \dnval{$\times$1.2} & 22.6 & \upmark & \upval{279\%} & 5.9 & \upmark & \upval{$\times$4.9} \\
    POI-Gauss & 1.36 & 1.1 & 1.27 & \dnmark & \dnval{7\%} & 0.92 & \dnmark & \dnval{$\times$1.2} & 4.37 & \upmark & \upval{222\%} & 4.8 & \upmark & \upval{$\times$4.4} \\
    \midrule
    \textbf{Mean} & {} & {} & {} & \upmark & \upval{\textbf{23\%}} & {} & \upmark & \upval{\textbf{$\times$2.1}} & {} & \upmark & \upval{\textbf{265\%}} & {} & \upmark & \upval{\textbf{$\times$8.3}} \\
    \textbf{Above\,/\,within\,/\,below Daedalus} & \multicolumn{2}{c}{} & \multicolumn{3}{c}{\textbf{20\,/\,4\,/\,4}} & \multicolumn{3}{c}{} & \multicolumn{3}{c}{\textbf{24\,/\,0\,/\,0}} & \multicolumn{3}{c}{} \\
    \bottomrule
  \end{tabular}}
\end{table}

\paragraph{Metric and selection.} The error of a sample is the relative $L^1$ error of each output channel on the raw fields, averaged over the channels, and Table~\ref{tab:benchmark} reports its median over the 128 test samples. Every model is tested once, at the checkpoint with the lowest validation median of the relative $L^2$ error. The original evaluations of RIGNO and GAOT differ slightly from this protocol, since they measure the error on globally normalized fields and report several rollout schedules, so their published numbers do not compare one to one with those reported here.

\paragraph{Training cost.} Every run uses four RTX~4090 GPUs and a consistent batch size of 128, or 64 on the $512^2$ grids, the Rayleigh--Taylor problem and CEU-BluffBody, except that RIGNO-12 uses 32 on the five other dynamic compressible problems on $128^2$ grids. The total training time (TTT) is the wall-clock time from the start of the process to the end of the test scoring, data loading and the building of caches included. For \daedalus{} it covers the downstream run alone and excludes the pre-training of the backbone, which is task-agnostic, carried out once and shared by all 28 problems.

\paragraph{RIGNO-12.} We use the twelve-layer variant rather than the eighteen-layer model of the original paper, in bfloat16 (bf16) arithmetic, which is $1.5$ times faster. The model predicts time derivatives, trains on every admissible pair of a trajectory in every epoch without a warmup of the lead time, places its regional nodes on a quarter of the points, and standardizes the inputs of steady problems with their own statistics. Two settings depart from the paper and leave it undertrained. The batch, mostly 128, is up to eight times the paper's, and the time-dependent problems run for 1000 epochs instead of 2500, while the steady problems keep the paper's 3000. Together these give five to twenty times fewer optimizer steps than in the paper at the same learning rate, so the reported errors overstate those RIGNO-12 attains with its original settings. A smaller batch would barely lengthen an epoch, since the throughput of RIGNO-12 hardly depends on the batch size, but its training time is already the longest of the three models, which is also why it is not trained on the four steady compressible-flow problems.

\paragraph{GAOT.} We use the architecture of the main table of the original paper, with 5.62 million parameters, a latent grid of $64\times64$ tokens merged in $2\times2$ blocks into 1024 transformer tokens, and five transformer layers, in float32 (fp32) arithmetic. The model predicts time derivatives, takes its time statistics on the retained snapshots, and keeps its own rescaling of the coordinates of every sample. The batch of 128 is split over the four GPUs by distributed data parallelism, against 64 on a single GPU in the paper, at the paper's learning rate and training length, 500 epochs on time-dependent and 1000 on steady problems, and with the learning-rate schedule of the released code. On the fixed-geometry problems GAOT runs in its fast mode, which computes its geometric preprocessing once and shares it across all samples, while on a variable geometry it precomputes the graphs of every sample.

\paragraph{\daedalus{}.} All problems share the frozen backbone and the processor of Appendix~\ref{app:architecture}, and the processor is trained with the curriculum of Appendix~\ref{app:operator-cooldown}. Every problem is cut into 1024 partitions, or 2048 on the $512^2$ grids and CEU-BluffBody, with an overlap that adds a fifth to the assignments. The processor has a width of 512 and eight heads and a single block of two units, each applied four times, and it attends over all partition embeddings at once, with a single attention window and no grouping ($W{=}M{=}1$). A warmup of 8192 functions on batches of eight and the main phase train on the latent loss with cached input and target embeddings. The cooldown trains the last fifth of the epochs on the physical loss, with the backbone still frozen, on an input cache built when it starts and at the training batch size, except on the $512^2$ grids, where a batch of 32 instead of 64 keeps the decoded fields within memory. The learning rate rises linearly to $10^{-3}$, decays along a cosine to $10^{-4}$ at nine tenths of the warmup and the main phase, and stays there for the rest of the main phase and the whole cooldown, under AdamW with a weight decay of $10^{-4}$ and with gradients clipped to a global norm of $1$ and by adaptive clipping at $0.1$. A run sees about $8.2$ million training pairs, one per trajectory and epoch, or $4.1$ million on POI-Gauss and HEL-Gauss. The model is evaluated at the start of the cooldown and at its end, besides the regular evaluations, and the checkpoint tested is the best over the whole run.

\paragraph{Results.} \daedalus{} is more accurate than GAOT on 23 of the 28 problems and more accurate than RIGNO-12 on all 24 problems on which it is trained. Averaged over the problems, the per-problem error of GAOT exceeds ours by 23\% and that of RIGNO-12 by 265\%, and their per-problem training times are $2.1$ and $8.3$ times ours.

\paragraph{Accuracy by family.} The margin over GAOT varies with the family of problems. It is largest on the steady compressible flows, where the error of GAOT is higher than ours on every problem and by 45\% on average, and it follows on the dynamic compressible flows, 26\% on average and higher on 8 of the 11 problems, on the acoustic waves and on solid mechanics, 19\% each and higher on every problem. It is smallest on incompressible flow, 9\%, where GAOT is more accurate on IEU-VortexSheet, and on the Helmholtz and Poisson problems, 7\%, where it is more accurate on POI-Gauss. Among the dynamic compressible flows the margin grows with the resolution, from 13\% on the five problems on $128^2$ grids to 53\% on the same problems on their native $512^2$ grids, on which \daedalus{} trains on every point and the baselines on a subset. The margin over RIGNO-12 exceeds 100\% in every family, from 148\% on solid mechanics to 602\% on incompressible flow.

\paragraph{Training time by geometry.} The training time of GAOT depends on whether the samples share their geometry. On the 19 problems with a fixed geometry it runs in its fast mode, computing its geometric preprocessing once for all samples, and its training time is on average $1.7$ times ours, while it trains faster than \daedalus{} (at most 1.2 times) on three of them, the Helmholtz and the Poisson problems, whose runs see half as many training pairs, and the Rayleigh--Taylor problem. On the nine problems whose geometry changes from sample to sample, GAOT preprocesses every sample, whereas cached embeddings remain available to \daedalus{}, and its training time is on average $2.7$ times ours, with an error 30\% higher, although it trains faster on CEU-BluffBody. The training time of \daedalus{} grows on the $512^2$ grids, to about eleven hours, since it trains on every point there, and remains below those of both baselines, by factors of $1.5$ and $2.7$. RIGNO-12 trains $7.5$ times longer than \daedalus{} on the fixed geometries and 11 times longer on the varying clouds of the elasticity problems.

\subsubsection{Training regimes of \daedalus{}}
\label{app:benchmark-regimes}

Table~\ref{tab:regimes} compares four ways of training the same processor on eight of the problems, five evolution problems and three steady ones, under the protocol above. The latent loss trains every epoch on the latent loss with input and output caches, and the physical loss every epoch on the physical loss with an input cache only. The cooldown is the curriculum of Appendix~\ref{app:operator-cooldown} used in Table~\ref{tab:benchmark}, and the unfrozen cooldown follows the same curriculum but unfreezes the encoder and the decoder during the cooldown, without a cache.

\begin{table}[!htbp]\centering\scriptsize
  \renewcommand{\arraystretch}{1.08}\setlength{\tabcolsep}{4pt}
  \caption{\textbf{Training regimes of \daedalus{}.} Median relative $L^1$ test error in percent and total training time (TTT) in hours on four RTX~4090 GPUs for one processor trained on the latent loss throughout, on the physical loss throughout, with a cooldown on the physical loss and the backbone frozen, and with a cooldown that also unfreezes the encoder and the decoder. The lowest error of every problem is set in bold, and a dagger marks a geometry that varies between samples. The last two rows give the geometric mean over each group of the ratio to the latent loss.}
  \label{tab:regimes}
  \resizebox{\textwidth}{!}{%
  \begin{tabular}{l *{4}{r r}}
    \toprule
     & \multicolumn{2}{c}{Latent loss} & \multicolumn{2}{c}{Physical loss} & \multicolumn{2}{c}{Cooldown} & \multicolumn{2}{c}{Unfrozen cooldown} \\
    \cmidrule(lr){2-3}\cmidrule(lr){4-5}\cmidrule(lr){6-7}\cmidrule(lr){8-9}
    Problem & {Error [\%]} & {TTT [h]} & {Error [\%]} & {TTT [h]} & {Error [\%]} & {TTT [h]} & {Error [\%]} & {TTT [h]} \\
    \midrule
    \multicolumn{9}{l}{\emph{Evolution problems}} \\
    AWA-Layer & 7.14 & 2.1 & 12.2 & 2.9 & \bfseries 7.06 & 2.4 & 7.15 & 6.0 \\
    IEU-BB & 2.27 & 2.2 & \bfseries 1.86 & 3.5 & 2.00 & 2.5 & 2.23 & 6.3 \\
    IEU-Gauss & 4.84 & 2.1 & 5.05 & 4.1 & \bfseries 4.63 & 2.7 & 4.74 & 6.5 \\
    IEU-VortexSheet & 2.04 & 2.2 & 2.13 & 3.5 & \bfseries 1.66 & 2.6 & 1.68 & 5.9 \\
    CEU-Gauss ($128^2$) & 5.07 & 2.3 & 8.83 & 4.8 & \bfseries 4.95 & 2.8 & 4.96 & 9.1 \\
    \midrule
    \multicolumn{9}{l}{\emph{Steady problems}} \\
    SEQ-CloudCircle-Hyp$^\dagger$ & 1.25 & 2.1 & \bfseries 0.682 & 3.6 & 0.768 & 2.4 & 0.685 & 5.5 \\
    SEQ-CloudCircle-Lin0$^\dagger$ & 0.900 & 2.3 & 0.541 & 3.6 & 0.589 & 2.5 & \bfseries 0.520 & 5.5 \\
    SEQ-CloudCircle-Lin1$^\dagger$ & 1.33 & 2.2 & 1.21 & 3.6 & 1.09 & 2.6 & \bfseries 1.08 & 5.8 \\
    \midrule
    Evolution, ratio to latent & 1.00 & 1.00 & 1.22 & 1.70 & 0.92 & 1.18 & 0.95 & 3.05 \\
    Steady, ratio to latent & 1.00 & 1.00 & 0.67 & 1.64 & 0.69 & 1.13 & 0.64 & 2.53 \\
    \bottomrule
  \end{tabular}}
\end{table}

\paragraph{Physical loss.} The physical loss consistently helps the steady problems, where its error lies 33\% below that of the latent loss in the geometric mean, but it can lead to a lower accuracy on the evolution problems, where its error lies 22\% above that of the latent loss in the geometric mean, and 71\% and 74\% above it on AWA-Layer and CEU-Gauss ($128^2$). We attribute this to the design of the training pipeline, namely PartNorm, the independent encoding and decoding of every channel, and the lack of residual connections in the physical space. On these problems an operator trained on the physical loss from the start can settle in a suboptimal local minimum or in an alternative representation of the same field (Appendix~\ref{app:operator-cooldown}). Since the physical loss cannot use the output cache, its training also takes $1.7$ times as long as that of the latent loss on average.

\paragraph{Cooldown.} The cooldown avoids these problems. On the steady problems it closes the proxy gap, with an error 31\% below that of the latent loss and only 3\% above that of the physical loss in the geometric mean. On the evolution problems it does not harm the accuracy, since the processor is already in the right basin when the loss changes, and its error is in fact 8\% below that of the latent loss. The bulk of the training uses the input and output caches and the physical loss is active only in the last fifth of the epochs, so the cooldown adds only 16\% to the total training time on average.

\paragraph{Unfrozen cooldown.} Unfreezing the encoder and the decoder in the cooldown helps marginally on the steady problems, where it lowers the error of the frozen cooldown by a further 8\% in the geometric mean, and not on the evolution problems. Without a cache it takes $2.4$ times as long as the frozen cooldown. It also comes at the cost of the invertibility of the encoding and decoding and of the interpretability of the embeddings, since it lets the FAE adapt to the task at hand rather than provide invertible, task-agnostic embeddings, and we therefore treat it as an option rather than the default.

\FloatBarrier
\subsection{Performance benchmark}
\label{app:perf}

We measure the speed and the memory of training and inference for \daedalus{}, GAOT \citep{wen2025geometry}, and RIGNO-12 \citep{mousavi2025rigno} on synthetic problems whose size can be set freely, with the same configurations, sweep points, timing protocol, and functions for all three models.

\subsubsection{Training and inference modes}
\label{app:perf-modes}

The three training modes of \daedalus{} differ in which parts of the pipeline a training step runs (Appendix~\ref{app:caching}). The \emph{latent mode} trains the processor on the latent loss with input and output caches, so a step never touches the points, and here the cached embeddings are synthetic ones of the shapes the backbone produces. The \emph{physical mode} trains the processor on the physical loss with an input cache, encoded beforehand and outside the measurement, so a step runs the processor and the frozen decoder. The \emph{end-to-end mode} uses no cache and partitions, encodes, processes, and decodes in every step, training the encoder and the decoder along with the processor.
GAOT and RIGNO have no reduced space that exists apart from the task, so they are trained end to end, but on graphs built before training, so that their geometric preprocessing is cached while every other part of the model runs in every step. Among our modes the closest counterpart is the physical mode, in which the backbone is frozen and its input side cached. Our end-to-end mode trains everything and partitions every input anew at every step, and is therefore slower than our two cached modes, except with a deep processor on the coarse problem, which dominates the step and leaves the three level.

Inference maps an input field to an output field. For \daedalus{} it consists of two compiled calls, partitioning and encoding followed by processing and decoding. The inference of GAOT includes the construction of its graphs, while that of RIGNO excludes it. A rollout applies the operator autoregressively. Our \emph{latent rollout} encodes the initial state once, advances it in the reduced space, and decodes only the final state (Appendix~\ref{app:operator-marching}), whereas our \emph{end-to-end rollout} partitions, encodes, processes, and decodes at every step. GAOT and RIGNO roll out in the physical space with time-derivative steps and build their graphs once per rollout.

The point clouds are either \emph{fixed}, one cloud and one graph shared by the whole batch, or \emph{variable}, one cloud per sample. \daedalus{} always treats them as variable and partitions every sample on its own, GAOT is measured in both settings, and RIGNO with fixed coordinates only.

\subsubsection{Protocol}
\label{app:perf-protocol}

The functions are multi-mode anisotropic sinusoidal functions with $25$ modes and frequencies in $[1,3]$, with one input and one output channel and the input as the target. By default a function has 16384 points and 1024 partitions or latent tokens, and the batch holds 1, 4, or 32 functions. Every measurement runs in a fresh process on a single RTX~4090 GPU with 24\,GB of memory. The data are sampled and the random keys drawn before the clock starts. The first, compiling call is excluded, three warmup calls follow, and the reported time is the median of five repetitions of ten calls each, every repetition synchronizing on all outputs, the updated parameters and optimizer state included. A rollout is timed as a whole, as the median of three repetitions, after a compiling call and three warmup calls of a single step. We report the latency in milliseconds per sample, the batch time divided by the batch size, the training throughput in samples per second, and the peak memory, the allocator's high-water mark including the scratch space of compilation and autotuning. Once a model runs out of device memory in a series, exhausts the memory of the host, or fails otherwise, its larger values are skipped, so a curve ends at the last size that runs.

The latent size maps the partitions of \daedalus{} to the transformer tokens of GAOT and the regional nodes of RIGNO, and the depth maps the processor units of \daedalus{} to the transformer layers of GAOT and the message-passing steps of RIGNO. Equal depth is not equal size, since a unit holds about 1.6 million parameters, a GAOT layer about 1.1 million, and a RIGNO step about 0.12 million. The model-size sweep varies the depth from 1 to 1024 on a coarse problem with the default sizes and on a dense one with 262144 points and 4096 partitions. The problem-size sweep grows the partition count at 16, 64, and 256 points per partition, so that the points grow with the tokens, and the space-size sweep grows the points per function from 4096 to about 67 million at 1024 partitions. The rollout sweep varies the rollout length from 1 to 4096 steps on the coarse and the dense problem.

\subsubsection{Settings of the three models}
\label{app:perf-settings}

\paragraph{\daedalus{}.} The backbone has the configuration of the low-dimensional backbone (\hyperref[tab:backbones]{appendix prelude}), as in the accuracy benchmark, and the processor has a width of 512, eight heads, and rotary position embeddings, with two units of four repeats each in all sweeps but the model-size sweep, where the depth sets the number of units and every unit is applied once. Together with the backbone the model holds 3.79 million parameters, 3.69 million of them in the processor, runs in bf16 mixed precision, and its processor is conditioned on the lead time only in rollouts. The partitioner is the inference partitioner of Appendix~\ref{app:partitioner}, with five local Lloyd iterations from the curve start and an overlap that adds a fifth to the assignments, except that it is fitted on a random quarter of the points. The optimizer is AdamW with the gradient clipping of the downstream runs. The implementation is in JAX, and the backbone's partitioning and encoding are timed in every mode that runs them.

\paragraph{GAOT.} We use the transformer of the main table of the original paper, with five layers, a width of 256, eight heads, and rotary position embeddings, between multiscale attentional graph neural operator (MAGNO) encoders and decoders with attention quadrature and geometric embeddings, in fp32 arithmetic, with 5.56 million parameters at 1024 tokens. Unlike the accuracy benchmark, the latent grid is not merged into blocks, so it is the token grid itself, $32\times32$ at 1024 tokens, and the radius of the encoder and the decoder is scaled with the grid so that every grid covers the domain as the original one does. With fixed coordinates GAOT runs in its fast mode. The nonlinearity of its encoder and decoder acts on the coordinates only, so the graph and the geometric features are computed once for the shared cloud and reused, and its neighbor cache is emptied before every timed inference call. With variable coordinates it builds a graph per sample on the GPU. It is trained with AdamW at a learning rate of $8\cdot10^{-4}$ on the mean squared error, with an attention dropout of $0.2$, in PyTorch.

\paragraph{RIGNO-12.} RIGNO-12 has a latent width of 128 and twelve message-passing steps, computes in bf16 with parameters and mean aggregations in fp32, and holds 1.77 million parameters. Its regional nodes equal the partition count, whereas the accuracy benchmark places them on a quarter of the points. It runs with fixed coordinates only, and the construction of its graphs is excluded from every timing. The Delaunay regional mesh and the support edges are built once on the CPU before the clock, which would add about 95\,ms per graph at 16384 points and makes RIGNO ill suited to variable geometries. Its processor is unrolled at compile time, so that compiling it at 1024 steps exhausts the memory of the host, in training and in inference alike. It is trained with AdamW on the mean squared error, with edge masking at a rate of $0.5$, in JAX.

\subsubsection{Results}
\label{app:perf-results}

The sweeps answer two questions of scaling, how the cost grows with the model at a fixed problem, and how it grows with the problem at a fixed model. The problem grows in two ways. In the \emph{problem-size} sweep the partitions and the points grow together at a fixed number of points per partition, so that the latent resolution follows the data, which is how a larger problem is met in practice and the basis of a fair comparison. In the \emph{space-size} sweep only the points grow, at a fixed 1024 partitions. Two properties shape every result below. The latent mode of \daedalus{} never reads the points, so its cost depends on the partitions and the processor alone. GAOT, in turn, is fast when all samples of a batch share one point cloud, because the nonlinearity of its encoder and decoder acts on the coordinates only, so that its graph and its geometric features are computed once and reused for the whole batch. With a point cloud of its own for every sample it has to build a graph and geometric features per sample, which scales with the batch and with the points, and this is why the advantage of \daedalus{} is largest with variable geometries, the setting of most practical problems. We report both settings, and the figures give the complete curves, including the regimes where a baseline is faster.

\begin{figure}[!htbp]\centering
  \includegraphics[width=0.8\textwidth]{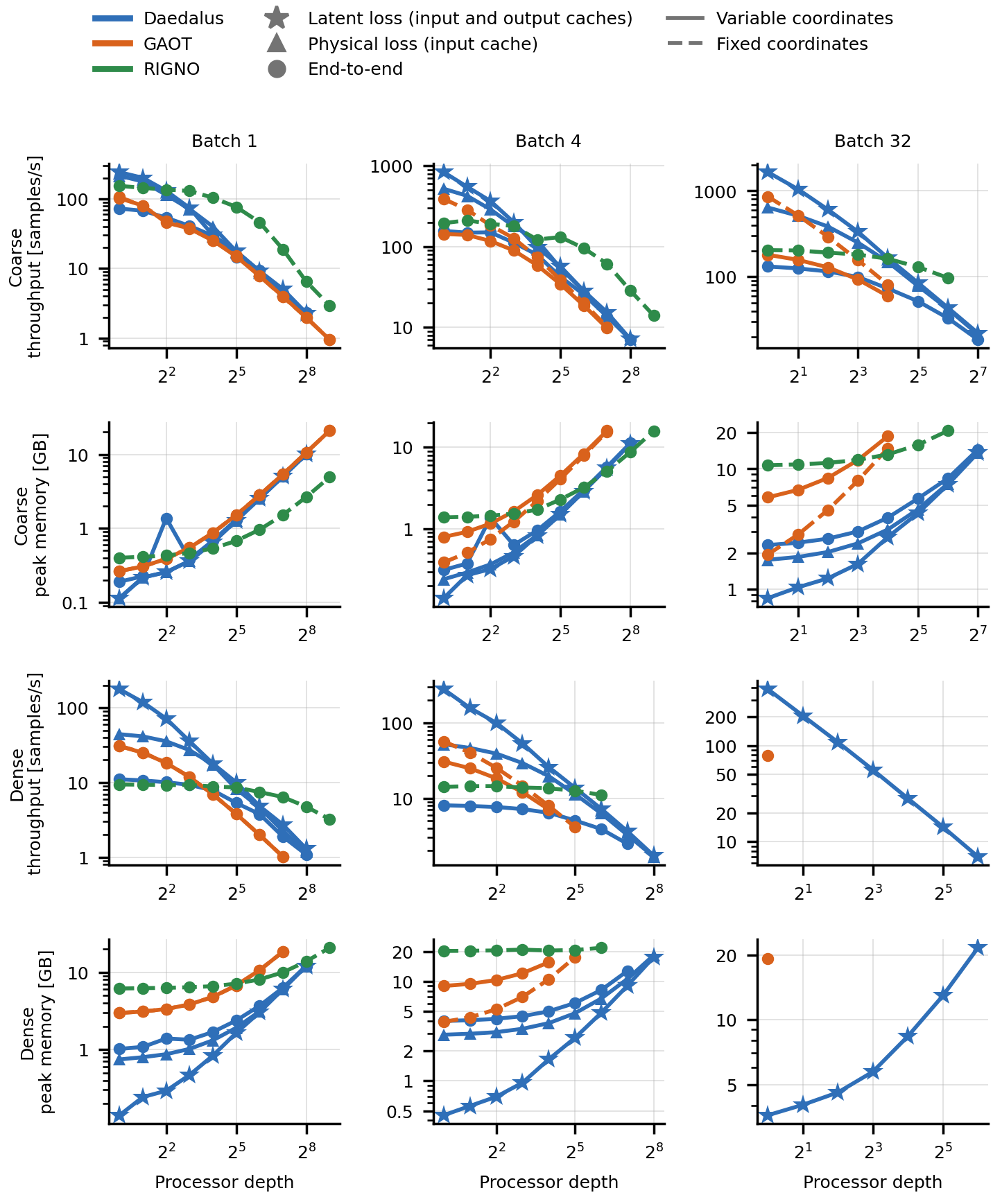}
  \caption{\textbf{Training cost against the model size.} The problem is fixed, either the coarse one (16384 points, 1024 partitions) or the dense one (262144 points, 4096 partitions), and the depth of the processor grows from 1 to 1024, counting the processor units of \daedalus{}, the transformer layers of GAOT, and the message-passing steps of RIGNO. Training throughput and peak memory at batches of 1, 4, and 32 on one RTX~4090. A line ends where its next depth runs out of device or host memory or fails.}
  \label{fig:perf-model-size-training}
\end{figure}

\begin{figure}[!htbp]\centering
  \includegraphics[width=0.8\textwidth]{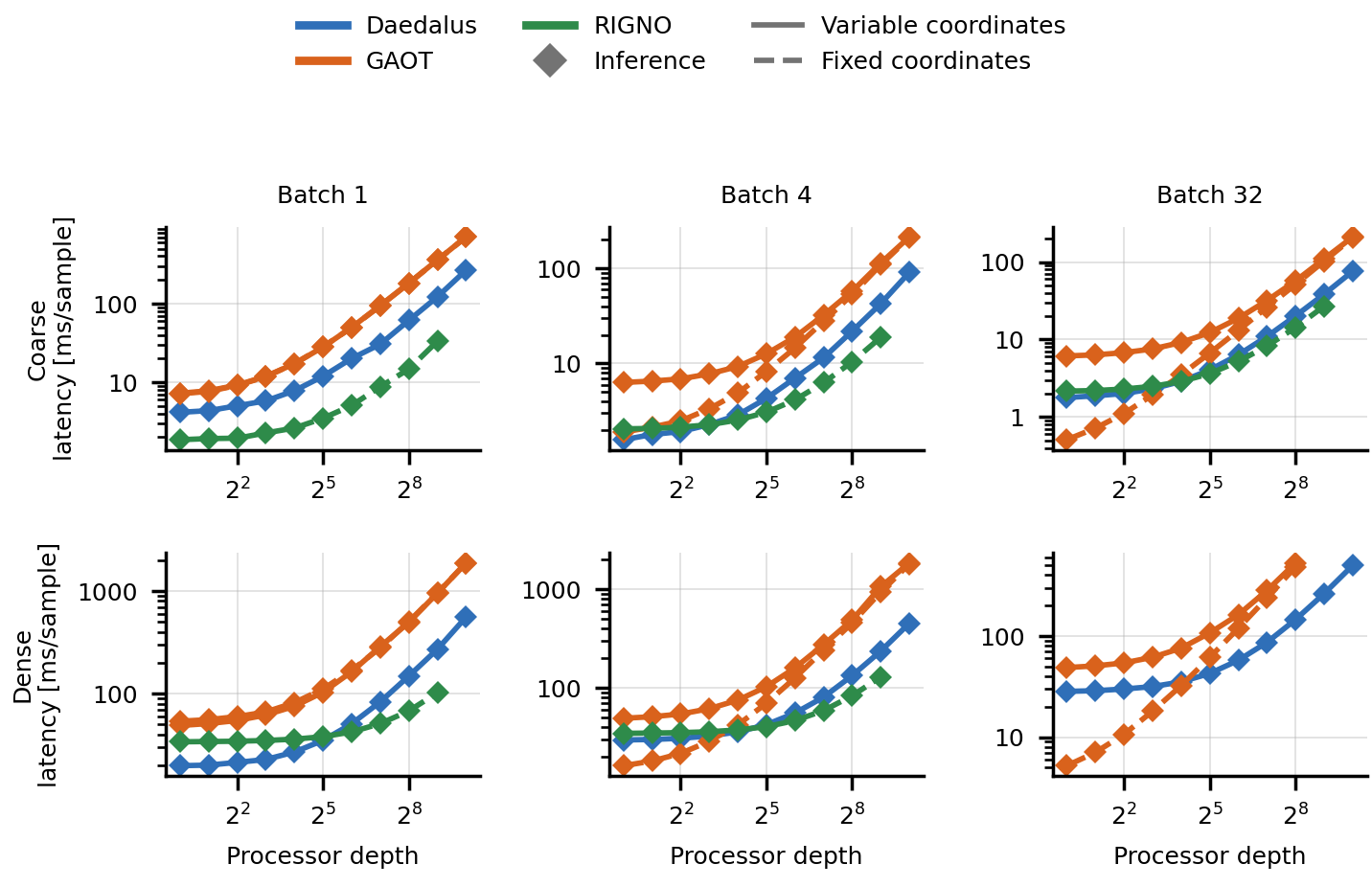}
  \caption{\textbf{Inference latency against the model size}, on the coarse and the dense problem, with the depth swept as in Figure~\ref{fig:perf-model-size-training}. The latency of \daedalus{} is nearly flat over shallow processors, where the backbone and the partitioning dominate, and grows linearly with the depth once its processor does.}
  \label{fig:perf-model-size-latency}
\end{figure}

\paragraph{Model size.} The depth of the processor sets how large a model a GPU can train, and the latent mode leaves almost the whole GPU to the processor (Figure~\ref{fig:perf-model-size-training}). On the dense problem at a batch of 32 it trains a processor of 64 units and 101 million parameters within 21.6\,GB, whereas end-to-end training fits at most a single layer, and only for GAOT with a fixed geometry, 92 times fewer parameters. At a batch of one, where both GAOT settings coincide, the latent mode trains a processor of 404 million parameters against 143 million for GAOT. Across the depths the latent mode trains 2.4 to 5.8 times faster than GAOT on the dense problem at a batch of one, in both settings, and at a batch of four 3.7 to 9.1 times faster with variable and 3.3 to 4.9 times with a fixed geometry. At inference the same holds (Figure~\ref{fig:perf-model-size-latency}). At a batch of one \daedalus{} is 1.7 to 3.5 times faster than GAOT at every depth in both settings, and at a batch of 32 on the dense problem it runs a processor of 1.6 billion parameters, where the largest GAOT has 285 million, while being 1.7 to 3.6 times faster than GAOT with variable geometries. With a fixed geometry, GAOT shares its geometric preprocessing across the batch and is faster for shallow processors, until the processor dominates and \daedalus{} overtakes it from 16 to 32 units.

\begin{figure}[!htbp]\centering
  \includegraphics[width=0.8\textwidth]{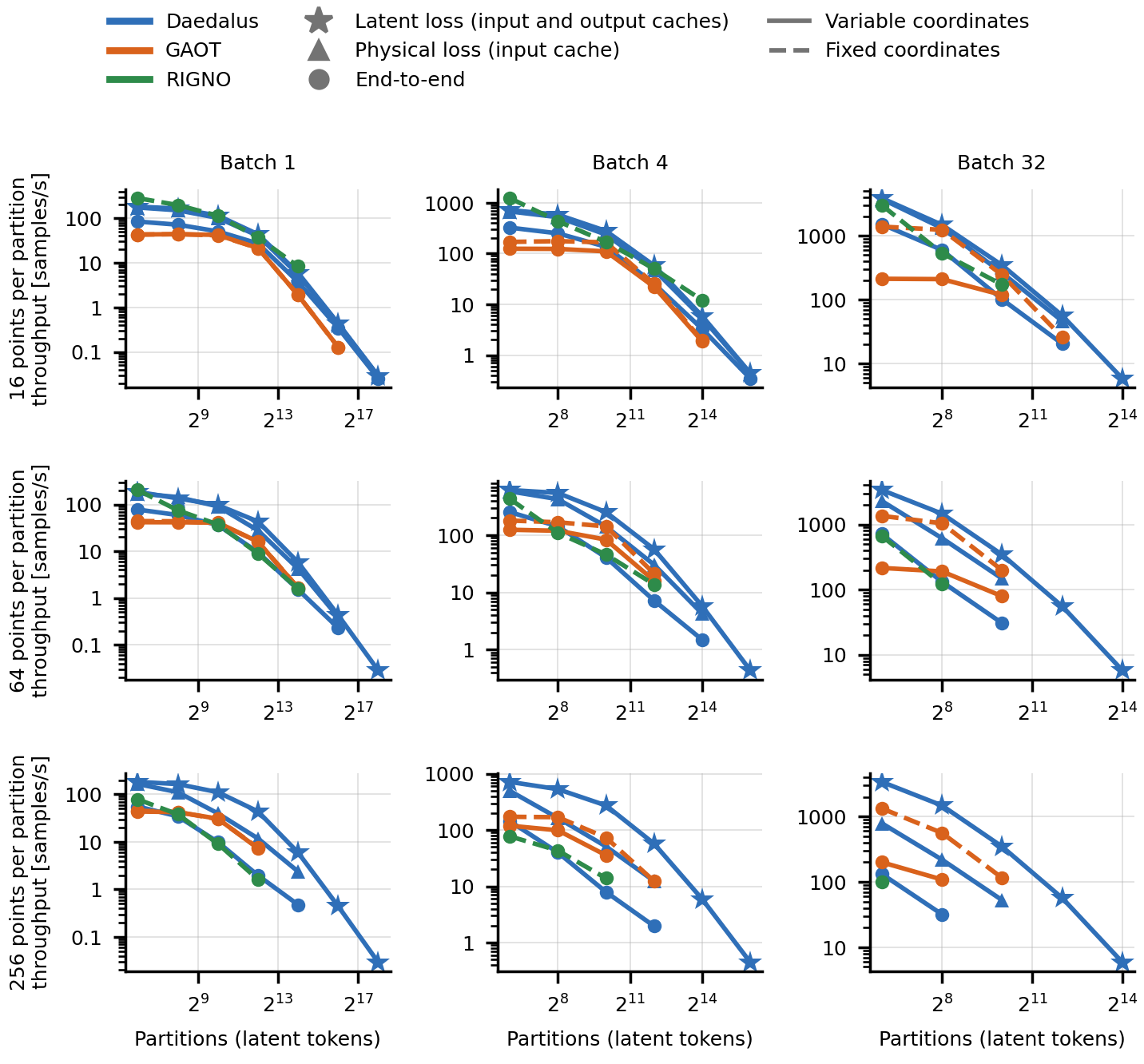}
  \caption{\textbf{Training throughput against the problem size.} The model is fixed, and the number of partitions, or of tokens and regional nodes for the baselines, grows from 64 to 262144 together with the points, at a fixed 16, 64, or 256 points per partition (one row each), so that the points per function grow from 1024 to about 67 million. Batches of 1, 4, and 32 on one RTX~4090. A line ends where its next size runs out of memory.}
  \label{fig:perf-problem-size-training}
\end{figure}

\begin{figure}[!htbp]\centering
  \includegraphics[width=0.8\textwidth]{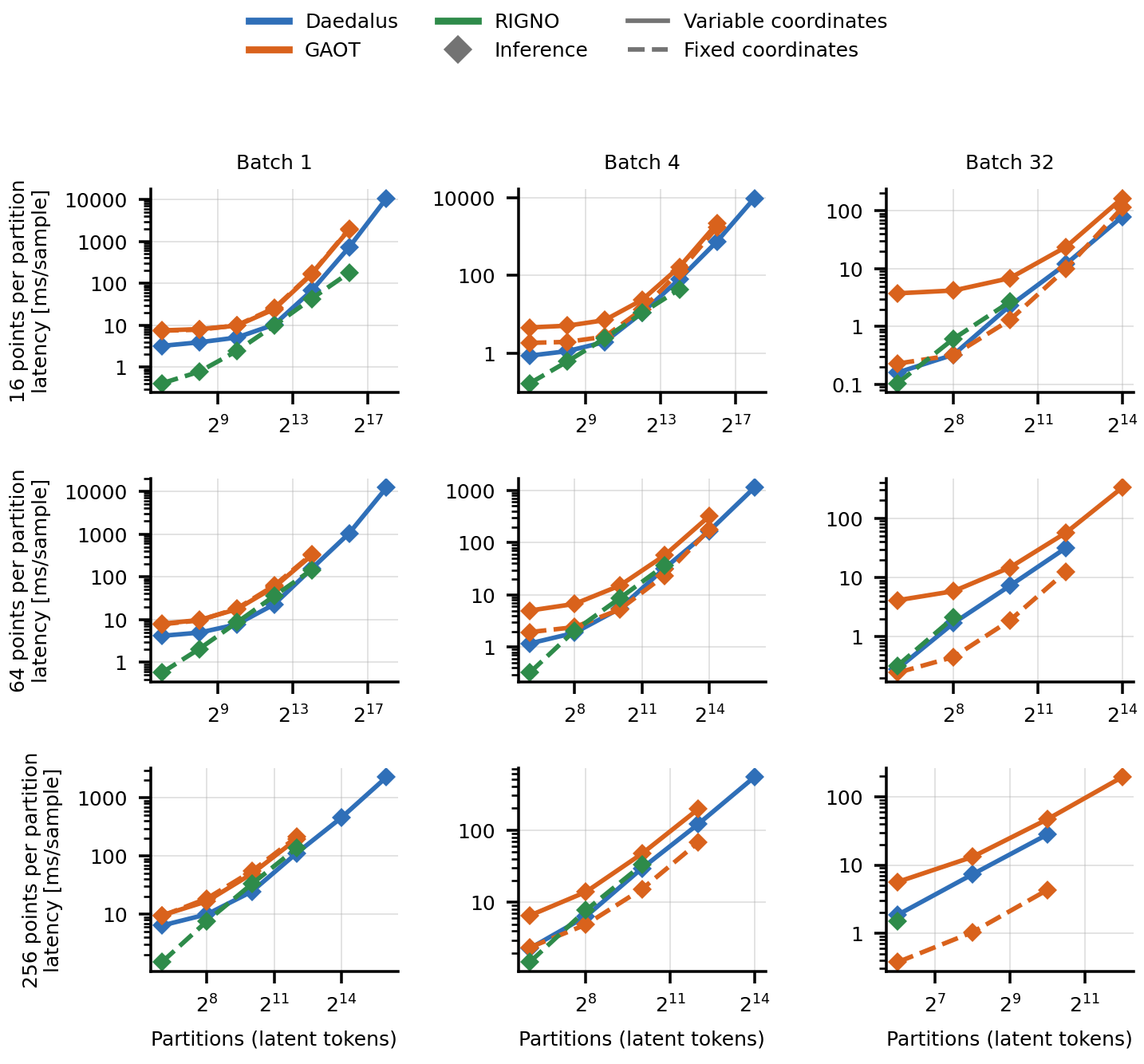}
  \caption{\textbf{Inference latency against the problem size}, with the number of partitions and of points grown together as in Figure~\ref{fig:perf-problem-size-training}. The memory of \daedalus{} grows linearly with the points. At batches of one and four its curves reach problems at least four times larger than those of GAOT and RIGNO.}
  \label{fig:perf-problem-size-latency}
\end{figure}

\paragraph{Problem size.} When the number of partitions grows with the points, the latent mode trains on far larger problems than any end-to-end setting (Figure~\ref{fig:perf-problem-size-training}). At a batch of 32 and 256 points per partition it trains on 16384 partitions and 4.2 million points, 64 times more than GAOT with variable geometries and 16 times more than with a fixed geometry, and at a batch of one it reaches 262144 partitions at every density. Over the sizes both models reach it trains 13.6 to 17 times faster than GAOT with variable geometries at a batch of 32 and 2.5 to 3.0 times faster with a fixed geometry, and the gap widens with the number of points per partition, since the points enter the cost of GAOT but not that of latent training. At inference the memory of \daedalus{} grows linearly with the points, and at a batch of one it is 1.5 to 2.8 times faster than GAOT at every size and infers on problems four to 16 times larger (Figure~\ref{fig:perf-problem-size-latency}).

\begin{figure}[!htbp]\centering
  \includegraphics[width=0.8\textwidth]{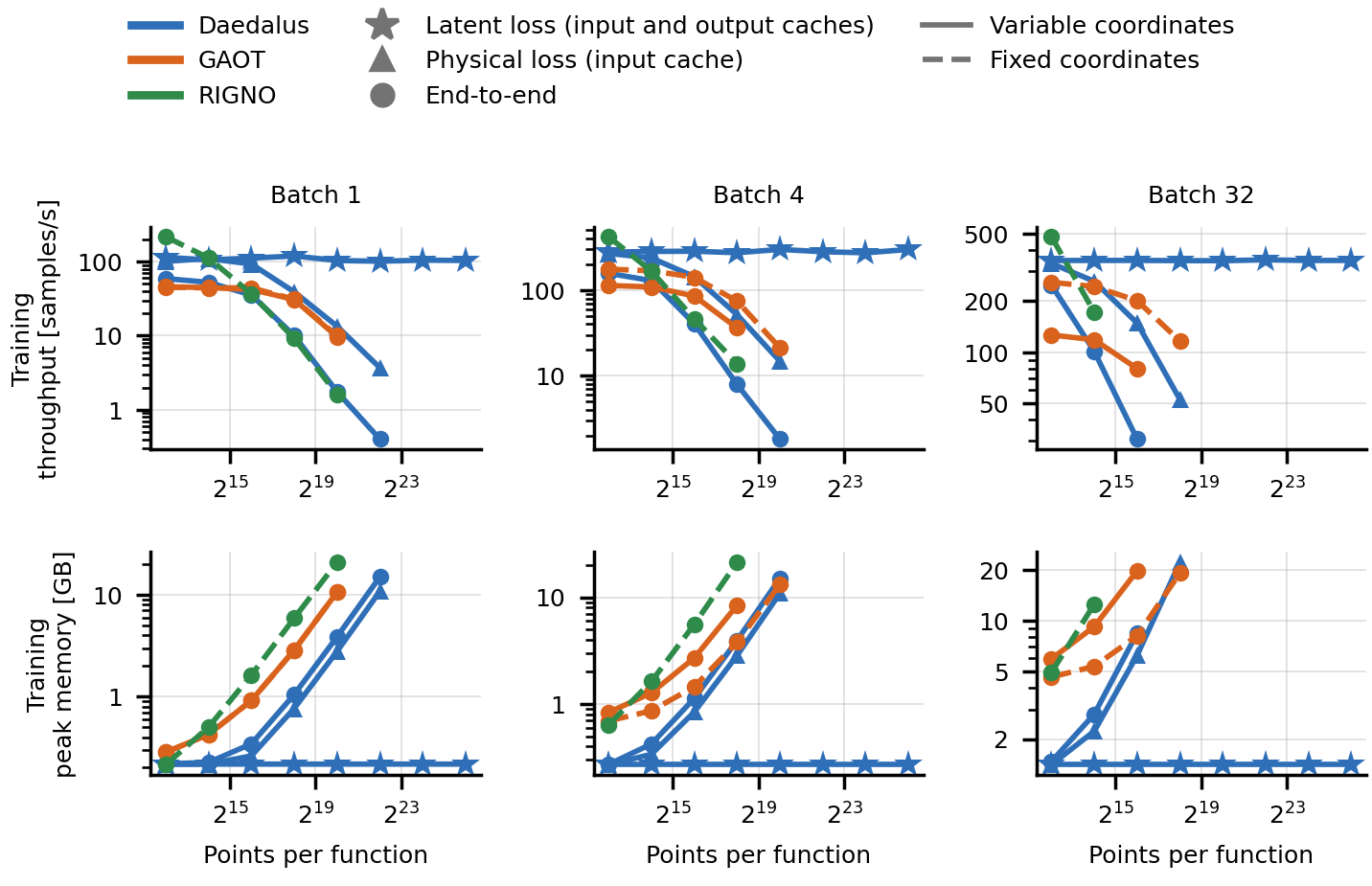}
  \caption{\textbf{Training cost against the space size.} The model and the latent size are fixed, at 1024 partitions, tokens, or regional nodes, and only the points per function grow, from 4096 to about 67 million. Training throughput and peak memory at batches of 1, 4, and 32 on one RTX~4090. A line ends where its next size runs out of memory or fails.}
  \label{fig:perf-space-size}
\end{figure}

\paragraph{Space size.} At a fixed 1024 partitions the latent mode is independent of the resolution of the data (Figure~\ref{fig:perf-space-size}). From 4 thousand to 67 million points per function it trains at the same rate, about 105, 280, and 347 samples per second at batches of 1, 4, and 32, within the same memory, 0.22\,GB at a batch of one and 1.4\,GB at a batch of 32, whereas every setting that encodes or decodes the points slows down and runs out of memory as they grow. At a million points and a batch of one it trains 10 times faster than GAOT, with fixed and variable geometries alike, within 50 times less memory, and the training data therefore need no spatial downsampling. At inference \daedalus{} handles functions of 16.8 million points in one piece, and chunked inference lifts this bound further (Appendix~\ref{app:perf-chunks}).

\begin{figure}[!htbp]\centering
  \includegraphics[width=0.8\textwidth]{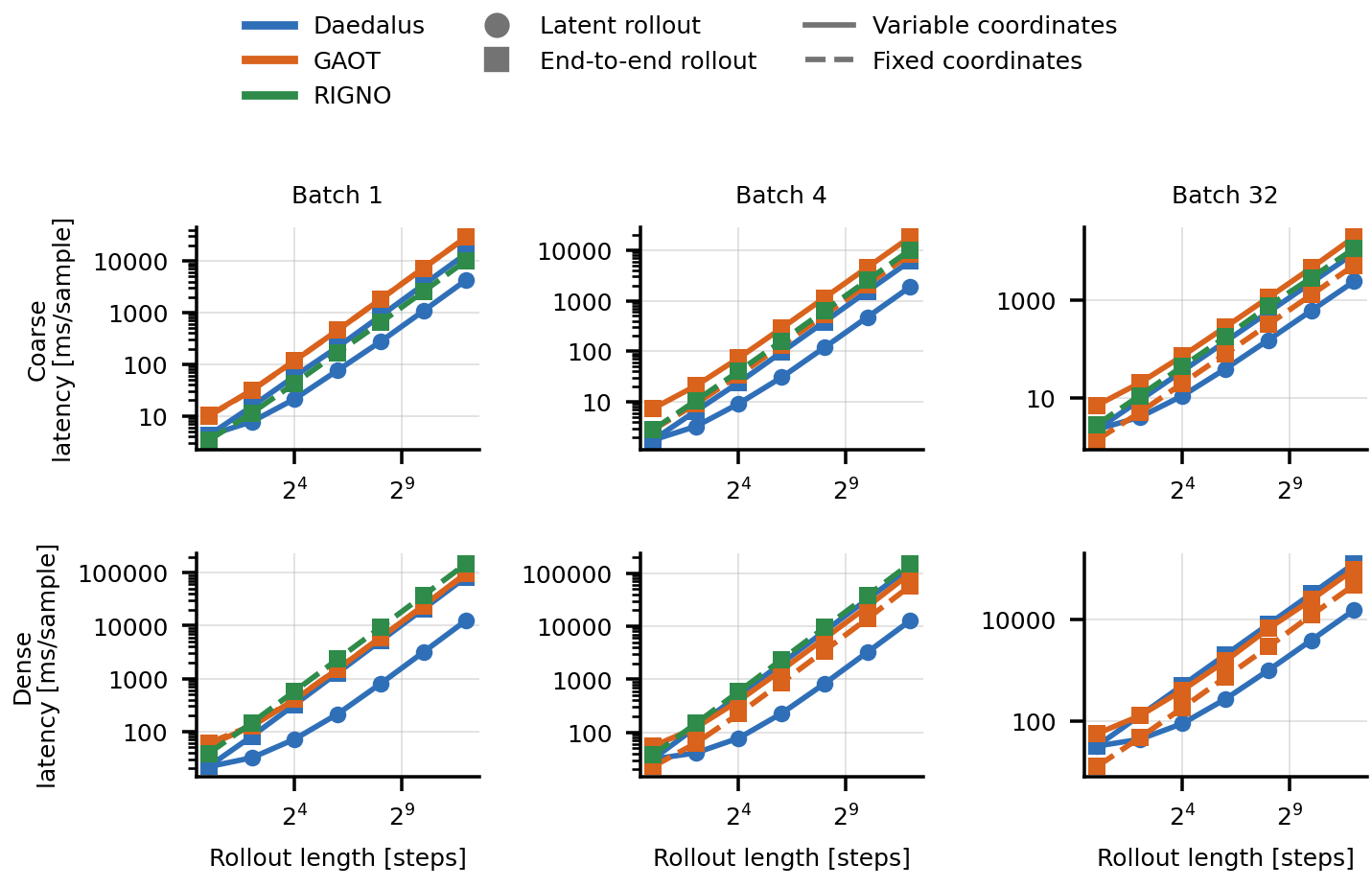}
  \caption{\textbf{Rollout latency against the rollout length.} The model and the problem are fixed, the coarse or the dense one, and the operator is applied autoregressively for 1 to 4096 steps. The latent rollout encodes once and decodes only the final state, whereas the end-to-end rollout of \daedalus{} partitions, encodes, and decodes at every step, and GAOT and RIGNO step in the physical space.}
  \label{fig:perf-rollout}
\end{figure}

\paragraph{Rollouts.} An end-to-end rollout costs an encoding, a processor step, and a decoding per step, so its latency grows linearly with its length (Figure~\ref{fig:perf-rollout}). A latent rollout encodes once and decodes once, and a step costs the processor alone. Over 4096 steps on the dense problem it is 6.6 to 10 times faster than the end-to-end rollout of \daedalus{} itself, and 6.4 to 7.8 times faster than GAOT with variable geometries and 3.1 to 7.6 times with a fixed geometry. The memory of a rollout does not grow with its length.

\subsubsection{Chunked inference}
\label{app:perf-chunks}

\begin{figure}[!htbp]\centering
  \includegraphics[width=\textwidth]{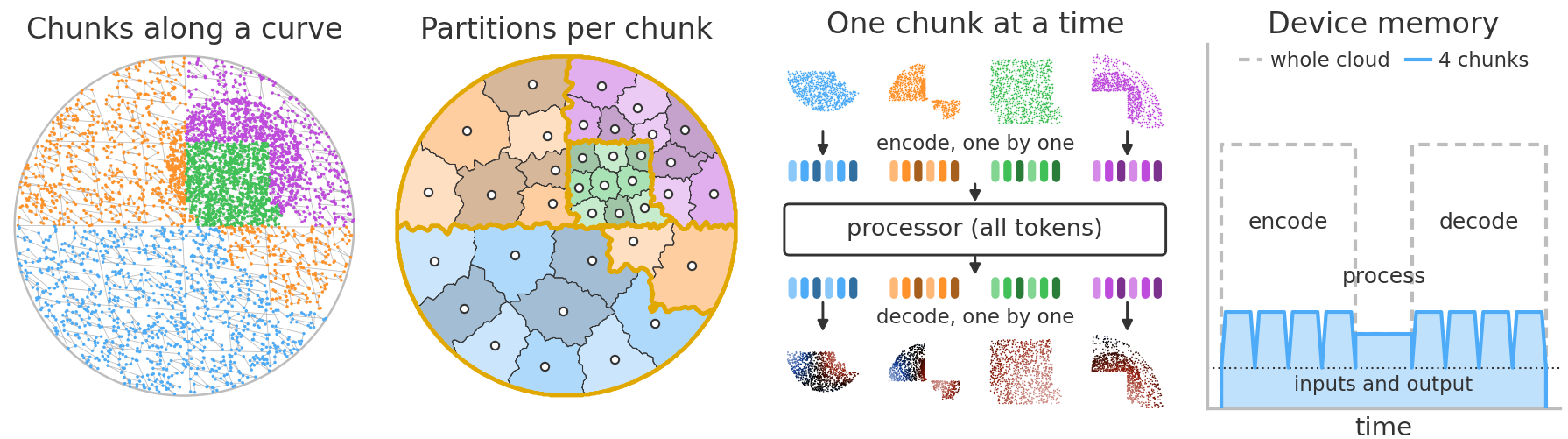}
  \caption{\textbf{Chunked inference.} (a) The points ordered along a Z-order curve and cut into four runs of equal length, the chunks, one color each. An equal run can be irregular in shape and come in more than one piece, as the orange chunk does. (b) Every chunk partitioned on its own, its partitions in shades of its color, with the chunk boundaries in yellow, which no partition crosses. (c) The chunks are encoded one after the other into their embeddings, the processor acts on the embeddings of all chunks together, and the chunks are decoded one after the other. (d) A sketch of the memory over one inference. In one piece the encoder and the decoder hold their intermediate features for every point at once, whereas in chunks they hold them for a single chunk, and only the inputs and the output remain for every point.}
  \label{fig:chunking}
\end{figure}

The memory of inference grows linearly with the points (Figure~\ref{fig:perf-problem-size-latency}), since the encoder and the decoder hold intermediate features for every pair of a point and a partition it is assigned to, so that a single GPU bounds the size of the function it can map. The segregation of the FAE from the processor lifts this bound. The encoder reduces every partition to its embedding independently of all others, and the decoder evaluates every point from its own coordinate and the embeddings of its partitions, so neither needs the whole cloud at once. Only the processor couples the partitions, and it acts on embeddings, which are fewer than the points by the size of a partition. In \emph{chunked inference} the points of a function are cut into $C$ chunks of equal size along a space-filling curve, every chunk is partitioned into $P/C$ partitions of its own, and the chunks are encoded one after the other (Figure~\ref{fig:chunking}). The processor then acts on the embeddings of all chunks together, exactly as without chunks, so that information still travels across the whole domain, and the decoder again treats one chunk at a time and writes its values into the output. Apart from the inputs and the output, the GPU holds the intermediate arrays of a single chunk at any moment, and the memory they take shrinks with the number of chunks. The assignments of the points to their partitions are not kept between the encoding and the decoding. They are recomputed chunk by chunk from the stored clustering, which reproduces the assignments of the encoding exactly. The mode changes no weight and is switched on at inference for a model trained without it.

A partition never spans two chunks, so the result depends slightly on their number. An equal run of the curve can end in a thin tail or include a small detached piece, and a partition there can be longer and thinner than the compact partitions the decoder is trained on. The share of such partitions falls as the number of partitions per chunk grows. With the low-dimensional backbone, on functions of about a million points in 4096 partitions, the median relative $L^2$ error is $0.208\%$ in one piece and $0.215\%$ in 64 chunks.

We measure the inference of the model of Appendix~\ref{app:perf-settings}, partitioning and encoding followed by processing and decoding, at a batch of one and for 1, 4, 16, 64, and 256 chunks, a single chunk being the plain inference. The number of partitions grows with the points, at 16, 64, and 256 points per partition. The processor groups $M$ embeddings into one token, $1024$ divided by the points per partition, so that it holds one token per 1024 points at every density and never bounds the memory, and the partitioner assigns every point to a single partition. The compiling call is excluded and the latency is the median of three calls, with no timeout. Synthetic functions of a billion points exceed the memory of the host, so the points are drawn uniformly on the disk and given cheap smooth values, which change neither the latency nor the memory.

\begin{figure}[!htbp]\centering
  \includegraphics[width=0.8\textwidth]{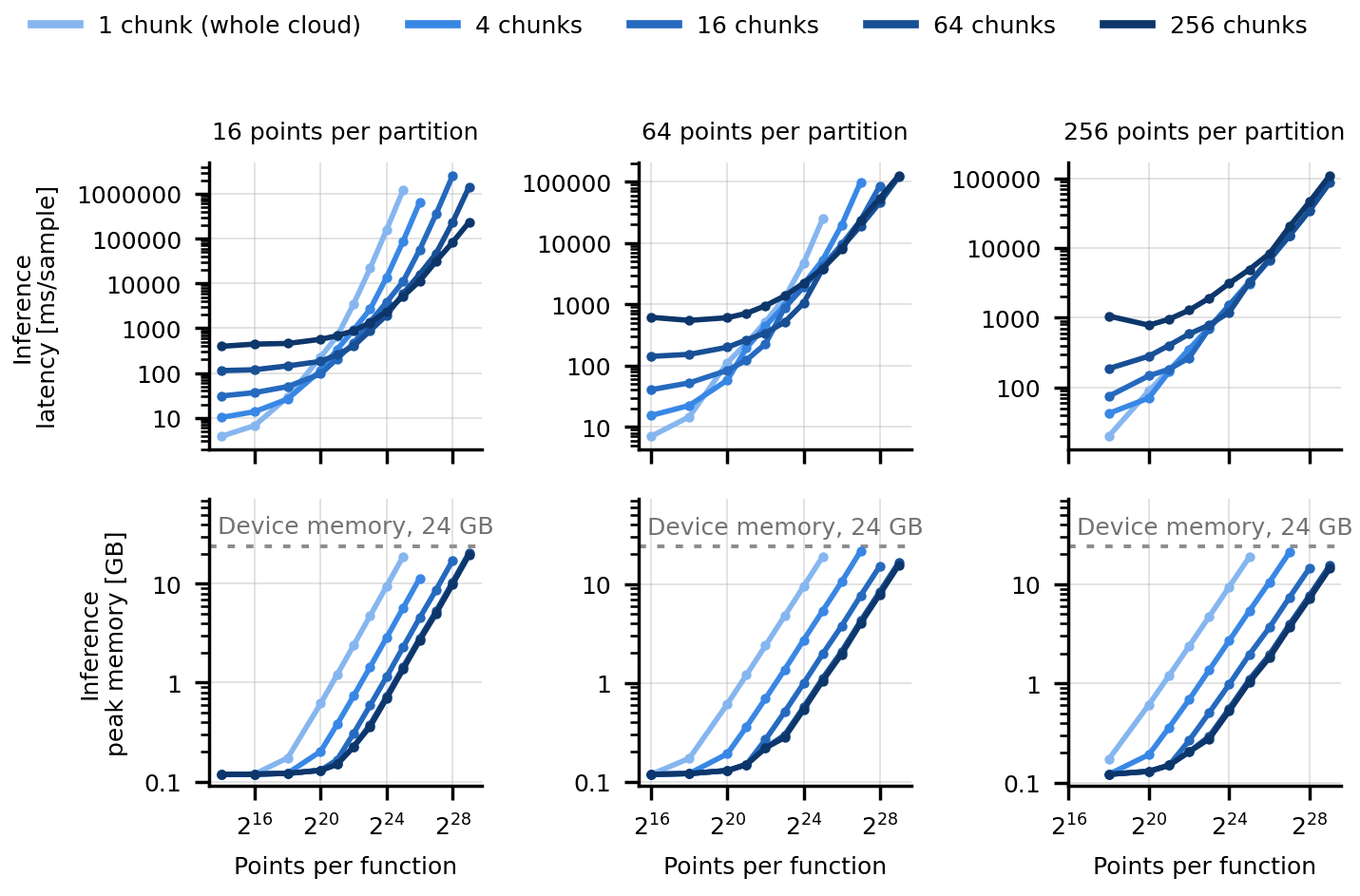}
  \caption{\textbf{Chunked inference against the problem size.} Inference latency and peak memory of \daedalus{} against the points per function, which grow with the number of partitions at 16, 64, and 256 points per partition, at a batch of one on one RTX~4090, one line per number of chunks, and the memory of the GPU dotted. A line ends at the last size that fits, the next size running out of memory. In one piece the largest function has 33.6 million points, and in 64 chunks it has 537 million at every density.}
  \label{fig:perf-chunks}
\end{figure}

\paragraph{Results.} In one piece the largest function that fits on the GPU has 33.6 million points at every density, at about 19\,GB (Figure~\ref{fig:perf-chunks}). Chunks lower the memory at that size, at 64 points per partition from 18.8\,GB to 5.3\,GB in four chunks, 2.0\,GB in 16, and 1.1\,GB in 64, 17 times less. Sixteen chunks reach 268 million points, and 64 chunks reach 537 million at every density, 16 times more than in one piece, in 86 seconds within 15.4\,GB at 256 points per partition and in 121 seconds within 16.5\,GB at 64. More chunks lower the memory only a little further, since what is held for every point then dominates, and 256 chunks reach the same size, while the next size, about 1.07 billion points, runs out of memory for every number of chunks. The chunks run one after the other, so a small function pays for them, 142\,ms in 64 chunks against 7\,ms in one piece at 65 thousand points and 64 points per partition. A large function does not, because the cost of the partitioner grows faster than linearly in the partitions it clusters at once. Sixty-four chunks infer faster than one piece from 4.2 million points at 64 points per partition and from 16.8 million at 256, and at 33.6 million points and 64 points per partition they take 3.8 seconds against 24.7. At 16 points per partition, where the same points form four to sixteen times as many partitions, chunks are above all a speed-up. At 33.6 million points one piece takes 1229 seconds and 64 chunks 5.9, and 256 chunks map 537 million points in 33.6 million partitions in 234 seconds, against 1442 in 64 chunks.

\FloatBarrier
\clearpage

\section{Architecture}
\label{app:architecture}
\setcounter{figure}{0}
\setcounter{table}{0}

\begin{figure}[!htbp]\centering
  \includegraphics[width=0.72\textwidth]{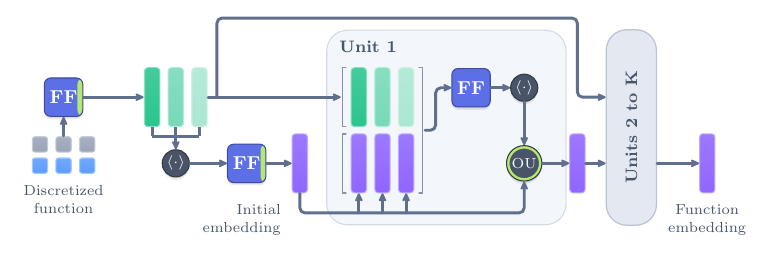}
  \caption{\textbf{Encoder.} Many points to one embedding, via edge-wise pairing, segment-mean aggregation, and the on-sphere orthogonal refinement of Equation~\eqref{eq:refine}, drawn as the orthogonalized update (OU), with FF marking a feed-forward layer.}
  \label{fig:arch-enc}
\end{figure}

\paragraph{Encoder.} The encoder turns the many points of one partition into a single embedding (Figure~\ref{fig:arch-enc}). Each point's coordinate and value are projected into a feature, the running embedding is paired with every point feature and each pair is processed into an update, the updates are mean-aggregated over the partition, and the embedding is refined by an orthogonalized, re-normalized correction. The embedding is always a direction of fixed magnitude, because an RMSNorm after every unit keeps it on the hypersphere. Writing $\embmorph$ for the current embedding and $g$ for the normalized aggregate, the correction drops the component of $g$ along $\embmorph$ before applying it,

\begin{equation}
  g_\perp = g - \frac{g\cdot\embmorph}{\embmorph\cdot\embmorph}\,\embmorph,
  \qquad
  \embmorph \leftarrow \mathrm{RMSNorm}\!\big(\embmorph + e^{\log\alpha}\,g_\perp\big),
  \label{eq:refine}
\end{equation}
with a learned scalar $\alpha$ shared by all units of the encoder ($\log\alpha$ is the trained parameter). Correcting rather than replacing the normalized embedding keeps its distribution stable from unit to unit, which is what allows the same unit to be applied many times over. Orthogonalizing the correction removes the component aligned with the current embedding, which would merely rescale the embedding and be undone by the normalization that follows, so that component does no work while still being able to destabilize the update. What remains is exactly the part that moves the embedding along the sphere. This is close to the normalized-transformer construction of \citet{loshchilov2025ngpt}, which also re-normalizes the hidden state onto the unit hypersphere after every layer, there with a learned per-dimension scale, but without the orthogonalization step.

\begin{figure}[!htbp]\centering
  \includegraphics[width=0.8\textwidth]{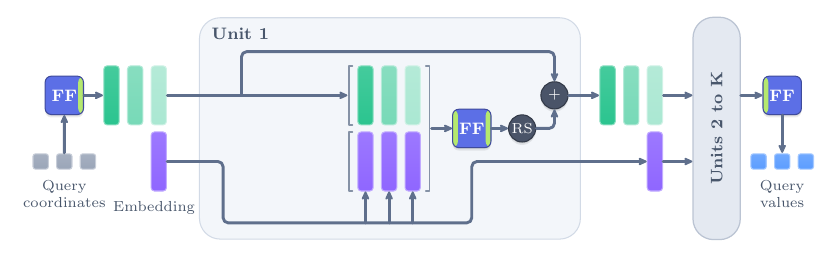}
  \caption{\textbf{Decoder.} One embedding to many points, by concatenating the embedding with each lifted query coordinate, merging through the units independently per point, and un-normalizing with the stored static frame, where FF marks a feed-forward layer and RS the residual scale.}
  \label{fig:arch-dec}
\end{figure}

\paragraph{Decoder.} The decoder maps one embedding back to many points (Figure~\ref{fig:arch-dec}). It lifts every query coordinate to a hidden state, merges the embedding into each query's hidden state through a stack of residual units that read their concatenation, refines every query independently into an output value, and un-normalizes with the per-partition static frame. Values are aggregated back to points by a segment mean over the partition assignments. Because a query is answered from its own coordinate alone, the decoded field is a neural field in the strict sense, and its spatial gradient is therefore available exactly, at the cost of one forward-mode pass per coordinate axis rather than of a finite difference.

\begin{figure}[!htbp]\centering
  \includegraphics[width=\textwidth]{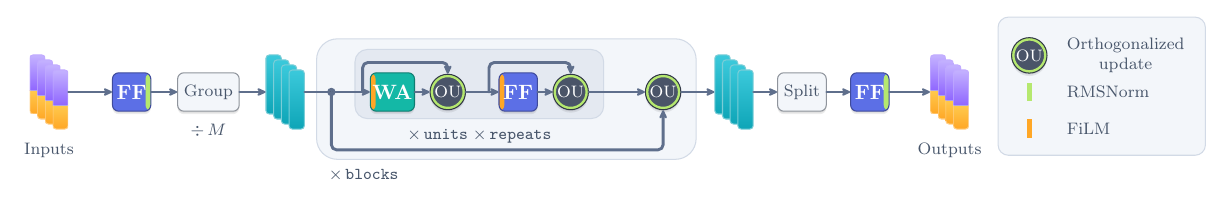}
  \caption{\textbf{Processor.} From the input embeddings to the output embeddings. The lift maps the embeddings to width $d/M$, the grouping concatenates $M$ neighbors along a random space-filling curve into tokens of width $d$, and the block is applied repeatedly, each block being its units (a windowed attention and a feed-forward layer, each folded back by its own orthogonalized update, OU) and a long-range OU from before the block to after it. The amber stripes mark the sublayer inputs that a conditioned processor modulates by the lead time (feature-wise linear modulation, FiLM), and the rotary position embeddings of the attention are not drawn. The tokens are then split and read out.}
  \label{fig:arch-proc}
\end{figure}

\begin{figure}[!htbp]\centering
  \includegraphics[width=\textwidth]{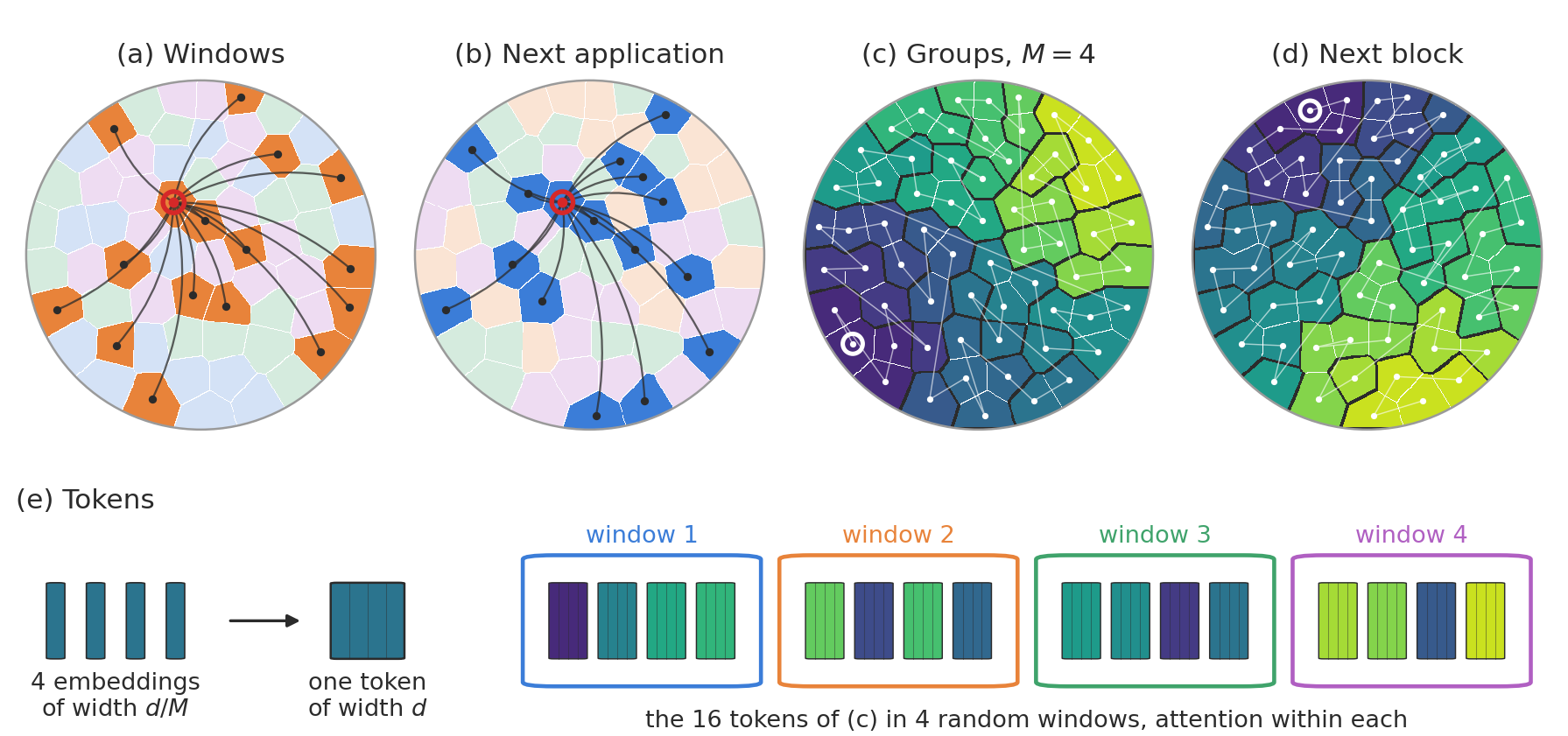}
  \caption{\textbf{Windows and groups of the processor}, on $64$ partition embeddings drawn as the cells of their partition centers. (a) Windowed attention without grouping ($M{=}1$). One unit application permutes the embeddings at random and cuts them into $W{=}4$ windows, and a query (ringed) attends only to the members of its window (colored and linked), which are scattered over the whole domain, so the attention remains global. (b) The next application draws a new permutation, and the window of the same query changes. (c) Grouping with $M{=}4$. The embeddings are ordered along a Z-order curve under a random rotation (thin line, its start ringed), and every run of $M$ consecutive embeddings, a compact group of neighbors, forms one token (outlined, shaded along the curve). (d) The next block rotates the curve anew, so the groups slide across depth. (e) The $M$ embeddings of a group, each of width $d/M$, are concatenated into one token of width $d$, and the $16$ tokens of (c) are cut into $W{=}4$ random windows, so that every window gathers groups from all along the curve.}
  \label{fig:swa}
  \label{fig:swa-members}
  \label{fig:swa-grouping}
\end{figure}

\paragraph{Processor.} For operator learning a processor maps the collection of embeddings to a new one (Figure~\ref{fig:arch-proc}). A lift, a feed-forward layer, maps the embeddings of every partition, all input channels with their static frames, to width $d/M$. The embeddings are then ordered along a space-filling curve drawn at random, and every run of $M$ consecutive embeddings is concatenated into one token of the fixed attention width $d$ (Figure~\ref{fig:swa-members}). The processor applies its block several times with shared weights. A block consists of a few distinct units, each applied several times in place, and a long-range residual that folds the output of the units back into the block's input tokens by one orthogonalized update. Each unit has two sublayers, a windowed attention and a feed-forward layer, and each is folded back into the tokens by its own orthogonalized update of the kind used in the encoder (Equation~\eqref{eq:refine}), the feed-forward layer reading the tokens that the attention has already updated. Every block groups along a new curve, so the groups slide across depth. After the last block the tokens are split back into embeddings, and a read-out writes a value frame and an RMS-normalized embedding for every output channel. The same read-out can also be taken after every few blocks. The geometry frame is copied from the input, and the number of output channels may differ from the input (for instance $\funcMain\!\to\!\nabla\funcMain$ maps one channel to two). All normalizations are RMS normalizations without learned weights, since every orthogonalized update takes its reference to be RMS-normalized.

\paragraph{Positions.} The random windows carry no notion of where their tokens lie, and the geometry that the lift reads from the static frame is progressively blurred with depth. Every windowed attention therefore rotates its queries and keys by rotary position embeddings \citep{su2024roformer} of the token positions, applied axially in two dimensions as by \citet{heo2024rotary}. The position of a token is the center of its partition, or, for a token of several members, the mean center of the partitions it concatenates along the curve of the current block. The first half of the dimensions of every head is rotated by the $x$-coordinate and the second half by the $y$-coordinate, each half as pairs at frequencies spaced geometrically from a wavelength of twice the domain $[-1,1]$ down to about the spacing of the tokens, so that a logit depends on the offset between the two tokens alone. The values are not rotated, the rotation adds no parameters, and the head dimension has to be divisible by four.

\paragraph{Lead-time conditioning.} A processor for a time-continuous operator (Appendix~\ref{app:operator-time}) receives the lead time $\tau$, normalized by the span of the time window, in two ways. The lead time is concatenated into every lift input, and it modulates both sublayers of every unit by feature-wise linear modulation \citep{perez2018film}. A small network per unit maps $\tau$ to a scale $\gamma$ and a shift $\beta$ for each sublayer, which then reads $(1+\gamma)\odot\mathbf{h}+\beta$ in place of the tokens $\mathbf{h}$. The last layer of that network is initialized at zero, so a freshly initialized unit is exactly the unconditioned one. The modulation acts on what a sublayer reads and never on the tokens into which the orthogonalized update folds its output, and these tokens therefore stay RMS-normalized, as the update requires. For the same reason the feed-forward layer carries no normalization of its own, since the tokens it reads are already normalized and a further normalization would undo part of the modulation, and its input is a normalized copy of the tokens that is scaled and shifted by the lead time. A processor for a steady operator receives no condition and carries no modulation.

In this work the processor is a transformer, one choice among several. What the processor receives is a set of embeddings anchored at the partition centers, which is a point cloud of $P$ embeddings carrying their own coordinates, so any architecture mapping such a set to a set of the same cardinality can take its place. A graph neural network built on a neighborhood graph of the partition centers is the closest alternative and has two properties worth noting. Its message passing is local by construction, which matches operators with finite propagation speed and avoids the all-to-all coupling that attention applies at every layer, and its cost grows with the number of edges rather than with $P^2$, so it scales differently as the partitioning is refined. Convolutional and spectral processors are admissible as well whenever the partition centers happen to lie on a regular grid. Since the backbone is frozen and the embeddings are cached, swapping the processor requires no retraining of the FAE, and the comparison of reduced-space architectures on a fixed latent space is left to future work.

\paragraph{Windowed attention.} Full self-attention over $T$ tokens costs $O(T^{2})$. Windowed attention permutes the tokens at random, splits them into $W$ windows, and attends within each window, which reduces the cost to $O(T^{2}/W)$, and $W{=}1$ recovers full attention (Figure~\ref{fig:swa}). Every window is spread across the whole cloud, so the attention remains global, and the windows are redrawn at every application of a unit, so they slide across depth.

Local mixing is provided by the grouping instead. The query, key, and value projections and the feed-forward layer act on a whole token, so they mix its $M$ members locally and cheaply, while the attention between tokens carries the long range. The attention runs over $T = P/M$ tokens, so one application costs $O(P^{2}/(M^{2}W))$. Windowing trades a global but sparser attention for a lower cost. Without it the attention scores of all token pairs dominate the cost of a step once the partitions number in the thousands, and every doubling of $W$ roughly halves that part of the cost. An attention kernel that never stores the scores removes them from the memory of a step, and windowing then reduces the computation only.

\paragraph{Weight sharing.} Each component holds a small number of distinct units and deepens the forward pass by two weight-shared repetition counts. Unit repetitions re-apply each unit in place, and block repetitions re-apply the whole sequence of units, so the depth grows without adding parameters, and the effective parameter count, the stored one times the number of applications, is far larger than the stored one. In the decoder the total number of applications sets a fixed $1/\sqrt{\text{depth}}$ residual scale, and in the processor the initial strength $1/(4\sqrt{\text{depth}})$ of the update of every unit, while the encoder learns one update strength shared by all its applications.

\FloatBarrier
\clearpage

\end{document}